\documentclass[paper=a4, pagesize=auto, abstract=true, twocolumn=true, fontsize=10pt]{scrarticle}

\usepackage[utf8]{inputenc}
\usepackage[T1]{fontenc}
\usepackage[english]{babel}

\usepackage{amsmath}
\usepackage{amssymb}
\usepackage{amsfonts}
\usepackage{amssymb}
\usepackage{amsthm}

\usepackage{times}

\usepackage[ruled,vlined]{algorithm2e}
\usepackage{booktabs}
\usepackage{dsfont}
\usepackage{footmisc}
\usepackage[a4paper, margin=18mm, tmargin=25mm, bmargin=27mm]{geometry}
\usepackage{graphicx}
\usepackage[pagebackref=true,breaklinks=true,colorlinks,bookmarks=false,hypertexnames=false]{hyperref}
\usepackage{lastpage}
\usepackage{longtable}
\usepackage{mathtools}
\usepackage{multirow}
\usepackage{stmaryrd}
\usepackage[dvipsnames]{xcolor}

\usepackage{tikz}
\usetikzlibrary{calc,fit}

\usepackage[numbers,square,sort]{natbib}
\usepackage{microtype} 

\addtokomafont{title}{\rmfamily}
\addtokomafont{section}{\rmfamily}
\addtokomafont{subsection}{\rmfamily}
\addtokomafont{subsubsection}{\rmfamily}
\addtokomafont{paragraph}{\rmfamily}
\addtokomafont{date}{\normalsize}
\addtokomafont{captionlabel}{\bfseries}
\setcapindent{0pt}
\makeatletter
\newcommand\Paragraph{\@startsection{paragraph}{4}{\z@}%
                                    {\z@}%
                                    {-.4em plus -1.0em}%
                                    {\normalfont\normalsize\bfseries}}
\newcommand\ParagraphRun{\@startsection{paragraph}{4}{\z@}%
                                       {\z@}%
                                       {-\fontdimen2\font plus -\fontdimen3\font minus -\fontdimen4\font}%
                                       {\normalfont\normalsize\bfseries}}
\makeatother

\newtheoremstyle{plain}%
  {2pt plus .2em}
  {2pt plus .2em}
  {\itshape}
  {}
  {\bfseries}
  {.}
  {7pt plus 10pt}
  {}

\theoremstyle{plain}
\newtheorem{proposition}{Proposition}

\newcommand{\cf}{cf.\xspace}
\newcommand{\eg}{e.\,g.\xspace}

\newcommand{\ie}{i.\,e.\xspace}
\newcommand{\wrt}{w.\,r.\,t.\xspace}

\newcommand\SV{\mathcal{V}}
\newcommand\SE{\mathcal{E}}
\newcommand\SG{\mathcal{G}}
\newcommand\SL{\mathcal{L}}
\newcommand\SR{\mathbb{R}}
\newcommand\SN{\mathcal{N}}

\newcommand\Sdata[1]{{\fontfamily{cmvtt}\selectfont #1}}
\newcommand\Salg[1]{{\fontfamily{cmvtt}\selectfont #1}}
\newcommand\Ssup[1]{\S\ref{#1}}

\newcommand{\Sopt}[1]{{\fontsize{6pt}{6pt}\selectfont #1}\hspace*{-1em}}

\DeclareMathOperator*\argmin{arg\,min}

\title{Fusion Moves for Graph Matching}

\author{
  Lisa Hutschenreiter$^*$,\hspace{1em}%
  Stefan Haller$^*$,\hspace{1em}%
  Lorenz Feineis$^*$,\\
  Carsten Rother$^*$,\hspace{1em}%
  Dagmar Kainmüller$^\dag$,\hspace{1em}%
  Bogdan Savchynskyy$^*$\\[1.1ex]
  \large
  $^*$\,Heidelberg University, $^\dag$\,Max Delbrück Center for Molecular Medicine, Berlin
}

\date{}

\titlehead{%
  \color{gray}\centering%
  \textbf{\footnotesize Before printing:}
  Note that this document has \pageref*{LastPage} pages.}

\begin{document}

\maketitle

\begin{abstract}
  We contribute to approximate algorithms for the quadratic assignment
  problem also known as graph matching.
  Inspired by the success of the fusion moves technique developed for
  multilabel discrete Markov random fields,
  we investigate its applicability to graph matching. In particular,
  we show how fusion moves can be efficiently combined with
  the dedicated state-of-the-art dual methods that have recently
  shown superior results in computer vision and bio-imaging applications.
  As our empirical evaluation on a wide variety of graph matching
  datasets suggests, fusion moves significantly improve performance of these
  methods in terms of speed and quality of the obtained solutions.
  Our method sets a new state-of-the-art with a notable margin with respect to its competitors.
\end{abstract}

\section{Introduction}\label{sec:intro}

The \emph{quadratic assignment problem} also known as \emph{graph matching} is one of the most prominent combinatorial problems having
numerous applications. In computer vision it is predominantly used for feature matching~\cite{GraphMatchingDDTorresaniEtAl}.
The modern approach to this application is \emph{deep graph matching}, see \eg~\cite{zanfir2018deep,rolinek2020deep,sarlin2020superglue}, which enjoys constantly growing attention in the community.
As follows from the name, deep graph matching combines neural networks with combinatorial matching techniques for inference and joint learning.
Whereas most of earlier deep graph matching approaches~\cite{jiang2019glmnet,wang2019neural,yu2019learning,wang2019learning,fey2020deep} employed a linear assignment problem (LAP) solver%
\footnote{A polynomial subclass of graph matching without quadratic costs.}
 to obtain matchings in their pipeline, the most promising state-of-the-art method~\cite{rolinek2020deep} uses a fully featured graph matching solver.
Being called in a loop on each training iteration, this solver must provide high-quality solutions within a very restricted time budget, typically less than a second.
Modern state-of-the-art methods~\cite{swoboda2017study,GraphMatchingDDTorresaniEtAl,HungarianBP} satisfy this requirement only if applied to relatively small problems, with a few dozen feature points at most. Hence, scalability of deep graph matching critically depends on the existence of highly-efficient graph matching solvers.

In this work we address this problem by introducing a new graph matching technique, which notably improves the state-of-the-art in terms of speed and attained accuracy. In particular, it provides highly accurate solutions for problems with more than 500~features in less than a second.

\begin{figure}
  \centering
  \includegraphics[width=0.9\linewidth]{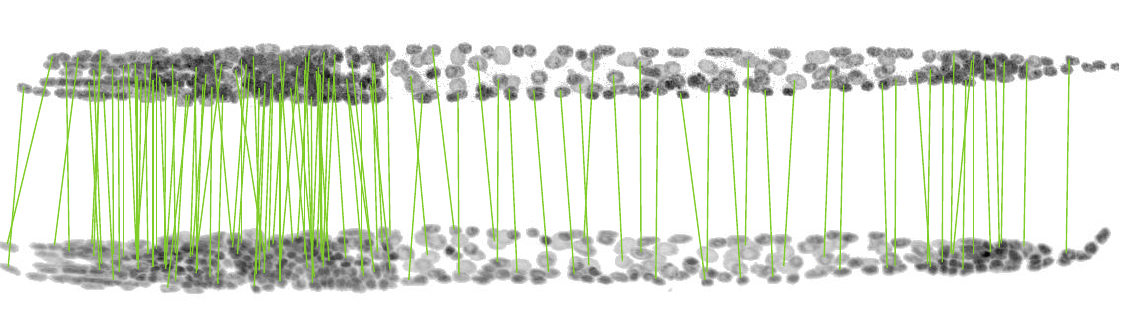}%

  \medskip
  \includegraphics[width=0.8\linewidth, viewport=12 14 275 131, clip]{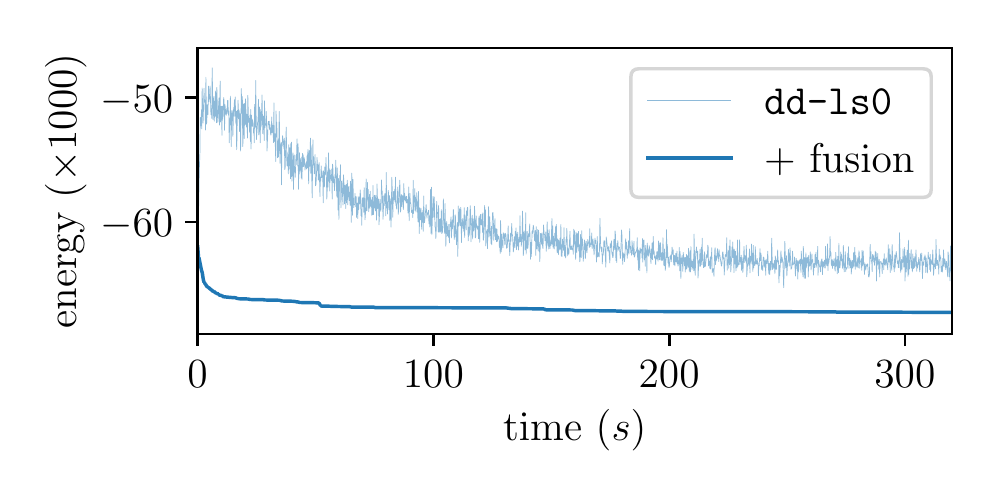}
  \caption{%
    \textbf{(Top)} Scalable graph matching is especially important for bio-imaging, where hundreds or even thousands of cells on different images must be matched to each other. An instance from the \Sdata{pairs} dataset (see Sec.~\ref{sec:experiments}), only each 5th matching is shown.
    \textbf{(bottom)} Convergence of the state-of-the-art method \Salg{dd-ls0}~\cite{GraphMatchingDDTorresaniEtAl}~(see Sec.~\ref{sec:experiments}) without and with fusion moves. Note that fusion moves attain much better energy in notably shorter time.
  }%
  \label{fig:teaser}
\end{figure}


\Paragraph{Related work.}

First formulated in 1957~\cite{QAPBeckman}, the graph matching problem plays a central role in combinatorial optimization. Due to its importance, nearly all possible optimization techniques were put to the test for it, see the surveys~\cite{BurkardQAP,AnalyticalSurveyQAP,cela2013quadratic} and references therein.

As usual for NP-hard problems, no single method can efficiently address all graph matching instances. Different applications require different methods, and we concentrate here on problem instances specific to computer vision. Traditionally, within this community primal heuristics%
\footnote{A common name for algorithms missing optimality guarantees.}%
~\cite{TabuSearchGraphMatching,GraduatedAssignmentGold,FactorizedGraphMatching,yu2020determinant,IntegerFixedPointGraphMatching,SpectralTechniqueAssignmentLeordeanu,cho2014finding,UmeyamaGraphMatching,EigenDecompositionGraphMatchingZhao,jiang2017graph} were used predominantly, since demand for low computation times usually dominates the need for optimality guarantees.
These also include methods that build upon spectral relaxations~\cite{SpectralTechniqueAssignmentLeordeanu,IntegerFixedPointGraphMatching,UmeyamaGraphMatching,EigenDecompositionGraphMatchingZhao}, or convex-to-concave path-following procedures~\cite{FactorizedGraphMatching,dym2017ds++,bernard2018ds}.
However, recent works~\cite{GraphMatchingDDTorresaniEtAl,HungarianBP,swoboda2017study} have shown that Lagrange duality-based methods attain significantly better accuracy, especially as problem size and complexity grow.
It is important to note that in operations research such Lagrange dual methods for graph matching are known at least since the 90s~\cite{adams1994improved,hahn1998lower,karisch1999dual}, and are widely used in branch-and-bound solvers. Although they address similar relaxations as~\cite{swoboda2017study,HungarianBP}, their iteration complexity is an order of magnitude higher than those of~\cite{swoboda2017study,HungarianBP}. This makes them prohibitively expensive for use in typical computer vision applications.

While branch-and-bound remains the main tool to obtain exact solutions, it has an exponential worst-case complexity and is often too expensive. Hence, dual methods like~\cite{GraphMatchingDDTorresaniEtAl,swoboda2017study} use simple primal heuristics
with low computational cost that can be called after each dual iteration.
Improving such heuristics to obtain high-quality primal solutions already after few dual iterations would allow to outperform the purely primal methods not only in accuracy but also in runtime.

\ParagraphRun{Fusion moves\normalfont,} as introduced by~\cite{lempitsky2010fusion}, is a primal heuristic proposed for maximum a posteriori inference in Markov random fields, known also as \emph{discrete labeling} or \emph{energy minimization problem}, see \eg~\cite{savchynskyy2019discrete}. For brevity we will refer to it as the \emph{MRF problem}.

In its most common setting the fusion moves method tries to improve a current approximate primal assignment by merging it with another assignment proposal. The merging constitutes a comparatively small two-label MRF problem, for which efficient exact and approximate techniques exist. As noted in~\cite{lempitsky2010fusion}, success of the method
significantly depends on the \emph{quality} and \emph{diversity} of proposals. A number of ways of generating generic proposals for MRF problems and (approximate) solvers for the corresponding auxiliary problem have been evaluated by~\cite{kappes2014map}. They also considered several instances of the graph matching problem treated as MRF. However, they found fusion moves
with their, typically infeasible, proposals to be inferior to other methods. A similar negative result was reported by~\cite{GraphMatchingDDTorresaniEtAl} with a simple but low quality local search-based proposal generator.

In operations research fusion moves is known since 1997 as \emph{optimized crossover} or \emph{recombination}, when it was proposed to address the independent set problem~\cite{aggarwal1997optimized}.
However, for the quadratic assignment problem it was reported as being inefficient, when used as a building block of a greedy genetic algorithm~\cite{ahuja2000greedy}.
This was attributed to the lack of diversity of the solution population resulting from this method.

\Paragraph{Contribution.}

We show how to use fusion moves to efficiently solve
graph matching problems,
and provide a theoretical rationale that efficient proposals
for fusion moves must be feasible, \ie satisfy the uniqueness constraints of
the graph matching problem. We ensure \emph{quality} of our proposals by
generating them based on reparametrized costs improved in the course
of dual optimization,
and enforce \emph{diversity} of proposals by making use of
either \emph{oscillating} dual updates, as in the dual subgradient method,
or our proposed efficient \emph{randomized} greedy algorithm.
Altogether, our method combines the accuracy of dual solvers with the speed
of dedicated primal heuristics.
We demonstrate the superior performance of our technique on multiple datasets.
Our code and datasets we used are available at \url{https://vislearn.github.io/libmpopt/iccv2021}.

\ParagraphRun{The appendix,} referred to as \S A1--\S A7 contains detailed proofs, dataset, experiment and algorithm descriptions.

\section{Preliminaries}

\Paragraph{Graph matching problem.}

Let $\SG=(\SV,\SE)$ be an undirected graph, where $\SV$ is the finite
set of \emph{nodes} and $\SE\subseteq\binom{\SV}{2}$ the set of
\emph{edges}.
For convenience we denote edges $\{u,v\}\in \SE$ simply by $uv$.
Let $\mathcal{L}$ be a finite set of \emph{labels}.
We associate with each node $u\in \SV$ a
subset of labels $\mathcal{L}_u\subseteq\mathcal{L}$, and a
\emph{unary cost function} $\theta_u\colon \SL^\#_u\to\mathbb{R}$,
where $\SL^\#_u:=\SL_u\cup\{\#\}$.
Here, $\#$ denotes a \emph{dummy label}
distinct from all labels in $\mathcal{L}$ to encode that no label is selected.
Likewise, for each edge $uv\in \SE$ let
$\theta_{uv}\colon \SL^\#_u\times\SL^\#_v \to\mathbb{R}$
be a \emph{pairwise cost function}.

Then the problem of finding an optimal assignment of labels to nodes,
known as \emph{graph matching} or \emph{quadratic assignment} problem,
can be stated as
\begin{align}
  &\min_{x\in X} \Big[E(x):= \sum_{u\in \SV}\theta_u(x_u) + \sum_{uv \in \SE}\theta_{uv}(x_u,x_v)\Big] \label{eq:qap}\\
  &\text{s.\,t.}\quad \forall\, u,v\in \SV, u\neq v:\; x_u \neq x_v \;\text{or}\; x_u = \# , \nonumber
\end{align}
where $X$ stands for the Cartesian product $\bigtimes _{u\in \SV} \SL^\#_u$.
The objective $E$ is referred to as \emph{energy}, and the
constraints in~\eqref{eq:qap} are known as \emph{uniqueness constraints}.
They allow each non-dummy label to be selected at most once.
The number of selected dummy labels is not limited.
Elements $x\in X$ are called \emph{assignments}. An assignment is
\emph{feasible} if it satisfies all uniqueness constraints.
So, essentially,~\eqref{eq:qap} corresponds to an
\emph{MRF problem with uniqueness constraints} for the labels.

Note that this formulation generalizes the classical quadratic
assignment problem, see \eg~\cite{BurkardQAP}, by allowing for \emph{incomplete}
assignments, \ie
not every label in $\mathcal{L}$ has to be
assigned to a node, and not necessarily every node is assigned a label
in $\mathcal{L}$. Instead, nodes can be assigned the dummy label.
Choosing a large constant
as unary cost for the dummy label in each node
enforces a \emph{complete} assignment.

Without pairwise costs $\theta_{uv}$ the quadratic assignment
problem~\eqref{eq:qap} reduces to the well-known
\emph{linear assignment problem} (LAP).
While the quadratic assignment problem is in general NP-hard, the LAP can be solved in polynomial time by \eg the Hungarian method.

\ParagraphRun{Fusion moves}\cite{lempitsky2010fusion} address the,
compared to~\eqref{eq:qap} unconstrained,
MRF problem $\min_{x\in X} E(x)$.
In the simplest, but most widely used scenario, on each iteration
of the algorithm
the currently best assignment $x'\in X$ is \emph{fused} with another
candidate assignment $x''\in X$ by solving the
\emph{auxiliary} minimization problem
\begin{equation}
  \min_{x\in X_{\text{aux}}} E(x)\,,\label{eq:mrf-fusion}
\end{equation}
where $X_{\text{aux}} := \{ x\in X \mid x_u \in \{ x'_u, x''_u\}, u\in \SV\}$.
Due to the considerably smaller size of the restricted
label space $X_{\text{aux}}$ the auxiliary problem~\eqref{eq:mrf-fusion}
can often be efficiently solved
approximately, or even exactly.
The solver only has to guarantee \emph{monotone improvement} of
the best assignment by assuring
\begin{equation}\label{eq:monotonicity-condition}
  E(x^*) \leq \min(E(x'), E(x''))
\end{equation}
for its output $x^*$, which is then further considered as the best
assignment, \ie in the next iteration $x':=x^*$.
Note that the monotonicity condition~\eqref{eq:monotonicity-condition}
automatically holds for any $x^*$
that is an exact solution of~\eqref{eq:mrf-fusion}.
For approximate methods the inequality~\eqref{eq:monotonicity-condition}
can be enforced by assigning $x^*$
to the proposal with lower energy if needed.
Each fusion operation is also referred to as a \emph{fusion move}.

We adopt this method
to the graph matching problem~\eqref{eq:qap} by extending
the auxiliary problem~\eqref{eq:mrf-fusion}:
\begin{align}
  &\min_{x\in X_{\text{aux}}} E(x) \label{eq:qap-fusion}\\
  &\text{s.\,t.}\quad \forall\, u,v\in \SV, u\neq v:\; x_u \neq x_v \;\text{or}\; x_u = \# \,. \nonumber
\end{align}
That is, compared to~\eqref{eq:mrf-fusion}, the uniqueness constraints
are taken into account during fusion,
which guarantees feasibility of the current best assignment.

There are two main questions that have to be answered
to apply fusion moves:
\textbf{(i)} How to generate proposals?
\textbf{(ii)} How to solve the auxiliary problem~\eqref{eq:qap-fusion}?
Starting with the second, we address these questions below.

\section{Solving the auxiliary problem}\label{sec:aux-subproblems}

\Paragraph{ILP formulation.}

The auxiliary problem~\eqref{eq:qap-fusion} can be formulated as an
\emph{integer linear program} (ILP) as follows.
For all $u\in\SV$
let $\hat \SL_u:=\{x'_u,x''_u\}$ be the restricted set of labels.%
\footnote{Without loss of generality we assume the non-trivial case $x'_u\neq x''_u$ for all $u\in\SV$.}
We introduce binary variables $\mu_{u,s}\in\{0,1\}$ for each
node $u\in\SV$ and label $s\in\hat\SL_u$,
and $\mu_{uv,st}\in\{0,1\}$ for each edge $uv\in\SE$ and each
label pair $(s,t)\in\hat\SL_u\times\hat\SL_v$.
Setting $\mu_{u,s}=\mu_{v,t}=\mu_{uv,st}=1$ corresponds to assigning coordinates $x_u=s$ and $x_v=t$ of the solution labeling $x$.
Together these variables form
a vector $\mu\in\{0,1\}^{N}$, where $N=2|\SV|+4|\SE|$.
Then the ILP
\begin{align}
  \min_{\mu\in\{0,1\}^N} &\sum_{\substack{u\in \SV\\s\in\hat \SL_u}} \mu_{u,s}\theta_u(s)
  	+\hspace{-10pt}\sum_{\substack{uv\in \SE\\(s,t)\in\hat\SL_u\times\hat\SL_v}}\hspace{-10pt} \mu_{uv,st}\theta_{uv}(s,t) \label{eq:fusion-ilp}\\
  \text{s.\,t. } &\forall\,u\in \SV\colon \mu_{u,x'_u}+\mu_{u,x''_u} = 1 \nonumber\\
  &\forall\,uv\in \SE,(s,t)\in\hat\SL_u\times\hat\SL_v\colon \nonumber\\
  &\quad \mu_{uv,st} \leq \mu_{u,s},\; \mu_{uv,st} \leq \mu_{v,t}\,, \nonumber \\
  &\quad \mu_{uv,st} \geq \mu_{u,s} + \mu_{v,t} - 1\,, \nonumber\\
  &\forall\,u,v\in \SV, u\neq v, s\in(\hat\SL_u\cap\hat\SL_v)\backslash\{\#\} \colon \label{eq:fusion-ilp-uniqueness-constraints} \\
  & \quad \mu_{u,s} + \mu_{v,s} \leq 1 \nonumber
\end{align}
is equivalent to~\eqref{eq:qap-fusion}.
In particular, the inequalities in~\eqref{eq:fusion-ilp-uniqueness-constraints}
enforce the uniqueness constraints.
Clearly, problem~\eqref{eq:fusion-ilp} without
the uniqueness
constraints~\eqref{eq:fusion-ilp-uniqueness-constraints} constitutes an
ILP representation of the MRF auxiliary problem~\eqref{eq:mrf-fusion}.

The ILP problem~\eqref{eq:fusion-ilp}-\eqref{eq:fusion-ilp-uniqueness-constraints}
can be addressed by off-the-shelf ILP solvers like Gurobi~\cite{gurobi}.
However, with growing problem size, such solvers become prohibitively slow, as they have exponential worst-case complexity.
Therefore, one has to resort to other exact or approximate optimization techniques, which we review now.

\Paragraph{Elimination of uniqueness constraints.}

The uniqueness constraints~\eqref{eq:fusion-ilp-uniqueness-constraints}
between nodes $u$ and $v$ can be eliminated by assigning a very large cost
$C_{\infty}$ to the pairwise cost function on the corresponding edge, \ie
\begin{equation}\label{eq:add-infty-pw-cost}
 \theta_{uv}(s,s):=C_{\infty},\ \forall s\in(\hat\SL_u\cap\hat\SL_v)\backslash\{\#\}\,.
\end{equation}
If $uv\notin\SE$, the edge $uv$ is added to $\SE$ together with pairwise
costs $\theta_{uv}(s,t): = C_{\infty}\cdot\llbracket s=t\neq\#\rrbracket$,
where $\llbracket A\rrbracket$ is equal to $1$ if $A$ holds,
and $0$ otherwise.

This way the graph matching auxiliary problem~\eqref{eq:qap-fusion} is reduced
to the MRF auxiliary problem~\eqref{eq:mrf-fusion}, on a, possibly different, graph.
This allows considering dedicated methods addressing the
MRF auxiliary problem~\eqref{eq:mrf-fusion}. Efficiency of these methods is very much dependent on the \emph{submodularity} of the pairwise costs $\theta_{uv}$. We review this property and the corresponding optimization methods below.

\Paragraph{Submodular case.}

In general, two-label MRF problems like~\eqref{eq:mrf-fusion} are NP-hard~\cite{BorosHammer02}.
However, they become efficiently solvable, if for all $u\in\SV$
there exists a bijective mapping
$\delta_u\colon\{0,1\}\to \hat\SL_u$ called \emph{ordering},
such that all pairwise
costs $\theta_{uv}$, $uv\in\SE$, in problem~\eqref{eq:mrf-fusion}
are \emph{submodular},  \ie
\begin{equation}\label{eq:submodular-def}
 \theta_{uv}(0,0) + \theta_{uv}(1,1) \le \theta_{uv}(0,1) + \theta_{uv}(1,0)\,,
\end{equation}
where we abbreviate $\theta_{uv}(\delta_u(0),\delta_v(1))$
by $\theta_{uv}(0,1)$.
It is known that in this case the natural linear program (LP) relaxation
of~\eqref{eq:fusion-ilp} is tight, and, moreover,
reducible to the efficiently solvable min-cut/max-flow problem~\cite{kolmogorov2007minimizing}.%
\footnote{The orderings $\delta_u$ can also be found
explicitly~\cite{schlesinger2007exact},
allowing for a more efficient
min-cut/max-flow reduction~\cite{kolmogorov2004energy}.}

\Paragraph{Non-submodular case.}

The pairwise costs not fulfilling the submodularity
condition~\eqref{eq:submodular-def} for a given mapping $\delta_u$
are called \emph{supermodular}.
Inequality~\eqref{eq:submodular-def} implies that swapping
the ``labels'' 1 and 0 turns submodular pairwise costs
into supermodular ones and vice versa. However, since a swap in one
node changes sub-/supermodularity of all incident
pairwise costs, we cannot always turn all supermodular
pairwise costs into submodular ones.
This is already impossible if the graph contains a triangular subgraph
with all pairwise costs being supermodular.

In these cases the mentioned LP relaxation is in general not tight.
However, it has the important \emph{persistency} property,
\ie all integer coordinates of a relaxed solution belong to
an optimal integer solution~\cite{kolmogorov2007minimizing}. This
allows for building efficient approximate methods
for~\eqref{eq:mrf-fusion} applicable also to the non-submodular case~\cite{rother07-cvpr}.
These methods are known
in the literature as \emph{quadratic pseudo-boolean
optimization} (QPBO) or \emph{roof duality}.
As an alternative, \emph{trust region-based} approximate optimization
algorithms for~\eqref{eq:mrf-fusion} have been suggested
by~\cite{gorelickboykov-pami17}. They are based on an iterative
approximation of the problem by submodular problems. To this end
the supermodular pairwise costs are approximated with unary costs.
Similar to the QPBO techniques, performance of trust-region
methods drops as the number of supermodular pairwise costs increases.
Contrary to the QPBO techniques they require an explicit
ordering of the label sets.

\section{Feasibility of proposals}\label{sec:feasibility}

Before we address the generation of proposals,
we theoretically substantiate the main
property of fusion move proposals
for graph matching problems: \emph{feasibility}.
In other words,
proposals should satisfy the uniqueness constraints
to allow the method to perform well.

\Paragraph{Size of the search space.}

In a nutshell, fusion moves is a local search method,
with the search space defined by
proposals. Performance of such methods critically depends
on the size of the search space.
Assuming that a better, or even the best, solution within
this space can be found efficiently,
this search space should be as large as possible to allow
for better approximations.
The following proposition sets the bounds on the size of
the search space:
\begin{proposition}\label{prop:search-space-size}
Let $x'$ be a feasible, and $x''$ a possibly infeasible assignment
for the graph matching problem~\eqref{eq:qap}.
Let $m$ be the number of dummy, and $n$ the
number of
different
non-dummy labels in $x''$.
Then the auxiliary problem~\eqref{eq:qap-fusion} has at most
$2^{m}\bigl(\frac{|\SV|}{n} + 1\bigr)^n$
feasible solutions.
\end{proposition}
In other words, for a fixed number of dummy labels in $x''$ the size of the search space exponentially
increases with the number of different labels in $x''$.
Feasible assignments maximize this number, see~\Ssup{supp:proof-prop-1} for a proof of Prop.~\ref{prop:search-space-size}.

The need for feasible assignments
distinguishes graph matching from the MRF problem,
where the space of possible solutions always grows as $2^n$,
where $n$ is the \emph{total} number of nodes where the proposals differ.
Therefore, a popular and quite efficient way to generate
MRF proposals known as $\alpha$-expansion~\cite{boykov2001fast},
where $x''_u=\alpha$ for all $u\in\SV$,
is completely ineffective for graph matching:
According to Proposition~\ref{prop:search-space-size} the
search space reduces to $|\SV|+1$ solutions.
Another popular method~\cite{lempitsky2010fusion,kappes2014map}
suggests constructing proposals from locally best labels returned
by, \eg, loopy belief propagation.
As empirically observed by~\cite{kappes2014map}, for the graph matching problem such proposals typically do not satisfy the uniqueness constraints and, therefore, lead to a non-competitive performance of fusion moves.

\Paragraph{Efficiency of approximate solvers.}

As noted in Section~\ref{sec:aux-subproblems}, performance of approximate
solvers for the auxiliary problem~\eqref{eq:mrf-fusion} drops with an increasing
proportion of supermodular pairwise costs. Since the uniqueness constraints
for the auxiliary problem~\eqref{eq:qap-fusion} are translated
into large pairwise costs, it is important to find an ordering where these large
costs do not lead to a violation of the submodularity constraint~\eqref{eq:submodular-def}.

Let the proposal $x''$ be infeasible,
\ie there exist $u,v\in\SV$, $u\neq v$, with $x''_u = x''_v \neq \#$.
Consider now the ordering where labels $x''_u$
are mapped to $0$ for all $u\in\SV$, and all $x'_u$ to $1$.
Then, according to~\eqref{eq:add-infty-pw-cost}, $\theta_{uv}(0,0)=C_{\infty}$,
which would lead to a supermodular pairwise cost $\theta_{uv}$.
Should there be multiple
nodes with equal labels, \ie $x''_u = x''_v= x''_w$, this would lead
to a fully connected subgraph with supermodular costs.
As discussed in Section~\ref{sec:aux-subproblems},
these costs cannot all be turned into submodular ones by swapping
the labels $0$ and $1$.
As a consequence, this leads to a deterioration in performance of approximate methods
for the graph matching auxiliary problem~\eqref{eq:qap-fusion}.
This case can be avoided by requiring $x''$ to be feasible.

Conversely, consider the practically inevitable case of equal labels in
\emph{different} proposals, \ie $x'_u=x''_v$ for some $u,v\in\SV$, $u\neq v$.
According to~\eqref{eq:add-infty-pw-cost}, with the same initial ordering
as above, $\theta_{uv}(1,0):=C_{\infty}$.
This, however, renders the corresponding pairwise cost submodular,
\cf~\eqref{eq:submodular-def},
which simplifies optimization.

\emph{\textbf{To summarize}, the feasibility of proposals increases the
search space for each fusion, while at the same time allowing for efficient
approximate solvers for the auxiliary problem.}

\section{Proposal generation}

As mentioned in Section~\ref{sec:intro} fusion moves
work best if the proposals
are of \emph{high quality} and \emph{diverse}.
Essentially, \emph{high quality} means low corresponding
energy $E$, and \emph{diversity} can be quantified by counting the number of nodes where two proposals differ.

\Paragraph{How to obtain high quality proposals?}

The natural idea to get high quality proposals is to employ some iterative optimization process which outputs solutions on each iteration.
As discussed in Section~\ref{sec:feasibility}, in the case of graph matching, these proposals should be feasible.
Dual methods equipped with efficiently computable primal heuristics are therefore natural candidates for proposal generators.
In Section~\ref{sec:dual-methods} we briefly describe two types of such methods, block-coordinate ascent- and subgradient-based ones.

\Paragraph{How to obtain diverse proposals?}

Diversity of proposals based on
dual optimization can either be induced by noisy dual
updates, or must be an intrinsic property of the primal heuristic.
We utilize both strategies.

The subgradient method is a representative of the first type. Due to
non-optimal step-sizes and update directions it usually demonstrates a
``zig-zag'' progress of the dual value that induces similar behavior in the
assignment scores obtained by a primal heuristic.

In contrast, block-coordinate ascent methods are based on optimal updates and
guarantee a monotone improvement of the dual value. As a consequence, the
corresponding assignments computed by deterministic primal heuristics often
lack diversity.

To address this issue, we suggest to use the \emph{randomized greedy heuristic}
described in Section~\ref{sec:greedy-generation} as a generic method to
generate diverse proposals. It combines \emph{diversity} due to randomization
of the node selection order with \emph{high quality} due to taking locally
optimal labels. Another important advantage of this method is that it can
profit from the dual optimization, and provides qualitatively better proposals
as the dual optimization progresses. In particular, it returns a globally
optimal assignment if the latter is unique and the dual bound is tight. We describe its use in
connection with a dual BCA solver in Section~\ref{sec:dual-methods}.

\subsection{Randomized greedy heuristic}\label{sec:greedy-generation}

Let $\SN(u):=\{v\in\SV\mid uv\in\SE\}$ be the neighborhood of $u$,
and $\SN(\SV'):=\left(\bigcup_{u\in\SV'}\SN(u)\right)\backslash\SV'$
the neighborhood of $\SV'\subseteq\SV$. Note, $\SN(\emptyset)=\emptyset$.
The \emph{randomized greedy heuristic} is defined
by Algorithm~\ref{alg:greedy-heuristic}.
In each step an unassigned node is randomly selected
from the neighborhood of the assigned nodes,
and a label is assigned to it such that the uniqueness constraints
are satisfied and the sum of its unary cost
and all pairwise costs on edges connecting it with assigned nodes minimized.

\begin{algorithm}[t]
  \addtolength{\hsize}{1.5em}%
  \DontPrintSemicolon
  \KwIn{graph $\SG = (\SV, \SE)$, labels $\SL$ and costs $\theta$}
  initialize $\;\SV':=\emptyset\;$ and $\;\SL':=\emptyset$\;
  \While{$\SV'\neq \SV$}{
   randomly select $u\!\in\!\SN(\SV')$ or $u\!\in\!\SV\backslash\kern-.5pt\SV'$ if $\SN(\SV'\kern -1pt)\kern-.8pt=\kern-.8pt\emptyset$\;
   set $\;x_u:=\!\!\argmin\limits_{s\in\SL_u\backslash \SL'\cup\{\#\}}\bigg[\theta_u(s)+\sum\limits_{v\in\SN\kern-.5pt(u)\cap\SV'}\hspace{-5pt}\theta_{uv}(s,x_{v})\bigg]$\;
   update $\;\SV':=\SV'\cup\{u\}\;$ and $\;\SL':=\SL'\cup\{x_u\}$\;
  }
  \KwOut{feasible assignment $x=(x_u)_{u\in \SV}$}
  \caption{Randomized greedy heuristic.}
  \label{alg:greedy-heuristic}
\end{algorithm}

\subsection{Dual solvers}\label{sec:dual-methods}

\Paragraph{Dual problem.}

The graph matching problem~\eqref{eq:qap} can
be represented in an ILP form similar
to that of the auxiliary problem~\eqref{eq:mrf-fusion}.
We define binary variables $\mu_{u,s}$ and $\mu_{uv,st}$
analogously.
The number of such variables is
$M:=\sum_{u\in\SV}|\SL^\#_u|+\sum_{uv\in \SE}|\SL^\#_u||\SL^\#_v|$.
By denoting the set of nodes containing a particular non-dummy
label $s$ as $\SV(s):=\{u\in\SV\mid s\in\SL_u\}$,
problem~\eqref{eq:qap} can be written as:
\begin{align}
  \min_{\mu\in\{0,1\}^M} &\sum_{\substack{u\in \SV\\s\in\SL^\#_u}} \mu_{u,s}\theta_u(s)
  	+\hspace{-15pt}\sum_{\substack{uv\in \SE\\(s,t)\in\SL^\#_u\times\SL^\#_v}} \hspace{-12pt}\mu_{uv,st}\theta_{uv}(s,t) \label{eq:qap-as-ILP-objective}\\
  \text{s.\,t. } &\forall\,uv\in \SE,\, t\in\SL^\#_v \colon \sum_{s\in\SL^\#_u}\mu_{uv,st} = \mu_{v,t} \label{eq:qap-as-C1}\\
  &\forall\,u\in \SV\colon \sum\nolimits_{s\in\SL^\#_u}\mu_{u,s} = 1 \label{eq:qap-as-C2}\\
  &\forall\,s\in \SL\colon \sum\nolimits_{u\in \SV(s)}\mu_{u,s} \le 1 \label{eq:qap-as-C3}
\end{align}
By introducing
$\xi^\lambda_u(s):=\frac{\theta_u(s)}{2}+\lambda_{u,s}$
and $\hat\xi^\lambda_u(s):=\frac{\theta_u(s)}{2}-\lambda_{u,s}$
for $u\in\SV$, $s\in\SL^\#_u$,
and arbitrary $\lambda_{u,s}\in\SR$,
we can equivalently rewrite the objective in~\eqref{eq:qap-as-ILP-objective} as
a sum of objectives of MRF and LAP subproblems denoted
as $E^{\text{MRF}}$ and $E^{\text{LAP}}$, respectively:
\begin{equation}
  \underbrace{\sum_{\substack{u\in \SV\\s\in\SL^\#_u}}\hspace{-2pt} \mu_{u,s}\xi^\lambda_u(s)
	+\hspace{-17pt}\sum_{\substack{uv\in \SE \\ (s,t)\in\SL^\#_u\times\SL^\#_v}} \hspace{-17pt}\mu_{uv,st}\theta_{uv}(s,t)}_{=:E^{\text{MRF}}(\mu,\lambda)}
 	+\hspace{-2pt} \underbrace{\sum_{\substack{u\in \SV\\ s\in\SL^\#_u}}\hspace{-2pt}\mu_{u,s}\hat\xi^\lambda_u(s)}_{=:E^{\text{LAP}}(\mu,\lambda)}.\nonumber
\end{equation}

Let $\Lambda$ be the set of all binary vectors $\mu\in\{0,1\}^M$
satisfying constraints~\eqref{eq:qap-as-C1}-\eqref{eq:qap-as-C2},
and $B$ the set of those satisfying~\eqref{eq:qap-as-C2}-\eqref{eq:qap-as-C3}.
Then the sum
\begin{equation}
 \min_{\mu\in\Lambda}E^{\text{MRF}}(\mu,\lambda)+\min_{\hat\mu\in B}E^{\text{LAP}}(\hat\mu,\lambda)
\end{equation}
of independent minimizations of the MRF and LAP
subproblems constitutes a lower bound
for~\eqref{eq:qap-as-ILP-objective}-\eqref{eq:qap-as-C3}.

While the second term can be minimized efficiently, \eg by the
Hungarian method, the first term is an NP-hard problem
by itself. By dualizing the constraints~\eqref{eq:qap-as-C1} one
obtains its Lagrange dual lower bound, \cf~\cite[Ch.6]{savchynskyy2019discrete},
\begin{equation}
 D(\phi,\lambda):=\sum_{u\in\SV}\min_{s\in\SL^\#_u}\xi^{\phi,\lambda}_{u}(s)
 +\!\!\! \sum_{uv\in \SE}\min_{(s,t)\in\SL^\#_u\times\SL^\#_v}\!\!\theta^{\phi}_{uv}(s,t)\,, \nonumber
\end{equation}
where
\begin{align}
 \xi^{\phi,\lambda}_u(s)   &  :=\xi^{\lambda}_u(s)\;-\hspace{-4pt}\sum_{v\in\SN(u)}\phi_{u,v}(s)\,, \\
 \theta^{\phi}_{uv}(s,t) &  :=\theta_{uv}(s,t)+\phi_{u,v}(s)+\phi_{v,u}(t)\,, \nonumber
\end{align}
are commonly referred to as \emph{reparametrized costs}.

All in all, the dual problem of~\eqref{eq:qap-as-ILP-objective}-\eqref{eq:qap-as-C3}
consists in the lower bound maximization
\begin{equation}\label{eq:qap-dual-problem}
 \max_{\phi,\lambda}\left[D(\phi,\lambda)+\min_{\hat\mu\in B}E^{\text{LAP}}(\hat\mu,\lambda)\right]\,.
\end{equation}

\Paragraph{Dual block-coordinate ascent, \Ssup{supp:dual-algorithm}.}\label{sec:bca-solvers}

Based on the ideas of~\cite{swoboda2017dual,HungarianBP},
and the recent progress in development
of dual solvers for MRFs~\cite{tourani2018mplp,tourani2020taxonomy},
we implemented a \emph{block-coordinate ascent} (BCA) solver
that also allows to output assignment proposals.

Our solver monotonically improves the
dual bound~\eqref{eq:qap-dual-problem} by interleaving
maximization \wrt $\phi$ and $\lambda$.
Each step on $\phi$ consists of maximizing the bound \wrt
the block of
variables $(\phi_{u,v}(s),\phi_{v,u}(t))$, $(s,t)\in\SL^\#_u\times\SL^\#_v$
associated with one edge $uv\in\SE$.
A sequence of these steps addressing all edges is equivalent
to one iteration of the MPLP++ algorithm of~\cite{tourani2018mplp}, that
notably outperforms the MPLP algorithm~\cite{MPLP} used by~\cite{HungarianBP}.
Each step on $\lambda$ consists of maximizing the
dual objective \wrt blocks $(\lambda_{u,s})$, $u\in\SV(s)$,
for each label $s\in\SL$ similar to how it was done by~\cite{swoboda2017dual}.

\ParagraphRun{For primal estimates} we either use the exact solution of
the LAP term $\min_{\hat\mu\in B}E^{\text{LAP}}(\hat\mu,\lambda)$
in~\eqref{eq:qap-dual-problem}
for the current value of $\lambda$,
or run our randomized greedy Algorithm~\ref{alg:greedy-heuristic}
on the graph matching problem with unary and pairwise
costs $\xi^{\phi,\lambda}_u$ and $\theta^{\phi}_{uv}$,
for current values of $\phi$ and $\lambda$.
For the LAP heuristic we use Gurobi~\cite{gurobi} as a solver.
While using \eg the Hungarian method for solving LAPs would be faster,
we found that the greedy heuristics combined with fusion moves consistently
outperforms its LAP counterpart in all our experiments in terms of run time and
quality.

\Paragraph{Subgradient method.}

We use the code of~\cite{GraphMatchingDDTorresaniEtAl} as a representative of the
dual subgradient methods. In its basic version denoted as \emph{dd-ls0},
which stands for \emph{dual decomposition with no local subproblems},
it optimizes the same bound as~\eqref{eq:qap-dual-problem} with
the difference
that instead of the dual MRF-term $D(\phi,\lambda)$ it
uses an equivalent \emph{tree-decomposition}
of the problem $\min_{\mu\in\Lambda}E^{\text{MRF}}(\mu,\lambda)$,
see \eg~\cite[Ch.9]{savchynskyy2019discrete}
for details. As the primal bound it uses a solution of the
LAP problem $\min_{\hat\mu\in B}E^{\text{LAP}}(\hat\mu,\lambda)$
for the current value of $\lambda$.

Two other versions of the solver we use in our experiments,
denoted as \emph{dd-ls3} and
\emph{dd-ls4}, additionally consider \emph{local subproblems}
on subgraphs of $\SG$
consisting of $3$ or $4$ neighboring nodes of the
original graph, respectively.
These modifications require more time per iteration, but optimize
tighter bounds than~\eqref{eq:qap-dual-problem}. Additionally,
these variants estimate primal solutions based on solutions
of the local subproblems.
We refer to~\cite{GraphMatchingDDTorresaniEtAl} for further details.

\ParagraphRun{Dual BCA algorithm complexity} per iteration is $O(\sum_{uv\in\SE}|\SL^\#_u||\SL^\#_v|)$, \eg linear in the size of the problem.
For a fully connected graph with $\SL^\#_u=\SL\cup\{\#\}$, $\forall u\in\SV$, this turns to $O(|\SV|^4)$. This is one degree of power less than the iteration complexity $O(|\SV|^5)$ of the dual ascent algorithms~\cite{adams1994improved,hahn1998lower,karisch1999dual} known in operations research.

\section{Experiments and analysis}\label{sec:experiments}

\Paragraph{Experimental setup.}

We evaluate the performance of all tested algorithms by measuring their total run
time and the obtained solution quality. Our experiments were run on a
compute cluster equipped with AMD EPYC 7702 2.0\,GHz processors and 512\,GB main memory.
For a fair comparison we used efficient implementations of all
discussed algorithms, and report the minimal runtime of 5 independently scheduled trials.

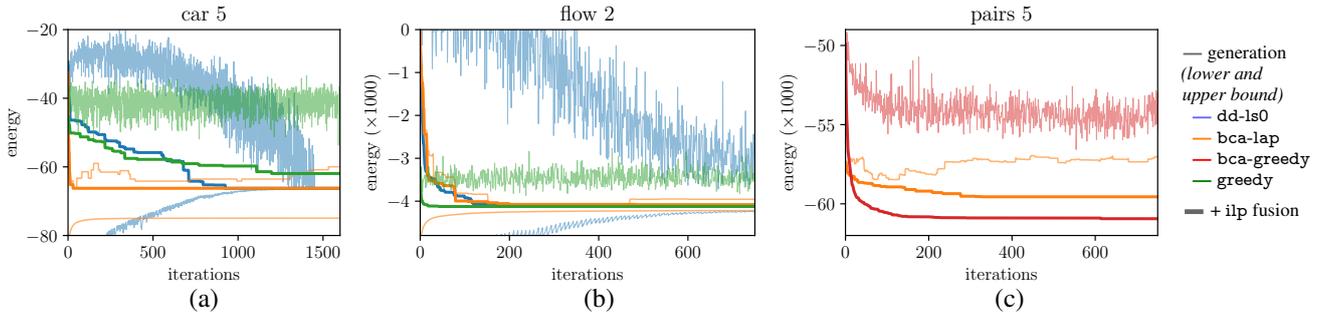
\begin{figure*}[t]
  \vspace*{-8pt}%
  \includegraphics[height=10.5em, viewport=16 16 239 203, clip]{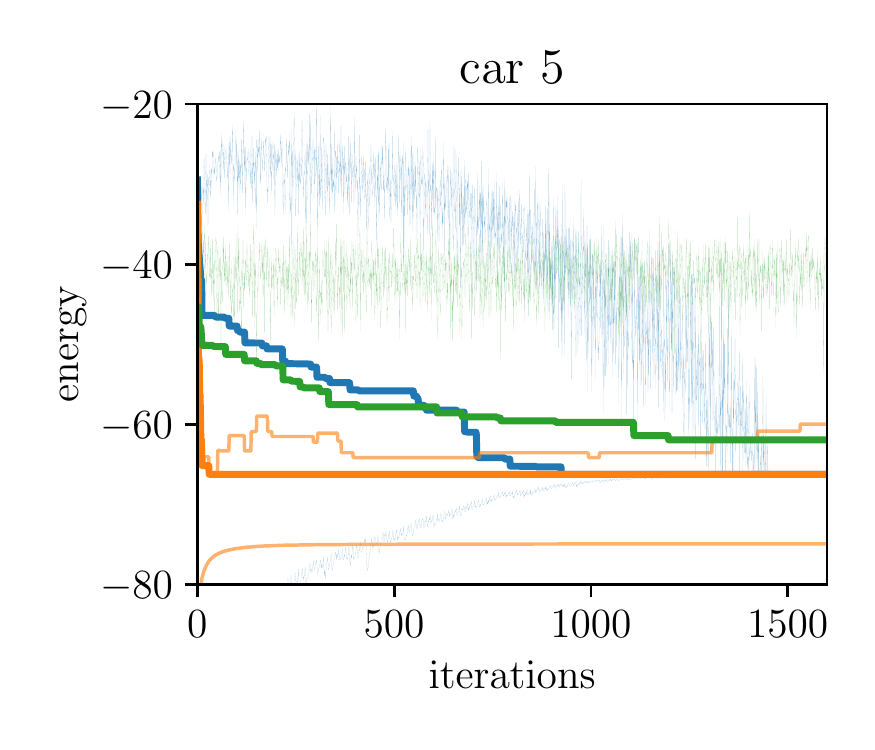}\hfill%
  \includegraphics[height=10.5em, viewport=14 16 275 203, clip]{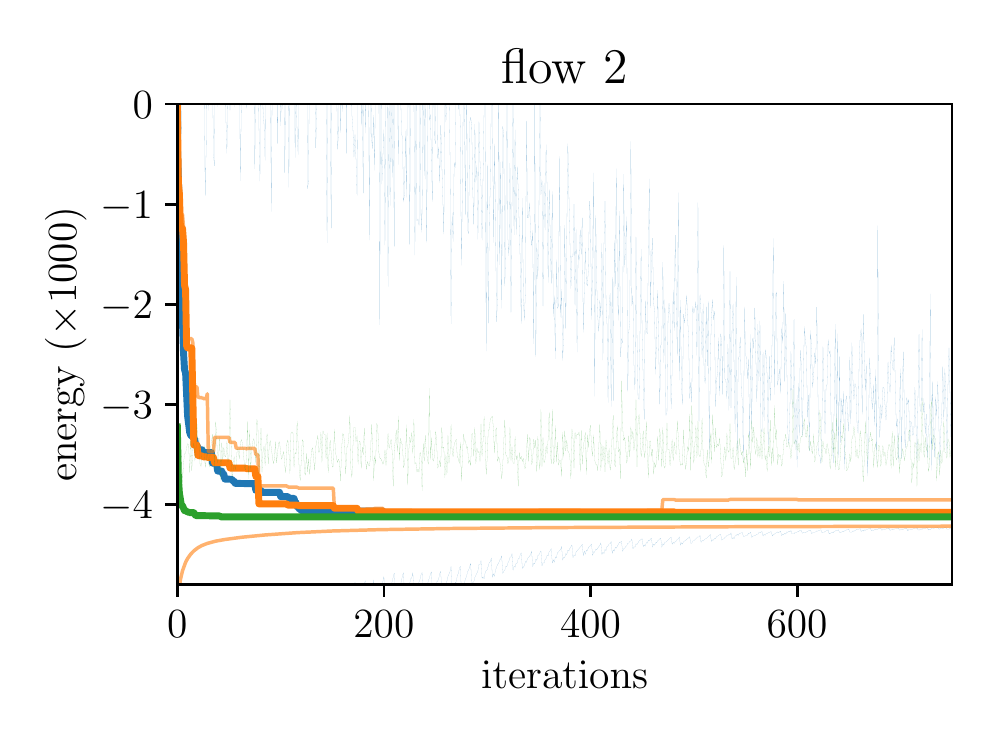}\hfill%
  \includegraphics[height=10.5em, viewport=22 16 275 203, clip]{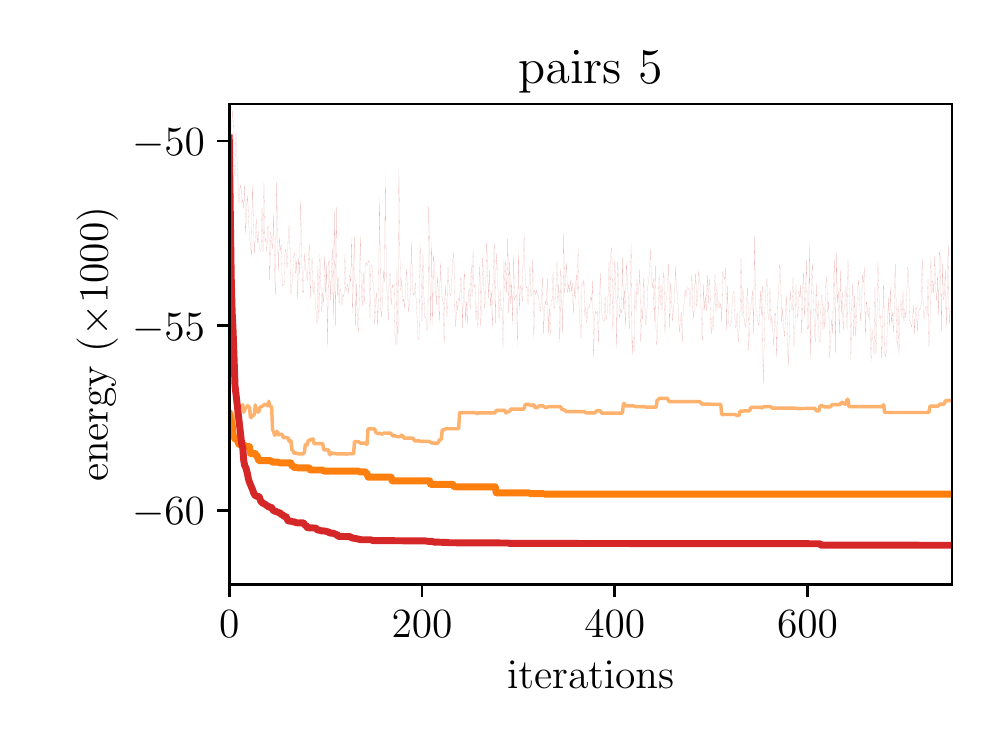}\hfill%
  \raisebox{19mm}{%
    \begin{minipage}{5.5em}
      \scriptsize
      \tikz[baseline=-0.5ex]{\draw [line width=0.7pt, black!60] (0,0) -- (1em,0);}~generation \\
        \textit{(lower and} \\
        \textit{upper bound)} \\
      \tikz[baseline=-0.5ex]{\draw [line width=0.7pt, blue!60] (0,0) (0.5em,0) -- (1.5em,0);}~\Salg{dd-ls0} \\
      \tikz[baseline=-0.5ex]{\draw [line width=0.7pt, orange] (0,0) (0.5em,0) -- (1.5em,0);}~\Salg{bca-lap} \\
      \tikz[baseline=-0.5ex]{\draw [line width=0.7pt, red] (0,0) (0.5em,0) -- (1.5em,0);}~\Salg{bca-greedy} \\
      \tikz[baseline=-0.5ex]{\draw [line width=0.7pt, green!50!black] (0,0) (0.5em,0) -- (1.5em,0);}~\Salg{greedy}
          \vspace*{.5em}

      \tikz[baseline=-0.5ex]{\draw [line width=1.8pt, black!60] (0,0) -- (1em,0);}~+ \Salg{ilp} fusion
    \end{minipage}}

    \vspace{-3pt}%
    \hspace{24mm}(a)%
    \hspace{48mm}(b)%
    \hspace{50mm}(c)

  \smallskip%
  \caption{
    \textbf{(a-b) Influence of fusion.} The plots show the energy of assignments generated by \Salg{dd-ls0} (blue), \Salg{bca-lap} (orange) and \Salg{greedy} (green) algorithms together with the dual bound where applicable. The thick line in matching color shows for each algorithm the achieved energy when using an \Salg{ilp} solver for fusion on top. Notably, fusion achieves very good quality with much less iterations. For some datasets, even greedily generated proposals suffice to obtain (almost) optimal solutions when fused.
    \textbf{(c) LAP vs. greedy heuristic.} The plot shows the quality of proposals generated by \Salg{bca-lap} (orange) and \Salg{bca-greedy} (red) for an exemplary instance of \Sdata{pairs}. The fused solutions on top of these generators are shown in the same color as a thick line. Fusion moves applied to \Salg{bca-greedy} yield significantly better results than when applied to \Salg{bca-lap}, even though the \Salg{bca-greedy} proposals are visibly worse than those of \Salg{bca-lap}.
  }
  \label{fig:general-performance}
\end{figure*}


\Paragraph{Datasets, \Ssup{supp:detailed-datasets}.}

Our experimental evaluation was conducted on 8~datasets with overall 316~problem instances from computer vision and
bio-imaging described in detail below. To demonstrate the scalability of our approach, along with the standard small-scale datasets for computer vision \Sdata{hotel}, \Sdata{house}, \Sdata{car}, \Sdata{motor} and \Sdata{opengm} with $|\SV| \le 52$, we consider the middle-sized ones \Sdata{flow}, $|\SV| \le 126$, and the large-scaled \Sdata{worms} and \Sdata{pairs} datasets with $|\SV| \le 565$. The latter are, to our knowledge, the \emph{largest graph matching problem instances ever considered in computer vision}.

\emph{Wide baseline matching} (\Sdata{hotel}, \Sdata{house})
is based on a series of images of the same object
from different view angles. We use the same image pairs, landmarks, and
cost structure as in~\cite{GraphMatchingDDTorresaniEtAl} based on
the work by~\cite{CaetanoMCLS09}.

\emph{Keypoint matching} (\Sdata{car}, \Sdata{motor})
contains \emph{car} and \emph{motor}bike instances from the
PASCAL VOC 2007 Challenge~\cite{VocPascal}
with the features and costs from~\cite{UnsupervisedLearningForGraphMatching}.
We preprocessed the models by removing edges
with zero cost, thereby reducing graph density substantially.

\emph{Large displacement flow} (\Sdata{flow})
was introduced by~\cite{GraphFlow} for key point matching
on scenes with large displacement flow.
We use the keypoints and costs from~\cite{swoboda2017study}.

\emph{OpenGM matching} (\Sdata{opengm})
is a set of non-rigid point matching
problems by~\cite{KomodakisP08}, now part of the
\emph{OpenGM} Benchmark~\cite{OpenGMBenchmark}.

\emph{Worm atlas matching} (\Sdata{worms})
has the goal to annotate nuclei of \emph{C.~elegans}, a famous model
organism used in developmental biology,
by assigning nuclei names from a precomputed atlas of
the organism.
We use the models from~\cite{kainmueller2014active,kainmueller2017graph}.

\emph{Worm-to-worm matching} (\Sdata{pairs}), see Fig.~\ref{fig:teaser} for illustration,
in contrast to the \Sdata{worms} dataset, directly matches the cell nuclei of
individual \emph{C.~elegans} worms to each other.
This alleviates the need to precompute an atlas based on
manual annotations.
Unary and pairwise costs of the respective graph matching problems
are derived by averaging the nucleus-(pair-)specific covariance
matrices captured by the atlas over all nuclei.
This coarsens the model to a level achievable without manual annotations.
For our experiments we randomly chose 16~instances out of
the $30\cdot 29 = 870$ non-trivial pairs of worms based on the same
data as \Sdata{worms}.

\Paragraph{Algorithms.}

As proposal generators we evaluate the three
dual subgradient-based algorithms
\Salg{dd-ls0}, \Salg{dd-ls3}, \Salg{dd-ls4} and our
BCA solver \Salg{bca} described in Section~\ref{sec:dual-methods}.
The latter is used with either the primal heuristics based on the
LAP solution or on the greedy Algorithm~\ref{alg:greedy-heuristic},
denoted as \Salg{bca-lap} and \Salg{bca-greedy}, respectively.
We also use the \Salg{greedy} Algorithm~\ref{alg:greedy-heuristic}
as a standalone baseline.

As proposal fusing methods we evaluate \emph{Gurobi}~\cite{gurobi} as an exact
ILP solver for the auxiliary problem~\eqref{eq:fusion-ilp} denoted
as \Salg{ilp},
the \emph{trust region}-based method of~\cite{gorelickboykov-pami17} denoted
as \Salg{lsatr}, as well as
different \emph{QPBO} variants denoted as \Salg{qpbo-XX}.
For the description of these variants see~\cite{rother07-cvpr}, and
the corresponding source code.

\begin{table*}
  \caption{
    \textbf{Summary of fusion moves performance.}
    Averaged energy of the best proposals for each dataset~(\textit{best gen.}), and time needed on average to generate it~($t_{\text{gen}}$) are shown.
    Furthermore, it shows for the \Salg{qpbo-i} fusion algorithm how long it took on average to beat the \Salg{dd-ls0} or \Salg{bca-greedy} proposals when fusing~($t_{\text{beat}}$), the average energy of the best proposal generated by fusion~(\textit{best fused}), and the average time after which this was obtained~($t_\text{fuse}$).
    All times are in seconds.
    The small numbers in front of the energies represent the number of instances solved to optimality by the respective method.
    Methods with fusion attain better energy values and are faster on average.}
  \label{tab:final-average-table}

  \centering
  \footnotesize
  \smallskip
  \begin{tabular}[l]{l*{14}{r}}
    \toprule
    \multirow{2}{*}[-0.5em]{\begin{minipage}{1.8cm}\centering dataset\\\tiny (number of instances)\end{minipage}} & \multicolumn{3}{c}{\Salg{dd-ls0}}
    & \multicolumn{4}{c}{+ \Salg{qpbo-i}}
    & \multicolumn{3}{c}{\Salg{bca-greedy}}
    & \multicolumn{4}{c}{+ \Salg{qpbo-i}} \\
    \cmidrule(lr){2-4} \cmidrule(lr){5-8} \cmidrule(lr){9-11} \cmidrule(lr){12-15}
    & & \textit{best gen.} & $t_{\text{gen.}}$
    & $t_{\text{beat}}$ & & \textit{best fused} & $t_{\text{fuse}}$
    & & \textit{best gen.} & $t_{\text{gen.}}$
    & $t_{\text{beat}}$ & & \textit{best fused} & $t_{\text{fuse}}$\\
    \midrule
    \Sdata{hotel} {\tiny(105)}\hspace*{-0.2em}
 & \Sopt{105} &   \textbf{-4293.00} &     0.07
 &     0.04
 & \Sopt{105} &   \textbf{-4293.00} &     0.04
 & \Sopt{103} &   -4291.21 &     0.81
 &     0.03
 & \Sopt{105} &   \textbf{-4293.00} &     \textbf{0.03}
\\
\Sdata{house} {\tiny(105)}\hspace*{-0.2em}
 & \Sopt{105} &   \textbf{-3778.13} &     \textbf{0.02}
 &     0.02
 & \Sopt{105} &   \textbf{-3778.13} &     \textbf{0.02}
 & \Sopt{105} &   \textbf{-3778.13} &     0.94
 &     0.09
 & \Sopt{105} &   \textbf{-3778.13} &     0.09
\\
\Sdata{car} {\tiny(30)}\hspace*{-0.2em}
 & \Sopt{29} &     -69.34 &     \textbf{0.17}
 &     0.17
 & \Sopt{29} &     \textbf{-69.37} &     \textbf{0.17}
 & \Sopt{27} &     -69.19 &     1.28
 &     0.12
 & \Sopt{29} &     \textbf{-69.37} &     \textbf{0.17}
\\
\Sdata{motor} {\tiny(20)}\hspace*{-0.2em}
 & \Sopt{20} &     \textbf{-62.95} &     0.06
 &     0.03
 & \Sopt{20} &     \textbf{-62.95} &     0.03
 & \Sopt{19} &     -62.93 &     0.80
 &     0.02
 & \Sopt{20} &     \textbf{-62.95} &     \textbf{0.02}
\\
\Sdata{flow} {\tiny(6)}\hspace*{-0.2em}
 & \Sopt{3} &   -2818.83 &     2.79
 &     1.12
 & \Sopt{5} &   -2835.84 &     1.91
 & \Sopt{4} &   -2837.82 &     9.65
 &     0.65
 & \Sopt{5} &   \textbf{-2840.00} &     \textbf{0.66}
\\
\Sdata{opengm} {\tiny(4)}\hspace*{-0.2em}
 & \Sopt{3} &      31.42 &     0.94
 &     0.77
 & \Sopt{3} &      26.18 &     0.87
 & \Sopt{4} &      \textbf{21.22} &     1.29
 &     0.04
 & \Sopt{4} &      \textbf{21.22} &     \textbf{0.04}
\\
\Sdata{worms} {\tiny(30)}\hspace*{-0.2em}
 & \Sopt{0} &  -43824.08 &   492.00
 &    41.57
 & \Sopt{1} &  -48347.09 &   428.79
 & \Sopt{9} &  -48454.89 &    54.53
 &    15.72
 & \Sopt{27} &  \textbf{-48465.28} &    \textbf{17.77}
\\
\Sdata{pairs} {\tiny(16)}\hspace*{-0.2em}
 & \Sopt{0} &  -63453.77 &   348.33
 &     7.87
 & \Sopt{0} &  -65936.89 &   \textbf{343.49}
 & \Sopt{0} &  -62696.28 &   783.83
 &     1.96
 & \Sopt{0} &  \textbf{-66005.55} &   953.09
\\
    \bottomrule
  \end{tabular}
\end{table*}


\begin{table*}
  \caption{
    \textbf{Comparison table.}
    \textit{our} denotes the proposed \Salg{bca-greedy+qpbo-i} method.
    For each method we state
    \textit{opt\,/\,t} denoting the number of optimally solved instances together with the average time in seconds to attain the optimal solutions ("--" if no instance was solved to optimality),
    the average solution energy $E$ (lower is better),
    and the average solution accuracy \textit{acc} in percent.
    The sign "--" in the $E$ or $acc$ column means that at least for one problem instance the respective method yielded no assignment.
    For datasets indicated by $\dagger$ no ground truth is known and, therefore, no accuracy reported.
    The best accuracy is not highlighted in bold, since algorithms do not have access to the ground truth and hence do not maximize accuracy explicitly.
    The relatively low accuracy of 86\% attained for \Sdata{worms} is explained by model misspecification. The original work~\cite{kainmueller2014active} reports 83\% accuracy achieved by \Salg{dd-ls4} without time restrictions.
    Since \textit{our} method is randomized, we report ranges where appropriate.
    }
  \label{tab:comparison-detailled}
  \newcommand\MyMRow[1]{\smash{\raisebox{-1em}{#1}}}
  \newcommand\MyCellFmt{\scriptsize}
  \centering
  \footnotesize
  \setlength{\tabcolsep}{2.4pt}
  \smallskip%
  \begin{tabular}{@{}l@{}r *{6}{rrr}@{}}
    \toprule
    \multirow{2}{*}[-0.5em]{\begin{minipage}{1.0cm}\centering dataset\\\tiny (number of\\ instances)\end{minipage}}
    & \multirow{2}{*}[-0.25em]{\hspace*{-1em}\begin{minipage}{.7cm}\centering\scriptsize time\\budget\end{minipage}}
    & \multicolumn{3}{c}{\Salg{dd-ls0}~\cite{GraphMatchingDDTorresaniEtAl}}
    & \multicolumn{3}{c}{\Salg{dd-ls3}~\cite{GraphMatchingDDTorresaniEtAl}}
    & \multicolumn{3}{c}{\Salg{HBP}~\cite{HungarianBP}}
    & \multicolumn{3}{c}{\Salg{AMP}~\cite{swoboda2017study}}
    & \multicolumn{3}{c}{\Salg{AMP-tight}~\cite{swoboda2017study}}
    & \multicolumn{3}{c}{\textit{our}} \\
    \cmidrule(lr){3-5}
    \cmidrule(lr){6-8}
    \cmidrule(lr){9-11}
    \cmidrule(lr){12-14}
    \cmidrule(lr){15-17}
    \cmidrule(l){18-20}
    &
    & opt\,/\,$t$ & \textit{E} & acc
    & opt\,/\,$t$ & \textit{E} & acc
    & opt\,/\,$t$ & \textit{E} & acc
    & opt\,/\,$t$ & \textit{E} & acc
    & opt\,/\,$t$ & \textit{E} & acc
    & opt\,/\,$t$ & \textit{E} & acc\\
    \midrule
    \Sdata{hotel} \tiny (105) &
1\,s &
\MyCellFmt \textbf{105}/\textbf{0.01} &
\MyCellFmt \textbf{-4293} &
\MyCellFmt 100 &
\MyCellFmt \textbf{105}/0.04 &
\MyCellFmt \textbf{-4293} &
\MyCellFmt 100 &
\MyCellFmt 102/0.11 &
\MyCellFmt --- &
\MyCellFmt --- &
\MyCellFmt 98/0.11 &
\MyCellFmt -4280 &
\MyCellFmt 99 &
\MyCellFmt 104/0.13 &
\MyCellFmt -4292 &
\MyCellFmt 100 &
\MyCellFmt 100--\textbf{105}/\textbf{0.01} &
\MyCellFmt \textbf{-4291\kern.4pt $\pm$\kern.4pt 2} &
\MyCellFmt 100 \\
\Sdata{house} \tiny (105) &
1\,s &
\MyCellFmt \textbf{105}/0.03 &
\MyCellFmt \textbf{-3778} &
\MyCellFmt 100 &
\MyCellFmt \textbf{105}/0.13 &
\MyCellFmt \textbf{-3778} &
\MyCellFmt 100 &
\MyCellFmt 104/0.20 &
\MyCellFmt --- &
\MyCellFmt --- &
\MyCellFmt 102/0.30 &
\MyCellFmt -3773 &
\MyCellFmt 100 &
\MyCellFmt \textbf{105}/0.19 &
\MyCellFmt \textbf{-3778} &
\MyCellFmt 100 &
\MyCellFmt \textbf{105}/\textbf{0.01} &
\MyCellFmt \textbf{-3778} &
\MyCellFmt 100 \\
\Sdata{car} \tiny (30) &
1\,s &
\MyCellFmt \textbf{28}/0.13 &
\MyCellFmt \textbf{-69} &
\MyCellFmt 92 &
\MyCellFmt 14/0.55 &
\MyCellFmt -57 &
\MyCellFmt 74 &
\MyCellFmt 23/0.12 &
\MyCellFmt --- &
\MyCellFmt --- &
\MyCellFmt 24/0.11 &
\MyCellFmt \textbf{-69} &
\MyCellFmt 92 &
\MyCellFmt 26/0.12 &
\MyCellFmt \textbf{-69} &
\MyCellFmt 91 &
\MyCellFmt 27--\textbf{28}/\textbf{0.01} &
\MyCellFmt \textbf{-69} &
\MyCellFmt 91\kern.4pt $\pm$\kern.4pt 1 \\
\Sdata{motor} \tiny (20) &
1\,s &
\MyCellFmt \textbf{20}/0.07 &
\MyCellFmt \textbf{-63} &
\MyCellFmt 97 &
\MyCellFmt 13/0.25 &
\MyCellFmt -57 &
\MyCellFmt 87 &
\MyCellFmt 19/0.14 &
\MyCellFmt --- &
\MyCellFmt --- &
\MyCellFmt 18/0.04 &
\MyCellFmt \textbf{-63} &
\MyCellFmt 96 &
\MyCellFmt 17/0.08 &
\MyCellFmt \textbf{-63} &
\MyCellFmt 97 &
\MyCellFmt 19--\textbf{20}/\textbf{0.01} &
\MyCellFmt \textbf{-63} &
\MyCellFmt 98\kern.4pt $\pm$\kern.4pt 1 \\
\Sdata{opengm}$^\dagger$\tiny (4) &
1\,s &
\MyCellFmt 1/0.81 &
\MyCellFmt -151 &
\MyCellFmt  &
\MyCellFmt 0/--- &
\MyCellFmt -118 &
\MyCellFmt  &
\MyCellFmt 0/--- &
\MyCellFmt --- &
\MyCellFmt  &
\MyCellFmt 0/--- &
\MyCellFmt -57 &
\MyCellFmt  &
\MyCellFmt 0/--- &
\MyCellFmt -150 &
\MyCellFmt  &
\MyCellFmt \textbf{4}/\textbf{0.004} &
\MyCellFmt \textbf{-171} &
\MyCellFmt  \\
 &
10\,s &
\MyCellFmt 3/1.08 &
\MyCellFmt -161 &
\MyCellFmt  &
\MyCellFmt \textbf{4}/2.61 &
\MyCellFmt \textbf{-171} &
\MyCellFmt  &
\MyCellFmt 2/2.71 &
\MyCellFmt -164 &
\MyCellFmt  &
\MyCellFmt 0/--- &
\MyCellFmt -57 &
\MyCellFmt  &
\MyCellFmt 0/--- &
\MyCellFmt -150 &
\MyCellFmt  &
\MyCellFmt \textbf{4}/\textbf{0.004} &
\MyCellFmt \textbf{-171} &
\MyCellFmt  \\
\Sdata{flow}$^\dagger$\tiny (6) &
1\,s &
\MyCellFmt 2/0.79 &
\MyCellFmt -2089 &
\MyCellFmt  &
\MyCellFmt 1/0.90 &
\MyCellFmt -1962 &
\MyCellFmt  &
\MyCellFmt 0/--- &
\MyCellFmt --- &
\MyCellFmt  &
\MyCellFmt 1/0.13 &
\MyCellFmt -2628 &
\MyCellFmt  &
\MyCellFmt 3/0.16 &
\MyCellFmt --- &
\MyCellFmt  &
\MyCellFmt 4--\textbf{5}/\textbf{0.06} &
\MyCellFmt \textbf{-2837\kern.4pt $\pm$\kern.4pt 1} &
\MyCellFmt  \\
 &
10\,s &
\MyCellFmt 3/1.66 &
\MyCellFmt -2819 &
\MyCellFmt  &
\MyCellFmt \textbf{5}/2.81 &
\MyCellFmt -2821 &
\MyCellFmt  &
\MyCellFmt 0/--- &
\MyCellFmt --- &
\MyCellFmt  &
\MyCellFmt 1/0.13 &
\MyCellFmt -2674 &
\MyCellFmt  &
\MyCellFmt 3/0.16 &
\MyCellFmt -2838 &
\MyCellFmt  &
\MyCellFmt \textbf{5}/\textbf{0.06} &
\MyCellFmt \textbf{-2838} &
\MyCellFmt  \\
\Sdata{worms} \tiny (30) &
1\,s &
\MyCellFmt 0/--- &
\MyCellFmt 60597 &
\MyCellFmt 26 &
\MyCellFmt 0/--- &
\MyCellFmt 64158 &
\MyCellFmt 23 &
\MyCellFmt 0/--- &
\MyCellFmt --- &
\MyCellFmt --- &
\MyCellFmt 0/--- &
\MyCellFmt --- &
\MyCellFmt --- &
\MyCellFmt 0/--- &
\MyCellFmt --- &
\MyCellFmt --- &
\MyCellFmt 10--\textbf{22}/\textbf{0.23} &
\MyCellFmt \textbf{-48461\kern.4pt $\pm$\kern.4pt 3} &
\MyCellFmt 86 \\
 &
10\,s &
\MyCellFmt 0/--- &
\MyCellFmt 50578 &
\MyCellFmt 24 &
\MyCellFmt 0/--- &
\MyCellFmt 49610 &
\MyCellFmt 24 &
\MyCellFmt 0/--- &
\MyCellFmt --- &
\MyCellFmt --- &
\MyCellFmt 1/6.45 &
\MyCellFmt -48389 &
\MyCellFmt 86 &
\MyCellFmt 0/--- &
\MyCellFmt --- &
\MyCellFmt --- &
\MyCellFmt 16--\textbf{25}/\textbf{0.39} &
\MyCellFmt \textbf{-48464\kern.4pt $\pm$\kern.4pt 1} &
\MyCellFmt 86 \\
\Sdata{pairs}$^\dagger$\tiny (16) &
10\,s &
\MyCellFmt 0/--- &
\MyCellFmt -61482 &
\MyCellFmt  &
\MyCellFmt 0/--- &
\MyCellFmt -61638 &
\MyCellFmt  &
\MyCellFmt 0/--- &
\MyCellFmt --- &
\MyCellFmt  &
\MyCellFmt 0/--- &
\MyCellFmt -64130 &
\MyCellFmt  &
\MyCellFmt 0/--- &
\MyCellFmt --- &
\MyCellFmt  &
\MyCellFmt 0/--- &
\MyCellFmt \textbf{-65259\kern.4pt $\pm$\kern.4pt 133} &
\MyCellFmt  \\
 &
30\,s &
\MyCellFmt 0/--- &
\MyCellFmt -61482 &
\MyCellFmt  &
\MyCellFmt 0/--- &
\MyCellFmt -61638 &
\MyCellFmt  &
\MyCellFmt 0/--- &
\MyCellFmt --- &
\MyCellFmt  &
\MyCellFmt 0/--- &
\MyCellFmt -64319 &
\MyCellFmt  &
\MyCellFmt 0/--- &
\MyCellFmt --- &
\MyCellFmt  &
\MyCellFmt 0/--- &
\MyCellFmt \textbf{-65594\kern.4pt $\pm$\kern.4pt 120} &
\\
    \bottomrule
  \end{tabular}%
  \vspace{5mm}%
\end{table*}


Additionally, in Section~\ref{sec:comparison-to-competing-methods} we compare our method
to a number of state-of-the-art algorithms.

\subsection{Influence of different components}\label{sec:ablation-study}

\Paragraph{Influence of fusion on solution quality, Fig.~\ref{fig:general-performance}\,(a-b), \Ssup{supp:ablation-study}, \Ssup{supp:detailed-exepriments-ablation}.}

For our first experiment we ran three methods to generate
proposals: \Salg{bca-lap}, \Salg{dd-ls0} and \Salg{greedy}.
The algorithms \Salg{bca-lap} and \Salg{dd-ls0} represent standard dual techniques
with a LAP-based primal heuristic, and \Salg{greedy} constitutes a baseline.
We fuse the generated proposals with the \Salg{ilp} method.

Fig.~\ref{fig:general-performance}\,(a-b) shows the results for these three
proposal generators before and after fusion for two exemplary instances
from the considered datasets.
Although the energy of  \Salg{dd-ls0} proposals is far from optimal,
their fusion immediately leads to much better results.

Although the energy of \Salg{bca-lap} proposals is often much lower than that
of \Salg{dd-ls0}, the energy of the fused solutions is not necessarily lower.
We explain this by lacking diversity
in the \Salg{bca-lap} proposals.
This explanation is confirmed by the performance
of the fused \Salg{greedy} proposals.
Even though the proposal quality for \Salg{greedy} is very dataset dependent,
in combination with fusion it often leads to good results.
While they are still worse than those obtained by dual methods in Fig.~\ref{fig:general-performance}\,(a), they can be very competitive as seen in Fig.~\ref{fig:general-performance}\,(b).

This experiment clearly shows that the overall solution quality can
be substantially improved by fusing generated proposals. Since fusion provides already
very good results with relatively few proposals, it promises a
significant speed-up compared to fusion-free methods as this can significantly reduce
the number of necessary iterations to achieve a certain solution quality.

\Paragraph{Exact vs.~approximate fusion, \Ssup{supp:ablation-study}, \Ssup{supp:detailed-exepriments-ablation}.}

To estimate the speed-up in runtime obtained by fusion we compared the exact \Salg{ilp} solver with the approximate \Salg{qpbo-i}, \Salg{qpbo-p}, \Salg{qpbo-pi} and \Salg{lsatr} solvers.
Among them we found \Salg{qpbo-i} to be best performing in terms of consistent quality and speed.
Despite a worst-case computational complexity of $O(|\SV||\SE|)$, \Salg{qpbo-i} was 10--50~times faster than the dual updates.

\Paragraph{LAP vs.~greedy heuristic for BCA, Fig.~\ref{fig:general-performance}\,(c).}

As noted above, fusion moves only marginally improve performance of
\Salg{bca-lap} because of the low diversity of proposals generated by this method.
This is easy to see if we compare it to \Salg{bca-greedy},
where the LAP heuristic is replaced by the greedy Algorithm~\ref{alg:greedy-heuristic}.
Indeed, for all datasets we observed that fusion of \Salg{bca-greedy} proposals produced
results at least as good as fusion of \Salg{bca-lap} proposals,
even when the \Salg{bca-greedy} proposals themselves had higher energies than those of \Salg{bca-lap}.

\Paragraph{Effect of relaxation tightening,~\Ssup{supp:ablation-study}, \Ssup{supp:detailed-exepriments-ablation}.}

In general, tighter relaxations provide better bounds in the long run.
However, one pays with a higher runtime per iteration for this.
Interestingly, due to fusion all three subgradient methods, \Salg{dd-ls0}, \Salg{dd-ls3} and \Salg{dd-ls4}, get close or even attain the global optimum in most of the datasets. Therefore, due to lower iteration time the method \Salg{dd-ls0} corresponding to the weakest relaxation converges first, and, hence, is preferable. Since fusion notably improves the energy of the found solutions, we claim that without fusion one would have to use tighter relaxations to attain the same result quality.

\Paragraph{Summary.}

Table~\ref{tab:final-average-table} summarizes our performance study.
We include \Salg{dd-ls0} and \Salg{bca-greedy} as the best representatives
of their algorithm classes.
We observed \Salg{qpbo-i}-based fusion to achieve solutions
with the same or lower energy as the underlying proposal generator,
while also on average converging faster than the proposal generator without fusion.
In other words, \emph{it is always sensible to use fusion moves}.

Although \Salg{bca-greedy+qpbo-i} outperforms \Salg{dd-ls0+qpbo-i}, the latter one is very competitive and notably outperforms its basic variant \Salg{dd-ls0}.

\subsection{Comparisons and conclusions}\label{sec:comparison-to-competing-methods}

Table~\ref{tab:comparison-detailled} compares our \Salg{bca-greedy+qpbo-i} method to several state-of-the-art techniques, see also~\Ssup{supp:comparison-study}.
We omitted a detailed comparison to~\cite{CaetanoMCLS09,RandomWalksForGraphMatching,GraduatedAssignmentGold,SpectralTechniqueAssignmentLeordeanu,IntegerFixedPointGraphMatching,FactorizedGraphMatching}, since the accuracy they attain is notably lower than that of the dual methods~\cite{GraphMatchingDDTorresaniEtAl,HungarianBP} as is shown in the latter papers. This conclusion is also supported by our own experiments. The more recent works~\cite{jiang2017graph,yu2020determinant} unfortunately compare only to the weak baselines above, and do not make their code publicly available. Therefore, we restrict our comparison to the duality-based techniques, as they have been shown to perform best on the computer vision datasets. Note that the \Salg{AMP} method~\cite{swoboda2017study} recently pushed up the state-of-the-art within a deep graph matching approach~\cite{rolinek2020deep}.

We distinguish between \emph{easy} problem instances (\Sdata{hotel}, \Sdata{house}, \Sdata{motor}, \Sdata{car}), \emph{mid-difficult} problems (\Sdata{opengm}, \Sdata{flow}, \Sdata{worms}) and \emph{difficult} ones (\Sdata{pairs}).
For the easy datasets we provide results in 1~second, for mid-difficult in 1 and 10 and for difficult in 10 and 30~seconds respectively, see also~\Ssup{supp:comparison-study} for other run-time settings and a comparison of memory consumption.
For qualitative results on \Sdata{worms} and \Sdata{pairs} see \Ssup{supp:qualitative-results}.

\Paragraph{Conclusions.}

As Table~\ref{tab:comparison-detailled} shows our method notably outperforms its competitors in terms of speed and accuracy. Since it practically solves all easy and mid-difficult problem instances in significantly less than a second, it can be efficiently used in deep graph matching pipelines.
The easy datasets \Sdata{hotel}, \Sdata{house}, \Sdata{motor}, \Sdata{car} are largely solved by all state-of-the-art methods and cannot be used to show progress of the solvers anymore.

\section*{Acknowledgements}

This work was supported by
the European Research Council
(ERC European Unions Horizon 2020 research and innovation program, grant 647769),
the German Research Foundation
(Exact Relaxation-Based Inference in Graphical Models, DFG SA 2640/1-1),
and
the Helmholtz Information \& Data Science School for Health (HIDSS4Health).
The computations were performed on an HPC Cluster at the Center for
Information Services and High Performance Computing (ZIH) at TU Dresden.

\bibliographystyle{plainnat}
\bibliography{venues-full,paper}

\clearpage
\onecolumn
\raggedbottom
\setlength\parindent{0pt}
\setlength\parskip\medskipamount
\setcounter{section}{0}
\setcounter{equation}{0}
\setcounter{figure}{0}
\setcounter{table}{0}
\renewcommand{\thesection}{A\arabic{section}}
\renewcommand{\theequation}{A\arabic{equation}}
\renewcommand{\thefigure}{A\arabic{figure}}
\renewcommand{\thetable}{A\arabic{table}}

\begin{center}
  \begin{minipage}{.7\textwidth}
    \centering
    \scshape\LARGE
    Please do not print the full PDF!

    \medskip
    \normalfont\normalsize
    (For reproducibilty and easier comparision of work in the future, this
    appendix contains a record of all experimental results, totaling over 100
    pages.)
  \end{minipage}
\end{center}

\bigskip
\begin{center}
  \huge\bfseries
  Appendix 
\end{center}
\bigskip

\section{Proof of Proposition 1}\label{supp:proof-prop-1}

\setcounter{proposition}{0}
\begin{proposition}[as stated in the main paper]
  Let $x'$ be a feasible, and $x''$ a possibly infeasible assignment
  for the graph matching problem~\eqref{eq:qap}.
  Let $m$ be the number of dummy, and $n$ the
  number of different non-dummy labels in $x''$.
  Then the auxiliary problem~\eqref{eq:qap-fusion}
  has at most
  $2^{m}\bigl(\frac{|\SV|}{n} + 1\bigr)^n$
  feasible solutions.
\end{proposition}

\begin{proof}
Let $s_1,\ldots,s_n \in \SL$ be the $n$ distinct non-dummy labels
occurring in $x''$, and let $m_1,\ldots,m_n \in \mathbb{N}$ be the
corresponding number of occurences. Obviously,
\begin{align}
    m_i & \geq 1 \hspace*{2em} \text{for all $i\in \{1,\ldots ,n\}$}, \label{eq:proof-positive} \\
    \sum_{i=1}^n m_i & \leq |\SV|. \label{eq:proof-bound}
\end{align}

The number of feasible solutions of the auxiliary problem~\eqref{eq:qap-fusion} is
bounded by the maximum number of choices possible between $x'$ and $x''$,
since any solution $\hat x$ of~\eqref{eq:qap-fusion} satisfies $\hat x_u = x'_u$ or
$\hat x_u = x''_u$ for all $u\in\SV$ by definition.

We can now observe the following:
\begin{itemize}
    \item For each node $u\in\SV$ where $x''$ is assigned the dummy label,
    \ie~$x''_u = \#$, we have at most 2 possible choices. We can either
    choose $x'_u$, or $\#$.

    \item Consider fixed $i\in\{1,\ldots,n\}$. For the set of all nodes
    that share label $s_i$ in $x''$, we have at most $m_i+1$ choices.
    We can either choose $s_i$ for exactly one of these $m_i$ nodes
    or for none, and stay with the label from $x'$ for the remaining of
    these nodes. $s_i$ cannot be chosen for more than one of these $m_i$
    nodes, as this would guarantee the resulting assignment to be
    infeasible.
\end{itemize}
Hence, by basic combinatorics the number of feasible solutions of the
auxiliary problem is at most
\begin{align}
    2^m\cdot\prod_{i=1}^n (m_i+1).
\end{align}
By the inequality of arithmetic and geometric means,
applicable due to~\eqref{eq:proof-positive},
together with~\eqref{eq:proof-bound}, we obtain
\begin{align}
    2^m\cdot\prod_{i=1}^n (m_i+1) & \leq 2^m\cdot \Biggl( \frac{1}{n}\sum_{i=1}^n(m_i+1) \Biggr) ^n = 2^m\cdot \Biggl( \frac{1}{n}\sum_{i=1}^n m_i + 1 \Biggr) ^n \\
        & \leq 2^m \cdot \Biggl( \frac{|\SV|}{n} + 1\Biggr )^n,
\end{align}
which proves Proposition~\ref{prop:search-space-size}.
\end{proof}

\section{Dual algorithm description}\label{supp:dual-algorithm}

Consider the dual problem~\eqref{eq:qap-dual-problem}:

\begin{equation}
 \max_{\phi,\lambda}\left[D(\phi,\lambda)+\min_{\hat\mu\in B}E^{\text{LAP}}(\hat\mu,\lambda)\right]\,.
\end{equation}

To avoid the necessity to minimize $E^{\text{LAP}}$ on each iteration we consider an equivalent dual

\begin{equation}\label{equ:simpler-dual-1}
 \max_{\phi,\lambda}\left[D(\phi,\lambda)+\min_{\hat\mu\in\{0,1\}^{I} \atop \forall s\in\SL\colon\hspace{-5pt} \sum\limits_{u\in\SV(s)}\hspace{-5pt}\hat\mu_{u,s} \le 1} \sum_{\substack{u\in \SV\\ s\in\SL^{\#}_u}}\hspace{-2pt}\hat\mu_{u,s}\hat\xi^\lambda_u(s) \right]\,,
\end{equation}
where $I=\sum_{s\in\SL}|\SV(s)|$, and the second term contains only the \emph{node uniqueness} constraints~\eqref{eq:qap-as-C3} and does not contain the \emph{label uniqueness} constraints~\eqref{eq:qap-as-C2}, as the latter are already included in the first term $D(\phi,\lambda)$. We also fix the values $\lambda_{u,\#}$ to $\theta_u(\#)/2$ implying that $\hat\xi^\lambda_u(\#)=0$, as the corresponding variable $\hat\mu_{u,\#}$ is not included in the uniqueness constraints in the second term of~\eqref{equ:simpler-dual-1}.

Note that the inequalities in the second term of~\eqref{equ:simpler-dual-1} are independent for different values of $s\in\SL$, and, therefore, the corresponding minimization can be solved in closed form. \eqref{equ:simpler-dual-1}~turns into
\begin{equation}\label{equ:simpler-dual-2}
 \max_{\phi,\lambda}\left[D(\phi,\lambda)+\sum_{s\in\SL_u}\min_{u\in \SV(s)}\{\hat\xi^\lambda_u(s),0\} \right]\,.
\end{equation}

For the sake of notation, similarly to the dummy label, we introduce a \emph{dummy node} $\#$ and  introduce the respective notations 
$\SV^\#(s):=\SV(s)\cup\{\#\}$ and $\hat\xi^\lambda_\#(s):=0$, $\forall s\in\SL$. This turns the problem~\eqref{equ:simpler-dual-2} into
\begin{equation}\label{equ:simpler-dual-3}
 \max_{\phi,\lambda}\left[D(\phi,\lambda)+\sum_{s\in\SL_u}\min_{u\in \SV^\#(s)}\hat\xi^\lambda_u(s) \right]\,.
\end{equation}

We maximize the objective of~\eqref{equ:simpler-dual-3} by interleaving $\phi$- and $\lambda$-steps, defined below. First we perform $\phi$-steps for all edges $uv\in\SE$ and then $\lambda$-steps for all $u\in\SV$ and all $s\in\SL$.
\begin{enumerate}
 \item A \textbf{$\phi$-step} maximizes $D(\phi,\lambda)$ \wrt $(\phi_{u,v}(s),\phi_{v,u}(l)\colon s\in\SL^\#_u, l\in\SL^\#_v)$ for each $uv\in\SE$. We denote the vector $\phi$ prior to the maximization as $\phi^t$ .
 The vector $\lambda$ is fixed. Each maximization consists of two blocks of operations:
 \begin{itemize}
    \item \textit{Accumulation}. The unary costs from the nodes $u$, $v$ are pushed to the pairwise costs of the edge $uv$:
 \begin{align}\label{equ:dual-redistribution-step}
 \forall s\in\SL^\#_u\,,\quad \phi^{t+1}_{u,v}(s):= & \phi^{t}_{u,v}(s) + \xi^{\phi^t,\lambda}_u(s)\,,\\
 \forall l\in\SL^\#_v\,,\quad \phi^{t+1}_{v,u}(l):= & \phi^{t}_{v,u}(l) + \xi^{\phi^t,\lambda}_v(l)\,.\nonumber
 \end{align}
    \item \textit{Redistribution}. The pairwise costs from the edge $uv$ are redistributed to the unary costs of the incident nodes $u$ and~$v$:
 \begin{align}\label{equ:dual-aggregation-step}
 \forall uv\in\SE\colon \\
 &\forall s\in\SL^\#_u\,, \quad \phi^{t+2}_{u,v}(s):=\phi^{t+1}_{u,v}(s)  - \frac{1}{2}\min_{l\in\SL^\#_v}\theta^{\phi^{t+1}}_{uv}(s,l)\,, \nonumber\\
 &\forall l\in\SL^\#_v\,, \quad \phi^{t+2}_{v,u}(l):=-\min_{s\in\SL^\#_u}\left(\theta_{uv}(s,l) + \phi^{t+2}_{u,v}(s)\right) \,,\nonumber \\
 &\forall s\in\SL^\#_u\,, \quad \phi^{t+3}_{u,v}(s):=\phi^{t+2}_{u,v}(s)  - \min_{l\in\SL^\#_v}\theta^{\phi^{t+2}}_{uv}(s,l)\,. \nonumber
 \end{align}
\end{itemize}
Operations~\eqref{equ:dual-redistribution-step}-\eqref{equ:dual-aggregation-step} are the updates of the MPLP++ algorithm for maximization of the MRF dual $D$, see~\cite{tourani2018mplp} for a theoretical substantiation, empirical evaluation, proof of convergence and the BCA property.

 \item A \textbf{$\lambda$-step} aims to increase the value of the dual objective by updating first the block of variables $(\lambda_{u,s}\colon s\in\SL^\#_u)$ for each $u\in\SV$, and then the block $(\lambda_{u,s}\colon u\in\SV(s))$ for each $s\in\SL$. The vector $\phi$ is fixed. We denote the vector $\lambda$ prior to the operations as $\lambda^t$ .
\begin{itemize}
 \item Consider the minimal label $s^*$ and the second minimal label $s'$ in a node $u$:
 \begin{equation}\label{equ:lambda-step-u-local-opt}
  s^*\in\argmin_{s\in\SL^\#_u}\xi^{\phi,\lambda^t}_{u}(s)\,;\quad s'\in\argmin_{s\in\SL^\#_u\backslash\{s^*\}}\xi^{\phi,\lambda^t}_{u}(s)\,.
 \end{equation}
 The first update step consists in setting the values of $\xi^{\phi,\lambda^{t+1}}_u(s)$ for all $s\in\SL_u$ to $\Delta_u:=\frac{\xi^{\phi,\lambda^t}_u(s') - \xi^{\phi,\lambda^t}_u(s^*)}{2}$:
\begin{equation}\label{equ:lambda-step-u}
\forall s\in\SL_u\colon\quad \lambda^{t+1}_{u,s}=\lambda^{t}_{u,s} - \xi^{\phi,\lambda^t}_u(s) + \Delta_u\,.
\end{equation}
\item Symmetrically for each $s\in\SL$ let
\begin{equation}
  u^*\in\argmin_{u\in\SV^\#(s)}\hat\xi^{\lambda^{t+1}}_{u}(s)\,;\quad u'\in\argmin_{u\in\SV^\#(s)\backslash\{u^*\}}\hat\xi^{\lambda^{t+1}}_{u}(s)\,.
\end{equation}
 The second update step sets the values of $\hat\xi^{\lambda^{t+2}}_u(s)$ for all $u\in\SV(s)$ to $\hat\Delta_s:=\frac{\hat\xi^{\lambda^{t+1}}_{u'}(s) - \hat\xi^{\lambda^{t+1}}_{u^*}(s)}{2}$:
\begin{equation}
\forall u\in\SV(s)\colon \quad \lambda^{t+2}_{u,s}=\lambda^{t+1}_{u,s} + \hat\xi^{\lambda^{t+1}}_u(s) - \hat\Delta_s\,.
\end{equation}
\end{itemize}
Each $\lambda$-step constitutes an \emph{admissible message}~\cite{swoboda2017dual}, and, therefore, guarantees a monotonic (non-decreasing) improvement of the dual objective, see~\cite{swoboda2017dual,swoboda2017study} for details.
\end{enumerate}

We execute the greedy Algorithm~\ref{alg:greedy-heuristic} between the $\phi$- and $\lambda$-steps.

\section{Details of used datasets}\label{supp:detailed-datasets}

Our experimental evaluation was conducted on 8~datasets from computer vision and
bio-imaging, whose characteristics are listed in Table~\ref{tab:datasets}, and which are described again in detail below, see also Section~\ref{sec:experiments}. To demonstrate the scalability of our approach, along with the standard small-scale datasets for computer vision \Sdata{hotel}, \Sdata{house}, \Sdata{car}, \Sdata{motor} and \Sdata{opengm} with $|\SV| \le 52$, we consider the middle-sized ones \Sdata{flow}, $|\SV| \le 126$, and the large-scaled \Sdata{worms} and \Sdata{pairs} datasets with $|\SV| \le 565$. The latter are, to our knowledge, the largest graph matching problem instances ever investigated in the literature.

We provide all dataset instances used in the paper on our project website:
\url{https://vislearn.github.io/libmpopt/iccv2021/}

\emph{Wide baseline matching} (\Sdata{hotel}, \Sdata{house})
is based on a series of images of the same object
from different view angles. We use the same image pairs, landmarks, and
cost structure as in~\cite{GraphMatchingDDTorresaniEtAl} based on
the work by~\cite{CaetanoMCLS09}.

\emph{Keypoint matching} (\Sdata{car}, \Sdata{motor})
contains \emph{car} and \emph{motor}bike instances from the
PASCAL VOC 2007 Challenge~\cite{VocPascal}
with the features and costs from~\cite{UnsupervisedLearningForGraphMatching}.
We preprocessed the models by removing edges
with zero cost, thereby reducing graph density substantially.

\emph{Large displacement flow} (\Sdata{flow})
was introduced by~\cite{GraphFlow} for key point matching
on scenes with large displacement flow.
We use the keypoints and costs from~\cite{swoboda2017study}.

\emph{OpenGM matching} (\Sdata{opengm})
is a set of non-rigid point matching
problems by~\cite{KomodakisP08}, now part of the
\emph{OpenGM} Benchmark~\cite{OpenGMBenchmark}.

\emph{Worm atlas matching} (\Sdata{worms})
has the goal to annotate nuclei of \emph{C.~elegans}, a famous model
organism used in developmental biology,
by assigning nuclei names from a precomputed atlas of
the organism.
We use the models from~\cite{kainmueller2014active,kainmueller2017graph}.

\emph{Worm-to-worm matching} (\Sdata{pairs}),
in contrast to the \Sdata{worms} dataset, directly matches the cell nuclei of
individual \emph{C.~elegans} worms to each other.
This alleviates the need to precompute an atlas based on
manual annotations.
Unary and pairwise costs of the respective graph matching problems
are derived by averaging the nucleus-(pair-)specific covariance
matrices captured by the atlas over all nuclei.
This coarsens the model to a level achievable without manual annotations.
For our experiments we randomly chose 16~instances out of
the $30\cdot 29 = 870$ non-trivial pairs of worms based on the same
data as \Sdata{worms}.

The \Sdata{pairs} dataset was constructed by ourselves on the basis of the \Sdata{worms} data~\cite{kainmueller2014active,kainmueller2017graph}.
The files are available for download on our project website:
\url{https://vislearn.github.io/libmpopt/iccv2021/}

\begin{table}
  \centering
  \caption{
    \textbf{Characteristics of datasets.}
    For all datasets used for evaluation we state number of instances (\emph{inst.}), number of nodes ($n$), number of labels ($|\mathcal{L}|$), and graph density in percent (\emph{dens.}).
  }
  \label{tab:datasets}

  \medskip
  \small
  \setlength\tabcolsep{14pt}%
  \begin{tabular}{l*4c}
    \toprule
    & \textit{inst.} & $n$ & $|\mathcal{L}|$ & \textit{dens. (\%)} \\ \midrule
    \Sdata{hotel} & 105 & 30 & $=n$ & 100  \\
    \Sdata{house} & 105 & 30 & $=n$ & 100 \\
    \Sdata{car}$^\dagger$ & 30 & 19--49 & $=n$ & 11--27 \\
    \Sdata{motor}$^\dagger$ & 20 & 15--52 & $=n$ & 10--32\\
    \Sdata{flow} & 6 & 48--126 & $\approx n$ & 45--98 \\
    \Sdata{opengm} & 4 & 19\,/\,20 & $=n$ & 66\,/\,100\\
    \Sdata{worms} & 30 & 558 & $\approx 2.4\, n$ & $\approx 1.5$ \\
    \Sdata{pairs} & 16 & 511--565 & $\approx n$ & $\approx 20$\\
    \bottomrule
  \end{tabular}\\
  {\small $^\dagger$\,Zero edges were removed. Prior to this, graph density was $100\,\%$.}
\end{table}


\section{Additional material for performance study}\label{supp:ablation-study}

\begin{table}
  \centering
  \caption{
    \textbf{Exact vs.~approximate fusion.}
    The table shows the averaged energy of the best proposals generated by \Salg{dd-ls0} for each dataset (\textit{best gen.}), and time in seconds needed on average to generate it ($t_{\text{gen}}$).
    Furthermore, it states for \Salg{ilp} and \Salg{qpbo-i} fusion algorithms how long it took on average to beat the \Salg{dd-ls0} proposal when fusing ($t_{\text{beat}}$), and the average energy of the best proposal generated by the respective fusion method (\textit{best fused}).
    Notably, the best fused proposals obtained with \Salg{qpbo-i} do not differ significantly from those obtained with \Salg{ilp}, while the time it takes \Salg{qpbo-i} to surpass the \Salg{dd-ls0} proposal quality is significantly lower than for \Salg{ilp} fusion.
    For each dataset, the best primal and best time are highlighted in bold.
  }
  \label{tab:exact-vs-approximate}

  \medskip
  \small
  \setlength{\tabcolsep}{12pt}%
  \begin{tabular}{l*6r}
  \toprule
    & \multicolumn{2}{c}{\Salg{dd-ls0}}
    & \multicolumn{2}{c}{+ \Salg{ilp}}
    & \multicolumn{2}{c}{+ \Salg{qpbo-i}} \\
    \cmidrule(lr){2-3}\cmidrule(lr){4-5}\cmidrule(lr){6-7}
    & \textit{best gen.} & $t_{\text{gen}}$
    & $t_{\text{beat}}$ & \textit{best fused}
    & $t_{\text{beat}}$ & \textit{best fused} \\
  \midrule
  \Sdata{hotel}
 &   \textbf{-4293.00} &     0.07
 &     0.09
 &   \textbf{-4293.00}
 &     \textbf{0.04}
 &   \textbf{-4293.00}
\\
\Sdata{house}
 &   \textbf{-3778.13} &     \textbf{0.02}
 &     0.09
 &   \textbf{-3778.13}
 &     \textbf{0.02}
 &   \textbf{-3778.13}
\\
\Sdata{car}
 &     -69.34 &     \textbf{0.17}
 &     0.81
 &     -69.35
 &     \textbf{0.17}
 &     \textbf{-69.37}
\\
\Sdata{motor}
 &     \textbf{-62.95} &     0.06
 &     0.23
 &     \textbf{-62.95}
 &     \textbf{0.03}
 &     \textbf{-62.95}
\\
\Sdata{flow}
 &   -2818.83 &     2.79
 &     4.83
 &   \textbf{-2835.84}
 &     \textbf{1.12}
 &   \textbf{-2835.84}
\\
\Sdata{opengm}
 &      31.42 &     0.94
 &    28.02
 &      \textbf{24.61}
 &     \textbf{0.77}
 &      26.18
\\
\Sdata{worms}
 &  -43824.08 &   492.00
 &   177.65
 &  \textbf{-48349.36}
 &    \textbf{41.57}
 &  -48347.09
\\
\Sdata{pairs}
 &  -63453.77 &   348.33
 &    20.37
 &  \textbf{-65975.40}
 &     \textbf{7.87}
 &  -65936.89
 \\
  \bottomrule
  \end{tabular}
\end{table}


\begin{table}
  \caption{
    \textbf{Maximum peak memory consumption per dataset.}
    The numbers indicate the maximum resident set size of the process.
    The values in the \emph{our} column are cleaned to only count the problem instance data once (otherwise it would be counted twice -- once for the Python interpreter memory and once in the native C++ library).
  }\label{tab:memory-footprint}
  \medskip
  \small
  \centerline{%
  \begin{tabular}{l rrrrrr}
    \toprule
    dataset            & \Salg{dd-ls0} & \Salg{dd-ls3} & \Salg{HBP} & \Salg{AMP} & \Salg{AMP-tight} & \emph{our} \\
    \midrule
    \Sdata{hotel}      &          9 MB &         16 MB &     777 MB &      10 MB &           259 MB &  40 MB \\
    \Sdata{house}      &          9 MB &         16 MB &     714 MB &      10 MB &           309 MB &  40 MB \\
    \Sdata{car}        &         11 MB &         21 MB &    1262 MB &       9 MB &           183 MB &  40 MB \\
    \Sdata{motor}      &         12 MB &         24 MB &     670 MB &      10 MB &           155 MB &  40 MB \\
    \Sdata{opengm}     &         13 MB &         15 MB &     851 MB &       9 MB &            15 MB &  38 MB \\
    \Sdata{flow}       &         28 MB &         40 MB &     ---    &      26 MB &           161 MB &  60 MB \\
    \Sdata{worms}      &        270 MB &        372 MB &     ---    &      78 MB &           509 MB &  58 MB \\
    \Sdata{pairs}      &        257 MB &        351 MB &     ---    &     217 MB &          1494 MB & 270 MB \\
    \bottomrule
  \end{tabular}}
\end{table}


\Paragraph{Exact vs.~approximate fusion, Table~\ref{tab:exact-vs-approximate}.}

Among all evaluated approximate fusion methods we found \Salg{qpbo-i} to be best performing in
terms of consistent quality and speed. Table~\ref{tab:exact-vs-approximate} shows the averaged results for each dataset for exact fusion with the \Salg{ilp} method vs.~approximate fusion with the \Salg{qpbo-i} method when applied on top of the \Salg{dd-ls0} proposal generation method.
More detailed results for all other fusion methods can be found in~\Ssup{supp:detailed-exepriments-ablation}.
Conclusively, Table~\ref{tab:exact-vs-approximate} shows that \Salg{qpbo-i} produces results significantly faster than \Salg{ilp},
while the achieved quality is on par.

\begin{figure}
  \hfill
  \includegraphics[height=11.5em, viewport=16 14 276 202, clip]{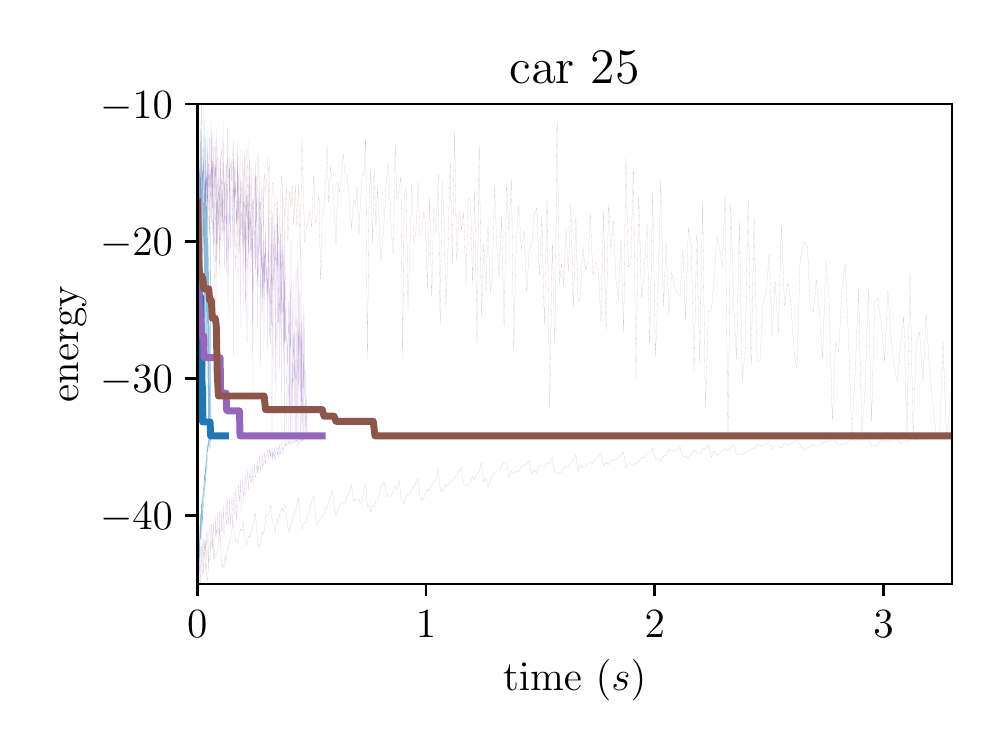}\hfill
  \includegraphics[height=11.5em, viewport=14 14 276 202, clip]{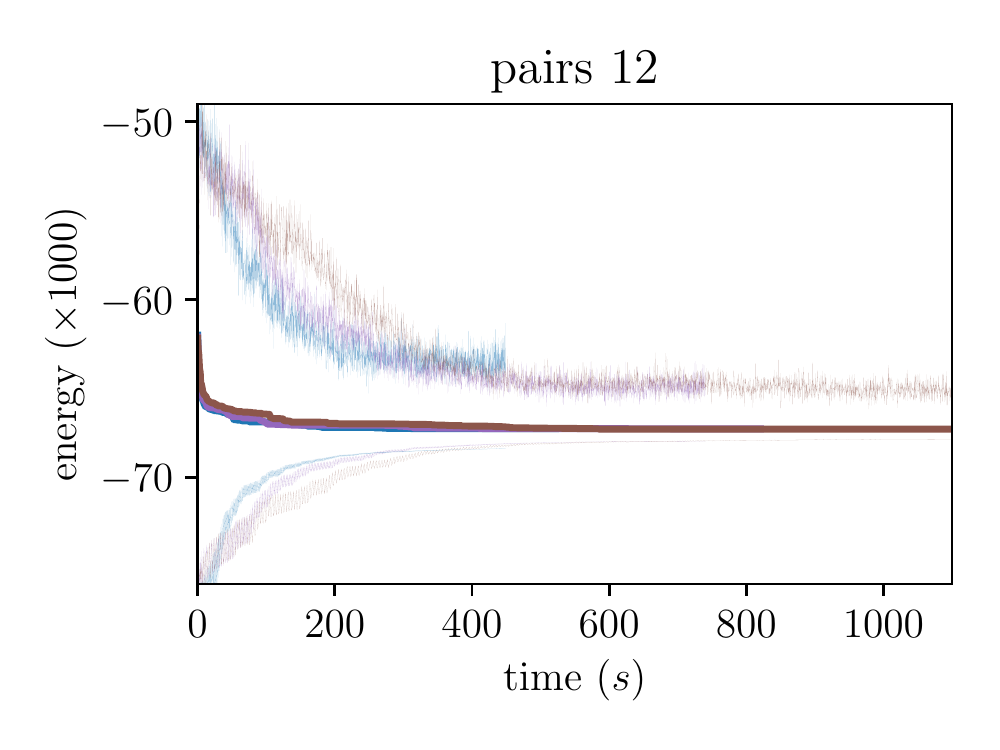}\hfill
  \raisebox{22mm}{%
    \begin{minipage}{5.5em}
      \scriptsize
      \tikz[baseline=-0.5ex]{\draw [line width=0.7pt, black!60] (0,0) -- (1em,0);}~generation\\
        \textit{(lower and\\upper bound)}\\
      \tikz[baseline=-0.5ex]{\draw [line width=0.7pt, blue!60] (0,0) (0.5em,0) -- (1.5em,0);}~\Salg{dd-ls0}\\
      \tikz[baseline=-0.5ex]{\draw [line width=0.7pt, violet!70] (0,0) (0.5em,0) -- (1.5em,0);}~\Salg{dd-ls3}\\
      \tikz[baseline=-0.5ex]{\draw [line width=0.7pt, brown!70] (0,0) (0.5em,0) -- (1.5em,0);}~\Salg{dd-ls4}
      \vspace*{.5em}

      \tikz[baseline=-0.5ex]{\draw [line width=1.8pt, black!60] (0,0) -- (1em,0);}~+ \Salg{qpbo-i}
    \end{minipage}}\hfill\hskip 0pt

  \caption{\textbf{Effect of relaxation tightening.}
    The plots show the obtained dual and primal bound together with the
    \Salg{qpbo-i} fused solutions for \Salg{dd-ls0} (blue), \Salg{dd-ls3} (violet), and
    \Salg{dd-ls4} (brown) over time for two instances from \Sdata{car} and
    \Sdata{pairs}. Unsurprisingly, the tighter relaxations take longer,
    but converge to better bounds for more difficult instances. For the
    \Sdata{car} instance on the left even the basic relaxation is tight,
    while for the \Sdata{pairs} instance on the right better bounds can be observed for
    the tighter relaxations. For all three generation variants, fusion added on top
    gives the same, or, if at all possible, even better bounds much faster.}\label{fig:tightening-effect}
\end{figure}


\Paragraph{Effect of relaxation tightening, Figure~\ref{fig:tightening-effect}.}

In general, tighter relaxations provide better bounds in the long run.
However, for that one pays with a higher runtime per iteration.
This can be clearly seen in Figure~\ref{fig:tightening-effect},
where the performance of \Salg{dd-ls0}, \Salg{dd-ls3} and \Salg{dd-ls4}
solvers is compared. Interestingly, due to fusion all three methods
attain the global optimum for most datasets. Therefore, due to lower iteration
time the method \Salg{dd-ls0} corresponding to the weakest relaxation
converges first, and, hence, is preferable. Only in rare cases, mainly in the \Sdata{pairs}
dataset, the optimum is not reached and methods
\Salg{dd-ls3} and \Salg{dd-ls4} opting for tighter relaxations
overtake \Salg{dd-ls0}. However, these better results come
at the price of significantly higher runtimes.
Since fusion notably improves the energy of the found solutions, we
claim that without it one would have to use tighter relaxations
to attain the same result quality.

\section{Details of the comparison study}\label{supp:comparison-study}

Paragraph{Comparision of peak memory consumption, Table~\ref{tab:memory-footprint}.}

The memory usages of the methods used in our comparison study are comparable, see Table~\ref{tab:memory-footprint}.
Our implementation uses Python bindings to conveniently construct a C++ solver instance, hence our memory consumption is larger by a constant.
Note that we delete the problem instance data from the Python interpreter and run the garbage collector before counting memory consumption for \emph{our} -- otherwise model data would be accounted twice, once in the Python interpreter and once in the C++ native library (only the C++ solver is necessary for optimizing the problem).
Other methods were benchmarked without modifications as to the best of our knowledge they directly construct the model only once in the solver's memory.
The auxiliary fusion problems are tiny when compared to the initial graph matching problems, and their memory consumption is essentially negligible.
Overall, the fusion method can be implemented without sacrifices in memory consumption.

\begin{table}
  \caption{\textbf{Comparison study for extended run-time settings.} The notation is the same as in Table~\ref{tab:comparison-detailled}. Our method remains competitive also in long-term 300 second runs, even when compared to \Salg{dd-ls3} and \Salg{AMP-tight} which are optimizing tighter relaxations than ours.}
  \label{tab:comparison-detailled-full}
  \newcommand\MyMRow[1]{\smash{\raisebox{-1em}{#1}}}
  \newcommand\MyCellFmt{\scriptsize}
  \centering
  \footnotesize
  \setlength{\tabcolsep}{2.15pt}
  \smallskip%
  \centering%
  \begin{tabular}{@{}l@{\hskip -5pt}r *{6}{rrr}@{}}
    \toprule
    \MyMRow{\rlap{dataset}}
    & \MyMRow{\llap{$t_\text{budget}$}}
    & \multicolumn{3}{c}{\Salg{dd-ls0}~\cite{GraphMatchingDDTorresaniEtAl}}
    & \multicolumn{3}{c}{\Salg{dd-ls3}~\cite{GraphMatchingDDTorresaniEtAl}}
    & \multicolumn{3}{c}{\Salg{HBP}~\cite{HungarianBP}}
    & \multicolumn{3}{c}{\Salg{AMP}~\cite{swoboda2017study}}
    & \multicolumn{3}{c}{\Salg{AMP-tight}~\cite{swoboda2017study}}
    & \multicolumn{3}{c}{\textit{our}} \\
    \cmidrule(lr){3-5}
    \cmidrule(lr){6-8}
    \cmidrule(lr){9-11}
    \cmidrule(lr){12-14}
    \cmidrule(lr){15-17}
    \cmidrule(l){18-20}
    &
    & opt\,/\,$t$ & \textit{E} & acc
    & opt\,/\,$t$ & \textit{E} & acc
    & opt\,/\,$t$ & \textit{E} & acc
    & opt\,/\,$t$ & \textit{E} & acc
    & opt\,/\,$t$ & \textit{E} & acc
    & opt\,/\,$t$ & \textit{E} & acc\\
    \midrule
    \Sdata{hotel} \tiny (105) &
1\,s &
\MyCellFmt \textbf{105}/\textbf{0.01} &
\MyCellFmt \textbf{-4293} &
\MyCellFmt 100 &
\MyCellFmt \textbf{105}/0.04 &
\MyCellFmt \textbf{-4293} &
\MyCellFmt 100 &
\MyCellFmt 102/0.11 &
\MyCellFmt --- &
\MyCellFmt --- &
\MyCellFmt 98/0.11 &
\MyCellFmt -4280 &
\MyCellFmt 99 &
\MyCellFmt 104/0.13 &
\MyCellFmt -4292 &
\MyCellFmt 100 &
\MyCellFmt 100--\textbf{105}/\textbf{0.01} &
\MyCellFmt \textbf{-4291\kern.4pt $\pm$\kern.4pt 2} &
\MyCellFmt 100 \\
 &
10\,s &
\MyCellFmt \textbf{105}/\textbf{0.01} &
\MyCellFmt \textbf{-4293} &
\MyCellFmt 100 &
\MyCellFmt \textbf{105}/0.04 &
\MyCellFmt \textbf{-4293} &
\MyCellFmt 100 &
\MyCellFmt 104/0.14 &
\MyCellFmt \textbf{-4293} &
\MyCellFmt 100 &
\MyCellFmt 99/0.13 &
\MyCellFmt -4281 &
\MyCellFmt 99 &
\MyCellFmt \textbf{105}/0.14 &
\MyCellFmt \textbf{-4293} &
\MyCellFmt 100 &
\MyCellFmt \textbf{105}/\textbf{0.01} &
\MyCellFmt \textbf{-4293} &
\MyCellFmt 100 \\
 &
30\,s &
\MyCellFmt \textbf{105}/\textbf{0.01} &
\MyCellFmt \textbf{-4293} &
\MyCellFmt 100 &
\MyCellFmt \textbf{105}/0.04 &
\MyCellFmt \textbf{-4293} &
\MyCellFmt 100 &
\MyCellFmt 104/0.14 &
\MyCellFmt \textbf{-4293} &
\MyCellFmt 100 &
\MyCellFmt 99/0.13 &
\MyCellFmt -4281 &
\MyCellFmt 99 &
\MyCellFmt \textbf{105}/0.14 &
\MyCellFmt \textbf{-4293} &
\MyCellFmt 100 &
\MyCellFmt \textbf{105}/\textbf{0.01} &
\MyCellFmt \textbf{-4293} &
\MyCellFmt 100 \\
 &
180\,s &
\MyCellFmt \textbf{105}/\textbf{0.01} &
\MyCellFmt \textbf{-4293} &
\MyCellFmt 100 &
\MyCellFmt \textbf{105}/0.04 &
\MyCellFmt \textbf{-4293} &
\MyCellFmt 100 &
\MyCellFmt 104/0.14 &
\MyCellFmt \textbf{-4293} &
\MyCellFmt 100 &
\MyCellFmt 99/0.13 &
\MyCellFmt -4281 &
\MyCellFmt 99 &
\MyCellFmt \textbf{105}/0.14 &
\MyCellFmt \textbf{-4293} &
\MyCellFmt 100 &
\MyCellFmt \textbf{105}/\textbf{0.01} &
\MyCellFmt \textbf{-4293} &
\MyCellFmt 100 \\
 &
300\,s &
\MyCellFmt \textbf{105}/\textbf{0.01} &
\MyCellFmt \textbf{-4293} &
\MyCellFmt 100 &
\MyCellFmt \textbf{105}/0.04 &
\MyCellFmt \textbf{-4293} &
\MyCellFmt 100 &
\MyCellFmt 104/0.14 &
\MyCellFmt \textbf{-4293} &
\MyCellFmt 100 &
\MyCellFmt 99/0.13 &
\MyCellFmt -4281 &
\MyCellFmt 99 &
\MyCellFmt \textbf{105}/0.14 &
\MyCellFmt \textbf{-4293} &
\MyCellFmt 100 &
\MyCellFmt \textbf{105}/\textbf{0.01} &
\MyCellFmt \textbf{-4293} &
\MyCellFmt 100 \\
\Sdata{house} \tiny (105) &
1\,s &
\MyCellFmt \textbf{105}/0.03 &
\MyCellFmt \textbf{-3778} &
\MyCellFmt 100 &
\MyCellFmt \textbf{105}/0.13 &
\MyCellFmt \textbf{-3778} &
\MyCellFmt 100 &
\MyCellFmt 104/0.20 &
\MyCellFmt --- &
\MyCellFmt --- &
\MyCellFmt 102/0.30 &
\MyCellFmt -3773 &
\MyCellFmt 100 &
\MyCellFmt \textbf{105}/0.19 &
\MyCellFmt \textbf{-3778} &
\MyCellFmt 100 &
\MyCellFmt \textbf{105}/\textbf{0.01} &
\MyCellFmt \textbf{-3778} &
\MyCellFmt 100 \\
 &
10\,s &
\MyCellFmt \textbf{105}/0.03 &
\MyCellFmt \textbf{-3778} &
\MyCellFmt 100 &
\MyCellFmt \textbf{105}/0.13 &
\MyCellFmt \textbf{-3778} &
\MyCellFmt 100 &
\MyCellFmt \textbf{105}/0.22 &
\MyCellFmt \textbf{-3778} &
\MyCellFmt 100 &
\MyCellFmt 104/0.31 &
\MyCellFmt -3777 &
\MyCellFmt 100 &
\MyCellFmt \textbf{105}/0.19 &
\MyCellFmt \textbf{-3778} &
\MyCellFmt 100 &
\MyCellFmt \textbf{105}/\textbf{0.01} &
\MyCellFmt \textbf{-3778} &
\MyCellFmt 100 \\
 &
30\,s &
\MyCellFmt \textbf{105}/0.03 &
\MyCellFmt \textbf{-3778} &
\MyCellFmt 100 &
\MyCellFmt \textbf{105}/0.13 &
\MyCellFmt \textbf{-3778} &
\MyCellFmt 100 &
\MyCellFmt \textbf{105}/0.22 &
\MyCellFmt \textbf{-3778} &
\MyCellFmt 100 &
\MyCellFmt 104/0.31 &
\MyCellFmt -3777 &
\MyCellFmt 100 &
\MyCellFmt \textbf{105}/0.19 &
\MyCellFmt \textbf{-3778} &
\MyCellFmt 100 &
\MyCellFmt \textbf{105}/\textbf{0.01} &
\MyCellFmt \textbf{-3778} &
\MyCellFmt 100 \\
 &
180\,s &
\MyCellFmt \textbf{105}/0.03 &
\MyCellFmt \textbf{-3778} &
\MyCellFmt 100 &
\MyCellFmt \textbf{105}/0.13 &
\MyCellFmt \textbf{-3778} &
\MyCellFmt 100 &
\MyCellFmt \textbf{105}/0.22 &
\MyCellFmt \textbf{-3778} &
\MyCellFmt 100 &
\MyCellFmt 104/0.31 &
\MyCellFmt -3777 &
\MyCellFmt 100 &
\MyCellFmt \textbf{105}/0.19 &
\MyCellFmt \textbf{-3778} &
\MyCellFmt 100 &
\MyCellFmt \textbf{105}/\textbf{0.01} &
\MyCellFmt \textbf{-3778} &
\MyCellFmt 100 \\
 &
300\,s &
\MyCellFmt \textbf{105}/0.03 &
\MyCellFmt \textbf{-3778} &
\MyCellFmt 100 &
\MyCellFmt \textbf{105}/0.13 &
\MyCellFmt \textbf{-3778} &
\MyCellFmt 100 &
\MyCellFmt \textbf{105}/0.22 &
\MyCellFmt \textbf{-3778} &
\MyCellFmt 100 &
\MyCellFmt 104/0.31 &
\MyCellFmt -3777 &
\MyCellFmt 100 &
\MyCellFmt \textbf{105}/0.19 &
\MyCellFmt \textbf{-3778} &
\MyCellFmt 100 &
\MyCellFmt \textbf{105}/\textbf{0.01} &
\MyCellFmt \textbf{-3778} &
\MyCellFmt 100 \\
\Sdata{car} \tiny (30) &
1\,s &
\MyCellFmt \textbf{28}/0.13 &
\MyCellFmt \textbf{-69} &
\MyCellFmt 92 &
\MyCellFmt 14/0.55 &
\MyCellFmt -57 &
\MyCellFmt 74 &
\MyCellFmt 23/0.12 &
\MyCellFmt --- &
\MyCellFmt --- &
\MyCellFmt 24/0.11 &
\MyCellFmt -69 &
\MyCellFmt 92 &
\MyCellFmt 26/0.12 &
\MyCellFmt -69 &
\MyCellFmt 91 &
\MyCellFmt 27--\textbf{28}/\textbf{0.01} &
\MyCellFmt \textbf{-69} &
\MyCellFmt 91\kern.4pt $\pm$\kern.4pt 1 \\
 &
10\,s &
\MyCellFmt \textbf{29}/0.18 &
\MyCellFmt \textbf{-69} &
\MyCellFmt 91 &
\MyCellFmt 26/1.83 &
\MyCellFmt -68 &
\MyCellFmt 89 &
\MyCellFmt 26/0.33 &
\MyCellFmt --- &
\MyCellFmt --- &
\MyCellFmt 24/0.11 &
\MyCellFmt -69 &
\MyCellFmt 91 &
\MyCellFmt \textbf{29}/0.51 &
\MyCellFmt \textbf{-69} &
\MyCellFmt 92 &
\MyCellFmt 27--\textbf{29}/\textbf{0.06} &
\MyCellFmt \textbf{-69} &
\MyCellFmt 91\kern.4pt $\pm$\kern.4pt 1 \\
 &
30\,s &
\MyCellFmt 29/0.18 &
\MyCellFmt \textbf{-69} &
\MyCellFmt 91 &
\MyCellFmt 29/3.31 &
\MyCellFmt \textbf{-69} &
\MyCellFmt 91 &
\MyCellFmt 26/0.33 &
\MyCellFmt -69 &
\MyCellFmt 91 &
\MyCellFmt 24/\textbf{0.11} &
\MyCellFmt -69 &
\MyCellFmt 91 &
\MyCellFmt \textbf{30}/1.16 &
\MyCellFmt \textbf{-69} &
\MyCellFmt 91 &
\MyCellFmt 27--\textbf{30}/0.71 &
\MyCellFmt \textbf{-69} &
\MyCellFmt 91\kern.4pt $\pm$\kern.4pt 1 \\
 &
180\,s &
\MyCellFmt 29/0.18 &
\MyCellFmt \textbf{-69} &
\MyCellFmt 91 &
\MyCellFmt 29/3.31 &
\MyCellFmt \textbf{-69} &
\MyCellFmt 91 &
\MyCellFmt 26/0.33 &
\MyCellFmt -69 &
\MyCellFmt 91 &
\MyCellFmt 24/\textbf{0.11} &
\MyCellFmt -69 &
\MyCellFmt 91 &
\MyCellFmt \textbf{30}/1.16 &
\MyCellFmt \textbf{-69} &
\MyCellFmt 91 &
\MyCellFmt 27--\textbf{30}/0.71 &
\MyCellFmt \textbf{-69} &
\MyCellFmt 91\kern.4pt $\pm$\kern.4pt 1 \\
 &
300\,s &
\MyCellFmt 29/0.18 &
\MyCellFmt \textbf{-69} &
\MyCellFmt 91 &
\MyCellFmt 29/3.31 &
\MyCellFmt \textbf{-69} &
\MyCellFmt 91 &
\MyCellFmt 26/0.33 &
\MyCellFmt -69 &
\MyCellFmt 91 &
\MyCellFmt 24/\textbf{0.11} &
\MyCellFmt -69 &
\MyCellFmt 91 &
\MyCellFmt \textbf{30}/1.16 &
\MyCellFmt \textbf{-69} &
\MyCellFmt 91 &
\MyCellFmt 27--\textbf{30}/0.71 &
\MyCellFmt \textbf{-69} &
\MyCellFmt 91\kern.4pt $\pm$\kern.4pt 1 \\
\Sdata{motor} \tiny (20) &
1\,s &
\MyCellFmt \textbf{20}/0.07 &
\MyCellFmt \textbf{-63} &
\MyCellFmt 97 &
\MyCellFmt 13/0.25 &
\MyCellFmt -57 &
\MyCellFmt 87 &
\MyCellFmt 19/0.14 &
\MyCellFmt --- &
\MyCellFmt --- &
\MyCellFmt 18/0.04 &
\MyCellFmt -63 &
\MyCellFmt 96 &
\MyCellFmt 17/0.08 &
\MyCellFmt -63 &
\MyCellFmt 97 &
\MyCellFmt 19--\textbf{20}/\textbf{0.01} &
\MyCellFmt \textbf{-63} &
\MyCellFmt 98\kern.4pt $\pm$\kern.4pt 1 \\
 &
10\,s &
\MyCellFmt \textbf{20}/0.07 &
\MyCellFmt \textbf{-63} &
\MyCellFmt 97 &
\MyCellFmt \textbf{20}/1.27 &
\MyCellFmt \textbf{-63} &
\MyCellFmt 97 &
\MyCellFmt \textbf{20}/0.30 &
\MyCellFmt \textbf{-63} &
\MyCellFmt 97 &
\MyCellFmt 18/0.04 &
\MyCellFmt -63 &
\MyCellFmt 96 &
\MyCellFmt 19/0.24 &
\MyCellFmt \textbf{-63} &
\MyCellFmt 99 &
\MyCellFmt \textbf{20}/\textbf{0.01} &
\MyCellFmt \textbf{-63} &
\MyCellFmt 97 \\
 &
30\,s &
\MyCellFmt \textbf{20}/0.07 &
\MyCellFmt \textbf{-63} &
\MyCellFmt 97 &
\MyCellFmt \textbf{20}/1.27 &
\MyCellFmt \textbf{-63} &
\MyCellFmt 97 &
\MyCellFmt \textbf{20}/0.30 &
\MyCellFmt \textbf{-63} &
\MyCellFmt 97 &
\MyCellFmt 18/0.04 &
\MyCellFmt -63 &
\MyCellFmt 96 &
\MyCellFmt 19/0.24 &
\MyCellFmt \textbf{-63} &
\MyCellFmt 99 &
\MyCellFmt \textbf{20}/\textbf{0.01} &
\MyCellFmt \textbf{-63} &
\MyCellFmt 97 \\
 &
180\,s &
\MyCellFmt \textbf{20}/0.07 &
\MyCellFmt \textbf{-63} &
\MyCellFmt 97 &
\MyCellFmt \textbf{20}/1.27 &
\MyCellFmt \textbf{-63} &
\MyCellFmt 97 &
\MyCellFmt \textbf{20}/0.30 &
\MyCellFmt \textbf{-63} &
\MyCellFmt 97 &
\MyCellFmt 18/0.04 &
\MyCellFmt -63 &
\MyCellFmt 96 &
\MyCellFmt 19/0.24 &
\MyCellFmt \textbf{-63} &
\MyCellFmt 99 &
\MyCellFmt \textbf{20}/\textbf{0.01} &
\MyCellFmt \textbf{-63} &
\MyCellFmt 97 \\
 &
300\,s &
\MyCellFmt \textbf{20}/0.07 &
\MyCellFmt \textbf{-63} &
\MyCellFmt 97 &
\MyCellFmt \textbf{20}/1.27 &
\MyCellFmt \textbf{-63} &
\MyCellFmt 97 &
\MyCellFmt \textbf{20}/0.30 &
\MyCellFmt \textbf{-63} &
\MyCellFmt 97 &
\MyCellFmt 18/0.04 &
\MyCellFmt -63 &
\MyCellFmt 96 &
\MyCellFmt 19/0.24 &
\MyCellFmt \textbf{-63} &
\MyCellFmt 99 &
\MyCellFmt \textbf{20}/\textbf{0.01} &
\MyCellFmt \textbf{-63} &
\MyCellFmt 97 \\
\Sdata{opengm}$^\dagger$\tiny (4) &
1\,s &
\MyCellFmt 1/0.81 &
\MyCellFmt -151 &
\MyCellFmt  &
\MyCellFmt 0/--- &
\MyCellFmt -118 &
\MyCellFmt  &
\MyCellFmt 0/--- &
\MyCellFmt --- &
\MyCellFmt  &
\MyCellFmt 0/--- &
\MyCellFmt -57 &
\MyCellFmt  &
\MyCellFmt 0/--- &
\MyCellFmt -150 &
\MyCellFmt  &
\MyCellFmt \textbf{4}/\textbf{0.004} &
\MyCellFmt \textbf{-171} &
\MyCellFmt  \\
 &
10\,s &
\MyCellFmt 3/1.08 &
\MyCellFmt -161 &
\MyCellFmt  &
\MyCellFmt \textbf{4}/2.61 &
\MyCellFmt \textbf{-171} &
\MyCellFmt  &
\MyCellFmt 2/2.71 &
\MyCellFmt -164 &
\MyCellFmt  &
\MyCellFmt 0/--- &
\MyCellFmt -57 &
\MyCellFmt  &
\MyCellFmt 0/--- &
\MyCellFmt -150 &
\MyCellFmt  &
\MyCellFmt \textbf{4}/\textbf{0.004} &
\MyCellFmt \textbf{-171} &
\MyCellFmt  \\
 &
30\,s &
\MyCellFmt 3/1.08 &
\MyCellFmt -161 &
\MyCellFmt  &
\MyCellFmt \textbf{4}/2.61 &
\MyCellFmt \textbf{-171} &
\MyCellFmt  &
\MyCellFmt 2/2.71 &
\MyCellFmt -164 &
\MyCellFmt  &
\MyCellFmt 0/--- &
\MyCellFmt -57 &
\MyCellFmt  &
\MyCellFmt 0/--- &
\MyCellFmt -150 &
\MyCellFmt  &
\MyCellFmt \textbf{4}/\textbf{0.004} &
\MyCellFmt \textbf{-171} &
\MyCellFmt  \\
 &
180\,s &
\MyCellFmt 3/1.08 &
\MyCellFmt -161 &
\MyCellFmt  &
\MyCellFmt \textbf{4}/2.61 &
\MyCellFmt \textbf{-171} &
\MyCellFmt  &
\MyCellFmt 2/2.71 &
\MyCellFmt -164 &
\MyCellFmt  &
\MyCellFmt 0/--- &
\MyCellFmt -57 &
\MyCellFmt  &
\MyCellFmt 0/--- &
\MyCellFmt -150 &
\MyCellFmt  &
\MyCellFmt \textbf{4}/\textbf{0.004} &
\MyCellFmt \textbf{-171} &
\MyCellFmt  \\
 &
300\,s &
\MyCellFmt 3/1.08 &
\MyCellFmt -161 &
\MyCellFmt  &
\MyCellFmt \textbf{4}/2.61 &
\MyCellFmt \textbf{-171} &
\MyCellFmt  &
\MyCellFmt 2/2.71 &
\MyCellFmt -164 &
\MyCellFmt  &
\MyCellFmt 0/--- &
\MyCellFmt -57 &
\MyCellFmt  &
\MyCellFmt 0/--- &
\MyCellFmt -150 &
\MyCellFmt  &
\MyCellFmt \textbf{4}/\textbf{0.004} &
\MyCellFmt \textbf{-171} &
\MyCellFmt  \\
\Sdata{flow}$^\dagger$\tiny (6) &
1\,s &
\MyCellFmt 2/0.79 &
\MyCellFmt -2089 &
\MyCellFmt  &
\MyCellFmt 1/0.90 &
\MyCellFmt -1962 &
\MyCellFmt  &
\MyCellFmt 0/--- &
\MyCellFmt --- &
\MyCellFmt  &
\MyCellFmt 1/0.13 &
\MyCellFmt -2628 &
\MyCellFmt  &
\MyCellFmt 3/0.16 &
\MyCellFmt --- &
\MyCellFmt  &
\MyCellFmt \textbf{4}--\textbf{5}/\textbf{0.06} &
\MyCellFmt \textbf{-2837\kern.4pt $\pm$\kern.4pt 1} &
\MyCellFmt  \\
 &
10\,s &
\MyCellFmt 3/1.66 &
\MyCellFmt -2819 &
\MyCellFmt  &
\MyCellFmt \textbf{5}/2.81 &
\MyCellFmt -2821 &
\MyCellFmt  &
\MyCellFmt 0/--- &
\MyCellFmt --- &
\MyCellFmt  &
\MyCellFmt 1/0.13 &
\MyCellFmt -2674 &
\MyCellFmt  &
\MyCellFmt 3/0.16 &
\MyCellFmt -2838 &
\MyCellFmt  &
\MyCellFmt \textbf{5}/\textbf{0.06} &
\MyCellFmt \textbf{-2838} &
\MyCellFmt  \\
 &
30\,s &
\MyCellFmt 3/1.66 &
\MyCellFmt -2819 &
\MyCellFmt  &
\MyCellFmt \textbf{5}/2.81 &
\MyCellFmt -2834 &
\MyCellFmt  &
\MyCellFmt 0/--- &
\MyCellFmt --- &
\MyCellFmt  &
\MyCellFmt 1/0.13 &
\MyCellFmt -2674 &
\MyCellFmt  &
\MyCellFmt 4/5.27 &
\MyCellFmt -2838 &
\MyCellFmt  &
\MyCellFmt \textbf{5}/\textbf{0.06} &
\MyCellFmt \textbf{-2838} &
\MyCellFmt  \\
 &
180\,s &
\MyCellFmt 3/1.66 &
\MyCellFmt -2819 &
\MyCellFmt  &
\MyCellFmt 5/2.81 &
\MyCellFmt -2834 &
\MyCellFmt  &
\MyCellFmt 0/--- &
\MyCellFmt --- &
\MyCellFmt  &
\MyCellFmt 1/\textbf{0.13} &
\MyCellFmt -2674 &
\MyCellFmt  &
\MyCellFmt 4/5.27 &
\MyCellFmt -2838 &
\MyCellFmt  &
\MyCellFmt \textbf{6}/14.61 &
\MyCellFmt \textbf{-2840} &
\MyCellFmt  \\
 &
300\,s &
\MyCellFmt 3/1.66 &
\MyCellFmt -2819 &
\MyCellFmt  &
\MyCellFmt 5/2.81 &
\MyCellFmt -2834 &
\MyCellFmt  &
\MyCellFmt 0/--- &
\MyCellFmt --- &
\MyCellFmt  &
\MyCellFmt 1/\textbf{0.13} &
\MyCellFmt -2674 &
\MyCellFmt  &
\MyCellFmt 4/5.27 &
\MyCellFmt -2838 &
\MyCellFmt  &
\MyCellFmt \textbf{6}/14.61 &
\MyCellFmt \textbf{-2840} &
\MyCellFmt  \\
\Sdata{worms} \tiny (30) &
1\,s &
\MyCellFmt 0/--- &
\MyCellFmt 60597 &
\MyCellFmt 26 &
\MyCellFmt 0/--- &
\MyCellFmt 64158 &
\MyCellFmt 23 &
\MyCellFmt 0/--- &
\MyCellFmt --- &
\MyCellFmt --- &
\MyCellFmt 0/--- &
\MyCellFmt --- &
\MyCellFmt --- &
\MyCellFmt 0/--- &
\MyCellFmt --- &
\MyCellFmt --- &
\MyCellFmt \textbf{10}--\textbf{22}/\textbf{0.23} &
\MyCellFmt \textbf{-48461\kern.4pt $\pm$\kern.4pt 3} &
\MyCellFmt 86 \\
 &
10\,s &
\MyCellFmt 0/--- &
\MyCellFmt 50578 &
\MyCellFmt 24 &
\MyCellFmt 0/--- &
\MyCellFmt 49610 &
\MyCellFmt 24 &
\MyCellFmt 0/--- &
\MyCellFmt --- &
\MyCellFmt --- &
\MyCellFmt 1/6.45 &
\MyCellFmt -48389 &
\MyCellFmt 86 &
\MyCellFmt 0/--- &
\MyCellFmt --- &
\MyCellFmt --- &
\MyCellFmt \textbf{16}--\textbf{25}/\textbf{0.39} &
\MyCellFmt \textbf{-48464\kern.4pt $\pm$\kern.4pt 1} &
\MyCellFmt 86 \\
 &
30\,s &
\MyCellFmt 0/--- &
\MyCellFmt 35020 &
\MyCellFmt 32 &
\MyCellFmt 0/--- &
\MyCellFmt 33948 &
\MyCellFmt 32 &
\MyCellFmt 0/--- &
\MyCellFmt --- &
\MyCellFmt --- &
\MyCellFmt 1/6.45 &
\MyCellFmt -48392 &
\MyCellFmt 86 &
\MyCellFmt 0/--- &
\MyCellFmt -48429 &
\MyCellFmt 85 &
\MyCellFmt \textbf{18}--\textbf{27}/\textbf{2.36} &
\MyCellFmt \textbf{-48464\kern.4pt $\pm$\kern.4pt 1} &
\MyCellFmt 86 \\
 &
180\,s &
\MyCellFmt 0/--- &
\MyCellFmt -36697 &
\MyCellFmt 79 &
\MyCellFmt 0/--- &
\MyCellFmt -17316 &
\MyCellFmt 63 &
\MyCellFmt 0/--- &
\MyCellFmt --- &
\MyCellFmt --- &
\MyCellFmt 1/6.45 &
\MyCellFmt -48392 &
\MyCellFmt 86 &
\MyCellFmt 0/--- &
\MyCellFmt -48440 &
\MyCellFmt 85 &
\MyCellFmt \textbf{18}--\textbf{27}/\textbf{2.36} &
\MyCellFmt \textbf{-48465\kern.4pt $\pm$\kern.4pt 1} &
\MyCellFmt 86 \\
 &
300\,s &
\MyCellFmt 0/--- &
\MyCellFmt -43805 &
\MyCellFmt 84 &
\MyCellFmt 0/--- &
\MyCellFmt -39283 &
\MyCellFmt 81 &
\MyCellFmt 0/--- &
\MyCellFmt --- &
\MyCellFmt --- &
\MyCellFmt 1/6.45 &
\MyCellFmt -48392 &
\MyCellFmt 86 &
\MyCellFmt 0/--- &
\MyCellFmt -48441 &
\MyCellFmt 85 &
\MyCellFmt \textbf{18}--\textbf{27}/\textbf{2.36} &
\MyCellFmt \textbf{-48465\kern.4pt $\pm$\kern.4pt 1} &
\MyCellFmt 86 \\
\Sdata{pairs}$^\dagger$\tiny (16) &
1\,s &
\MyCellFmt 0/--- &
\MyCellFmt -61482 &
\MyCellFmt  &
\MyCellFmt 0/--- &
\MyCellFmt \textbf{-61638} &
\MyCellFmt  &
\MyCellFmt 0/--- &
\MyCellFmt --- &
\MyCellFmt  &
\MyCellFmt 0/--- &
\MyCellFmt --- &
\MyCellFmt  &
\MyCellFmt 0/--- &
\MyCellFmt --- &
\MyCellFmt  &
\MyCellFmt 0/--- &
\MyCellFmt -61162\kern.4pt $\pm$\kern.4pt 1059 &
\MyCellFmt  \\
 &
10\,s &
\MyCellFmt 0/--- &
\MyCellFmt -61482 &
\MyCellFmt  &
\MyCellFmt 0/--- &
\MyCellFmt -61638 &
\MyCellFmt  &
\MyCellFmt 0/--- &
\MyCellFmt --- &
\MyCellFmt  &
\MyCellFmt 0/--- &
\MyCellFmt -64130 &
\MyCellFmt  &
\MyCellFmt 0/--- &
\MyCellFmt --- &
\MyCellFmt  &
\MyCellFmt 0/--- &
\MyCellFmt \textbf{-65259\kern.4pt $\pm$\kern.4pt 133} &
\MyCellFmt  \\
 &
30\,s &
\MyCellFmt 0/--- &
\MyCellFmt -61482 &
\MyCellFmt  &
\MyCellFmt 0/--- &
\MyCellFmt -61638 &
\MyCellFmt  &
\MyCellFmt 0/--- &
\MyCellFmt --- &
\MyCellFmt  &
\MyCellFmt 0/--- &
\MyCellFmt -64319 &
\MyCellFmt  &
\MyCellFmt 0/--- &
\MyCellFmt --- &
\MyCellFmt  &
\MyCellFmt 0/--- &
\MyCellFmt \textbf{-65594\kern.4pt $\pm$\kern.4pt 120} &
\MyCellFmt  \\
 &
180\,s &
\MyCellFmt 0/--- &
\MyCellFmt -63436 &
\MyCellFmt  &
\MyCellFmt 0/--- &
\MyCellFmt -63800 &
\MyCellFmt  &
\MyCellFmt 0/--- &
\MyCellFmt --- &
\MyCellFmt  &
\MyCellFmt 0/--- &
\MyCellFmt -64380 &
\MyCellFmt  &
\MyCellFmt 0/--- &
\MyCellFmt -65786 &
\MyCellFmt  &
\MyCellFmt 0/--- &
\MyCellFmt \textbf{-65798\kern.4pt $\pm$\kern.4pt 86} &
\MyCellFmt  \\
 &
300\,s &
\MyCellFmt 0/--- &
\MyCellFmt -63454 &
\MyCellFmt  &
\MyCellFmt 0/--- &
\MyCellFmt -64503 &
\MyCellFmt  &
\MyCellFmt 0/--- &
\MyCellFmt --- &
\MyCellFmt  &
\MyCellFmt 0/--- &
\MyCellFmt -64380 &
\MyCellFmt  &
\MyCellFmt 0/--- &
\MyCellFmt -65827 &
\MyCellFmt  &
\MyCellFmt 0/--- &
\MyCellFmt \textbf{-65821\kern.4pt $\pm$\kern.4pt 84} &
\\
    \bottomrule
  \end{tabular}
\end{table}


\Paragraph{Extended comparison study, Table~\ref{tab:comparison-detailled-full}.}

It can be seen that our method remains competitive also in longer runs of 300 seconds.
This is the case even when compared to \Salg{dd-ls3} and \Salg{AMP-tight} which are optimizing tighter relaxations than ours.

\Paragraph{Computation of \Sdata{worms} accuracies.}

To make results comparable, we computed accuracies for \Sdata{worms} in the same way as done in~\cite{kainmueller2014active,long20093d}.
Out of 558~nuclei only~357 were faithfully segmented and annotated in most instances. In some instances ground truth is not available in the dataset for all of them.
To measure accuracy we count the fraction of correctly matched nuclei out of the total 357 that are part of the annotated atlas. Due to incomplete annotations, 100\% accuracy is for most instances impossible to obtain.

\Paragraph{Run-time settings for the comparison study.}

Below we provide a description of the run-time settings of all algorithms that have been used in the comparison study.
We also specifiy the source where we obtained the algorithms if applicable.

\begin{itemize}
  \item
    \Salg{bca-greedy+qpbo-i}: See \url{https://vislearn.github.io/libmpopt/iccv2021/}.

    The command used for the benchmark was:

    \verb|qap_dd --max-batches 50000 --batch-size 1 --greedy-generations 1 input.dd|

    Note that the command line tool allows to switch to use Gurobi for solving the fusion move problems.
    The suffix \verb|_dd| selects the \verb|*.dd| file format parser.
    The format was established in~\cite{GraphMatchingDDTorresaniEtAl}.



  \item
    \Salg{dd-ls0}, \Salg{dd-ls3} and \Salg{dd-ls4}:
    We used the source code that the authors provide for~\cite{GraphMatchingDDTorresaniEtAl}.
    The original source code can be obtained at \url{https://pub.ist.ac.at/~vnk/software.html}.
    We modified the code to output more detailed timing, bound and assignment information,
    see \url{https://github.com/vislearn/tkr-graphmatching}.

    The commands used for the benchmark were:

    \Salg{dd-ls0}: \verb|dd --linear --tree input.dd|

    \Salg{dd-ls3}: \verb|dd --linear --local 3 --tree input.dd|

    \Salg{dd-ls4}: \verb|dd --linear --local 4 --tree input.dd|

    \emph{Note:} The \verb|--local| parameter determines the number of nodes in each local subproblem.
    In the original implementation by~\cite{GraphMatchingDDTorresaniEtAl} a setting of \texttt{1} results in two nodes per subproblem (so one more node than the setting suggests).
    We found the original interpretation of the parameter confusing and hence adjusted the behaviour in our command line wrapper.
    As one node per subproblem does not make sense, setting \verb|--local 1| is invalid in our version.

  \item
    \Salg{AMP} and \Salg{AMP-tight}:
    The source code for~\cite{swoboda2017study} was obtained from~\url{https://github.com/LPMP/LPMP}.
    Parameters have been selected after correspondence with the authors of~\cite{swoboda2017study}. \Salg{AMP} optimizes the same relaxation as \Salg{bca} and \Salg{dd-ls0}, and uses a LAP solver as primal heuristic. \Salg{AMP-tight} optimizes the relaxation equivalent to that of \Salg{dd-ls3}, and uses a Frank-Wolfe algorithm as a primal heuristic.

    The commands used for the benchmark were:

    \Salg{AMP}: \verb|graph_matching_mp -i input.dd --roundingReparametrization uniform|

    \Salg{AMP-tight}: \verb|graph_matching_mp_tightening -i input.dd --tighten \|\\
    \hspace*{24pt}\verb|--tightenInterval 50 \|\\
    \hspace*{24pt}\verb|--tightenIteration 200 \|\\
    \hspace*{24pt}\verb|--tightenConstraintsPercentage 0.01 \|\\
    \hspace*{24pt}\verb|--tightenReparametrization uniform:0.5 \|\\
    \hspace*{24pt}\verb|--graphMatchingRounding fw|

  \item
    \Salg{HBP}:
    The Matlab source code for~\cite{HungarianBP} was taken from \url{https://github.com/zzhang1987/HungarianBP}.
    Note that the initialization scripts where unable to download the \Salg{fgm} dependency and we replaced it by \url{https://github.com/zhfe99/fgm}.
    A working source copy can be obtained by running:

    \verb|git clone --recurse-submodules https://github.com/zzhang1987/HungarianBP|\\
    \verb|cd HungarianBP|\\
    \verb|git clone https://github.com/zhfe99/fgm|\\
    \verb|matlab -nodisplay -nojvm -r compiling|

    We modified the code slightly in order to be able to load the datasets and make the output more suitable.
    We converted the datasets into matrix form and saved them in Matlab format.
    The graph matching problem for computer vision is formulated as a minimization problem, but \Salg{HBP} expects a maximization problem.
    Therefore we flipped the sign for all costs before passing them to \Salg{HBP}.
\end{itemize}

\section{Detailed experimental results for the performance study}\label{supp:detailed-exepriments-ablation}
\label{sec:supp-experiments}

To minimize computational time for the performance study we implemented a tool that takes a list of assignments and fuses them one after another. This makes the fusion move operation independent of the dual algorithms and, thereby, avoids having to run the comparably slow dual algorithms for each fusion algorithm anew. We used this tool to report results for all algorithms in the performance study including our winning method \Salg{bca-greedy+qpbo-i}. This is in contrast to the comparison study, see Sections~\ref{sec:comparison-to-competing-methods}, \Ssup{supp:comparison-study}, where an optimized version of the latter method was used.

The command for this tool is: \verb|qap_dd_fuse input.dd proposals.txt fusion_results.txt|

Below we list our complete experimental results on which the tables in
the main paper are based. For each dataset we repeat how many instances
there are in the dataset (\cf Table~\ref{tab:datasets}), and state for
how many iterations the respective generating algorithms ran
for each instance.

Then, for each instance in the dataset we first specify the simplified
name we used, and, if applicable, the name used in previous publications
in brackets.
If the energy of the optimal solution is known, we indicate it.
Optima were taken
from~\cite{swoboda2017study} for \Sdata{hotel}, \Sdata{house},
\Sdata{car}, \Sdata{motor}, \Sdata{flow},
from~\cite{OpenGMBenchmark} for \Sdata{opengm}, and
obtained with CombiLP~\cite{HallerCombilpAAAI2018} for \Sdata{worms}.
A few previously missing optima for \Sdata{car} and \Sdata{motor} are due to
tight lower bounds in our own experiments.

In the table below the instance name, we state in the second column for all evaluated
proposal generating methods, \ie for \Salg{dd-ls0}, \Salg{dd-ls3},
\Salg{dd-ls4}, \Salg{bca-lap}, \Salg{bca-greedy}, and \Salg{greedy},
\begin{itemize}
    \item the best energy the respective method obtained within the
        given maximum number of iterations (\textit{best generated}),
        and
    \item how long the generation ran in seconds until reaching
        this energy ($t_{\text{gen}}$).
\end{itemize}
We then state for all evaluated fusion methods, namely for \Salg{ilp},
\Salg{qpbo}, \Salg{qpbo-i}, \Salg{qpbo-p}, \Salg{qpbo-pi}, and \Salg{lsatr}, in columns four to nine,
\begin{itemize}
    \item how long it took in seconds, when adding the fusion method on top of the
        respective generation method, to obtain an energy at least as good
        as the best energy obtained purely by the underlying generation
        method ($t_{\text{beat}}$),
    \item the overall best energy the respective method obtained when fusing
        the given generated proposals (\textit{best fused}), and
    \item how long fusion ran in seconds on top of generation until
        reaching this energy ($t_{\text{fuse}}$).
\end{itemize}
If the optimal solution is known, it is highlighted in bold if
obtained by the respective generation or fusion method.

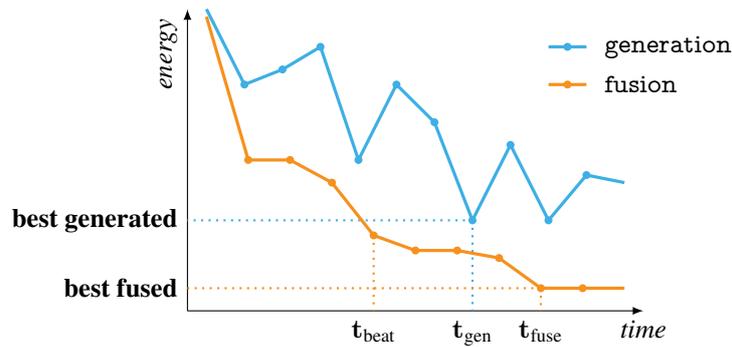
\begin{figure}[t]
  \centering
  \begin{tikzpicture}
    \draw [-latex] (0.25,0) -- node [at end, below] {\textit{time}} ++(6,0);
    \draw [-latex] (0.25,0) -- node [at end, left=.7em, rotate=90] {\textit{energy}} ++(0,4);
    \draw [CornflowerBlue, very thick] (0.5,4)
      \foreach \x/\y in {1/3, 1.5/3.2, 2/3.5, 2.5/2, 3/3, 3.5/2.5, 4/1.2, 4.5/2.2, 5/1.2, 5.5/1.8}
      { -- (\x,\y) node [fill, circle, inner sep=0pt, minimum size=3pt] {}}
      -- (6,1.7);
    \draw [CornflowerBlue, very thick]
      (5,3.5) -- ++(0.25,0) node [fill, circle, inner sep=0pt, minimum size=3pt] {}
      -- ++(0.25,0) ++(0.1,0) node [anchor=west, black] {\Salg{generation}};
    \draw [CornflowerBlue, dotted, thick]
      (4,1.2) -- (4,0) node [anchor=north, black] {$\mathbf{t_{\text{gen}}}$}
      (4,1.2) -- (0.25,1.2) node [anchor=east, black] {\textbf{best generated}};
    \draw [BurntOrange, very thick] (0.5,3.9)
      \foreach \x/\y in {1.05/2, 1.6/2, 2.15/1.7, 2.7/1, 3.25/0.8, 3.8/0.8, 4.35/0.7, 4.9/0.3, 5.45/0.3}
      { -- (\x,\y) node [fill, circle, inner sep=0pt, minimum size=3pt] {}}
      -- (6,0.3);
    \draw [BurntOrange, very thick]
      (5,3) -- ++(0.25,0) node [fill, circle, inner sep=0pt, minimum size=3pt] {}
      -- ++(0.25,0) ++(0.1,0) node [anchor=west, black] {\Salg{fusion}};
    \draw [BurntOrange, dotted, thick]
      (2.7,1) -- (2.7,0) node [anchor=north, black] {$\mathbf{t_{\text{beat}}}$}
      (4.9,0.3) -- (4.9,0) node [anchor=north, black] {$\mathbf{t_{\text{fuse}}}$}
      (4.9,0.3) -- (0.25,0.3) node [anchor=east, black] {\textbf{best fused}};
  \end{tikzpicture}

  \caption{
    \textbf{Fictional generation and fusion data} to illustrate the meaning of
    \textit{best generated}, \textit{best fused}, $t_{\text{gen}}$,
    $t_{\text{beat}}$, and $t_{\text{fuse}}$ as provided in the tables in
    Section~\ref{sec:supp-experiments} for each combination of a
    \Salg{generation} and \Salg{fusion} method.
  }
  \label{fig:supp-table-content}
\end{figure}

The meaning of table entries is
illustrated in Figure~\ref{fig:supp-table-content}.
As explained above, for each generation method each individual table contains four lines.
The entries to the left of ``\textit{best generated}'' and ``$t_{\text{gen}}$''
correspond to the generation method itself, the entries to the right of
``$t_{\text{beat}}$'', ``\textit{best fused}'' and ``$t_{\text{fuse}}$''
correspond columnwise to the respective fusion method applied on top of
the generation method.

\setlength{\parindent}{0pt}
\newcommand{\Sfirstsubsection}[1]{%
    \subsection{\Sdata{#1}}}
\newcommand{\Ssubsection}[1]{\newpage
    \subsection{\Sdata{#1}}}
\newcommand{\Snrinst}[1]{
    {\footnotesize number of instances: #1}}
\newcommand{\Snriter}[1]{
    {\footnotesize maximum number of iterations during generation: #1}}
\newcommand{\Sinstance}[1]{\vspace{1em}
    \begin{minipage}{\textwidth}
    \footnotesize
    \textbf{#1}}
\newcommand{\Salias}[1]{~(\textit{#1})}
\newcommand{\Soptimum}[1]{, known optimum: #1}
\newcommand{\Soptimal}[1]{\mathbf{#1}}
\newcommand{\Sstarttable}{%
    \scriptsize
    \begin{tabular}{l*8r}
    \toprule
        \multicolumn{2}{c}{generation\hspace*{-4em}} & & \multicolumn{6}{c}{+ fusion\hspace*{3.5em}} \\
    \cmidrule(lr{6.5em}){1-3}
    \cmidrule(l{6em}r){3-9}
        & & & \Salg{ilp} & \Salg{qpbo} & \Salg{qpbo-i} & \Salg{qpbo-p} & \Salg{qpbo-pi} & \Salg{lsatr} \\}
\newcommand{\Sendtable}{%
    \bottomrule
    \end{tabular}
    \
    \end{minipage}}
\newcommand{\SgenE}[2]{%
    \midrule
    \Salg{#1} & $#2$ & \multicolumn{1}{l}{\textit{best generated}} \\}
\newcommand{\Sgent}[7]{%
    & $#1$ & $t_{\text{gen}} (s)$\hspace*{1em}|\hspace*{1em}$t_{\text{beat}} (s)$ & $#2$ & $#3$ & $#4$ & $#5$ & $#6$ & $#7$ \\}
\newcommand{\SfusE}[6]{%
    \cmidrule(lr){3-3}
    & & \multicolumn{1}{r}{\textit{best fused}} & $#1$ & $#2$ & $#3$ & $#4$ & $#5$ & $#6$ \\}
\newcommand{\Sfust}[6]{%
    & & \multicolumn{1}{r}{$t_{\text{fuse}} (s)$} & $#1$ & $#2$ & $#3$ & $#4$ & $#5$ & $#6$ \\}

\input{tables/hotel-big}
\input{tables/house-big}
\Ssubsection{car}

\Snrinst{30}

\Snriter{2500}

\Sinstance{car1}\Soptimum{-34.88}\\
\Sstarttable
\SgenE{dd-ls0    }{\Soptimal{    -34.88}}
\Sgent{     0.011}{     0.169}{     0.007}{     0.007}{     0.028}{     0.027}{     0.034}
\SfusE{\Soptimal{    -34.88}}{\Soptimal{    -34.88}}{\Soptimal{    -34.88}}{\Soptimal{    -34.88}}{\Soptimal{    -34.88}}{\Soptimal{    -34.88}}
\Sfust{     0.169}{     0.007}{     0.007}{     0.028}{     0.027}{     0.034}
\SgenE{dd-ls3    }{\Soptimal{    -34.88}}
\Sgent{     0.081}{     0.251}{     0.049}{     0.050}{     0.106}{     0.104}{     0.052}
\SfusE{\Soptimal{    -34.88}}{\Soptimal{    -34.88}}{\Soptimal{    -34.88}}{\Soptimal{    -34.88}}{\Soptimal{    -34.88}}{\Soptimal{    -34.88}}
\Sfust{     0.251}{     0.049}{     0.050}{     0.106}{     0.104}{     0.052}
\SgenE{dd-ls4    }{\Soptimal{    -34.88}}
\Sgent{     1.012}{     0.276}{     0.145}{     0.146}{     1.043}{     1.044}{     0.147}
\SfusE{\Soptimal{    -34.88}}{\Soptimal{    -34.88}}{\Soptimal{    -34.88}}{\Soptimal{    -34.88}}{\Soptimal{    -34.88}}{\Soptimal{    -34.88}}
\Sfust{     0.276}{     0.145}{     0.146}{     1.043}{     1.044}{     0.147}
\SgenE{bca-lap   }{\Soptimal{    -34.88}}
\Sgent{     4.010}{     0.595}{     0.519}{     0.519}{     0.518}{     0.518}{     0.524}
\SfusE{\Soptimal{    -34.88}}{\Soptimal{    -34.88}}{\Soptimal{    -34.88}}{\Soptimal{    -34.88}}{\Soptimal{    -34.88}}{\Soptimal{    -34.88}}
\Sfust{     0.595}{     0.519}{     0.519}{     0.518}{     0.518}{     0.524}
\SgenE{bca-greedy}{\Soptimal{    -34.88}}
\Sgent{     0.525}{     0.069}{     0.012}{     0.013}{     0.014}{     0.013}{     0.012}
\SfusE{\Soptimal{    -34.88}}{\Soptimal{    -34.88}}{\Soptimal{    -34.88}}{\Soptimal{    -34.88}}{\Soptimal{    -34.88}}{\Soptimal{    -34.88}}
\Sfust{     0.069}{     0.012}{     0.013}{     0.014}{     0.013}{     0.012}
\SgenE{greedy    }{\Soptimal{    -34.88}}
\Sgent{     0.030}{     1.143}{     0.090}{     0.098}{     0.098}{     0.093}{     0.117}
\SfusE{\Soptimal{    -34.88}}{\Soptimal{    -34.88}}{\Soptimal{    -34.88}}{\Soptimal{    -34.88}}{\Soptimal{    -34.88}}{\Soptimal{    -34.88}}
\Sfust{     1.143}{     0.090}{     0.098}{     0.098}{     0.093}{     0.117}
\Sendtable

\Sinstance{car2}\Soptimum{-48.85}\\
\Sstarttable
\SgenE{dd-ls0    }{\Soptimal{    -48.85}}
\Sgent{     0.038}{     0.116}{     0.018}{     0.019}{     0.070}{     0.070}{     0.021}
\SfusE{\Soptimal{    -48.85}}{\Soptimal{    -48.85}}{\Soptimal{    -48.85}}{\Soptimal{    -48.85}}{\Soptimal{    -48.85}}{\Soptimal{    -48.85}}
\Sfust{     0.116}{     0.018}{     0.019}{     0.070}{     0.070}{     0.021}
\SgenE{dd-ls3    }{\Soptimal{    -48.85}}
\Sgent{     0.571}{     0.727}{     0.049}{     0.246}{     0.626}{     0.630}{     0.113}
\SfusE{\Soptimal{    -48.85}}{\Soptimal{    -48.85}}{\Soptimal{    -48.85}}{\Soptimal{    -48.85}}{\Soptimal{    -48.85}}{\Soptimal{    -48.85}}
\Sfust{     0.727}{     0.049}{     0.246}{     0.626}{     0.630}{     0.113}
\SgenE{dd-ls4    }{\Soptimal{    -48.85}}
\Sgent{     2.925}{     2.922}{     2.564}{     2.566}{     2.978}{     2.978}{     2.582}
\SfusE{\Soptimal{    -48.85}}{\Soptimal{    -48.85}}{\Soptimal{    -48.85}}{\Soptimal{    -48.85}}{\Soptimal{    -48.85}}{\Soptimal{    -48.85}}
\Sfust{     2.922}{     2.564}{     2.566}{     2.978}{     2.978}{     2.582}
\SgenE{bca-lap   }{\Soptimal{    -48.85}}
\Sgent{     1.987}{     0.133}{     0.084}{     0.083}{     0.083}{     0.083}{     0.084}
\SfusE{\Soptimal{    -48.85}}{\Soptimal{    -48.85}}{\Soptimal{    -48.85}}{\Soptimal{    -48.85}}{\Soptimal{    -48.85}}{\Soptimal{    -48.85}}
\Sfust{     0.133}{     0.084}{     0.083}{     0.083}{     0.083}{     0.084}
\SgenE{bca-greedy}{\Soptimal{    -48.85}}
\Sgent{     0.100}{     0.163}{     0.045}{     0.046}{     0.046}{     0.047}{     0.050}
\SfusE{\Soptimal{    -48.85}}{\Soptimal{    -48.85}}{\Soptimal{    -48.85}}{\Soptimal{    -48.85}}{\Soptimal{    -48.85}}{\Soptimal{    -48.85}}
\Sfust{     0.163}{     0.045}{     0.046}{     0.046}{     0.047}{     0.050}
\SgenE{greedy    }{    -46.57}
\Sgent{     0.086}{     0.683}{     0.032}{     0.033}{     0.272}{     0.280}{     0.290}
\SfusE{\Soptimal{    -48.85}}{\Soptimal{    -48.85}}{\Soptimal{    -48.85}}{    -46.57}{    -46.57}{\Soptimal{    -48.85}}
\Sfust{     2.710}{     0.144}{     0.149}{     0.272}{     0.280}{     0.294}
\Sendtable

\Sinstance{car3}\Soptimum{-86.55}\\
\Sstarttable
\SgenE{dd-ls0    }{\Soptimal{    -86.55}}
\Sgent{     0.373}{     1.339}{     0.286}{     0.293}{     0.519}{     0.515}{     0.438}
\SfusE{\Soptimal{    -86.55}}{\Soptimal{    -86.55}}{\Soptimal{    -86.55}}{\Soptimal{    -86.55}}{\Soptimal{    -86.55}}{\Soptimal{    -86.55}}
\Sfust{     1.339}{     0.286}{     0.293}{     0.519}{     0.515}{     0.438}
\SgenE{dd-ls3    }{\Soptimal{    -86.55}}
\Sgent{     7.842}{     6.645}{     5.102}{     5.109}{     8.025}{     8.027}{     5.175}
\SfusE{\Soptimal{    -86.55}}{\Soptimal{    -86.55}}{\Soptimal{    -86.55}}{\Soptimal{    -86.55}}{\Soptimal{    -86.55}}{\Soptimal{    -86.55}}
\Sfust{     6.645}{     5.102}{     5.109}{     8.025}{     8.027}{     5.175}
\SgenE{dd-ls4    }{\Soptimal{    -86.55}}
\Sgent{    64.664}{    20.677}{    22.817}{    42.422}{    64.995}{    64.991}{    42.504}
\SfusE{\Soptimal{    -86.55}}{\Soptimal{    -86.55}}{\Soptimal{    -86.55}}{\Soptimal{    -86.55}}{\Soptimal{    -86.55}}{\Soptimal{    -86.55}}
\Sfust{    20.677}{    22.817}{    42.422}{    64.995}{    64.991}{    42.504}
\SgenE{bca-lap   }{\Soptimal{    -86.55}}
\Sgent{    34.533}{     0.731}{     0.670}{     0.671}{     0.670}{     0.671}{     0.673}
\SfusE{\Soptimal{    -86.55}}{\Soptimal{    -86.55}}{\Soptimal{    -86.55}}{\Soptimal{    -86.55}}{\Soptimal{    -86.55}}{\Soptimal{    -86.55}}
\Sfust{     0.731}{     0.670}{     0.671}{     0.670}{     0.671}{     0.673}
\SgenE{bca-greedy}{\Soptimal{    -86.55}}
\Sgent{     4.479}{     0.137}{     0.089}{     0.092}{     0.117}{     0.117}{     0.090}
\SfusE{\Soptimal{    -86.55}}{\Soptimal{    -86.55}}{\Soptimal{    -86.55}}{\Soptimal{    -86.55}}{\Soptimal{    -86.55}}{\Soptimal{    -86.55}}
\Sfust{     0.137}{     0.089}{     0.092}{     0.117}{     0.117}{     0.090}
\SgenE{greedy    }{    -76.58}
\Sgent{     0.194}{     1.958}{     0.170}{     0.181}{     0.630}{     0.615}{     0.137}
\SfusE{    -86.34}{    -86.34}{    -86.34}{    -76.58}{    -76.58}{    -86.34}
\Sfust{    14.873}{     0.332}{     0.354}{     0.630}{     0.615}{     1.009}
\Sendtable

\Sinstance{car4}\\
\Sstarttable
\SgenE{dd-ls0    }{    -51.25}
\Sgent{     0.575}{     2.813}{     0.438}{     0.750}{     0.903}{     0.904}{     0.369}
\SfusE{    -51.76}{    -51.43}{    -52.20}{    -51.25}{    -51.25}{    -52.34}
\Sfust{     3.374}{     0.438}{     0.821}{     0.903}{     0.904}{     0.412}
\SgenE{dd-ls3    }{    -52.09}
\Sgent{    15.093}{    18.353}{    15.497}{    15.529}{    15.554}{    15.613}{     8.613}
\SfusE{    -52.23}{    -52.23}{    -52.23}{    -52.09}{    -52.09}{    -52.23}
\Sfust{    19.097}{    15.768}{    15.801}{    15.554}{    15.613}{     8.613}
\SgenE{dd-ls4    }{    -52.33}
\Sgent{   180.823}{   185.827}{   181.274}{   181.317}{   181.308}{   181.309}{   181.605}
\SfusE{    -52.34}{    -52.34}{    -52.34}{    -52.33}{    -52.33}{    -52.34}
\Sfust{   198.557}{   193.873}{   193.918}{   181.308}{   181.309}{   194.222}
\SgenE{bca-lap   }{    -48.81}
\Sgent{     0.905}{     0.797}{     0.719}{     0.720}{     0.866}{     0.866}{     0.722}
\SfusE{    -49.11}{    -49.11}{    -49.11}{    -48.81}{    -48.81}{    -49.11}
\Sfust{     0.950}{     0.866}{     0.867}{     0.866}{     0.866}{     0.869}
\SgenE{bca-greedy}{    -51.49}
\Sgent{     0.484}{     0.311}{     0.057}{     0.059}{     0.591}{     0.592}{     0.151}
\SfusE{    -52.34}{    -52.34}{    -52.34}{    -51.49}{    -51.49}{    -52.27}
\Sfust{     2.016}{     0.462}{     0.482}{     0.591}{     0.592}{     0.151}
\SgenE{greedy    }{    -48.39}
\Sgent{     0.177}{     0.609}{     0.054}{     0.056}{     0.589}{     0.592}{     0.147}
\SfusE{    -52.27}{    -50.92}{    -50.92}{    -48.39}{    -48.39}{    -50.97}
\Sfust{     5.110}{     0.144}{     0.152}{     0.589}{     0.592}{     0.871}
\Sendtable

\Sinstance{car5}\Soptimum{-66.25}\\
\Sstarttable
\SgenE{dd-ls0    }{\Soptimal{    -66.25}}
\Sgent{     0.740}{     4.488}{     0.757}{     1.056}{     1.085}{     1.093}{     0.927}
\SfusE{\Soptimal{    -66.25}}{\Soptimal{    -66.25}}{\Soptimal{    -66.25}}{\Soptimal{    -66.25}}{\Soptimal{    -66.25}}{\Soptimal{    -66.25}}
\Sfust{     4.488}{     0.757}{     1.056}{     1.085}{     1.093}{     0.927}
\SgenE{dd-ls3    }{\Soptimal{    -66.25}}
\Sgent{    17.706}{    21.695}{    16.933}{    13.695}{    18.118}{    18.139}{    16.181}
\SfusE{\Soptimal{    -66.25}}{\Soptimal{    -66.25}}{\Soptimal{    -66.25}}{\Soptimal{    -66.25}}{\Soptimal{    -66.25}}{\Soptimal{    -66.25}}
\Sfust{    21.695}{    16.933}{    13.695}{    18.118}{    18.139}{    16.181}
\SgenE{dd-ls4    }{\Soptimal{    -66.25}}
\Sgent{   216.790}{   187.113}{   217.343}{   165.012}{   217.414}{   217.412}{   200.825}
\SfusE{\Soptimal{    -66.25}}{\Soptimal{    -66.25}}{\Soptimal{    -66.25}}{\Soptimal{    -66.25}}{\Soptimal{    -66.25}}{\Soptimal{    -66.25}}
\Sfust{   187.113}{   217.343}{   165.012}{   217.414}{   217.412}{   200.825}
\SgenE{bca-lap   }{\Soptimal{    -66.25}}
\Sgent{     0.953}{     0.803}{     0.735}{     0.736}{     0.736}{     0.736}{     0.738}
\SfusE{\Soptimal{    -66.25}}{\Soptimal{    -66.25}}{\Soptimal{    -66.25}}{\Soptimal{    -66.25}}{\Soptimal{    -66.25}}{\Soptimal{    -66.25}}
\Sfust{     0.803}{     0.735}{     0.736}{     0.736}{     0.736}{     0.738}
\SgenE{bca-greedy}{\Soptimal{    -66.25}}
\Sgent{     2.549}{     2.715}{     0.651}{     0.658}{     1.385}{     1.396}{     0.318}
\SfusE{\Soptimal{    -66.25}}{\Soptimal{    -66.25}}{\Soptimal{    -66.25}}{\Soptimal{    -66.25}}{\Soptimal{    -66.25}}{\Soptimal{    -66.25}}
\Sfust{     2.715}{     0.651}{     0.658}{     1.385}{     1.396}{     0.318}
\SgenE{greedy    }{    -55.23}
\Sgent{     0.197}{     1.628}{     0.284}{     0.315}{     0.641}{     0.637}{     0.395}
\SfusE{    -62.81}{    -62.27}{    -62.27}{    -55.23}{    -55.23}{    -62.44}
\Sfust{    15.642}{     0.555}{     0.613}{     0.641}{     0.637}{     0.749}
\Sendtable

\Sinstance{car6}\Soptimum{-83.78}\\
\Sstarttable
\SgenE{dd-ls0    }{\Soptimal{    -83.78}}
\Sgent{     0.305}{     1.159}{     0.111}{     0.145}{     0.423}{     0.433}{     0.242}
\SfusE{\Soptimal{    -83.78}}{\Soptimal{    -83.78}}{\Soptimal{    -83.78}}{\Soptimal{    -83.78}}{\Soptimal{    -83.78}}{\Soptimal{    -83.78}}
\Sfust{     1.159}{     0.111}{     0.145}{     0.423}{     0.433}{     0.242}
\SgenE{dd-ls3    }{\Soptimal{    -83.78}}
\Sgent{    10.159}{     3.531}{     8.455}{     8.464}{    10.327}{    10.331}{     4.405}
\SfusE{\Soptimal{    -83.78}}{\Soptimal{    -83.78}}{\Soptimal{    -83.78}}{\Soptimal{    -83.78}}{\Soptimal{    -83.78}}{\Soptimal{    -83.78}}
\Sfust{     3.531}{     8.455}{     8.464}{    10.327}{    10.331}{     4.405}
\SgenE{dd-ls4    }{\Soptimal{    -83.78}}
\Sgent{    97.725}{    56.181}{    54.707}{    54.711}{    98.152}{    98.148}{    54.786}
\SfusE{\Soptimal{    -83.78}}{\Soptimal{    -83.78}}{\Soptimal{    -83.78}}{\Soptimal{    -83.78}}{\Soptimal{    -83.78}}{\Soptimal{    -83.78}}
\Sfust{    56.181}{    54.707}{    54.711}{    98.152}{    98.148}{    54.786}
\SgenE{bca-lap   }{    -83.10}
\Sgent{     0.370}{     0.242}{     0.202}{     0.201}{     0.202}{     0.201}{     0.202}
\SfusE{    -83.10}{    -83.10}{    -83.10}{    -83.10}{    -83.10}{    -83.10}
\Sfust{     0.242}{     0.202}{     0.201}{     0.202}{     0.201}{     0.202}
\SgenE{bca-greedy}{\Soptimal{    -83.78}}
\Sgent{     1.192}{     0.308}{     0.113}{     0.114}{     0.502}{     0.504}{     0.118}
\SfusE{\Soptimal{    -83.78}}{\Soptimal{    -83.78}}{\Soptimal{    -83.78}}{\Soptimal{    -83.78}}{\Soptimal{    -83.78}}{\Soptimal{    -83.78}}
\Sfust{     0.308}{     0.113}{     0.114}{     0.502}{     0.504}{     0.118}
\SgenE{greedy    }{    -70.39}
\Sgent{     0.132}{     1.624}{     0.083}{     0.091}{     0.432}{     0.417}{     0.369}
\SfusE{    -83.02}{    -83.02}{    -83.02}{    -70.39}{    -70.39}{    -80.55}
\Sfust{    12.093}{     0.691}{     0.744}{     0.432}{     0.417}{     0.944}
\Sendtable

\Sinstance{car7}\Soptimum{-81.95}\\
\Sstarttable
\SgenE{dd-ls0    }{\Soptimal{    -81.95}}
\Sgent{     0.032}{     0.199}{     0.036}{     0.038}{     0.050}{     0.050}{     0.042}
\SfusE{\Soptimal{    -81.95}}{\Soptimal{    -81.95}}{\Soptimal{    -81.95}}{\Soptimal{    -81.95}}{\Soptimal{    -81.95}}{\Soptimal{    -81.95}}
\Sfust{     0.199}{     0.036}{     0.038}{     0.050}{     0.050}{     0.042}
\SgenE{dd-ls3    }{\Soptimal{    -81.95}}
\Sgent{     0.703}{     0.189}{     0.191}{     0.086}{     0.743}{     0.742}{     0.085}
\SfusE{\Soptimal{    -81.95}}{\Soptimal{    -81.95}}{\Soptimal{    -81.95}}{\Soptimal{    -81.95}}{\Soptimal{    -81.95}}{\Soptimal{    -81.95}}
\Sfust{     0.189}{     0.191}{     0.086}{     0.743}{     0.742}{     0.085}
\SgenE{dd-ls4    }{\Soptimal{    -81.95}}
\Sgent{     3.896}{     0.472}{     0.339}{     0.340}{     3.961}{     3.961}{     0.342}
\SfusE{\Soptimal{    -81.95}}{\Soptimal{    -81.95}}{\Soptimal{    -81.95}}{\Soptimal{    -81.95}}{\Soptimal{    -81.95}}{\Soptimal{    -81.95}}
\Sfust{     0.472}{     0.339}{     0.340}{     3.961}{     3.961}{     0.342}
\SgenE{bca-lap   }{\Soptimal{    -81.95}}
\Sgent{    13.281}{     0.278}{     0.231}{     0.231}{     0.231}{     0.231}{     0.233}
\SfusE{\Soptimal{    -81.95}}{\Soptimal{    -81.95}}{\Soptimal{    -81.95}}{\Soptimal{    -81.95}}{\Soptimal{    -81.95}}{\Soptimal{    -81.95}}
\Sfust{     0.278}{     0.231}{     0.231}{     0.231}{     0.231}{     0.233}
\SgenE{bca-greedy}{\Soptimal{    -81.95}}
\Sgent{     0.944}{     0.088}{     0.031}{     0.031}{     0.030}{     0.031}{     0.032}
\SfusE{\Soptimal{    -81.95}}{\Soptimal{    -81.95}}{\Soptimal{    -81.95}}{\Soptimal{    -81.95}}{\Soptimal{    -81.95}}{\Soptimal{    -81.95}}
\Sfust{     0.088}{     0.031}{     0.031}{     0.030}{     0.031}{     0.032}
\SgenE{greedy    }{\Soptimal{    -81.95}}
\Sgent{     0.109}{     2.382}{     0.125}{     0.117}{     0.354}{     0.358}{     0.163}
\SfusE{\Soptimal{    -81.95}}{\Soptimal{    -81.95}}{\Soptimal{    -81.95}}{\Soptimal{    -81.95}}{\Soptimal{    -81.95}}{\Soptimal{    -81.95}}
\Sfust{     2.382}{     0.125}{     0.117}{     0.354}{     0.358}{     0.163}
\Sendtable

\Sinstance{car8}\Soptimum{-44.00}\\
\Sstarttable
\SgenE{dd-ls0    }{\Soptimal{    -44.00}}
\Sgent{     0.156}{     1.245}{     0.154}{     0.162}{     0.282}{     0.278}{     0.280}
\SfusE{\Soptimal{    -44.00}}{\Soptimal{    -44.00}}{\Soptimal{    -44.00}}{\Soptimal{    -44.00}}{\Soptimal{    -44.00}}{\Soptimal{    -44.00}}
\Sfust{     1.245}{     0.154}{     0.162}{     0.282}{     0.278}{     0.280}
\SgenE{dd-ls3    }{\Soptimal{    -44.00}}
\Sgent{     2.136}{     1.230}{     2.023}{     0.926}{     2.284}{     2.288}{     2.122}
\SfusE{\Soptimal{    -44.00}}{\Soptimal{    -44.00}}{\Soptimal{    -44.00}}{\Soptimal{    -44.00}}{\Soptimal{    -44.00}}{\Soptimal{    -44.00}}
\Sfust{     1.230}{     2.023}{     0.926}{     2.284}{     2.288}{     2.122}
\SgenE{dd-ls4    }{\Soptimal{    -44.00}}
\Sgent{    28.144}{     9.161}{    20.333}{    20.339}{    28.338}{    28.337}{    23.300}
\SfusE{\Soptimal{    -44.00}}{\Soptimal{    -44.00}}{\Soptimal{    -44.00}}{\Soptimal{    -44.00}}{\Soptimal{    -44.00}}{\Soptimal{    -44.00}}
\Sfust{     9.161}{    20.333}{    20.339}{    28.338}{    28.337}{    23.300}
\SgenE{bca-lap   }{    -43.33}
\Sgent{     0.328}{     0.381}{     0.333}{     0.333}{     0.334}{     0.333}{     0.335}
\SfusE{    -43.56}{    -43.56}{    -43.56}{    -43.33}{    -43.33}{    -43.56}
\Sfust{     0.451}{     0.397}{     0.398}{     0.334}{     0.333}{     0.400}
\SgenE{bca-greedy}{    -43.53}
\Sgent{     1.039}{     7.162}{     1.460}{     1.466}{     1.329}{     1.333}{     1.649}
\SfusE{\Soptimal{    -44.00}}{\Soptimal{    -44.00}}{\Soptimal{    -44.00}}{    -43.53}{    -43.53}{\Soptimal{    -44.00}}
\Sfust{     7.223}{     1.474}{     1.481}{     1.329}{     1.333}{     1.670}
\SgenE{greedy    }{    -40.33}
\Sgent{     0.009}{     1.002}{     0.015}{     0.016}{     0.032}{     0.031}{     0.019}
\SfusE{\Soptimal{    -44.00}}{\Soptimal{    -44.00}}{\Soptimal{    -44.00}}{    -40.33}{    -40.33}{\Soptimal{    -44.00}}
\Sfust{     6.886}{     0.456}{     0.484}{     0.032}{     0.031}{     0.587}
\Sendtable

\Sinstance{car9}\Soptimum{-62.40}\\
\Sstarttable
\SgenE{dd-ls0    }{\Soptimal{    -62.40}}
\Sgent{     0.004}{     0.146}{     0.006}{     0.006}{     0.008}{     0.008}{     0.007}
\SfusE{\Soptimal{    -62.40}}{\Soptimal{    -62.40}}{\Soptimal{    -62.40}}{\Soptimal{    -62.40}}{\Soptimal{    -62.40}}{\Soptimal{    -62.40}}
\Sfust{     0.146}{     0.006}{     0.006}{     0.008}{     0.008}{     0.007}
\SgenE{dd-ls3    }{\Soptimal{    -62.40}}
\Sgent{     0.060}{     0.092}{     0.022}{     0.018}{     0.071}{     0.072}{     0.026}
\SfusE{\Soptimal{    -62.40}}{\Soptimal{    -62.40}}{\Soptimal{    -62.40}}{\Soptimal{    -62.40}}{\Soptimal{    -62.40}}{\Soptimal{    -62.40}}
\Sfust{     0.092}{     0.022}{     0.018}{     0.071}{     0.072}{     0.026}
\SgenE{dd-ls4    }{\Soptimal{    -62.40}}
\Sgent{     0.323}{     0.100}{     0.047}{     0.047}{     0.340}{     0.340}{     0.068}
\SfusE{\Soptimal{    -62.40}}{\Soptimal{    -62.40}}{\Soptimal{    -62.40}}{\Soptimal{    -62.40}}{\Soptimal{    -62.40}}{\Soptimal{    -62.40}}
\Sfust{     0.100}{     0.047}{     0.047}{     0.340}{     0.340}{     0.068}
\SgenE{bca-lap   }{\Soptimal{    -62.40}}
\Sgent{     5.510}{     0.126}{     0.094}{     0.094}{     0.094}{     0.094}{     0.094}
\SfusE{\Soptimal{    -62.40}}{\Soptimal{    -62.40}}{\Soptimal{    -62.40}}{\Soptimal{    -62.40}}{\Soptimal{    -62.40}}{\Soptimal{    -62.40}}
\Sfust{     0.126}{     0.094}{     0.094}{     0.094}{     0.094}{     0.094}
\SgenE{bca-greedy}{\Soptimal{    -62.40}}
\Sgent{     0.362}{     0.041}{     0.006}{     0.006}{     0.007}{     0.007}{     0.006}
\SfusE{\Soptimal{    -62.40}}{\Soptimal{    -62.40}}{\Soptimal{    -62.40}}{\Soptimal{    -62.40}}{\Soptimal{    -62.40}}{\Soptimal{    -62.40}}
\Sfust{     0.041}{     0.006}{     0.006}{     0.007}{     0.007}{     0.006}
\SgenE{greedy    }{\Soptimal{    -62.40}}
\Sgent{     0.020}{     0.082}{     0.005}{     0.005}{     0.067}{     0.065}{     0.007}
\SfusE{\Soptimal{    -62.40}}{\Soptimal{    -62.40}}{\Soptimal{    -62.40}}{\Soptimal{    -62.40}}{\Soptimal{    -62.40}}{\Soptimal{    -62.40}}
\Sfust{     0.082}{     0.005}{     0.005}{     0.067}{     0.065}{     0.007}
\Sendtable

\Sinstance{car10}\Soptimum{-57.62}\\
\Sstarttable
\SgenE{dd-ls0    }{\Soptimal{    -57.62}}
\Sgent{     0.063}{     0.124}{     0.026}{     0.026}{     0.110}{     0.112}{     0.032}
\SfusE{\Soptimal{    -57.62}}{\Soptimal{    -57.62}}{\Soptimal{    -57.62}}{\Soptimal{    -57.62}}{\Soptimal{    -57.62}}{\Soptimal{    -57.62}}
\Sfust{     0.124}{     0.026}{     0.026}{     0.110}{     0.112}{     0.032}
\SgenE{dd-ls3    }{\Soptimal{    -57.62}}
\Sgent{     0.802}{     0.448}{     0.192}{     0.094}{     0.859}{     0.861}{     0.335}
\SfusE{\Soptimal{    -57.62}}{\Soptimal{    -57.62}}{\Soptimal{    -57.62}}{\Soptimal{    -57.62}}{\Soptimal{    -57.62}}{\Soptimal{    -57.62}}
\Sfust{     0.448}{     0.192}{     0.094}{     0.859}{     0.861}{     0.335}
\SgenE{dd-ls4    }{\Soptimal{    -57.62}}
\Sgent{     8.442}{     0.559}{     0.408}{     0.409}{     8.544}{     8.545}{     0.683}
\SfusE{\Soptimal{    -57.62}}{\Soptimal{    -57.62}}{\Soptimal{    -57.62}}{\Soptimal{    -57.62}}{\Soptimal{    -57.62}}{\Soptimal{    -57.62}}
\Sfust{     0.559}{     0.408}{     0.409}{     8.544}{     8.545}{     0.683}
\SgenE{bca-lap   }{\Soptimal{    -57.62}}
\Sgent{    15.613}{     0.179}{     0.143}{     0.143}{     0.143}{     0.142}{     0.143}
\SfusE{\Soptimal{    -57.62}}{\Soptimal{    -57.62}}{\Soptimal{    -57.62}}{\Soptimal{    -57.62}}{\Soptimal{    -57.62}}{\Soptimal{    -57.62}}
\Sfust{     0.179}{     0.143}{     0.143}{     0.143}{     0.142}{     0.143}
\SgenE{bca-greedy}{\Soptimal{    -57.62}}
\Sgent{     0.974}{     0.165}{     0.012}{     0.013}{     0.029}{     0.030}{     0.018}
\SfusE{\Soptimal{    -57.62}}{\Soptimal{    -57.62}}{\Soptimal{    -57.62}}{\Soptimal{    -57.62}}{\Soptimal{    -57.62}}{\Soptimal{    -57.62}}
\Sfust{     0.165}{     0.012}{     0.013}{     0.029}{     0.030}{     0.018}
\SgenE{greedy    }{    -51.98}
\Sgent{     0.102}{     2.744}{     0.101}{     0.106}{     0.327}{     0.339}{     0.260}
\SfusE{\Soptimal{    -57.62}}{\Soptimal{    -57.62}}{\Soptimal{    -57.62}}{    -51.98}{    -51.98}{\Soptimal{    -57.62}}
\Sfust{     7.689}{     0.303}{     0.319}{     0.327}{     0.339}{     0.631}
\Sendtable

\Sinstance{car11}\Soptimum{-63.06}\\
\Sstarttable
\SgenE{dd-ls0    }{\Soptimal{    -63.06}}
\Sgent{     0.015}{     0.156}{     0.009}{     0.015}{     0.026}{     0.025}{     0.011}
\SfusE{\Soptimal{    -63.06}}{\Soptimal{    -63.06}}{\Soptimal{    -63.06}}{\Soptimal{    -63.06}}{\Soptimal{    -63.06}}{\Soptimal{    -63.06}}
\Sfust{     0.156}{     0.009}{     0.015}{     0.026}{     0.025}{     0.011}
\SgenE{dd-ls3    }{\Soptimal{    -63.06}}
\Sgent{     0.273}{     0.236}{     0.103}{     0.106}{     0.299}{     0.299}{     0.088}
\SfusE{\Soptimal{    -63.06}}{\Soptimal{    -63.06}}{\Soptimal{    -63.06}}{\Soptimal{    -63.06}}{\Soptimal{    -63.06}}{\Soptimal{    -63.06}}
\Sfust{     0.236}{     0.103}{     0.106}{     0.299}{     0.299}{     0.088}
\SgenE{dd-ls4    }{\Soptimal{    -63.06}}
\Sgent{     1.774}{     0.348}{     0.165}{     0.148}{     1.818}{     1.818}{     0.167}
\SfusE{\Soptimal{    -63.06}}{\Soptimal{    -63.06}}{\Soptimal{    -63.06}}{\Soptimal{    -63.06}}{\Soptimal{    -63.06}}{\Soptimal{    -63.06}}
\Sfust{     0.348}{     0.165}{     0.148}{     1.818}{     1.818}{     0.167}
\SgenE{bca-lap   }{\Soptimal{    -63.06}}
\Sgent{    20.064}{     0.092}{     0.060}{     0.060}{     0.060}{     0.060}{     0.060}
\SfusE{\Soptimal{    -63.06}}{\Soptimal{    -63.06}}{\Soptimal{    -63.06}}{\Soptimal{    -63.06}}{\Soptimal{    -63.06}}{\Soptimal{    -63.06}}
\Sfust{     0.092}{     0.060}{     0.060}{     0.060}{     0.060}{     0.060}
\SgenE{bca-greedy}{\Soptimal{    -63.06}}
\Sgent{     1.143}{     0.039}{     0.006}{     0.006}{     0.006}{     0.006}{     0.007}
\SfusE{\Soptimal{    -63.06}}{\Soptimal{    -63.06}}{\Soptimal{    -63.06}}{\Soptimal{    -63.06}}{\Soptimal{    -63.06}}{\Soptimal{    -63.06}}
\Sfust{     0.039}{     0.006}{     0.006}{     0.006}{     0.006}{     0.007}
\SgenE{greedy    }{    -61.90}
\Sgent{     0.078}{     2.412}{     0.113}{     0.130}{     0.254}{     0.242}{     0.150}
\SfusE{\Soptimal{    -63.06}}{\Soptimal{    -63.06}}{\Soptimal{    -63.06}}{    -61.90}{    -61.90}{\Soptimal{    -63.06}}
\Sfust{     2.744}{     0.129}{     0.148}{     0.254}{     0.242}{     0.170}
\Sendtable

\Sinstance{car12}\Soptimum{-57.36}\\
\Sstarttable
\SgenE{dd-ls0    }{\Soptimal{    -57.36}}
\Sgent{     0.062}{     0.310}{     0.033}{     0.058}{     0.107}{     0.107}{     0.120}
\SfusE{\Soptimal{    -57.36}}{\Soptimal{    -57.36}}{\Soptimal{    -57.36}}{\Soptimal{    -57.36}}{\Soptimal{    -57.36}}{\Soptimal{    -57.36}}
\Sfust{     0.310}{     0.033}{     0.058}{     0.107}{     0.107}{     0.120}
\SgenE{dd-ls3    }{\Soptimal{    -57.36}}
\Sgent{     0.710}{     0.743}{     0.612}{     0.378}{     0.759}{     0.764}{     0.433}
\SfusE{\Soptimal{    -57.36}}{\Soptimal{    -57.36}}{\Soptimal{    -57.36}}{\Soptimal{    -57.36}}{\Soptimal{    -57.36}}{\Soptimal{    -57.36}}
\Sfust{     0.743}{     0.612}{     0.378}{     0.759}{     0.764}{     0.433}
\SgenE{dd-ls4    }{\Soptimal{    -57.36}}
\Sgent{     7.125}{     0.839}{     2.056}{     0.668}{     7.214}{     7.215}{     0.671}
\SfusE{\Soptimal{    -57.36}}{\Soptimal{    -57.36}}{\Soptimal{    -57.36}}{\Soptimal{    -57.36}}{\Soptimal{    -57.36}}{\Soptimal{    -57.36}}
\Sfust{     0.839}{     2.056}{     0.668}{     7.214}{     7.215}{     0.671}
\SgenE{bca-lap   }{\Soptimal{    -57.36}}
\Sgent{    11.461}{     0.821}{     0.751}{     0.751}{     0.751}{     0.752}{     0.754}
\SfusE{\Soptimal{    -57.36}}{\Soptimal{    -57.36}}{\Soptimal{    -57.36}}{\Soptimal{    -57.36}}{\Soptimal{    -57.36}}{\Soptimal{    -57.36}}
\Sfust{     0.821}{     0.751}{     0.751}{     0.751}{     0.752}{     0.754}
\SgenE{bca-greedy}{\Soptimal{    -57.36}}
\Sgent{     0.651}{     0.068}{     0.008}{     0.008}{     0.019}{     0.019}{     0.009}
\SfusE{\Soptimal{    -57.36}}{\Soptimal{    -57.36}}{\Soptimal{    -57.36}}{\Soptimal{    -57.36}}{\Soptimal{    -57.36}}{\Soptimal{    -57.36}}
\Sfust{     0.068}{     0.008}{     0.008}{     0.019}{     0.019}{     0.009}
\SgenE{greedy    }{    -50.86}
\Sgent{     0.177}{     4.152}{     0.221}{     0.233}{     0.557}{     0.566}{     0.283}
\SfusE{\Soptimal{    -57.36}}{\Soptimal{    -57.36}}{\Soptimal{    -57.36}}{    -50.86}{    -50.86}{    -54.89}
\Sfust{     9.597}{     0.502}{     0.530}{     0.557}{     0.566}{     0.692}
\Sendtable

\Sinstance{car13}\Soptimum{-72.15}\\
\Sstarttable
\SgenE{dd-ls0    }{\Soptimal{    -72.15}}
\Sgent{     0.057}{     0.266}{     0.020}{     0.030}{     0.093}{     0.092}{     0.023}
\SfusE{\Soptimal{    -72.15}}{\Soptimal{    -72.15}}{\Soptimal{    -72.15}}{\Soptimal{    -72.15}}{\Soptimal{    -72.15}}{\Soptimal{    -72.15}}
\Sfust{     0.266}{     0.020}{     0.030}{     0.093}{     0.092}{     0.023}
\SgenE{dd-ls3    }{\Soptimal{    -72.15}}
\Sgent{     0.686}{     0.266}{     0.479}{     0.487}{     0.732}{     0.739}{     0.131}
\SfusE{\Soptimal{    -72.15}}{\Soptimal{    -72.15}}{\Soptimal{    -72.15}}{\Soptimal{    -72.15}}{\Soptimal{    -72.15}}{\Soptimal{    -72.15}}
\Sfust{     0.266}{     0.479}{     0.487}{     0.732}{     0.739}{     0.131}
\SgenE{dd-ls4    }{\Soptimal{    -72.15}}
\Sgent{     4.502}{     0.561}{     0.588}{     0.255}{     4.578}{     4.579}{     1.129}
\SfusE{\Soptimal{    -72.15}}{\Soptimal{    -72.15}}{\Soptimal{    -72.15}}{\Soptimal{    -72.15}}{\Soptimal{    -72.15}}{\Soptimal{    -72.15}}
\Sfust{     0.561}{     0.588}{     0.255}{     4.578}{     4.579}{     1.129}
\SgenE{bca-lap   }{\Soptimal{    -72.15}}
\Sgent{     0.244}{     0.256}{     0.214}{     0.214}{     0.214}{     0.214}{     0.215}
\SfusE{\Soptimal{    -72.15}}{\Soptimal{    -72.15}}{\Soptimal{    -72.15}}{\Soptimal{    -72.15}}{\Soptimal{    -72.15}}{\Soptimal{    -72.15}}
\Sfust{     0.256}{     0.214}{     0.214}{     0.214}{     0.214}{     0.215}
\SgenE{bca-greedy}{\Soptimal{    -72.15}}
\Sgent{     0.017}{     0.072}{     0.011}{     0.011}{     0.017}{     0.018}{     0.013}
\SfusE{\Soptimal{    -72.15}}{\Soptimal{    -72.15}}{\Soptimal{    -72.15}}{\Soptimal{    -72.15}}{\Soptimal{    -72.15}}{\Soptimal{    -72.15}}
\Sfust{     0.072}{     0.011}{     0.011}{     0.017}{     0.018}{     0.013}
\SgenE{greedy    }{    -68.93}
\Sgent{     0.053}{     2.276}{     0.042}{     0.176}{     0.178}{     0.175}{     0.138}
\SfusE{\Soptimal{    -72.15}}{\Soptimal{    -72.15}}{\Soptimal{    -72.15}}{    -68.93}{    -68.93}{\Soptimal{    -72.15}}
\Sfust{     3.684}{     0.103}{     0.185}{     0.178}{     0.175}{     0.220}
\Sendtable

\Sinstance{car14}\Soptimum{-97.96}\\
\Sstarttable
\SgenE{dd-ls0    }{\Soptimal{    -97.96}}
\Sgent{     0.222}{     1.015}{     0.299}{     0.304}{     0.304}{     0.305}{     0.333}
\SfusE{\Soptimal{    -97.96}}{\Soptimal{    -97.96}}{\Soptimal{    -97.96}}{\Soptimal{    -97.96}}{\Soptimal{    -97.96}}{\Soptimal{    -97.96}}
\Sfust{     1.015}{     0.299}{     0.304}{     0.304}{     0.305}{     0.333}
\SgenE{dd-ls3    }{\Soptimal{    -97.96}}
\Sgent{     4.993}{     6.112}{     5.109}{     4.338}{     5.117}{     5.115}{     5.166}
\SfusE{\Soptimal{    -97.96}}{\Soptimal{    -97.96}}{\Soptimal{    -97.96}}{\Soptimal{    -97.96}}{\Soptimal{    -97.96}}{\Soptimal{    -97.96}}
\Sfust{     6.112}{     5.109}{     4.338}{     5.117}{     5.115}{     5.166}
\SgenE{dd-ls4    }{\Soptimal{    -97.96}}
\Sgent{    58.771}{    41.132}{    59.053}{    59.058}{    59.060}{    59.065}{    59.120}
\SfusE{\Soptimal{    -97.96}}{\Soptimal{    -97.96}}{\Soptimal{    -97.96}}{\Soptimal{    -97.96}}{\Soptimal{    -97.96}}{\Soptimal{    -97.96}}
\Sfust{    41.132}{    59.053}{    59.058}{    59.060}{    59.065}{    59.120}
\SgenE{bca-lap   }{\Soptimal{    -97.96}}
\Sgent{     2.321}{     0.177}{     0.141}{     0.142}{     2.141}{     2.142}{     0.142}
\SfusE{\Soptimal{    -97.96}}{\Soptimal{    -97.96}}{\Soptimal{    -97.96}}{\Soptimal{    -97.96}}{\Soptimal{    -97.96}}{\Soptimal{    -97.96}}
\Sfust{     0.177}{     0.141}{     0.142}{     2.141}{     2.142}{     0.142}
\SgenE{bca-greedy}{\Soptimal{    -97.96}}
\Sgent{     0.718}{     0.890}{     0.514}{     0.517}{     0.516}{     0.516}{     0.536}
\SfusE{\Soptimal{    -97.96}}{\Soptimal{    -97.96}}{\Soptimal{    -97.96}}{\Soptimal{    -97.96}}{\Soptimal{    -97.96}}{\Soptimal{    -97.96}}
\Sfust{     0.890}{     0.514}{     0.517}{     0.516}{     0.516}{     0.536}
\SgenE{greedy    }{    -92.53}
\Sgent{     0.290}{     0.830}{     0.041}{     0.043}{     0.938}{     0.950}{     0.041}
\SfusE{\Soptimal{    -97.96}}{\Soptimal{    -97.96}}{\Soptimal{    -97.96}}{    -92.53}{    -92.53}{\Soptimal{    -97.96}}
\Sfust{    15.996}{     0.891}{     0.948}{     0.938}{     0.950}{     1.159}
\Sendtable

\Sinstance{car15}\Soptimum{-66.89}\\
\Sstarttable
\SgenE{dd-ls0    }{\Soptimal{    -66.89}}
\Sgent{     0.029}{     0.079}{     0.019}{     0.018}{     0.050}{     0.050}{     0.011}
\SfusE{\Soptimal{    -66.89}}{\Soptimal{    -66.89}}{\Soptimal{    -66.89}}{\Soptimal{    -66.89}}{\Soptimal{    -66.89}}{\Soptimal{    -66.89}}
\Sfust{     0.079}{     0.019}{     0.018}{     0.050}{     0.050}{     0.011}
\SgenE{dd-ls3    }{\Soptimal{    -66.89}}
\Sgent{     0.333}{     0.155}{     0.033}{     0.035}{     0.362}{     0.365}{     0.035}
\SfusE{\Soptimal{    -66.89}}{\Soptimal{    -66.89}}{\Soptimal{    -66.89}}{\Soptimal{    -66.89}}{\Soptimal{    -66.89}}{\Soptimal{    -66.89}}
\Sfust{     0.155}{     0.033}{     0.035}{     0.362}{     0.365}{     0.035}
\SgenE{dd-ls4    }{\Soptimal{    -66.89}}
\Sgent{     1.145}{     0.865}{     0.694}{     0.694}{     1.176}{     1.176}{     0.699}
\SfusE{\Soptimal{    -66.89}}{\Soptimal{    -66.89}}{\Soptimal{    -66.89}}{\Soptimal{    -66.89}}{\Soptimal{    -66.89}}{\Soptimal{    -66.89}}
\Sfust{     0.865}{     0.694}{     0.694}{     1.176}{     1.176}{     0.699}
\SgenE{bca-lap   }{\Soptimal{    -66.89}}
\Sgent{    13.721}{     0.162}{     0.127}{     0.127}{     0.127}{     0.127}{     0.127}
\SfusE{\Soptimal{    -66.89}}{\Soptimal{    -66.89}}{\Soptimal{    -66.89}}{\Soptimal{    -66.89}}{\Soptimal{    -66.89}}{\Soptimal{    -66.89}}
\Sfust{     0.162}{     0.127}{     0.127}{     0.127}{     0.127}{     0.127}
\SgenE{bca-greedy}{\Soptimal{    -66.89}}
\Sgent{     0.735}{     0.048}{     0.006}{     0.006}{     0.008}{     0.007}{     0.006}
\SfusE{\Soptimal{    -66.89}}{\Soptimal{    -66.89}}{\Soptimal{    -66.89}}{\Soptimal{    -66.89}}{\Soptimal{    -66.89}}{\Soptimal{    -66.89}}
\Sfust{     0.048}{     0.006}{     0.006}{     0.008}{     0.007}{     0.006}
\SgenE{greedy    }{\Soptimal{    -66.89}}
\Sgent{     0.071}{     0.438}{     0.032}{     0.045}{     0.226}{     0.225}{     0.042}
\SfusE{\Soptimal{    -66.89}}{\Soptimal{    -66.89}}{\Soptimal{    -66.89}}{\Soptimal{    -66.89}}{\Soptimal{    -66.89}}{\Soptimal{    -66.89}}
\Sfust{     0.438}{     0.032}{     0.045}{     0.226}{     0.225}{     0.042}
\Sendtable

\Sinstance{car16}\Soptimum{-68.21}\\
\Sstarttable
\SgenE{dd-ls0    }{\Soptimal{    -68.21}}
\Sgent{     0.036}{     0.111}{     0.014}{     0.014}{     0.059}{     0.060}{     0.016}
\SfusE{\Soptimal{    -68.21}}{\Soptimal{    -68.21}}{\Soptimal{    -68.21}}{\Soptimal{    -68.21}}{\Soptimal{    -68.21}}{\Soptimal{    -68.21}}
\Sfust{     0.111}{     0.014}{     0.014}{     0.059}{     0.060}{     0.016}
\SgenE{dd-ls3    }{\Soptimal{    -68.21}}
\Sgent{     0.798}{     0.366}{     0.060}{     0.134}{     0.851}{     0.853}{     0.067}
\SfusE{\Soptimal{    -68.21}}{\Soptimal{    -68.21}}{\Soptimal{    -68.21}}{\Soptimal{    -68.21}}{\Soptimal{    -68.21}}{\Soptimal{    -68.21}}
\Sfust{     0.366}{     0.060}{     0.134}{     0.851}{     0.853}{     0.067}
\SgenE{dd-ls4    }{\Soptimal{    -68.21}}
\Sgent{     4.843}{     0.223}{     0.246}{     0.246}{     4.915}{     4.916}{     1.662}
\SfusE{\Soptimal{    -68.21}}{\Soptimal{    -68.21}}{\Soptimal{    -68.21}}{\Soptimal{    -68.21}}{\Soptimal{    -68.21}}{\Soptimal{    -68.21}}
\Sfust{     0.223}{     0.246}{     0.246}{     4.915}{     4.916}{     1.662}
\SgenE{bca-lap   }{\Soptimal{    -68.21}}
\Sgent{     5.927}{     0.151}{     0.116}{     0.116}{     0.116}{     0.116}{     0.117}
\SfusE{\Soptimal{    -68.21}}{\Soptimal{    -68.21}}{\Soptimal{    -68.21}}{\Soptimal{    -68.21}}{\Soptimal{    -68.21}}{\Soptimal{    -68.21}}
\Sfust{     0.151}{     0.116}{     0.116}{     0.116}{     0.116}{     0.117}
\SgenE{bca-greedy}{\Soptimal{    -68.21}}
\Sgent{     0.391}{     0.050}{     0.008}{     0.007}{     0.008}{     0.008}{     0.007}
\SfusE{\Soptimal{    -68.21}}{\Soptimal{    -68.21}}{\Soptimal{    -68.21}}{\Soptimal{    -68.21}}{\Soptimal{    -68.21}}{\Soptimal{    -68.21}}
\Sfust{     0.050}{     0.008}{     0.007}{     0.008}{     0.008}{     0.007}
\SgenE{greedy    }{    -64.94}
\Sgent{     0.044}{     0.469}{     0.028}{     0.030}{     0.146}{     0.148}{     0.047}
\SfusE{\Soptimal{    -68.21}}{\Soptimal{    -68.21}}{\Soptimal{    -68.21}}{    -64.94}{    -64.94}{\Soptimal{    -68.21}}
\Sfust{     0.469}{     0.028}{     0.030}{     0.146}{     0.148}{     0.057}
\Sendtable

\Sinstance{car17}\Soptimum{-57.09}\\
\Sstarttable
\SgenE{dd-ls0    }{\Soptimal{    -57.09}}
\Sgent{     0.048}{     0.159}{     0.026}{     0.027}{     0.083}{     0.082}{     0.045}
\SfusE{\Soptimal{    -57.09}}{\Soptimal{    -57.09}}{\Soptimal{    -57.09}}{\Soptimal{    -57.09}}{\Soptimal{    -57.09}}{\Soptimal{    -57.09}}
\Sfust{     0.159}{     0.026}{     0.027}{     0.083}{     0.082}{     0.045}
\SgenE{dd-ls3    }{\Soptimal{    -57.09}}
\Sgent{     1.389}{     0.348}{     0.119}{     0.121}{     1.444}{     1.446}{     0.472}
\SfusE{\Soptimal{    -57.09}}{\Soptimal{    -57.09}}{\Soptimal{    -57.09}}{\Soptimal{    -57.09}}{\Soptimal{    -57.09}}{\Soptimal{    -57.09}}
\Sfust{     0.348}{     0.119}{     0.121}{     1.444}{     1.446}{     0.472}
\SgenE{dd-ls4    }{\Soptimal{    -57.09}}
\Sgent{     9.613}{     1.802}{     1.130}{     1.472}{     9.713}{     9.710}{     2.428}
\SfusE{\Soptimal{    -57.09}}{\Soptimal{    -57.09}}{\Soptimal{    -57.09}}{\Soptimal{    -57.09}}{\Soptimal{    -57.09}}{\Soptimal{    -57.09}}
\Sfust{     1.802}{     1.130}{     1.472}{     9.713}{     9.710}{     2.428}
\SgenE{bca-lap   }{\Soptimal{    -57.09}}
\Sgent{    11.365}{     0.491}{     0.417}{     0.417}{     0.416}{     0.417}{     0.419}
\SfusE{\Soptimal{    -57.09}}{\Soptimal{    -57.09}}{\Soptimal{    -57.09}}{\Soptimal{    -57.09}}{\Soptimal{    -57.09}}{\Soptimal{    -57.09}}
\Sfust{     0.491}{     0.417}{     0.417}{     0.416}{     0.417}{     0.419}
\SgenE{bca-greedy}{\Soptimal{    -57.09}}
\Sgent{     0.596}{     0.123}{     0.029}{     0.029}{     0.030}{     0.030}{     0.031}
\SfusE{\Soptimal{    -57.09}}{\Soptimal{    -57.09}}{\Soptimal{    -57.09}}{\Soptimal{    -57.09}}{\Soptimal{    -57.09}}{\Soptimal{    -57.09}}
\Sfust{     0.123}{     0.029}{     0.029}{     0.030}{     0.030}{     0.031}
\SgenE{greedy    }{\Soptimal{    -57.09}}
\Sgent{     0.123}{     0.347}{     0.021}{     0.023}{     0.391}{     0.396}{     0.029}
\SfusE{\Soptimal{    -57.09}}{\Soptimal{    -57.09}}{\Soptimal{    -57.09}}{\Soptimal{    -57.09}}{\Soptimal{    -57.09}}{\Soptimal{    -57.09}}
\Sfust{     0.347}{     0.021}{     0.023}{     0.391}{     0.396}{     0.029}
\Sendtable

\Sinstance{car18}\Soptimum{-92.18}\\
\Sstarttable
\SgenE{dd-ls0    }{\Soptimal{    -92.18}}
\Sgent{     0.097}{     0.137}{     0.025}{     0.025}{     0.140}{     0.142}{     0.027}
\SfusE{\Soptimal{    -92.18}}{\Soptimal{    -92.18}}{\Soptimal{    -92.18}}{\Soptimal{    -92.18}}{\Soptimal{    -92.18}}{\Soptimal{    -92.18}}
\Sfust{     0.137}{     0.025}{     0.025}{     0.140}{     0.142}{     0.027}
\SgenE{dd-ls3    }{\Soptimal{    -92.18}}
\Sgent{     2.168}{     0.400}{     0.155}{     0.236}{     2.249}{     2.243}{     0.288}
\SfusE{\Soptimal{    -92.18}}{\Soptimal{    -92.18}}{\Soptimal{    -92.18}}{\Soptimal{    -92.18}}{\Soptimal{    -92.18}}{\Soptimal{    -92.18}}
\Sfust{     0.400}{     0.155}{     0.236}{     2.249}{     2.243}{     0.288}
\SgenE{dd-ls4    }{\Soptimal{    -92.18}}
\Sgent{    11.498}{     0.855}{     0.727}{     0.727}{    11.628}{    11.627}{     1.249}
\SfusE{\Soptimal{    -92.18}}{\Soptimal{    -92.18}}{\Soptimal{    -92.18}}{\Soptimal{    -92.18}}{\Soptimal{    -92.18}}{\Soptimal{    -92.18}}
\Sfust{     0.855}{     0.727}{     0.727}{    11.628}{    11.627}{     1.249}
\SgenE{bca-lap   }{\Soptimal{    -92.18}}
\Sgent{    21.160}{     0.159}{     0.125}{     0.125}{     0.125}{     0.125}{     0.126}
\SfusE{\Soptimal{    -92.18}}{\Soptimal{    -92.18}}{\Soptimal{    -92.18}}{\Soptimal{    -92.18}}{\Soptimal{    -92.18}}{\Soptimal{    -92.18}}
\Sfust{     0.159}{     0.125}{     0.125}{     0.125}{     0.125}{     0.126}
\SgenE{bca-greedy}{\Soptimal{    -92.18}}
\Sgent{     1.534}{     0.051}{     0.012}{     0.012}{     0.049}{     0.049}{     0.013}
\SfusE{\Soptimal{    -92.18}}{\Soptimal{    -92.18}}{\Soptimal{    -92.18}}{\Soptimal{    -92.18}}{\Soptimal{    -92.18}}{\Soptimal{    -92.18}}
\Sfust{     0.051}{     0.012}{     0.012}{     0.049}{     0.049}{     0.013}
\SgenE{greedy    }{    -89.92}
\Sgent{     0.069}{     1.183}{     0.051}{     0.058}{     0.225}{     0.221}{     0.072}
\SfusE{\Soptimal{    -92.18}}{\Soptimal{    -92.18}}{\Soptimal{    -92.18}}{    -89.92}{    -89.92}{\Soptimal{    -92.18}}
\Sfust{     3.277}{     0.150}{     0.168}{     0.225}{     0.221}{     0.203}
\Sendtable

\Sinstance{car19}\Soptimum{-115.11}\\
\Sstarttable
\SgenE{dd-ls0    }{\Soptimal{   -115.11}}
\Sgent{     0.209}{     0.418}{     0.101}{     0.104}{     0.276}{     0.275}{     0.111}
\SfusE{\Soptimal{   -115.11}}{\Soptimal{   -115.11}}{\Soptimal{   -115.11}}{\Soptimal{   -115.11}}{\Soptimal{   -115.11}}{\Soptimal{   -115.11}}
\Sfust{     0.418}{     0.101}{     0.104}{     0.276}{     0.275}{     0.111}
\SgenE{dd-ls3    }{\Soptimal{   -115.11}}
\Sgent{     5.996}{     1.495}{     2.337}{     1.899}{     6.141}{     6.130}{     1.388}
\SfusE{\Soptimal{   -115.11}}{\Soptimal{   -115.11}}{\Soptimal{   -115.11}}{\Soptimal{   -115.11}}{\Soptimal{   -115.11}}{\Soptimal{   -115.11}}
\Sfust{     1.495}{     2.337}{     1.899}{     6.141}{     6.130}{     1.388}
\SgenE{dd-ls4    }{\Soptimal{   -115.11}}
\Sgent{    33.063}{     6.958}{     2.591}{     2.593}{    33.375}{    33.375}{     2.305}
\SfusE{\Soptimal{   -115.11}}{\Soptimal{   -115.11}}{\Soptimal{   -115.11}}{\Soptimal{   -115.11}}{\Soptimal{   -115.11}}{\Soptimal{   -115.11}}
\Sfust{     6.958}{     2.591}{     2.593}{    33.375}{    33.375}{     2.305}
\SgenE{bca-lap   }{\Soptimal{   -115.11}}
\Sgent{    76.316}{     0.344}{     0.300}{     0.301}{     0.301}{     0.301}{     0.303}
\SfusE{\Soptimal{   -115.11}}{\Soptimal{   -115.11}}{\Soptimal{   -115.11}}{\Soptimal{   -115.11}}{\Soptimal{   -115.11}}{\Soptimal{   -115.11}}
\Sfust{     0.344}{     0.300}{     0.301}{     0.301}{     0.301}{     0.303}
\SgenE{bca-greedy}{\Soptimal{   -115.11}}
\Sgent{     6.226}{     0.140}{     0.054}{     0.054}{     0.086}{     0.086}{     0.059}
\SfusE{\Soptimal{   -115.11}}{\Soptimal{   -115.11}}{\Soptimal{   -115.11}}{\Soptimal{   -115.11}}{\Soptimal{   -115.11}}{\Soptimal{   -115.11}}
\Sfust{     0.140}{     0.054}{     0.054}{     0.086}{     0.086}{     0.059}
\SgenE{greedy    }{    -92.48}
\Sgent{     0.128}{     1.589}{     0.023}{     0.027}{     0.435}{     0.429}{     0.030}
\SfusE{\Soptimal{   -115.11}}{\Soptimal{   -115.11}}{\Soptimal{   -115.11}}{    -92.48}{    -92.48}{\Soptimal{   -115.11}}
\Sfust{     9.431}{     0.942}{     1.001}{     0.435}{     0.429}{     1.229}
\Sendtable

\Sinstance{car20}\Soptimum{-106.69}\\
\Sstarttable
\SgenE{dd-ls0    }{\Soptimal{   -106.69}}
\Sgent{     0.062}{     0.080}{     0.020}{     0.013}{     0.088}{     0.087}{     0.015}
\SfusE{\Soptimal{   -106.69}}{\Soptimal{   -106.69}}{\Soptimal{   -106.69}}{\Soptimal{   -106.69}}{\Soptimal{   -106.69}}{\Soptimal{   -106.69}}
\Sfust{     0.080}{     0.020}{     0.013}{     0.088}{     0.087}{     0.015}
\SgenE{dd-ls3    }{\Soptimal{   -106.69}}
\Sgent{     1.323}{     0.344}{     0.210}{     0.212}{     1.378}{     1.377}{     0.226}
\SfusE{\Soptimal{   -106.69}}{\Soptimal{   -106.69}}{\Soptimal{   -106.69}}{\Soptimal{   -106.69}}{\Soptimal{   -106.69}}{\Soptimal{   -106.69}}
\Sfust{     0.344}{     0.210}{     0.212}{     1.378}{     1.377}{     0.226}
\SgenE{dd-ls4    }{\Soptimal{   -106.69}}
\Sgent{     6.800}{     0.443}{     0.357}{     0.357}{     6.904}{     6.904}{     0.671}
\SfusE{\Soptimal{   -106.69}}{\Soptimal{   -106.69}}{\Soptimal{   -106.69}}{\Soptimal{   -106.69}}{\Soptimal{   -106.69}}{\Soptimal{   -106.69}}
\Sfust{     0.443}{     0.357}{     0.357}{     6.904}{     6.904}{     0.671}
\SgenE{bca-lap   }{\Soptimal{   -106.69}}
\Sgent{    29.776}{     0.131}{     0.098}{     0.098}{     0.097}{     0.098}{     0.098}
\SfusE{\Soptimal{   -106.69}}{\Soptimal{   -106.69}}{\Soptimal{   -106.69}}{\Soptimal{   -106.69}}{\Soptimal{   -106.69}}{\Soptimal{   -106.69}}
\Sfust{     0.131}{     0.098}{     0.098}{     0.097}{     0.098}{     0.098}
\SgenE{bca-greedy}{\Soptimal{   -106.69}}
\Sgent{     3.402}{     0.066}{     0.016}{     0.016}{     0.023}{     0.023}{     0.021}
\SfusE{\Soptimal{   -106.69}}{\Soptimal{   -106.69}}{\Soptimal{   -106.69}}{\Soptimal{   -106.69}}{\Soptimal{   -106.69}}{\Soptimal{   -106.69}}
\Sfust{     0.066}{     0.016}{     0.016}{     0.023}{     0.023}{     0.021}
\SgenE{greedy    }{\Soptimal{   -106.69}}
\Sgent{     0.229}{     3.277}{     0.159}{     0.170}{     0.699}{     0.702}{     0.219}
\SfusE{\Soptimal{   -106.69}}{\Soptimal{   -106.69}}{\Soptimal{   -106.69}}{\Soptimal{   -106.69}}{\Soptimal{   -106.69}}{\Soptimal{   -106.69}}
\Sfust{     3.277}{     0.159}{     0.170}{     0.699}{     0.702}{     0.219}
\Sendtable

\Sinstance{car21}\Soptimum{-94.55}\\
\Sstarttable
\SgenE{dd-ls0    }{\Soptimal{    -94.55}}
\Sgent{     0.164}{     0.312}{     0.076}{     0.078}{     0.229}{     0.230}{     0.116}
\SfusE{\Soptimal{    -94.55}}{\Soptimal{    -94.55}}{\Soptimal{    -94.55}}{\Soptimal{    -94.55}}{\Soptimal{    -94.55}}{\Soptimal{    -94.55}}
\Sfust{     0.312}{     0.076}{     0.078}{     0.229}{     0.230}{     0.116}
\SgenE{dd-ls3    }{\Soptimal{    -94.55}}
\Sgent{     4.651}{     1.526}{     0.368}{     2.339}{     4.776}{     4.777}{     0.583}
\SfusE{\Soptimal{    -94.55}}{\Soptimal{    -94.55}}{\Soptimal{    -94.55}}{\Soptimal{    -94.55}}{\Soptimal{    -94.55}}{\Soptimal{    -94.55}}
\Sfust{     1.526}{     0.368}{     2.339}{     4.776}{     4.777}{     0.583}
\SgenE{dd-ls4    }{\Soptimal{    -94.55}}
\Sgent{    20.363}{     1.554}{     2.365}{     3.380}{    20.548}{    20.549}{     2.914}
\SfusE{\Soptimal{    -94.55}}{\Soptimal{    -94.55}}{\Soptimal{    -94.55}}{\Soptimal{    -94.55}}{\Soptimal{    -94.55}}{\Soptimal{    -94.55}}
\Sfust{     1.554}{     2.365}{     3.380}{    20.548}{    20.549}{     2.914}
\SgenE{bca-lap   }{\Soptimal{    -94.55}}
\Sgent{    56.521}{     0.192}{     0.153}{     0.153}{     0.153}{     0.154}{     0.154}
\SfusE{\Soptimal{    -94.55}}{\Soptimal{    -94.55}}{\Soptimal{    -94.55}}{\Soptimal{    -94.55}}{\Soptimal{    -94.55}}{\Soptimal{    -94.55}}
\Sfust{     0.192}{     0.153}{     0.153}{     0.153}{     0.154}{     0.154}
\SgenE{bca-greedy}{\Soptimal{    -94.55}}
\Sgent{     4.336}{     0.124}{     0.034}{     0.034}{     0.059}{     0.059}{     0.036}
\SfusE{\Soptimal{    -94.55}}{\Soptimal{    -94.55}}{\Soptimal{    -94.55}}{\Soptimal{    -94.55}}{\Soptimal{    -94.55}}{\Soptimal{    -94.55}}
\Sfust{     0.124}{     0.034}{     0.034}{     0.059}{     0.059}{     0.036}
\SgenE{greedy    }{    -85.05}
\Sgent{     0.207}{     1.916}{     0.145}{     0.153}{     0.642}{     0.668}{     0.146}
\SfusE{\Soptimal{    -94.55}}{\Soptimal{    -94.55}}{\Soptimal{    -94.55}}{    -85.05}{    -85.05}{\Soptimal{    -94.55}}
\Sfust{    13.492}{     0.196}{     0.207}{     0.642}{     0.668}{     0.942}
\Sendtable

\Sinstance{car22}\Soptimum{-55.58}\\
\Sstarttable
\SgenE{dd-ls0    }{\Soptimal{    -55.58}}
\Sgent{     0.036}{     0.095}{     0.013}{     0.013}{     0.063}{     0.062}{     0.016}
\SfusE{\Soptimal{    -55.58}}{\Soptimal{    -55.58}}{\Soptimal{    -55.58}}{\Soptimal{    -55.58}}{\Soptimal{    -55.58}}{\Soptimal{    -55.58}}
\Sfust{     0.095}{     0.013}{     0.013}{     0.063}{     0.062}{     0.016}
\SgenE{dd-ls3    }{\Soptimal{    -55.58}}
\Sgent{     0.304}{     0.092}{     0.028}{     0.023}{     0.336}{     0.337}{     0.025}
\SfusE{\Soptimal{    -55.58}}{\Soptimal{    -55.58}}{\Soptimal{    -55.58}}{\Soptimal{    -55.58}}{\Soptimal{    -55.58}}{\Soptimal{    -55.58}}
\Sfust{     0.092}{     0.028}{     0.023}{     0.336}{     0.337}{     0.025}
\SgenE{dd-ls4    }{\Soptimal{    -55.58}}
\Sgent{     1.329}{     0.237}{     0.336}{     0.140}{     1.364}{     1.365}{     0.141}
\SfusE{\Soptimal{    -55.58}}{\Soptimal{    -55.58}}{\Soptimal{    -55.58}}{\Soptimal{    -55.58}}{\Soptimal{    -55.58}}{\Soptimal{    -55.58}}
\Sfust{     0.237}{     0.336}{     0.140}{     1.364}{     1.365}{     0.141}
\SgenE{bca-lap   }{\Soptimal{    -55.58}}
\Sgent{     6.768}{     0.094}{     0.058}{     0.058}{     0.058}{     0.058}{     0.058}
\SfusE{\Soptimal{    -55.58}}{\Soptimal{    -55.58}}{\Soptimal{    -55.58}}{\Soptimal{    -55.58}}{\Soptimal{    -55.58}}{\Soptimal{    -55.58}}
\Sfust{     0.094}{     0.058}{     0.058}{     0.058}{     0.058}{     0.058}
\SgenE{bca-greedy}{\Soptimal{    -55.58}}
\Sgent{     0.392}{     0.046}{     0.006}{     0.006}{     0.008}{     0.007}{     0.011}
\SfusE{\Soptimal{    -55.58}}{\Soptimal{    -55.58}}{\Soptimal{    -55.58}}{\Soptimal{    -55.58}}{\Soptimal{    -55.58}}{\Soptimal{    -55.58}}
\Sfust{     0.046}{     0.006}{     0.006}{     0.008}{     0.007}{     0.011}
\SgenE{greedy    }{\Soptimal{    -55.58}}
\Sgent{     0.040}{     0.606}{     0.031}{     0.036}{     0.134}{     0.129}{     0.025}
\SfusE{\Soptimal{    -55.58}}{\Soptimal{    -55.58}}{\Soptimal{    -55.58}}{\Soptimal{    -55.58}}{\Soptimal{    -55.58}}{\Soptimal{    -55.58}}
\Sfust{     0.606}{     0.031}{     0.036}{     0.134}{     0.129}{     0.025}
\Sendtable

\Sinstance{car23}\Soptimum{-70.20}\\
\Sstarttable
\SgenE{dd-ls0    }{\Soptimal{    -70.20}}
\Sgent{     0.101}{     0.326}{     0.053}{     0.053}{     0.158}{     0.157}{     0.121}
\SfusE{\Soptimal{    -70.20}}{\Soptimal{    -70.20}}{\Soptimal{    -70.20}}{\Soptimal{    -70.20}}{\Soptimal{    -70.20}}{\Soptimal{    -70.20}}
\Sfust{     0.326}{     0.053}{     0.053}{     0.158}{     0.157}{     0.121}
\SgenE{dd-ls3    }{\Soptimal{    -70.20}}
\Sgent{     1.438}{     0.987}{     0.574}{     0.577}{     1.511}{     1.508}{     0.867}
\SfusE{\Soptimal{    -70.20}}{\Soptimal{    -70.20}}{\Soptimal{    -70.20}}{\Soptimal{    -70.20}}{\Soptimal{    -70.20}}{\Soptimal{    -70.20}}
\Sfust{     0.987}{     0.574}{     0.577}{     1.511}{     1.508}{     0.867}
\SgenE{dd-ls4    }{\Soptimal{    -70.20}}
\Sgent{    16.127}{     1.735}{     2.461}{     2.461}{    16.270}{    16.269}{     0.750}
\SfusE{\Soptimal{    -70.20}}{\Soptimal{    -70.20}}{\Soptimal{    -70.20}}{\Soptimal{    -70.20}}{\Soptimal{    -70.20}}{\Soptimal{    -70.20}}
\Sfust{     1.735}{     2.461}{     2.461}{    16.270}{    16.269}{     0.750}
\SgenE{bca-lap   }{\Soptimal{    -70.20}}
\Sgent{    29.696}{     0.217}{     0.151}{     0.151}{     0.152}{     0.152}{     0.152}
\SfusE{\Soptimal{    -70.20}}{\Soptimal{    -70.20}}{\Soptimal{    -70.20}}{\Soptimal{    -70.20}}{\Soptimal{    -70.20}}{\Soptimal{    -70.20}}
\Sfust{     0.217}{     0.151}{     0.151}{     0.152}{     0.152}{     0.152}
\SgenE{bca-greedy}{\Soptimal{    -70.20}}
\Sgent{     1.889}{     0.054}{     0.010}{     0.010}{     0.029}{     0.029}{     0.011}
\SfusE{\Soptimal{    -70.20}}{\Soptimal{    -70.20}}{\Soptimal{    -70.20}}{\Soptimal{    -70.20}}{\Soptimal{    -70.20}}{\Soptimal{    -70.20}}
\Sfust{     0.054}{     0.010}{     0.010}{     0.029}{     0.029}{     0.011}
\SgenE{greedy    }{    -59.84}
\Sgent{     0.088}{     1.327}{     0.018}{     0.020}{     0.295}{     0.288}{     0.103}
\SfusE{\Soptimal{    -70.20}}{\Soptimal{    -70.20}}{\Soptimal{    -70.20}}{    -59.84}{    -59.84}{\Soptimal{    -70.20}}
\Sfust{     8.343}{     0.525}{     0.572}{     0.295}{     0.288}{     0.600}
\Sendtable

\Sinstance{car24}\Soptimum{-64.58}\\
\Sstarttable
\SgenE{dd-ls0    }{\Soptimal{    -64.58}}
\Sgent{     0.095}{     0.212}{     0.033}{     0.034}{     0.151}{     0.153}{     0.039}
\SfusE{\Soptimal{    -64.58}}{\Soptimal{    -64.58}}{\Soptimal{    -64.58}}{\Soptimal{    -64.58}}{\Soptimal{    -64.58}}{\Soptimal{    -64.58}}
\Sfust{     0.212}{     0.033}{     0.034}{     0.151}{     0.153}{     0.039}
\SgenE{dd-ls3    }{\Soptimal{    -64.58}}
\Sgent{     1.631}{     0.513}{     0.270}{     0.272}{     1.714}{     1.712}{     0.503}
\SfusE{\Soptimal{    -64.58}}{\Soptimal{    -64.58}}{\Soptimal{    -64.58}}{\Soptimal{    -64.58}}{\Soptimal{    -64.58}}{\Soptimal{    -64.58}}
\Sfust{     0.513}{     0.270}{     0.272}{     1.714}{     1.712}{     0.503}
\SgenE{dd-ls4    }{\Soptimal{    -64.58}}
\Sgent{     7.923}{     1.524}{     1.265}{     1.266}{     8.025}{     8.022}{     3.200}
\SfusE{\Soptimal{    -64.58}}{\Soptimal{    -64.58}}{\Soptimal{    -64.58}}{\Soptimal{    -64.58}}{\Soptimal{    -64.58}}{\Soptimal{    -64.58}}
\Sfust{     1.524}{     1.265}{     1.266}{     8.025}{     8.022}{     3.200}
\SgenE{bca-lap   }{\Soptimal{    -64.58}}
\Sgent{     6.717}{     2.953}{     2.768}{     2.769}{     2.769}{     2.768}{     2.779}
\SfusE{\Soptimal{    -64.58}}{\Soptimal{    -64.58}}{\Soptimal{    -64.58}}{\Soptimal{    -64.58}}{\Soptimal{    -64.58}}{\Soptimal{    -64.58}}
\Sfust{     2.953}{     2.768}{     2.769}{     2.769}{     2.768}{     2.779}
\SgenE{bca-greedy}{\Soptimal{    -64.58}}
\Sgent{     0.391}{     0.270}{     0.104}{     0.109}{     0.106}{     0.105}{     0.116}
\SfusE{\Soptimal{    -64.58}}{\Soptimal{    -64.58}}{\Soptimal{    -64.58}}{\Soptimal{    -64.58}}{\Soptimal{    -64.58}}{\Soptimal{    -64.58}}
\Sfust{     0.270}{     0.104}{     0.109}{     0.106}{     0.105}{     0.116}
\SgenE{greedy    }{    -60.07}
\Sgent{     0.193}{     0.711}{     0.043}{     0.046}{     0.621}{     0.611}{     0.082}
\SfusE{\Soptimal{    -64.58}}{\Soptimal{    -64.58}}{\Soptimal{    -64.58}}{    -60.07}{    -60.07}{\Soptimal{    -64.58}}
\Sfust{     1.094}{     0.066}{     0.070}{     0.621}{     0.611}{     0.359}
\Sendtable

\Sinstance{car25}\Soptimum{-34.19}\\
\Sstarttable
\SgenE{dd-ls0    }{\Soptimal{    -34.19}}
\Sgent{     0.047}{     0.532}{     0.056}{     0.058}{     0.097}{     0.097}{     0.050}
\SfusE{\Soptimal{    -34.19}}{\Soptimal{    -34.19}}{\Soptimal{    -34.19}}{\Soptimal{    -34.19}}{\Soptimal{    -34.19}}{\Soptimal{    -34.19}}
\Sfust{     0.532}{     0.056}{     0.058}{     0.097}{     0.097}{     0.050}
\SgenE{dd-ls3    }{\Soptimal{    -34.19}}
\Sgent{     0.326}{     0.608}{     0.185}{     0.188}{     0.376}{     0.377}{     0.205}
\SfusE{\Soptimal{    -34.19}}{\Soptimal{    -34.19}}{\Soptimal{    -34.19}}{\Soptimal{    -34.19}}{\Soptimal{    -34.19}}{\Soptimal{    -34.19}}
\Sfust{     0.608}{     0.185}{     0.188}{     0.376}{     0.377}{     0.205}
\SgenE{dd-ls4    }{\Soptimal{    -34.19}}
\Sgent{     2.320}{     1.086}{     0.783}{     0.784}{     2.370}{     2.368}{     0.218}
\SfusE{\Soptimal{    -34.19}}{\Soptimal{    -34.19}}{\Soptimal{    -34.19}}{\Soptimal{    -34.19}}{\Soptimal{    -34.19}}{\Soptimal{    -34.19}}
\Sfust{     1.086}{     0.783}{     0.784}{     2.370}{     2.368}{     0.218}
\SgenE{bca-lap   }{\Soptimal{    -34.19}}
\Sgent{     0.060}{     0.169}{     0.069}{     0.069}{     0.069}{     0.068}{     0.069}
\SfusE{\Soptimal{    -34.19}}{\Soptimal{    -34.19}}{\Soptimal{    -34.19}}{\Soptimal{    -34.19}}{\Soptimal{    -34.19}}{\Soptimal{    -34.19}}
\Sfust{     0.169}{     0.069}{     0.069}{     0.069}{     0.068}{     0.069}
\SgenE{bca-greedy}{\Soptimal{    -34.19}}
\Sgent{     0.708}{     0.084}{     0.006}{     0.006}{     0.018}{     0.019}{     0.006}
\SfusE{\Soptimal{    -34.19}}{\Soptimal{    -34.19}}{\Soptimal{    -34.19}}{\Soptimal{    -34.19}}{\Soptimal{    -34.19}}{\Soptimal{    -34.19}}
\Sfust{     0.084}{     0.006}{     0.006}{     0.018}{     0.019}{     0.006}
\SgenE{greedy    }{    -30.07}
\Sgent{     0.111}{     0.138}{     0.005}{     0.006}{     0.354}{     0.365}{     0.006}
\SfusE{\Soptimal{    -34.19}}{\Soptimal{    -34.19}}{\Soptimal{    -34.19}}{    -30.07}{    -30.07}{\Soptimal{    -34.19}}
\Sfust{     3.586}{     0.204}{     0.215}{     0.354}{     0.365}{     0.272}
\Sendtable

\Sinstance{car26}\Soptimum{-59.92}\\
\Sstarttable
\SgenE{dd-ls0    }{\Soptimal{    -59.92}}
\Sgent{     0.066}{     0.284}{     0.046}{     0.047}{     0.111}{     0.111}{     0.123}
\SfusE{\Soptimal{    -59.92}}{\Soptimal{    -59.92}}{\Soptimal{    -59.92}}{\Soptimal{    -59.92}}{\Soptimal{    -59.92}}{\Soptimal{    -59.92}}
\Sfust{     0.284}{     0.046}{     0.047}{     0.111}{     0.111}{     0.123}
\SgenE{dd-ls3    }{\Soptimal{    -59.92}}
\Sgent{     1.060}{     1.494}{     0.853}{     0.858}{     1.123}{     1.126}{     0.838}
\SfusE{\Soptimal{    -59.92}}{\Soptimal{    -59.92}}{\Soptimal{    -59.92}}{\Soptimal{    -59.92}}{\Soptimal{    -59.92}}{\Soptimal{    -59.92}}
\Sfust{     1.494}{     0.853}{     0.858}{     1.123}{     1.126}{     0.838}
\SgenE{dd-ls4    }{\Soptimal{    -59.92}}
\Sgent{     6.481}{     4.200}{     3.725}{     3.727}{     6.567}{     6.568}{     6.610}
\SfusE{\Soptimal{    -59.92}}{\Soptimal{    -59.92}}{\Soptimal{    -59.92}}{\Soptimal{    -59.92}}{\Soptimal{    -59.92}}{\Soptimal{    -59.92}}
\Sfust{     4.200}{     3.725}{     3.727}{     6.567}{     6.568}{     6.610}
\SgenE{bca-lap   }{\Soptimal{    -59.92}}
\Sgent{    11.076}{     1.104}{     1.019}{     1.020}{     1.019}{     1.019}{     1.024}
\SfusE{\Soptimal{    -59.92}}{\Soptimal{    -59.92}}{\Soptimal{    -59.92}}{\Soptimal{    -59.92}}{\Soptimal{    -59.92}}{\Soptimal{    -59.92}}
\Sfust{     1.104}{     1.019}{     1.020}{     1.019}{     1.019}{     1.024}
\SgenE{bca-greedy}{\Soptimal{    -59.92}}
\Sgent{     0.657}{     0.109}{     0.033}{     0.033}{     0.099}{     0.097}{     0.045}
\SfusE{\Soptimal{    -59.92}}{\Soptimal{    -59.92}}{\Soptimal{    -59.92}}{\Soptimal{    -59.92}}{\Soptimal{    -59.92}}{\Soptimal{    -59.92}}
\Sfust{     0.109}{     0.033}{     0.033}{     0.099}{     0.097}{     0.045}
\SgenE{greedy    }{    -58.36}
\Sgent{     0.015}{     0.323}{     0.023}{     0.024}{     0.049}{     0.051}{     0.056}
\SfusE{\Soptimal{    -59.92}}{\Soptimal{    -59.92}}{\Soptimal{    -59.92}}{    -58.36}{    -58.36}{\Soptimal{    -59.92}}
\Sfust{     0.887}{     0.059}{     0.063}{     0.049}{     0.051}{     0.086}
\Sendtable

\Sinstance{car27}\Soptimum{-67.95}\\
\Sstarttable
\SgenE{dd-ls0    }{\Soptimal{    -67.95}}
\Sgent{     0.067}{     0.444}{     0.082}{     0.085}{     0.108}{     0.106}{     0.096}
\SfusE{\Soptimal{    -67.95}}{\Soptimal{    -67.95}}{\Soptimal{    -67.95}}{\Soptimal{    -67.95}}{\Soptimal{    -67.95}}{\Soptimal{    -67.95}}
\Sfust{     0.444}{     0.082}{     0.085}{     0.108}{     0.106}{     0.096}
\SgenE{dd-ls3    }{\Soptimal{    -67.95}}
\Sgent{     0.910}{     0.698}{     0.877}{     0.879}{     0.975}{     0.965}{     0.896}
\SfusE{\Soptimal{    -67.95}}{\Soptimal{    -67.95}}{\Soptimal{    -67.95}}{\Soptimal{    -67.95}}{\Soptimal{    -67.95}}{\Soptimal{    -67.95}}
\Sfust{     0.698}{     0.877}{     0.879}{     0.975}{     0.965}{     0.896}
\SgenE{dd-ls4    }{\Soptimal{    -67.95}}
\Sgent{     4.748}{     4.510}{     0.487}{     0.487}{     4.821}{     4.822}{     4.844}
\SfusE{\Soptimal{    -67.95}}{\Soptimal{    -67.95}}{\Soptimal{    -67.95}}{\Soptimal{    -67.95}}{\Soptimal{    -67.95}}{\Soptimal{    -67.95}}
\Sfust{     4.510}{     0.487}{     0.487}{     4.821}{     4.822}{     4.844}
\SgenE{bca-lap   }{\Soptimal{    -67.95}}
\Sgent{     4.481}{     0.266}{     0.222}{     0.222}{     0.222}{     0.223}{     0.224}
\SfusE{\Soptimal{    -67.95}}{\Soptimal{    -67.95}}{\Soptimal{    -67.95}}{\Soptimal{    -67.95}}{\Soptimal{    -67.95}}{\Soptimal{    -67.95}}
\Sfust{     0.266}{     0.222}{     0.222}{     0.222}{     0.223}{     0.224}
\SgenE{bca-greedy}{\Soptimal{    -67.95}}
\Sgent{     0.285}{     0.160}{     0.014}{     0.014}{     0.020}{     0.019}{     0.015}
\SfusE{\Soptimal{    -67.95}}{\Soptimal{    -67.95}}{\Soptimal{    -67.95}}{\Soptimal{    -67.95}}{\Soptimal{    -67.95}}{\Soptimal{    -67.95}}
\Sfust{     0.160}{     0.014}{     0.014}{     0.020}{     0.019}{     0.015}
\SgenE{greedy    }{    -66.00}
\Sgent{     0.080}{     6.099}{     0.282}{     0.307}{     0.262}{     0.255}{     0.366}
\SfusE{\Soptimal{    -67.95}}{\Soptimal{    -67.95}}{\Soptimal{    -67.95}}{    -66.00}{    -66.00}{\Soptimal{    -67.95}}
\Sfust{     6.546}{     0.304}{     0.331}{     0.262}{     0.255}{     0.394}
\Sendtable

\Sinstance{car28}\Soptimum{-75.70}\\
\Sstarttable
\SgenE{dd-ls0    }{\Soptimal{    -75.70}}
\Sgent{     1.332}{     6.301}{     1.520}{     1.553}{     1.828}{     1.812}{     1.909}
\SfusE{\Soptimal{    -75.70}}{\Soptimal{    -75.70}}{\Soptimal{    -75.70}}{\Soptimal{    -75.70}}{\Soptimal{    -75.70}}{\Soptimal{    -75.70}}
\Sfust{     6.301}{     1.520}{     1.553}{     1.828}{     1.812}{     1.909}
\SgenE{dd-ls3    }{\Soptimal{    -75.70}}
\Sgent{    20.054}{    20.837}{    20.379}{    20.394}{    20.454}{    20.449}{    20.909}
\SfusE{\Soptimal{    -75.70}}{\Soptimal{    -75.70}}{\Soptimal{    -75.70}}{\Soptimal{    -75.70}}{\Soptimal{    -75.70}}{\Soptimal{    -75.70}}
\Sfust{    20.837}{    20.379}{    20.394}{    20.454}{    20.449}{    20.909}
\SgenE{dd-ls4    }{\Soptimal{    -75.70}}
\Sgent{   572.740}{   310.976}{   331.099}{   331.116}{   573.676}{   573.686}{   271.501}
\SfusE{\Soptimal{    -75.70}}{\Soptimal{    -75.70}}{\Soptimal{    -75.70}}{\Soptimal{    -75.70}}{\Soptimal{    -75.70}}{\Soptimal{    -75.70}}
\Sfust{   310.976}{   331.099}{   331.116}{   573.676}{   573.686}{   271.501}
\SgenE{bca-lap   }{    -72.63}
\Sgent{     3.898}{     0.849}{     0.782}{     0.783}{     3.631}{     3.632}{     0.785}
\SfusE{    -73.85}{    -73.85}{    -73.85}{    -72.63}{    -72.63}{    -72.93}
\Sfust{     1.873}{     1.746}{     1.748}{     3.631}{     3.632}{     0.785}
\SgenE{bca-greedy}{    -71.58}
\Sgent{     0.146}{     0.285}{     0.069}{     0.069}{     0.175}{     0.174}{     0.207}
\SfusE{\Soptimal{    -75.70}}{\Soptimal{    -75.70}}{\Soptimal{    -75.70}}{    -71.58}{    -71.58}{    -75.51}
\Sfust{     3.641}{     1.082}{     1.092}{     0.175}{     0.174}{     1.722}
\SgenE{greedy    }{    -65.80}
\Sgent{     0.264}{     9.054}{     0.369}{     0.385}{     0.866}{     0.859}{     0.605}
\SfusE{    -73.70}{    -72.36}{    -72.36}{    -65.80}{    -65.80}{    -69.31}
\Sfust{    17.659}{     0.880}{     0.932}{     0.866}{     0.859}{     1.149}
\Sendtable

\Sinstance{car29}\Soptimum{-84.31}\\
\Sstarttable
\SgenE{dd-ls0    }{\Soptimal{    -84.31}}
\Sgent{     0.134}{     0.865}{     0.053}{     0.054}{     0.191}{     0.193}{     0.211}
\SfusE{\Soptimal{    -84.31}}{\Soptimal{    -84.31}}{\Soptimal{    -84.31}}{\Soptimal{    -84.31}}{\Soptimal{    -84.31}}{\Soptimal{    -84.31}}
\Sfust{     0.865}{     0.053}{     0.054}{     0.191}{     0.193}{     0.211}
\SgenE{dd-ls3    }{\Soptimal{    -84.31}}
\Sgent{     3.904}{     3.026}{     2.440}{     0.327}{     4.031}{     4.013}{     3.394}
\SfusE{\Soptimal{    -84.31}}{\Soptimal{    -84.31}}{\Soptimal{    -84.31}}{\Soptimal{    -84.31}}{\Soptimal{    -84.31}}{\Soptimal{    -84.31}}
\Sfust{     3.026}{     2.440}{     0.327}{     4.031}{     4.013}{     3.394}
\SgenE{dd-ls4    }{\Soptimal{    -84.31}}
\Sgent{    39.826}{     9.998}{     2.836}{     2.838}{    40.072}{    40.073}{     3.973}
\SfusE{\Soptimal{    -84.31}}{\Soptimal{    -84.31}}{\Soptimal{    -84.31}}{\Soptimal{    -84.31}}{\Soptimal{    -84.31}}{\Soptimal{    -84.31}}
\Sfust{     9.998}{     2.836}{     2.838}{    40.072}{    40.073}{     3.973}
\SgenE{bca-lap   }{\Soptimal{    -84.31}}
\Sgent{    10.448}{    10.924}{    10.366}{    10.369}{    10.369}{    10.367}{    10.404}
\SfusE{\Soptimal{    -84.31}}{\Soptimal{    -84.31}}{\Soptimal{    -84.31}}{\Soptimal{    -84.31}}{\Soptimal{    -84.31}}{\Soptimal{    -84.31}}
\Sfust{    10.924}{    10.366}{    10.369}{    10.369}{    10.367}{    10.404}
\SgenE{bca-greedy}{\Soptimal{    -84.31}}
\Sgent{     0.210}{     0.111}{     0.038}{     0.038}{     0.137}{     0.134}{     0.041}
\SfusE{\Soptimal{    -84.31}}{\Soptimal{    -84.31}}{\Soptimal{    -84.31}}{\Soptimal{    -84.31}}{\Soptimal{    -84.31}}{\Soptimal{    -84.31}}
\Sfust{     0.111}{     0.038}{     0.038}{     0.137}{     0.134}{     0.041}
\SgenE{greedy    }{    -76.00}
\Sgent{     0.262}{     0.592}{     0.031}{     0.034}{     0.850}{     0.822}{     0.042}
\SfusE{\Soptimal{    -84.31}}{\Soptimal{    -84.31}}{\Soptimal{    -84.31}}{    -76.00}{    -76.00}{\Soptimal{    -84.31}}
\Sfust{     0.993}{     0.053}{     0.058}{     0.850}{     0.822}{     0.071}
\Sendtable

\Sinstance{car30}\Soptimum{-58.85}\\
\Sstarttable
\SgenE{dd-ls0    }{\Soptimal{    -58.85}}
\Sgent{     0.059}{     0.493}{     0.077}{     0.080}{     0.100}{     0.101}{     0.089}
\SfusE{\Soptimal{    -58.85}}{\Soptimal{    -58.85}}{\Soptimal{    -58.85}}{\Soptimal{    -58.85}}{\Soptimal{    -58.85}}{\Soptimal{    -58.85}}
\Sfust{     0.493}{     0.077}{     0.080}{     0.100}{     0.101}{     0.089}
\SgenE{dd-ls3    }{\Soptimal{    -58.85}}
\Sgent{     0.972}{     1.506}{     0.848}{     0.850}{     1.050}{     1.036}{     0.822}
\SfusE{\Soptimal{    -58.85}}{\Soptimal{    -58.85}}{\Soptimal{    -58.85}}{\Soptimal{    -58.85}}{\Soptimal{    -58.85}}{\Soptimal{    -58.85}}
\Sfust{     1.506}{     0.848}{     0.850}{     1.050}{     1.036}{     0.822}
\SgenE{dd-ls4    }{\Soptimal{    -58.85}}
\Sgent{     5.785}{     1.395}{     3.239}{     3.242}{     5.865}{     5.865}{     5.878}
\SfusE{\Soptimal{    -58.85}}{\Soptimal{    -58.85}}{\Soptimal{    -58.85}}{\Soptimal{    -58.85}}{\Soptimal{    -58.85}}{\Soptimal{    -58.85}}
\Sfust{     1.395}{     3.239}{     3.242}{     5.865}{     5.865}{     5.878}
\SgenE{bca-lap   }{    -58.34}
\Sgent{     0.795}{     0.407}{     0.356}{     0.356}{     0.356}{     0.356}{     0.358}
\SfusE{\Soptimal{    -58.85}}{\Soptimal{    -58.85}}{\Soptimal{    -58.85}}{    -58.34}{    -58.34}{\Soptimal{    -58.85}}
\Sfust{     0.407}{     0.356}{     0.356}{     0.356}{     0.356}{     0.358}
\SgenE{bca-greedy}{\Soptimal{    -58.85}}
\Sgent{     1.466}{     0.069}{     0.016}{     0.017}{     0.017}{     0.018}{     0.017}
\SfusE{\Soptimal{    -58.85}}{\Soptimal{    -58.85}}{\Soptimal{    -58.85}}{\Soptimal{    -58.85}}{\Soptimal{    -58.85}}{\Soptimal{    -58.85}}
\Sfust{     0.069}{     0.016}{     0.017}{     0.017}{     0.018}{     0.017}
\SgenE{greedy    }{    -56.65}
\Sgent{     0.142}{     0.448}{     0.024}{     0.027}{     0.446}{     0.450}{     0.138}
\SfusE{\Soptimal{    -58.85}}{\Soptimal{    -58.85}}{\Soptimal{    -58.85}}{    -56.65}{    -56.65}{\Soptimal{    -58.85}}
\Sfust{     6.991}{     0.111}{     0.121}{     0.446}{     0.450}{     0.522}
\Sendtable

\Ssubsection{motor}

\Snrinst{20}

\Snriter{1500}

\Sinstance{motor1}\Soptimum{-94.18}\\
\Sstarttable
\SgenE{dd-ls0    }{\Soptimal{    -94.18}}
\Sgent{     0.367}{     0.414}{     0.085}{     0.088}{     0.497}{     0.502}{     0.100}
\SfusE{\Soptimal{    -94.18}}{\Soptimal{    -94.18}}{\Soptimal{    -94.18}}{\Soptimal{    -94.18}}{\Soptimal{    -94.18}}{\Soptimal{    -94.18}}
\Sfust{     0.414}{     0.085}{     0.088}{     0.497}{     0.502}{     0.100}
\SgenE{dd-ls3    }{\Soptimal{    -94.18}}
\Sgent{     7.162}{     4.322}{     3.336}{     5.467}{     7.317}{     7.319}{     3.465}
\SfusE{\Soptimal{    -94.18}}{\Soptimal{    -94.18}}{\Soptimal{    -94.18}}{\Soptimal{    -94.18}}{\Soptimal{    -94.18}}{\Soptimal{    -94.18}}
\Sfust{     4.322}{     3.336}{     5.467}{     7.317}{     7.319}{     3.465}
\SgenE{dd-ls4    }{\Soptimal{    -94.18}}
\Sgent{    55.664}{    44.008}{    28.969}{    44.550}{    56.006}{    56.004}{    42.815}
\SfusE{\Soptimal{    -94.18}}{\Soptimal{    -94.18}}{\Soptimal{    -94.18}}{\Soptimal{    -94.18}}{\Soptimal{    -94.18}}{\Soptimal{    -94.18}}
\Sfust{    44.008}{    28.969}{    44.550}{    56.006}{    56.004}{    42.815}
\SgenE{bca-lap   }{\Soptimal{    -94.18}}
\Sgent{     2.342}{     0.211}{     0.154}{     0.155}{     0.154}{     0.154}{     0.155}
\SfusE{\Soptimal{    -94.18}}{\Soptimal{    -94.18}}{\Soptimal{    -94.18}}{\Soptimal{    -94.18}}{\Soptimal{    -94.18}}{\Soptimal{    -94.18}}
\Sfust{     0.211}{     0.154}{     0.155}{     0.154}{     0.154}{     0.155}
\SgenE{bca-greedy}{\Soptimal{    -94.18}}
\Sgent{     3.315}{     0.174}{     0.071}{     0.072}{     0.177}{     0.177}{     0.079}
\SfusE{\Soptimal{    -94.18}}{\Soptimal{    -94.18}}{\Soptimal{    -94.18}}{\Soptimal{    -94.18}}{\Soptimal{    -94.18}}{\Soptimal{    -94.18}}
\Sfust{     0.174}{     0.071}{     0.072}{     0.177}{     0.177}{     0.079}
\SgenE{greedy    }{    -85.38}
\Sgent{     0.145}{     2.376}{     0.142}{     0.150}{     0.484}{     0.482}{     0.193}
\SfusE{    -91.09}{    -91.09}{    -91.09}{    -85.38}{    -85.38}{    -91.09}
\Sfust{     8.181}{     0.428}{     0.453}{     0.484}{     0.482}{     0.577}
\Sendtable

\Sinstance{motor2}\Soptimum{-94.93}\\
\Sstarttable
\SgenE{dd-ls0    }{\Soptimal{    -94.93}}
\Sgent{     0.027}{     0.046}{     0.006}{     0.007}{     0.042}{     0.043}{     0.008}
\SfusE{\Soptimal{    -94.93}}{\Soptimal{    -94.93}}{\Soptimal{    -94.93}}{\Soptimal{    -94.93}}{\Soptimal{    -94.93}}{\Soptimal{    -94.93}}
\Sfust{     0.046}{     0.006}{     0.007}{     0.042}{     0.043}{     0.008}
\SgenE{dd-ls3    }{\Soptimal{    -94.93}}
\Sgent{     0.301}{     0.066}{     0.019}{     0.020}{     0.327}{     0.327}{     0.020}
\SfusE{\Soptimal{    -94.93}}{\Soptimal{    -94.93}}{\Soptimal{    -94.93}}{\Soptimal{    -94.93}}{\Soptimal{    -94.93}}{\Soptimal{    -94.93}}
\Sfust{     0.066}{     0.019}{     0.020}{     0.327}{     0.327}{     0.020}
\SgenE{dd-ls4    }{\Soptimal{    -94.93}}
\Sgent{     0.079}{     0.127}{     0.091}{     0.091}{     0.092}{     0.091}{     0.083}
\SfusE{\Soptimal{    -94.93}}{\Soptimal{    -94.93}}{\Soptimal{    -94.93}}{\Soptimal{    -94.93}}{\Soptimal{    -94.93}}{\Soptimal{    -94.93}}
\Sfust{     0.127}{     0.091}{     0.091}{     0.092}{     0.091}{     0.083}
\SgenE{bca-lap   }{\Soptimal{    -94.93}}
\Sgent{    17.194}{     0.131}{     0.089}{     0.089}{     0.089}{     0.089}{     0.089}
\SfusE{\Soptimal{    -94.93}}{\Soptimal{    -94.93}}{\Soptimal{    -94.93}}{\Soptimal{    -94.93}}{\Soptimal{    -94.93}}{\Soptimal{    -94.93}}
\Sfust{     0.131}{     0.089}{     0.089}{     0.089}{     0.089}{     0.089}
\SgenE{bca-greedy}{\Soptimal{    -94.93}}
\Sgent{     1.702}{     0.042}{     0.008}{     0.008}{     0.009}{     0.009}{     0.009}
\SfusE{\Soptimal{    -94.93}}{\Soptimal{    -94.93}}{\Soptimal{    -94.93}}{\Soptimal{    -94.93}}{\Soptimal{    -94.93}}{\Soptimal{    -94.93}}
\Sfust{     0.042}{     0.008}{     0.008}{     0.009}{     0.009}{     0.009}
\SgenE{greedy    }{\Soptimal{    -94.93}}
\Sgent{     0.008}{     0.108}{     0.009}{     0.010}{     0.024}{     0.026}{     0.012}
\SfusE{\Soptimal{    -94.93}}{\Soptimal{    -94.93}}{\Soptimal{    -94.93}}{\Soptimal{    -94.93}}{\Soptimal{    -94.93}}{\Soptimal{    -94.93}}
\Sfust{     0.108}{     0.009}{     0.010}{     0.024}{     0.026}{     0.012}
\Sendtable

\Sinstance{motor3}\Soptimum{-45.37}\\
\Sstarttable
\SgenE{dd-ls0    }{\Soptimal{    -45.37}}
\Sgent{     0.015}{     0.056}{     0.005}{     0.005}{     0.031}{     0.032}{     0.006}
\SfusE{\Soptimal{    -45.37}}{\Soptimal{    -45.37}}{\Soptimal{    -45.37}}{\Soptimal{    -45.37}}{\Soptimal{    -45.37}}{\Soptimal{    -45.37}}
\Sfust{     0.056}{     0.005}{     0.005}{     0.031}{     0.032}{     0.006}
\SgenE{dd-ls3    }{\Soptimal{    -45.37}}
\Sgent{     0.093}{     0.096}{     0.021}{     0.021}{     0.109}{     0.109}{     0.033}
\SfusE{\Soptimal{    -45.37}}{\Soptimal{    -45.37}}{\Soptimal{    -45.37}}{\Soptimal{    -45.37}}{\Soptimal{    -45.37}}{\Soptimal{    -45.37}}
\Sfust{     0.096}{     0.021}{     0.021}{     0.109}{     0.109}{     0.033}
\SgenE{dd-ls4    }{\Soptimal{    -45.37}}
\Sgent{     0.671}{     0.109}{     0.052}{     0.052}{     0.696}{     0.696}{     0.050}
\SfusE{\Soptimal{    -45.37}}{\Soptimal{    -45.37}}{\Soptimal{    -45.37}}{\Soptimal{    -45.37}}{\Soptimal{    -45.37}}{\Soptimal{    -45.37}}
\Sfust{     0.109}{     0.052}{     0.052}{     0.696}{     0.696}{     0.050}
\SgenE{bca-lap   }{\Soptimal{    -45.37}}
\Sgent{     8.183}{     0.505}{     0.404}{     0.404}{     0.403}{     0.404}{     0.406}
\SfusE{\Soptimal{    -45.37}}{\Soptimal{    -45.37}}{\Soptimal{    -45.37}}{\Soptimal{    -45.37}}{\Soptimal{    -45.37}}{\Soptimal{    -45.37}}
\Sfust{     0.505}{     0.404}{     0.404}{     0.403}{     0.404}{     0.406}
\SgenE{bca-greedy}{\Soptimal{    -45.37}}
\Sgent{     0.445}{     0.066}{     0.010}{     0.010}{     0.011}{     0.010}{     0.011}
\SfusE{\Soptimal{    -45.37}}{\Soptimal{    -45.37}}{\Soptimal{    -45.37}}{\Soptimal{    -45.37}}{\Soptimal{    -45.37}}{\Soptimal{    -45.37}}
\Sfust{     0.066}{     0.010}{     0.010}{     0.011}{     0.010}{     0.011}
\SgenE{greedy    }{\Soptimal{    -45.37}}
\Sgent{     0.016}{     0.111}{     0.008}{     0.008}{     0.052}{     0.053}{     0.009}
\SfusE{\Soptimal{    -45.37}}{\Soptimal{    -45.37}}{\Soptimal{    -45.37}}{\Soptimal{    -45.37}}{\Soptimal{    -45.37}}{\Soptimal{    -45.37}}
\Sfust{     0.111}{     0.008}{     0.008}{     0.052}{     0.053}{     0.009}
\Sendtable

\Sinstance{motor4}\Soptimum{-27.50}\\
\Sstarttable
\SgenE{dd-ls0    }{\Soptimal{    -27.50}}
\Sgent{     0.001}{     0.055}{     0.004}{     0.004}{     0.005}{     0.005}{     0.005}
\SfusE{\Soptimal{    -27.50}}{\Soptimal{    -27.50}}{\Soptimal{    -27.50}}{\Soptimal{    -27.50}}{\Soptimal{    -27.50}}{\Soptimal{    -27.50}}
\Sfust{     0.055}{     0.004}{     0.004}{     0.005}{     0.005}{     0.005}
\SgenE{dd-ls3    }{\Soptimal{    -27.50}}
\Sgent{     0.006}{     0.053}{     0.005}{     0.007}{     0.013}{     0.010}{     0.005}
\SfusE{\Soptimal{    -27.50}}{\Soptimal{    -27.50}}{\Soptimal{    -27.50}}{\Soptimal{    -27.50}}{\Soptimal{    -27.50}}{\Soptimal{    -27.50}}
\Sfust{     0.053}{     0.005}{     0.007}{     0.013}{     0.010}{     0.005}
\SgenE{dd-ls4    }{\Soptimal{    -27.50}}
\Sgent{     0.018}{     0.073}{     0.019}{     0.021}{     0.023}{     0.023}{     0.020}
\SfusE{\Soptimal{    -27.50}}{\Soptimal{    -27.50}}{\Soptimal{    -27.50}}{\Soptimal{    -27.50}}{\Soptimal{    -27.50}}{\Soptimal{    -27.50}}
\Sfust{     0.073}{     0.019}{     0.021}{     0.023}{     0.023}{     0.020}
\SgenE{bca-lap   }{\Soptimal{    -27.50}}
\Sgent{     3.483}{     0.098}{     0.043}{     0.044}{     0.043}{     0.043}{     0.043}
\SfusE{\Soptimal{    -27.50}}{\Soptimal{    -27.50}}{\Soptimal{    -27.50}}{\Soptimal{    -27.50}}{\Soptimal{    -27.50}}{\Soptimal{    -27.50}}
\Sfust{     0.098}{     0.043}{     0.044}{     0.043}{     0.043}{     0.043}
\SgenE{bca-greedy}{\Soptimal{    -27.50}}
\Sgent{     0.197}{     0.058}{     0.007}{     0.008}{     0.009}{     0.008}{     0.009}
\SfusE{\Soptimal{    -27.50}}{\Soptimal{    -27.50}}{\Soptimal{    -27.50}}{\Soptimal{    -27.50}}{\Soptimal{    -27.50}}{\Soptimal{    -27.50}}
\Sfust{     0.058}{     0.007}{     0.008}{     0.009}{     0.008}{     0.009}
\SgenE{greedy    }{    -26.50}
\Sgent{     0.036}{     0.078}{     0.005}{     0.005}{     0.116}{     0.114}{     0.006}
\SfusE{\Soptimal{    -27.50}}{\Soptimal{    -27.50}}{\Soptimal{    -27.50}}{    -26.50}{    -26.50}{\Soptimal{    -27.50}}
\Sfust{     0.106}{     0.009}{     0.009}{     0.116}{     0.114}{     0.009}
\Sendtable

\Sinstance{motor5}\Soptimum{-29.91}\\
\Sstarttable
\SgenE{dd-ls0    }{\Soptimal{    -29.91}}
\Sgent{     0.000}{     0.034}{     0.003}{     0.003}{     0.004}{     0.003}{     0.002}
\SfusE{\Soptimal{    -29.91}}{\Soptimal{    -29.91}}{\Soptimal{    -29.91}}{\Soptimal{    -29.91}}{\Soptimal{    -29.91}}{\Soptimal{    -29.91}}
\Sfust{     0.034}{     0.003}{     0.003}{     0.004}{     0.003}{     0.002}
\SgenE{dd-ls3    }{\Soptimal{    -29.91}}
\Sgent{     0.006}{     0.044}{     0.004}{     0.005}{     0.011}{     0.012}{     0.006}
\SfusE{\Soptimal{    -29.91}}{\Soptimal{    -29.91}}{\Soptimal{    -29.91}}{\Soptimal{    -29.91}}{\Soptimal{    -29.91}}{\Soptimal{    -29.91}}
\Sfust{     0.044}{     0.004}{     0.005}{     0.011}{     0.012}{     0.006}
\SgenE{dd-ls4    }{\Soptimal{    -29.91}}
\Sgent{     0.003}{     0.035}{     0.006}{     0.005}{     0.005}{     0.005}{     0.007}
\SfusE{\Soptimal{    -29.91}}{\Soptimal{    -29.91}}{\Soptimal{    -29.91}}{\Soptimal{    -29.91}}{\Soptimal{    -29.91}}{\Soptimal{    -29.91}}
\Sfust{     0.035}{     0.006}{     0.005}{     0.005}{     0.005}{     0.007}
\SgenE{bca-lap   }{\Soptimal{    -29.91}}
\Sgent{     3.227}{     0.183}{     0.042}{     0.041}{     0.042}{     0.041}{     0.041}
\SfusE{\Soptimal{    -29.91}}{\Soptimal{    -29.91}}{\Soptimal{    -29.91}}{\Soptimal{    -29.91}}{\Soptimal{    -29.91}}{\Soptimal{    -29.91}}
\Sfust{     0.183}{     0.042}{     0.041}{     0.042}{     0.041}{     0.041}
\SgenE{bca-greedy}{\Soptimal{    -29.91}}
\Sgent{     0.173}{     0.032}{     0.003}{     0.005}{     0.004}{     0.004}{     0.005}
\SfusE{\Soptimal{    -29.91}}{\Soptimal{    -29.91}}{\Soptimal{    -29.91}}{\Soptimal{    -29.91}}{\Soptimal{    -29.91}}{\Soptimal{    -29.91}}
\Sfust{     0.032}{     0.003}{     0.005}{     0.004}{     0.004}{     0.005}
\SgenE{greedy    }{\Soptimal{    -29.91}}
\Sgent{     0.001}{     0.054}{     0.006}{     0.005}{     0.007}{     0.007}{     0.006}
\SfusE{\Soptimal{    -29.91}}{\Soptimal{    -29.91}}{\Soptimal{    -29.91}}{\Soptimal{    -29.91}}{\Soptimal{    -29.91}}{\Soptimal{    -29.91}}
\Sfust{     0.054}{     0.006}{     0.005}{     0.007}{     0.007}{     0.006}
\Sendtable

\Sinstance{motor6}\Soptimum{-51.51}\\
\Sstarttable
\SgenE{dd-ls0    }{\Soptimal{    -51.51}}
\Sgent{     0.013}{     0.093}{     0.006}{     0.007}{     0.026}{     0.026}{     0.008}
\SfusE{\Soptimal{    -51.51}}{\Soptimal{    -51.51}}{\Soptimal{    -51.51}}{\Soptimal{    -51.51}}{\Soptimal{    -51.51}}{\Soptimal{    -51.51}}
\Sfust{     0.093}{     0.006}{     0.007}{     0.026}{     0.026}{     0.008}
\SgenE{dd-ls3    }{\Soptimal{    -51.51}}
\Sgent{     0.093}{     0.070}{     0.012}{     0.013}{     0.108}{     0.108}{     0.021}
\SfusE{\Soptimal{    -51.51}}{\Soptimal{    -51.51}}{\Soptimal{    -51.51}}{\Soptimal{    -51.51}}{\Soptimal{    -51.51}}{\Soptimal{    -51.51}}
\Sfust{     0.070}{     0.012}{     0.013}{     0.108}{     0.108}{     0.021}
\SgenE{dd-ls4    }{\Soptimal{    -51.51}}
\Sgent{     0.590}{     0.100}{     0.068}{     0.069}{     0.614}{     0.614}{     0.045}
\SfusE{\Soptimal{    -51.51}}{\Soptimal{    -51.51}}{\Soptimal{    -51.51}}{\Soptimal{    -51.51}}{\Soptimal{    -51.51}}{\Soptimal{    -51.51}}
\Sfust{     0.100}{     0.068}{     0.069}{     0.614}{     0.614}{     0.045}
\SgenE{bca-lap   }{\Soptimal{    -51.51}}
\Sgent{     4.323}{     0.094}{     0.047}{     0.047}{     0.046}{     0.046}{     0.047}
\SfusE{\Soptimal{    -51.51}}{\Soptimal{    -51.51}}{\Soptimal{    -51.51}}{\Soptimal{    -51.51}}{\Soptimal{    -51.51}}{\Soptimal{    -51.51}}
\Sfust{     0.094}{     0.047}{     0.047}{     0.046}{     0.046}{     0.047}
\SgenE{bca-greedy}{\Soptimal{    -51.51}}
\Sgent{     0.240}{     0.038}{     0.004}{     0.004}{     0.005}{     0.005}{     0.005}
\SfusE{\Soptimal{    -51.51}}{\Soptimal{    -51.51}}{\Soptimal{    -51.51}}{\Soptimal{    -51.51}}{\Soptimal{    -51.51}}{\Soptimal{    -51.51}}
\Sfust{     0.038}{     0.004}{     0.004}{     0.005}{     0.005}{     0.005}
\SgenE{greedy    }{\Soptimal{    -51.51}}
\Sgent{     0.002}{     0.060}{     0.004}{     0.003}{     0.008}{     0.008}{     0.004}
\SfusE{\Soptimal{    -51.51}}{\Soptimal{    -51.51}}{\Soptimal{    -51.51}}{\Soptimal{    -51.51}}{\Soptimal{    -51.51}}{\Soptimal{    -51.51}}
\Sfust{     0.060}{     0.004}{     0.003}{     0.008}{     0.008}{     0.004}
\Sendtable

\Sinstance{motor7}\Soptimum{-64.93}\\
\Sstarttable
\SgenE{dd-ls0    }{\Soptimal{    -64.93}}
\Sgent{     0.044}{     0.147}{     0.025}{     0.026}{     0.077}{     0.074}{     0.026}
\SfusE{\Soptimal{    -64.93}}{\Soptimal{    -64.93}}{\Soptimal{    -64.93}}{\Soptimal{    -64.93}}{\Soptimal{    -64.93}}{\Soptimal{    -64.93}}
\Sfust{     0.147}{     0.025}{     0.026}{     0.077}{     0.074}{     0.026}
\SgenE{dd-ls3    }{\Soptimal{    -64.93}}
\Sgent{     0.537}{     0.349}{     0.162}{     0.164}{     0.576}{     0.577}{     0.167}
\SfusE{\Soptimal{    -64.93}}{\Soptimal{    -64.93}}{\Soptimal{    -64.93}}{\Soptimal{    -64.93}}{\Soptimal{    -64.93}}{\Soptimal{    -64.93}}
\Sfust{     0.349}{     0.162}{     0.164}{     0.576}{     0.577}{     0.167}
\SgenE{dd-ls4    }{\Soptimal{    -64.93}}
\Sgent{     2.652}{     1.632}{     1.409}{     1.411}{     2.709}{     2.711}{     1.416}
\SfusE{\Soptimal{    -64.93}}{\Soptimal{    -64.93}}{\Soptimal{    -64.93}}{\Soptimal{    -64.93}}{\Soptimal{    -64.93}}{\Soptimal{    -64.93}}
\Sfust{     1.632}{     1.409}{     1.411}{     2.709}{     2.711}{     1.416}
\SgenE{bca-lap   }{\Soptimal{    -64.93}}
\Sgent{     6.751}{     0.200}{     0.163}{     0.162}{     0.163}{     0.162}{     0.164}
\SfusE{\Soptimal{    -64.93}}{\Soptimal{    -64.93}}{\Soptimal{    -64.93}}{\Soptimal{    -64.93}}{\Soptimal{    -64.93}}{\Soptimal{    -64.93}}
\Sfust{     0.200}{     0.163}{     0.162}{     0.163}{     0.162}{     0.164}
\SgenE{bca-greedy}{\Soptimal{    -64.93}}
\Sgent{     0.402}{     0.069}{     0.018}{     0.019}{     0.019}{     0.019}{     0.020}
\SfusE{\Soptimal{    -64.93}}{\Soptimal{    -64.93}}{\Soptimal{    -64.93}}{\Soptimal{    -64.93}}{\Soptimal{    -64.93}}{\Soptimal{    -64.93}}
\Sfust{     0.069}{     0.018}{     0.019}{     0.019}{     0.019}{     0.020}
\SgenE{greedy    }{    -64.06}
\Sgent{     0.089}{     0.200}{     0.063}{     0.066}{     0.273}{     0.277}{     0.017}
\SfusE{\Soptimal{    -64.93}}{\Soptimal{    -64.93}}{\Soptimal{    -64.93}}{    -64.06}{    -64.06}{\Soptimal{    -64.93}}
\Sfust{     0.604}{     0.081}{     0.085}{     0.273}{     0.277}{     0.054}
\Sendtable

\Sinstance{motor8}\Soptimum{-79.71}\\
\Sstarttable
\SgenE{dd-ls0    }{\Soptimal{    -79.71}}
\Sgent{     0.113}{     0.306}{     0.035}{     0.036}{     0.166}{     0.165}{     0.039}
\SfusE{\Soptimal{    -79.71}}{\Soptimal{    -79.71}}{\Soptimal{    -79.71}}{\Soptimal{    -79.71}}{\Soptimal{    -79.71}}{\Soptimal{    -79.71}}
\Sfust{     0.306}{     0.035}{     0.036}{     0.166}{     0.165}{     0.039}
\SgenE{dd-ls3    }{\Soptimal{    -79.71}}
\Sgent{     2.757}{     1.520}{     1.034}{     1.036}{     2.848}{     2.848}{     0.596}
\SfusE{\Soptimal{    -79.71}}{\Soptimal{    -79.71}}{\Soptimal{    -79.71}}{\Soptimal{    -79.71}}{\Soptimal{    -79.71}}{\Soptimal{    -79.71}}
\Sfust{     1.520}{     1.034}{     1.036}{     2.848}{     2.848}{     0.596}
\SgenE{dd-ls4    }{\Soptimal{    -79.71}}
\Sgent{    16.918}{     0.845}{     0.984}{     0.730}{    17.064}{    17.067}{     1.290}
\SfusE{\Soptimal{    -79.71}}{\Soptimal{    -79.71}}{\Soptimal{    -79.71}}{\Soptimal{    -79.71}}{\Soptimal{    -79.71}}{\Soptimal{    -79.71}}
\Sfust{     0.845}{     0.984}{     0.730}{    17.064}{    17.067}{     1.290}
\SgenE{bca-lap   }{\Soptimal{    -79.71}}
\Sgent{    28.974}{     0.679}{     0.608}{     0.609}{     0.608}{     0.608}{     0.611}
\SfusE{\Soptimal{    -79.71}}{\Soptimal{    -79.71}}{\Soptimal{    -79.71}}{\Soptimal{    -79.71}}{\Soptimal{    -79.71}}{\Soptimal{    -79.71}}
\Sfust{     0.679}{     0.608}{     0.609}{     0.608}{     0.608}{     0.611}
\SgenE{bca-greedy}{\Soptimal{    -79.71}}
\Sgent{     1.918}{     0.062}{     0.015}{     0.015}{     0.042}{     0.042}{     0.016}
\SfusE{\Soptimal{    -79.71}}{\Soptimal{    -79.71}}{\Soptimal{    -79.71}}{\Soptimal{    -79.71}}{\Soptimal{    -79.71}}{\Soptimal{    -79.71}}
\Sfust{     0.062}{     0.015}{     0.015}{     0.042}{     0.042}{     0.016}
\SgenE{greedy    }{    -71.63}
\Sgent{     0.021}{     0.540}{     0.075}{     0.075}{     0.070}{     0.067}{     0.059}
\SfusE{\Soptimal{    -79.71}}{\Soptimal{    -79.71}}{\Soptimal{    -79.71}}{    -71.63}{    -71.63}{\Soptimal{    -79.71}}
\Sfust{     1.478}{     0.339}{     0.347}{     0.070}{     0.067}{     0.111}
\Sendtable

\Sinstance{motor9}\Soptimum{-55.17}\\
\Sstarttable
\SgenE{dd-ls0    }{\Soptimal{    -55.17}}
\Sgent{     0.026}{     0.072}{     0.007}{     0.007}{     0.047}{     0.049}{     0.008}
\SfusE{\Soptimal{    -55.17}}{\Soptimal{    -55.17}}{\Soptimal{    -55.17}}{\Soptimal{    -55.17}}{\Soptimal{    -55.17}}{\Soptimal{    -55.17}}
\Sfust{     0.072}{     0.007}{     0.007}{     0.047}{     0.049}{     0.008}
\SgenE{dd-ls3    }{\Soptimal{    -55.17}}
\Sgent{     0.306}{     0.107}{     0.029}{     0.029}{     0.339}{     0.340}{     0.031}
\SfusE{\Soptimal{    -55.17}}{\Soptimal{    -55.17}}{\Soptimal{    -55.17}}{\Soptimal{    -55.17}}{\Soptimal{    -55.17}}{\Soptimal{    -55.17}}
\Sfust{     0.107}{     0.029}{     0.029}{     0.339}{     0.340}{     0.031}
\SgenE{dd-ls4    }{\Soptimal{    -55.17}}
\Sgent{     1.504}{     0.256}{     0.136}{     0.137}{     1.541}{     1.542}{     0.138}
\SfusE{\Soptimal{    -55.17}}{\Soptimal{    -55.17}}{\Soptimal{    -55.17}}{\Soptimal{    -55.17}}{\Soptimal{    -55.17}}{\Soptimal{    -55.17}}
\Sfust{     0.256}{     0.136}{     0.137}{     1.541}{     1.542}{     0.138}
\SgenE{bca-lap   }{\Soptimal{    -55.17}}
\Sgent{     6.078}{     0.113}{     0.080}{     0.080}{     0.079}{     0.079}{     0.080}
\SfusE{\Soptimal{    -55.17}}{\Soptimal{    -55.17}}{\Soptimal{    -55.17}}{\Soptimal{    -55.17}}{\Soptimal{    -55.17}}{\Soptimal{    -55.17}}
\Sfust{     0.113}{     0.080}{     0.080}{     0.079}{     0.079}{     0.080}
\SgenE{bca-greedy}{\Soptimal{    -55.17}}
\Sgent{     0.326}{     0.059}{     0.008}{     0.008}{     0.025}{     0.025}{     0.009}
\SfusE{\Soptimal{    -55.17}}{\Soptimal{    -55.17}}{\Soptimal{    -55.17}}{\Soptimal{    -55.17}}{\Soptimal{    -55.17}}{\Soptimal{    -55.17}}
\Sfust{     0.059}{     0.008}{     0.008}{     0.025}{     0.025}{     0.009}
\SgenE{greedy    }{    -52.94}
\Sgent{     0.023}{     0.155}{     0.031}{     0.032}{     0.074}{     0.073}{     0.016}
\SfusE{\Soptimal{    -55.17}}{\Soptimal{    -55.17}}{\Soptimal{    -55.17}}{    -52.94}{    -52.94}{\Soptimal{    -55.17}}
\Sfust{     0.161}{     0.063}{     0.066}{     0.074}{     0.073}{     0.017}
\Sendtable

\Sinstance{motor10}\Soptimum{-48.95}\\
\Sstarttable
\SgenE{dd-ls0    }{\Soptimal{    -48.95}}
\Sgent{     0.009}{     0.065}{     0.008}{     0.008}{     0.018}{     0.019}{     0.009}
\SfusE{\Soptimal{    -48.95}}{\Soptimal{    -48.95}}{\Soptimal{    -48.95}}{\Soptimal{    -48.95}}{\Soptimal{    -48.95}}{\Soptimal{    -48.95}}
\Sfust{     0.065}{     0.008}{     0.008}{     0.018}{     0.019}{     0.009}
\SgenE{dd-ls3    }{\Soptimal{    -48.95}}
\Sgent{     0.150}{     0.084}{     0.019}{     0.019}{     0.174}{     0.174}{     0.020}
\SfusE{\Soptimal{    -48.95}}{\Soptimal{    -48.95}}{\Soptimal{    -48.95}}{\Soptimal{    -48.95}}{\Soptimal{    -48.95}}{\Soptimal{    -48.95}}
\Sfust{     0.084}{     0.019}{     0.019}{     0.174}{     0.174}{     0.020}
\SgenE{dd-ls4    }{\Soptimal{    -48.95}}
\Sgent{     0.251}{     0.099}{     0.082}{     0.082}{     0.267}{     0.267}{     0.040}
\SfusE{\Soptimal{    -48.95}}{\Soptimal{    -48.95}}{\Soptimal{    -48.95}}{\Soptimal{    -48.95}}{\Soptimal{    -48.95}}{\Soptimal{    -48.95}}
\Sfust{     0.099}{     0.082}{     0.082}{     0.267}{     0.267}{     0.040}
\SgenE{bca-lap   }{\Soptimal{    -48.95}}
\Sgent{     3.843}{     0.099}{     0.069}{     0.069}{     0.069}{     0.069}{     0.069}
\SfusE{\Soptimal{    -48.95}}{\Soptimal{    -48.95}}{\Soptimal{    -48.95}}{\Soptimal{    -48.95}}{\Soptimal{    -48.95}}{\Soptimal{    -48.95}}
\Sfust{     0.099}{     0.069}{     0.069}{     0.069}{     0.069}{     0.069}
\SgenE{bca-greedy}{\Soptimal{    -48.95}}
\Sgent{     0.216}{     0.041}{     0.006}{     0.006}{     0.006}{     0.006}{     0.007}
\SfusE{\Soptimal{    -48.95}}{\Soptimal{    -48.95}}{\Soptimal{    -48.95}}{\Soptimal{    -48.95}}{\Soptimal{    -48.95}}{\Soptimal{    -48.95}}
\Sfust{     0.041}{     0.006}{     0.006}{     0.006}{     0.006}{     0.007}
\SgenE{greedy    }{\Soptimal{    -48.95}}
\Sgent{     0.000}{     0.062}{     0.003}{     0.003}{     0.003}{     0.004}{     0.003}
\SfusE{\Soptimal{    -48.95}}{\Soptimal{    -48.95}}{\Soptimal{    -48.95}}{\Soptimal{    -48.95}}{\Soptimal{    -48.95}}{\Soptimal{    -48.95}}
\Sfust{     0.062}{     0.003}{     0.003}{     0.003}{     0.004}{     0.003}
\Sendtable

\Sinstance{motor11}\Soptimum{-50.16}\\
\Sstarttable
\SgenE{dd-ls0    }{\Soptimal{    -50.16}}
\Sgent{     0.010}{     0.060}{     0.006}{     0.006}{     0.020}{     0.021}{     0.008}
\SfusE{\Soptimal{    -50.16}}{\Soptimal{    -50.16}}{\Soptimal{    -50.16}}{\Soptimal{    -50.16}}{\Soptimal{    -50.16}}{\Soptimal{    -50.16}}
\Sfust{     0.060}{     0.006}{     0.006}{     0.020}{     0.021}{     0.008}
\SgenE{dd-ls3    }{\Soptimal{    -50.16}}
\Sgent{     0.174}{     0.052}{     0.028}{     0.029}{     0.196}{     0.196}{     0.011}
\SfusE{\Soptimal{    -50.16}}{\Soptimal{    -50.16}}{\Soptimal{    -50.16}}{\Soptimal{    -50.16}}{\Soptimal{    -50.16}}{\Soptimal{    -50.16}}
\Sfust{     0.052}{     0.028}{     0.029}{     0.196}{     0.196}{     0.011}
\SgenE{dd-ls4    }{\Soptimal{    -50.16}}
\Sgent{     0.173}{     0.087}{     0.082}{     0.082}{     0.186}{     0.186}{     0.084}
\SfusE{\Soptimal{    -50.16}}{\Soptimal{    -50.16}}{\Soptimal{    -50.16}}{\Soptimal{    -50.16}}{\Soptimal{    -50.16}}{\Soptimal{    -50.16}}
\Sfust{     0.087}{     0.082}{     0.082}{     0.186}{     0.186}{     0.084}
\SgenE{bca-lap   }{\Soptimal{    -50.16}}
\Sgent{     4.270}{     0.111}{     0.051}{     0.051}{     0.051}{     0.051}{     0.051}
\SfusE{\Soptimal{    -50.16}}{\Soptimal{    -50.16}}{\Soptimal{    -50.16}}{\Soptimal{    -50.16}}{\Soptimal{    -50.16}}{\Soptimal{    -50.16}}
\Sfust{     0.111}{     0.051}{     0.051}{     0.051}{     0.051}{     0.051}
\SgenE{bca-greedy}{\Soptimal{    -50.16}}
\Sgent{     0.289}{     0.042}{     0.005}{     0.005}{     0.008}{     0.008}{     0.005}
\SfusE{\Soptimal{    -50.16}}{\Soptimal{    -50.16}}{\Soptimal{    -50.16}}{\Soptimal{    -50.16}}{\Soptimal{    -50.16}}{\Soptimal{    -50.16}}
\Sfust{     0.042}{     0.005}{     0.005}{     0.008}{     0.008}{     0.005}
\SgenE{greedy    }{\Soptimal{    -50.16}}
\Sgent{     0.065}{     0.278}{     0.010}{     0.010}{     0.204}{     0.206}{     0.013}
\SfusE{\Soptimal{    -50.16}}{\Soptimal{    -50.16}}{\Soptimal{    -50.16}}{\Soptimal{    -50.16}}{\Soptimal{    -50.16}}{\Soptimal{    -50.16}}
\Sfust{     0.278}{     0.010}{     0.010}{     0.204}{     0.206}{     0.013}
\Sendtable

\Sinstance{motor12}\Soptimum{-55.81}\\
\Sstarttable
\SgenE{dd-ls0    }{\Soptimal{    -55.81}}
\Sgent{     0.035}{     0.088}{     0.011}{     0.011}{     0.064}{     0.065}{     0.012}
\SfusE{\Soptimal{    -55.81}}{\Soptimal{    -55.81}}{\Soptimal{    -55.81}}{\Soptimal{    -55.81}}{\Soptimal{    -55.81}}{\Soptimal{    -55.81}}
\Sfust{     0.088}{     0.011}{     0.011}{     0.064}{     0.065}{     0.012}
\SgenE{dd-ls3    }{\Soptimal{    -55.81}}
\Sgent{     0.530}{     0.340}{     0.129}{     0.130}{     0.573}{     0.572}{     0.134}
\SfusE{\Soptimal{    -55.81}}{\Soptimal{    -55.81}}{\Soptimal{    -55.81}}{\Soptimal{    -55.81}}{\Soptimal{    -55.81}}{\Soptimal{    -55.81}}
\Sfust{     0.340}{     0.129}{     0.130}{     0.573}{     0.572}{     0.134}
\SgenE{dd-ls4    }{\Soptimal{    -55.81}}
\Sgent{     4.146}{     0.179}{     0.119}{     0.119}{     4.211}{     4.212}{     0.120}
\SfusE{\Soptimal{    -55.81}}{\Soptimal{    -55.81}}{\Soptimal{    -55.81}}{\Soptimal{    -55.81}}{\Soptimal{    -55.81}}{\Soptimal{    -55.81}}
\Sfust{     0.179}{     0.119}{     0.119}{     4.211}{     4.212}{     0.120}
\SgenE{bca-lap   }{\Soptimal{    -55.81}}
\Sgent{    10.781}{     0.351}{     0.216}{     0.216}{     3.042}{     3.043}{     0.217}
\SfusE{\Soptimal{    -55.81}}{\Soptimal{    -55.81}}{\Soptimal{    -55.81}}{\Soptimal{    -55.81}}{\Soptimal{    -55.81}}{\Soptimal{    -55.81}}
\Sfust{     0.351}{     0.216}{     0.216}{     3.042}{     3.043}{     0.217}
\SgenE{bca-greedy}{\Soptimal{    -55.81}}
\Sgent{     0.538}{     0.040}{     0.006}{     0.006}{     0.034}{     0.035}{     0.006}
\SfusE{\Soptimal{    -55.81}}{\Soptimal{    -55.81}}{\Soptimal{    -55.81}}{\Soptimal{    -55.81}}{\Soptimal{    -55.81}}{\Soptimal{    -55.81}}
\Sfust{     0.040}{     0.006}{     0.006}{     0.034}{     0.035}{     0.006}
\SgenE{greedy    }{    -52.52}
\Sgent{     0.024}{     0.364}{     0.023}{     0.024}{     0.078}{     0.075}{     0.138}
\SfusE{\Soptimal{    -55.81}}{\Soptimal{    -55.81}}{\Soptimal{    -55.81}}{    -52.52}{    -52.52}{\Soptimal{    -55.81}}
\Sfust{     4.339}{     0.329}{     0.353}{     0.078}{     0.075}{     0.456}
\Sendtable

\Sinstance{motor13}\Soptimum{-36.35}\\
\Sstarttable
\SgenE{dd-ls0    }{\Soptimal{    -36.35}}
\Sgent{     0.046}{     0.651}{     0.039}{     0.065}{     0.097}{     0.097}{     0.090}
\SfusE{\Soptimal{    -36.35}}{\Soptimal{    -36.35}}{\Soptimal{    -36.35}}{\Soptimal{    -36.35}}{\Soptimal{    -36.35}}{\Soptimal{    -36.35}}
\Sfust{     0.651}{     0.039}{     0.065}{     0.097}{     0.097}{     0.090}
\SgenE{dd-ls3    }{\Soptimal{    -36.35}}
\Sgent{     0.432}{     0.739}{     0.252}{     0.256}{     0.486}{     0.487}{     0.134}
\SfusE{\Soptimal{    -36.35}}{\Soptimal{    -36.35}}{\Soptimal{    -36.35}}{\Soptimal{    -36.35}}{\Soptimal{    -36.35}}{\Soptimal{    -36.35}}
\Sfust{     0.739}{     0.252}{     0.256}{     0.486}{     0.487}{     0.134}
\SgenE{dd-ls4    }{\Soptimal{    -36.35}}
\Sgent{     4.047}{     2.728}{     2.195}{     2.198}{     4.118}{     4.117}{     0.513}
\SfusE{\Soptimal{    -36.35}}{\Soptimal{    -36.35}}{\Soptimal{    -36.35}}{\Soptimal{    -36.35}}{\Soptimal{    -36.35}}{\Soptimal{    -36.35}}
\Sfust{     2.728}{     2.195}{     2.198}{     4.118}{     4.117}{     0.513}
\SgenE{bca-lap   }{    -35.45}
\Sgent{     1.615}{     0.382}{     0.317}{     0.317}{     0.497}{     0.499}{     0.319}
\SfusE{\Soptimal{    -36.35}}{\Soptimal{    -36.35}}{\Soptimal{    -36.35}}{    -35.45}{    -35.45}{\Soptimal{    -36.35}}
\Sfust{     2.044}{     1.860}{     1.863}{     0.497}{     0.499}{     1.873}
\SgenE{bca-greedy}{\Soptimal{    -36.35}}
\Sgent{     0.235}{     0.172}{     0.028}{     0.028}{     0.028}{     0.028}{     0.036}
\SfusE{\Soptimal{    -36.35}}{\Soptimal{    -36.35}}{\Soptimal{    -36.35}}{\Soptimal{    -36.35}}{\Soptimal{    -36.35}}{\Soptimal{    -36.35}}
\Sfust{     0.172}{     0.028}{     0.028}{     0.028}{     0.028}{     0.036}
\SgenE{greedy    }{    -32.03}
\Sgent{     0.036}{     0.585}{     0.048}{     0.053}{     0.129}{     0.123}{     0.040}
\SfusE{    -33.69}{    -33.69}{    -33.69}{    -32.03}{    -32.03}{    -33.69}
\Sfust{     3.406}{     0.244}{     0.270}{     0.129}{     0.123}{     0.326}
\Sendtable

\Sinstance{motor14}\Soptimum{-43.80}\\
\Sstarttable
\SgenE{dd-ls0    }{\Soptimal{    -43.80}}
\Sgent{     0.082}{     1.026}{     0.140}{     0.145}{     0.147}{     0.148}{     0.206}
\SfusE{\Soptimal{    -43.80}}{\Soptimal{    -43.80}}{\Soptimal{    -43.80}}{\Soptimal{    -43.80}}{\Soptimal{    -43.80}}{\Soptimal{    -43.80}}
\Sfust{     1.026}{     0.140}{     0.145}{     0.147}{     0.148}{     0.206}
\SgenE{dd-ls3    }{\Soptimal{    -43.80}}
\Sgent{     1.273}{     2.603}{     1.354}{     1.358}{     1.368}{     1.367}{     1.419}
\SfusE{\Soptimal{    -43.80}}{\Soptimal{    -43.80}}{\Soptimal{    -43.80}}{\Soptimal{    -43.80}}{\Soptimal{    -43.80}}{\Soptimal{    -43.80}}
\Sfust{     2.603}{     1.354}{     1.358}{     1.368}{     1.367}{     1.419}
\SgenE{dd-ls4    }{\Soptimal{    -43.80}}
\Sgent{    13.576}{    11.638}{    10.679}{     8.982}{    13.694}{    13.697}{    13.766}
\SfusE{\Soptimal{    -43.80}}{\Soptimal{    -43.80}}{\Soptimal{    -43.80}}{\Soptimal{    -43.80}}{\Soptimal{    -43.80}}{\Soptimal{    -43.80}}
\Sfust{    11.638}{    10.679}{     8.982}{    13.694}{    13.697}{    13.766}
\SgenE{bca-lap   }{    -42.92}
\Sgent{     0.370}{     0.424}{     0.366}{     0.367}{     0.366}{     0.367}{     0.368}
\SfusE{    -42.92}{    -42.92}{    -42.92}{    -42.92}{    -42.92}{    -42.92}
\Sfust{     0.424}{     0.366}{     0.367}{     0.366}{     0.367}{     0.368}
\SgenE{bca-greedy}{    -43.47}
\Sgent{     0.264}{     0.196}{     0.034}{     0.035}{     0.345}{     0.346}{     0.031}
\SfusE{\Soptimal{    -43.80}}{\Soptimal{    -43.80}}{\Soptimal{    -43.80}}{    -43.47}{    -43.47}{\Soptimal{    -43.80}}
\Sfust{     0.196}{     0.034}{     0.035}{     0.345}{     0.346}{     0.031}
\SgenE{greedy    }{    -42.37}
\Sgent{     0.018}{     0.434}{     0.056}{     0.060}{     0.057}{     0.060}{     0.072}
\SfusE{    -43.70}{    -43.72}{    -43.72}{    -42.37}{    -42.37}{    -43.72}
\Sfust{     4.739}{     0.056}{     0.060}{     0.057}{     0.060}{     0.072}
\Sendtable

\Sinstance{motor15}\Soptimum{-32.30}\\
\Sstarttable
\SgenE{dd-ls0    }{\Soptimal{    -32.30}}
\Sgent{     0.040}{     0.267}{     0.027}{     0.028}{     0.088}{     0.086}{     0.110}
\SfusE{\Soptimal{    -32.30}}{\Soptimal{    -32.30}}{\Soptimal{    -32.30}}{\Soptimal{    -32.30}}{\Soptimal{    -32.30}}{\Soptimal{    -32.30}}
\Sfust{     0.267}{     0.027}{     0.028}{     0.088}{     0.086}{     0.110}
\SgenE{dd-ls3    }{\Soptimal{    -32.30}}
\Sgent{     0.356}{     0.200}{     0.159}{     0.043}{     0.409}{     0.410}{     0.128}
\SfusE{\Soptimal{    -32.30}}{\Soptimal{    -32.30}}{\Soptimal{    -32.30}}{\Soptimal{    -32.30}}{\Soptimal{    -32.30}}{\Soptimal{    -32.30}}
\Sfust{     0.200}{     0.159}{     0.043}{     0.409}{     0.410}{     0.128}
\SgenE{dd-ls4    }{\Soptimal{    -32.30}}
\Sgent{     3.162}{     1.619}{     0.116}{     1.188}{     3.227}{     3.229}{     3.292}
\SfusE{\Soptimal{    -32.30}}{\Soptimal{    -32.30}}{\Soptimal{    -32.30}}{\Soptimal{    -32.30}}{\Soptimal{    -32.30}}{\Soptimal{    -32.30}}
\Sfust{     1.619}{     0.116}{     1.188}{     3.227}{     3.229}{     3.292}
\SgenE{bca-lap   }{\Soptimal{    -32.30}}
\Sgent{     0.152}{     0.095}{     0.060}{     0.059}{     0.059}{     0.059}{     0.060}
\SfusE{\Soptimal{    -32.30}}{\Soptimal{    -32.30}}{\Soptimal{    -32.30}}{\Soptimal{    -32.30}}{\Soptimal{    -32.30}}{\Soptimal{    -32.30}}
\Sfust{     0.095}{     0.060}{     0.059}{     0.059}{     0.059}{     0.060}
\SgenE{bca-greedy}{\Soptimal{    -32.30}}
\Sgent{     0.314}{     0.084}{     0.013}{     0.014}{     0.031}{     0.030}{     0.031}
\SfusE{\Soptimal{    -32.30}}{\Soptimal{    -32.30}}{\Soptimal{    -32.30}}{\Soptimal{    -32.30}}{\Soptimal{    -32.30}}{\Soptimal{    -32.30}}
\Sfust{     0.084}{     0.013}{     0.014}{     0.031}{     0.030}{     0.031}
\SgenE{greedy    }{    -29.23}
\Sgent{     0.057}{     0.489}{     0.028}{     0.030}{     0.184}{     0.189}{     0.037}
\SfusE{    -32.06}{    -32.06}{    -32.06}{    -29.23}{    -29.23}{    -32.06}
\Sfust{     4.211}{     0.233}{     0.260}{     0.184}{     0.189}{     0.309}
\Sendtable

\Sinstance{motor16}\Soptimum{-75.13}\\
\Sstarttable
\SgenE{dd-ls0    }{\Soptimal{    -75.13}}
\Sgent{     0.037}{     0.111}{     0.016}{     0.016}{     0.059}{     0.058}{     0.018}
\SfusE{\Soptimal{    -75.13}}{\Soptimal{    -75.13}}{\Soptimal{    -75.13}}{\Soptimal{    -75.13}}{\Soptimal{    -75.13}}{\Soptimal{    -75.13}}
\Sfust{     0.111}{     0.016}{     0.016}{     0.059}{     0.058}{     0.018}
\SgenE{dd-ls3    }{\Soptimal{    -75.13}}
\Sgent{     2.403}{     0.147}{     0.072}{     0.073}{     2.464}{     2.466}{     0.074}
\SfusE{\Soptimal{    -75.13}}{\Soptimal{    -75.13}}{\Soptimal{    -75.13}}{\Soptimal{    -75.13}}{\Soptimal{    -75.13}}{\Soptimal{    -75.13}}
\Sfust{     0.147}{     0.072}{     0.073}{     2.464}{     2.466}{     0.074}
\SgenE{dd-ls4    }{\Soptimal{    -75.13}}
\Sgent{     6.265}{     1.211}{     1.063}{     1.063}{     6.365}{     6.364}{     2.766}
\SfusE{\Soptimal{    -75.13}}{\Soptimal{    -75.13}}{\Soptimal{    -75.13}}{\Soptimal{    -75.13}}{\Soptimal{    -75.13}}{\Soptimal{    -75.13}}
\Sfust{     1.211}{     1.063}{     1.063}{     6.365}{     6.364}{     2.766}
\SgenE{bca-lap   }{\Soptimal{    -75.13}}
\Sgent{    12.469}{     0.123}{     0.081}{     0.081}{     0.081}{     0.081}{     0.081}
\SfusE{\Soptimal{    -75.13}}{\Soptimal{    -75.13}}{\Soptimal{    -75.13}}{\Soptimal{    -75.13}}{\Soptimal{    -75.13}}{\Soptimal{    -75.13}}
\Sfust{     0.123}{     0.081}{     0.081}{     0.081}{     0.081}{     0.081}
\SgenE{bca-greedy}{\Soptimal{    -75.13}}
\Sgent{     0.840}{     0.054}{     0.007}{     0.008}{     0.015}{     0.015}{     0.007}
\SfusE{\Soptimal{    -75.13}}{\Soptimal{    -75.13}}{\Soptimal{    -75.13}}{\Soptimal{    -75.13}}{\Soptimal{    -75.13}}{\Soptimal{    -75.13}}
\Sfust{     0.054}{     0.007}{     0.008}{     0.015}{     0.015}{     0.007}
\SgenE{greedy    }{    -67.85}
\Sgent{     0.061}{     1.296}{     0.061}{     0.069}{     0.200}{     0.203}{     0.083}
\SfusE{\Soptimal{    -75.13}}{\Soptimal{    -75.13}}{\Soptimal{    -75.13}}{    -67.85}{    -67.85}{\Soptimal{    -75.13}}
\Sfust{     5.896}{     0.168}{     0.186}{     0.200}{     0.203}{     0.227}
\Sendtable

\Sinstance{motor17}\Soptimum{-84.38}\\
\Sstarttable
\SgenE{dd-ls0    }{\Soptimal{    -84.38}}
\Sgent{     0.120}{     0.606}{     0.148}{     0.151}{     0.176}{     0.175}{     0.166}
\SfusE{\Soptimal{    -84.38}}{\Soptimal{    -84.38}}{\Soptimal{    -84.38}}{\Soptimal{    -84.38}}{\Soptimal{    -84.38}}{\Soptimal{    -84.38}}
\Sfust{     0.606}{     0.148}{     0.151}{     0.176}{     0.175}{     0.166}
\SgenE{dd-ls3    }{\Soptimal{    -84.38}}
\Sgent{     1.899}{     1.565}{     1.105}{     1.108}{     1.972}{     1.974}{     1.168}
\SfusE{\Soptimal{    -84.38}}{\Soptimal{    -84.38}}{\Soptimal{    -84.38}}{\Soptimal{    -84.38}}{\Soptimal{    -84.38}}{\Soptimal{    -84.38}}
\Sfust{     1.565}{     1.105}{     1.108}{     1.972}{     1.974}{     1.168}
\SgenE{dd-ls4    }{\Soptimal{    -84.38}}
\Sgent{    12.288}{     4.846}{     4.449}{     4.450}{    12.416}{    12.417}{     4.469}
\SfusE{\Soptimal{    -84.38}}{\Soptimal{    -84.38}}{\Soptimal{    -84.38}}{\Soptimal{    -84.38}}{\Soptimal{    -84.38}}{\Soptimal{    -84.38}}
\Sfust{     4.846}{     4.449}{     4.450}{    12.416}{    12.417}{     4.469}
\SgenE{bca-lap   }{\Soptimal{    -84.38}}
\Sgent{     0.089}{     0.141}{     0.106}{     0.105}{     0.106}{     0.105}{     0.106}
\SfusE{\Soptimal{    -84.38}}{\Soptimal{    -84.38}}{\Soptimal{    -84.38}}{\Soptimal{    -84.38}}{\Soptimal{    -84.38}}{\Soptimal{    -84.38}}
\Sfust{     0.141}{     0.106}{     0.105}{     0.106}{     0.105}{     0.106}
\SgenE{bca-greedy}{\Soptimal{    -84.38}}
\Sgent{     1.239}{     0.080}{     0.024}{     0.024}{     0.024}{     0.024}{     0.030}
\SfusE{\Soptimal{    -84.38}}{\Soptimal{    -84.38}}{\Soptimal{    -84.38}}{\Soptimal{    -84.38}}{\Soptimal{    -84.38}}{\Soptimal{    -84.38}}
\Sfust{     0.080}{     0.024}{     0.024}{     0.024}{     0.024}{     0.030}
\SgenE{greedy    }{    -78.95}
\Sgent{     0.147}{     0.847}{     0.045}{     0.051}{     0.455}{     0.470}{     0.040}
\SfusE{\Soptimal{    -84.38}}{\Soptimal{    -84.38}}{\Soptimal{    -84.38}}{    -78.95}{    -78.95}{\Soptimal{    -84.38}}
\Sfust{     5.790}{     0.335}{     0.367}{     0.455}{     0.470}{     0.441}
\Sendtable

\Sinstance{motor18}\Soptimum{-131.41}\\
\Sstarttable
\SgenE{dd-ls0    }{\Soptimal{   -131.41}}
\Sgent{     0.130}{     0.123}{     0.023}{     0.024}{     0.169}{     0.168}{     0.032}
\SfusE{\Soptimal{   -131.41}}{\Soptimal{   -131.41}}{\Soptimal{   -131.41}}{\Soptimal{   -131.41}}{\Soptimal{   -131.41}}{\Soptimal{   -131.41}}
\Sfust{     0.123}{     0.023}{     0.024}{     0.169}{     0.168}{     0.032}
\SgenE{dd-ls3    }{\Soptimal{   -131.41}}
\Sgent{     3.104}{     0.572}{     0.596}{     0.382}{     3.189}{     3.189}{     0.603}
\SfusE{\Soptimal{   -131.41}}{\Soptimal{   -131.41}}{\Soptimal{   -131.41}}{\Soptimal{   -131.41}}{\Soptimal{   -131.41}}{\Soptimal{   -131.41}}
\Sfust{     0.572}{     0.596}{     0.382}{     3.189}{     3.189}{     0.603}
\SgenE{dd-ls4    }{\Soptimal{   -131.41}}
\Sgent{    14.619}{     1.815}{     3.406}{     3.408}{    14.801}{    14.801}{     1.527}
\SfusE{\Soptimal{   -131.41}}{\Soptimal{   -131.41}}{\Soptimal{   -131.41}}{\Soptimal{   -131.41}}{\Soptimal{   -131.41}}{\Soptimal{   -131.41}}
\Sfust{     1.815}{     3.406}{     3.408}{    14.801}{    14.801}{     1.527}
\SgenE{bca-lap   }{\Soptimal{   -131.41}}
\Sgent{    17.871}{     0.187}{     0.130}{     0.130}{     0.131}{     0.130}{     0.131}
\SfusE{\Soptimal{   -131.41}}{\Soptimal{   -131.41}}{\Soptimal{   -131.41}}{\Soptimal{   -131.41}}{\Soptimal{   -131.41}}{\Soptimal{   -131.41}}
\Sfust{     0.187}{     0.130}{     0.130}{     0.131}{     0.130}{     0.131}
\SgenE{bca-greedy}{\Soptimal{   -131.41}}
\Sgent{     2.249}{     0.062}{     0.017}{     0.017}{     0.038}{     0.037}{     0.017}
\SfusE{\Soptimal{   -131.41}}{\Soptimal{   -131.41}}{\Soptimal{   -131.41}}{\Soptimal{   -131.41}}{\Soptimal{   -131.41}}{\Soptimal{   -131.41}}
\Sfust{     0.062}{     0.017}{     0.017}{     0.038}{     0.037}{     0.017}
\SgenE{greedy    }{   -123.38}
\Sgent{     0.100}{     0.269}{     0.029}{     0.029}{     0.311}{     0.308}{     0.021}
\SfusE{\Soptimal{   -131.41}}{\Soptimal{   -131.41}}{\Soptimal{   -131.41}}{   -123.38}{   -123.38}{\Soptimal{   -131.41}}
\Sfust{     1.064}{     0.073}{     0.076}{     0.311}{     0.308}{     0.086}
\Sendtable

\Sinstance{motor19}\Soptimum{-75.29}\\
\Sstarttable
\SgenE{dd-ls0    }{\Soptimal{    -75.29}}
\Sgent{     0.017}{     0.051}{     0.007}{     0.007}{     0.030}{     0.030}{     0.007}
\SfusE{\Soptimal{    -75.29}}{\Soptimal{    -75.29}}{\Soptimal{    -75.29}}{\Soptimal{    -75.29}}{\Soptimal{    -75.29}}{\Soptimal{    -75.29}}
\Sfust{     0.051}{     0.007}{     0.007}{     0.030}{     0.030}{     0.007}
\SgenE{dd-ls3    }{\Soptimal{    -75.29}}
\Sgent{     0.101}{     0.118}{     0.016}{     0.016}{     0.115}{     0.115}{     0.017}
\SfusE{\Soptimal{    -75.29}}{\Soptimal{    -75.29}}{\Soptimal{    -75.29}}{\Soptimal{    -75.29}}{\Soptimal{    -75.29}}{\Soptimal{    -75.29}}
\Sfust{     0.118}{     0.016}{     0.016}{     0.115}{     0.115}{     0.017}
\SgenE{dd-ls4    }{\Soptimal{    -75.29}}
\Sgent{     1.770}{     0.141}{     0.082}{     0.082}{     1.813}{     1.813}{     0.119}
\SfusE{\Soptimal{    -75.29}}{\Soptimal{    -75.29}}{\Soptimal{    -75.29}}{\Soptimal{    -75.29}}{\Soptimal{    -75.29}}{\Soptimal{    -75.29}}
\Sfust{     0.141}{     0.082}{     0.082}{     1.813}{     1.813}{     0.119}
\SgenE{bca-lap   }{\Soptimal{    -75.29}}
\Sgent{     6.908}{     0.102}{     0.069}{     0.068}{     0.069}{     0.068}{     0.069}
\SfusE{\Soptimal{    -75.29}}{\Soptimal{    -75.29}}{\Soptimal{    -75.29}}{\Soptimal{    -75.29}}{\Soptimal{    -75.29}}{\Soptimal{    -75.29}}
\Sfust{     0.102}{     0.069}{     0.068}{     0.069}{     0.068}{     0.069}
\SgenE{bca-greedy}{\Soptimal{    -75.29}}
\Sgent{     0.428}{     0.050}{     0.012}{     0.012}{     0.011}{     0.012}{     0.012}
\SfusE{\Soptimal{    -75.29}}{\Soptimal{    -75.29}}{\Soptimal{    -75.29}}{\Soptimal{    -75.29}}{\Soptimal{    -75.29}}{\Soptimal{    -75.29}}
\Sfust{     0.050}{     0.012}{     0.012}{     0.011}{     0.012}{     0.012}
\SgenE{greedy    }{\Soptimal{    -75.29}}
\Sgent{     0.060}{     0.444}{     0.025}{     0.027}{     0.197}{     0.191}{     0.032}
\SfusE{\Soptimal{    -75.29}}{\Soptimal{    -75.29}}{\Soptimal{    -75.29}}{\Soptimal{    -75.29}}{\Soptimal{    -75.29}}{\Soptimal{    -75.29}}
\Sfust{     0.444}{     0.025}{     0.027}{     0.197}{     0.191}{     0.032}
\Sendtable

\Sinstance{motor20}\Soptimum{-82.10}\\
\Sstarttable
\SgenE{dd-ls0    }{\Soptimal{    -82.10}}
\Sgent{     0.084}{     0.247}{     0.045}{     0.046}{     0.126}{     0.124}{     0.081}
\SfusE{\Soptimal{    -82.10}}{\Soptimal{    -82.10}}{\Soptimal{    -82.10}}{\Soptimal{    -82.10}}{\Soptimal{    -82.10}}{\Soptimal{    -82.10}}
\Sfust{     0.247}{     0.045}{     0.046}{     0.126}{     0.124}{     0.081}
\SgenE{dd-ls3    }{\Soptimal{    -82.10}}
\Sgent{     2.292}{     0.442}{     0.241}{     0.241}{     2.377}{     2.378}{     0.186}
\SfusE{\Soptimal{    -82.10}}{\Soptimal{    -82.10}}{\Soptimal{    -82.10}}{\Soptimal{    -82.10}}{\Soptimal{    -82.10}}{\Soptimal{    -82.10}}
\Sfust{     0.442}{     0.241}{     0.241}{     2.377}{     2.378}{     0.186}
\SgenE{dd-ls4    }{\Soptimal{    -82.10}}
\Sgent{    14.748}{     1.347}{     1.166}{     1.167}{    14.894}{    14.894}{     2.653}
\SfusE{\Soptimal{    -82.10}}{\Soptimal{    -82.10}}{\Soptimal{    -82.10}}{\Soptimal{    -82.10}}{\Soptimal{    -82.10}}{\Soptimal{    -82.10}}
\Sfust{     1.347}{     1.166}{     1.167}{    14.894}{    14.894}{     2.653}
\SgenE{bca-lap   }{\Soptimal{    -82.10}}
\Sgent{     8.677}{     0.338}{     0.294}{     0.294}{     0.294}{     0.294}{     0.296}
\SfusE{\Soptimal{    -82.10}}{\Soptimal{    -82.10}}{\Soptimal{    -82.10}}{\Soptimal{    -82.10}}{\Soptimal{    -82.10}}{\Soptimal{    -82.10}}
\Sfust{     0.338}{     0.294}{     0.294}{     0.294}{     0.294}{     0.296}
\SgenE{bca-greedy}{\Soptimal{    -82.10}}
\Sgent{     0.611}{     0.146}{     0.054}{     0.054}{     0.054}{     0.054}{     0.057}
\SfusE{\Soptimal{    -82.10}}{\Soptimal{    -82.10}}{\Soptimal{    -82.10}}{\Soptimal{    -82.10}}{\Soptimal{    -82.10}}{\Soptimal{    -82.10}}
\Sfust{     0.146}{     0.054}{     0.054}{     0.054}{     0.054}{     0.057}
\SgenE{greedy    }{    -80.99}
\Sgent{     0.045}{     1.862}{     0.031}{     0.035}{     0.161}{     0.150}{     0.212}
\SfusE{\Soptimal{    -82.10}}{\Soptimal{    -82.10}}{\Soptimal{    -82.10}}{    -80.99}{    -80.99}{    -81.55}
\Sfust{     1.862}{     0.031}{     0.035}{     0.161}{     0.150}{     0.212}
\Sendtable

\Ssubsection{flow}

\Snrinst{6}

\Snriter{2000}

\Sinstance{flow1}\Salias{board}\Soptimum{-2262.66}\\
\Sstarttable
\SgenE{dd-ls0    }{\Soptimal{  -2262.66}}
\Sgent{     0.616}{     3.323}{     0.365}{     0.427}{     1.228}{     1.300}{     2.026}
\SfusE{\Soptimal{  -2262.66}}{\Soptimal{  -2262.66}}{\Soptimal{  -2262.66}}{\Soptimal{  -2262.66}}{\Soptimal{  -2262.66}}{\Soptimal{  -2262.66}}
\Sfust{     3.323}{     0.365}{     0.427}{     1.228}{     1.300}{     2.026}
\SgenE{dd-ls3    }{\Soptimal{  -2262.66}}
\Sgent{     0.922}{     1.618}{     0.268}{     0.277}{     1.361}{     1.307}{     2.070}
\SfusE{\Soptimal{  -2262.66}}{\Soptimal{  -2262.66}}{\Soptimal{  -2262.66}}{\Soptimal{  -2262.66}}{\Soptimal{  -2262.66}}{\Soptimal{  -2262.66}}
\Sfust{     1.618}{     0.268}{     0.277}{     1.361}{     1.307}{     2.070}
\SgenE{dd-ls4    }{\Soptimal{  -2262.66}}
\Sgent{     1.593}{     1.878}{     0.603}{     0.622}{     1.927}{     1.909}{     2.227}
\SfusE{\Soptimal{  -2262.66}}{\Soptimal{  -2262.66}}{\Soptimal{  -2262.66}}{\Soptimal{  -2262.66}}{\Soptimal{  -2262.66}}{\Soptimal{  -2262.66}}
\Sfust{     1.878}{     0.603}{     0.622}{     1.927}{     1.909}{     2.227}
\SgenE{bca-lap   }{\Soptimal{  -2262.66}}
\Sgent{    15.045}{     4.603}{     4.080}{     4.124}{     4.194}{     4.171}{     4.360}
\SfusE{\Soptimal{  -2262.66}}{\Soptimal{  -2262.66}}{\Soptimal{  -2262.66}}{\Soptimal{  -2262.66}}{\Soptimal{  -2262.66}}{\Soptimal{  -2262.66}}
\Sfust{     4.603}{     4.080}{     4.124}{     4.194}{     4.171}{     4.360}
\SgenE{bca-greedy}{\Soptimal{  -2262.66}}
\Sgent{     3.294}{     0.092}{     0.024}{     0.024}{     0.166}{     0.171}{     0.212}
\SfusE{\Soptimal{  -2262.66}}{\Soptimal{  -2262.66}}{\Soptimal{  -2262.66}}{\Soptimal{  -2262.66}}{\Soptimal{  -2262.66}}{\Soptimal{  -2262.66}}
\Sfust{     0.092}{     0.024}{     0.024}{     0.166}{     0.171}{     0.212}
\SgenE{greedy    }{  -2261.62}
\Sgent{     0.056}{     0.127}{     0.007}{     0.008}{     0.248}{     0.246}{     0.702}
\SfusE{\Soptimal{  -2262.66}}{\Soptimal{  -2262.66}}{\Soptimal{  -2262.66}}{  -2261.62}{  -2261.62}{\Soptimal{  -2262.66}}
\Sfust{     0.175}{     0.013}{     0.017}{     0.248}{     0.246}{     2.691}
\Sendtable

\Sinstance{flow2}\Salias{books}\\
\Sstarttable
\SgenE{dd-ls0    }{  -4021.66}
\Sgent{     6.127}{     4.848}{     0.733}{     0.833}{    10.838}{    10.753}{     9.081}
\SfusE{  -4110.35}{  -4110.35}{  -4110.35}{  -4021.66}{  -4021.66}{  -4058.88}
\Sfust{    18.774}{     4.993}{     5.543}{    10.838}{    10.753}{    13.169}
\SgenE{dd-ls3    }{  -4100.28}
\Sgent{    16.043}{     8.441}{     2.841}{     2.939}{    22.195}{    21.860}{    14.642}
\SfusE{  -4135.27}{  -4134.77}{  -4134.77}{  -4100.28}{  -4100.28}{  -4112.27}
\Sfust{    34.243}{    19.953}{    17.708}{    22.195}{    21.860}{    21.152}
\SgenE{dd-ls4    }{  -4103.72}
\Sgent{    34.242}{     7.285}{     3.133}{    15.697}{    39.493}{    39.579}{    28.200}
\SfusE{  -4135.27}{  -4134.77}{  -4134.77}{  -4103.72}{  -4103.72}{  -4134.75}
\Sfust{    51.039}{    33.232}{    33.894}{    39.493}{    39.579}{    49.819}
\SgenE{bca-lap   }{  -4076.69}
\Sgent{    25.417}{    17.971}{    16.757}{    17.008}{    28.194}{    28.313}{    28.669}
\SfusE{  -4127.84}{  -4127.84}{  -4127.84}{  -4076.69}{  -4076.69}{  -4081.90}
\Sfust{    57.972}{    54.322}{    55.137}{    28.194}{    28.313}{    74.822}
\SgenE{bca-greedy}{  -4133.46}
\Sgent{    26.953}{     4.493}{     3.246}{     3.357}{    33.823}{    33.658}{    20.385}
\SfusE{  -4135.27}{  -4135.27}{  -4135.27}{  -4133.46}{  -4133.46}{  -4135.27}
\Sfust{     4.493}{     3.246}{     3.357}{    33.823}{    33.658}{    23.380}
\SgenE{greedy    }{  -3912.25}
\Sgent{     1.138}{     0.131}{     0.013}{     0.015}{     5.923}{     6.507}{     3.385}
\SfusE{  -4124.40}{  -4124.40}{  -4124.40}{  -3912.25}{  -3912.25}{  -4052.33}
\Sfust{     0.811}{     0.102}{     0.127}{     5.923}{     6.507}{    16.245}
\Sendtable

\Sinstance{flow3}\Salias{hammer}\Soptimum{-2097.78}\\
\Sstarttable
\SgenE{dd-ls0    }{\Soptimal{  -2097.78}}
\Sgent{     3.189}{     3.974}{     0.629}{     1.857}{     5.555}{     5.549}{     5.033}
\SfusE{\Soptimal{  -2097.78}}{\Soptimal{  -2097.78}}{\Soptimal{  -2097.78}}{\Soptimal{  -2097.78}}{\Soptimal{  -2097.78}}{\Soptimal{  -2097.78}}
\Sfust{     3.974}{     0.629}{     1.857}{     5.555}{     5.549}{     5.033}
\SgenE{dd-ls3    }{\Soptimal{  -2097.78}}
\Sgent{     3.744}{     2.721}{     0.495}{     0.471}{     4.910}{     4.922}{     5.219}
\SfusE{\Soptimal{  -2097.78}}{\Soptimal{  -2097.78}}{\Soptimal{  -2097.78}}{\Soptimal{  -2097.78}}{\Soptimal{  -2097.78}}{\Soptimal{  -2097.78}}
\Sfust{     2.721}{     0.495}{     0.471}{     4.910}{     4.922}{     5.219}
\SgenE{dd-ls4    }{\Soptimal{  -2097.78}}
\Sgent{    11.117}{     4.040}{     1.689}{     1.718}{    12.536}{    12.365}{    14.143}
\SfusE{\Soptimal{  -2097.78}}{\Soptimal{  -2097.78}}{\Soptimal{  -2097.78}}{\Soptimal{  -2097.78}}{\Soptimal{  -2097.78}}{\Soptimal{  -2097.78}}
\Sfust{     4.040}{     1.689}{     1.718}{    12.536}{    12.365}{    14.143}
\SgenE{bca-lap   }{  -1907.12}
\Sgent{     3.714}{     3.397}{     3.061}{     3.084}{     3.179}{     3.154}{     3.271}
\SfusE{  -2029.20}{  -2011.13}{  -2029.20}{  -1907.12}{  -1907.12}{  -1934.52}
\Sfust{    38.231}{     3.061}{    35.582}{     3.179}{     3.154}{    38.298}
\SgenE{bca-greedy}{\Soptimal{  -2097.78}}
\Sgent{    12.600}{     0.339}{     0.173}{     0.179}{     5.427}{     5.541}{     5.529}
\SfusE{\Soptimal{  -2097.78}}{\Soptimal{  -2097.78}}{\Soptimal{  -2097.78}}{\Soptimal{  -2097.78}}{\Soptimal{  -2097.78}}{\Soptimal{  -2097.78}}
\Sfust{     0.339}{     0.173}{     0.179}{     5.427}{     5.541}{     5.529}
\SgenE{greedy    }{  -1995.84}
\Sgent{     0.190}{     0.087}{     0.010}{     0.010}{     1.164}{     1.040}{     4.471}
\SfusE{\Soptimal{  -2097.78}}{\Soptimal{  -2097.78}}{\Soptimal{  -2097.78}}{  -1995.84}{  -1995.84}{  -2036.68}
\Sfust{     0.237}{     0.030}{     0.032}{     1.164}{     1.040}{    14.117}
\Sendtable

\Sinstance{flow4}\Salias{party}\Soptimum{-3629.91}\\
\Sstarttable
\SgenE{dd-ls0    }{  -3629.72}
\Sgent{     3.250}{     9.113}{     1.731}{     1.862}{     6.141}{     6.249}{    10.374}
\SfusE{\Soptimal{  -3629.91}}{\Soptimal{  -3629.91}}{\Soptimal{  -3629.91}}{  -3629.72}{  -3629.72}{  -3629.72}
\Sfust{     9.113}{     1.731}{     1.862}{     6.141}{     6.249}{    10.374}
\SgenE{dd-ls3    }{\Soptimal{  -3629.91}}
\Sgent{     3.772}{     9.318}{     3.490}{     3.719}{     5.250}{     5.321}{     7.906}
\SfusE{\Soptimal{  -3629.91}}{\Soptimal{  -3629.91}}{\Soptimal{  -3629.91}}{\Soptimal{  -3629.91}}{\Soptimal{  -3629.91}}{\Soptimal{  -3629.91}}
\Sfust{     9.318}{     3.490}{     3.719}{     5.250}{     5.321}{     7.906}
\SgenE{dd-ls4    }{\Soptimal{  -3629.91}}
\Sgent{    10.735}{    10.045}{     6.939}{     5.711}{    12.627}{    12.752}{    15.942}
\SfusE{\Soptimal{  -3629.91}}{\Soptimal{  -3629.91}}{\Soptimal{  -3629.91}}{\Soptimal{  -3629.91}}{\Soptimal{  -3629.91}}{\Soptimal{  -3629.91}}
\Sfust{    10.045}{     6.939}{     5.711}{    12.627}{    12.752}{    15.942}
\SgenE{bca-lap   }{\Soptimal{  -3629.91}}
\Sgent{    31.380}{    23.529}{    20.891}{    21.185}{    21.860}{    22.181}{    22.905}
\SfusE{\Soptimal{  -3629.91}}{\Soptimal{  -3629.91}}{\Soptimal{  -3629.91}}{\Soptimal{  -3629.91}}{\Soptimal{  -3629.91}}{\Soptimal{  -3629.91}}
\Sfust{    23.529}{    20.891}{    21.185}{    21.860}{    22.181}{    22.905}
\SgenE{bca-greedy}{\Soptimal{  -3629.91}}
\Sgent{    13.599}{     0.406}{     0.199}{     0.213}{     1.561}{     1.637}{     1.932}
\SfusE{\Soptimal{  -3629.91}}{\Soptimal{  -3629.91}}{\Soptimal{  -3629.91}}{\Soptimal{  -3629.91}}{\Soptimal{  -3629.91}}{\Soptimal{  -3629.91}}
\Sfust{     0.406}{     0.199}{     0.213}{     1.561}{     1.637}{     1.932}
\SgenE{greedy    }{  -3570.66}
\Sgent{     0.248}{     0.458}{     0.017}{     0.022}{     1.696}{     1.636}{     5.235}
\SfusE{\Soptimal{  -3629.91}}{\Soptimal{  -3629.91}}{\Soptimal{  -3629.91}}{  -3570.66}{  -3570.66}{  -3601.79}
\Sfust{     3.884}{     0.455}{     0.609}{     1.696}{     1.636}{     6.750}
\Sendtable

\Sinstance{flow5}\Salias{table}\Soptimum{-3288.51}\\
\Sstarttable
\SgenE{dd-ls0    }{  -3275.31}
\Sgent{     2.894}{     6.528}{     1.553}{     1.698}{     5.548}{     5.750}{     5.008}
\SfusE{\Soptimal{  -3288.51}}{  -3288.10}{\Soptimal{  -3288.51}}{  -3275.31}{  -3275.31}{  -3277.98}
\Sfust{     6.528}{     1.553}{     1.698}{     5.548}{     5.750}{     6.786}
\SgenE{dd-ls3    }{\Soptimal{  -3288.51}}
\Sgent{     3.358}{     7.077}{     3.140}{     3.304}{     4.683}{     4.705}{     5.909}
\SfusE{\Soptimal{  -3288.51}}{\Soptimal{  -3288.51}}{\Soptimal{  -3288.51}}{\Soptimal{  -3288.51}}{\Soptimal{  -3288.51}}{\Soptimal{  -3288.51}}
\Sfust{     7.077}{     3.140}{     3.304}{     4.683}{     4.705}{     5.909}
\SgenE{dd-ls4    }{\Soptimal{  -3288.51}}
\Sgent{     5.406}{     8.123}{     4.812}{     4.931}{     6.366}{     6.454}{     7.720}
\SfusE{\Soptimal{  -3288.51}}{\Soptimal{  -3288.51}}{\Soptimal{  -3288.51}}{\Soptimal{  -3288.51}}{\Soptimal{  -3288.51}}{\Soptimal{  -3288.51}}
\Sfust{     8.123}{     4.812}{     4.931}{     6.366}{     6.454}{     7.720}
\SgenE{bca-lap   }{  -3147.98}
\Sgent{     7.543}{     6.609}{     5.814}{     5.917}{     8.113}{     8.118}{     8.469}
\SfusE{  -3170.68}{  -3170.68}{  -3170.68}{  -3147.98}{  -3147.98}{  -3148.29}
\Sfust{    16.886}{    15.077}{    15.340}{     8.113}{     8.118}{     8.469}
\SgenE{bca-greedy}{  -3277.26}
\Sgent{     0.818}{     0.301}{     0.120}{     0.123}{     1.074}{     1.048}{     0.308}
\SfusE{\Soptimal{  -3288.51}}{\Soptimal{  -3288.51}}{\Soptimal{  -3288.51}}{  -3277.26}{  -3277.26}{\Soptimal{  -3288.51}}
\Sfust{     0.384}{     0.163}{     0.167}{     1.074}{     1.048}{    13.547}
\SgenE{greedy    }{  -3192.41}
\Sgent{     0.341}{     0.197}{     0.010}{     0.009}{     1.751}{     1.609}{     3.164}
\SfusE{\Soptimal{  -3288.51}}{\Soptimal{  -3288.51}}{\Soptimal{  -3288.51}}{  -3192.41}{  -3192.41}{  -3276.50}
\Sfust{     0.638}{     0.073}{     0.085}{     1.751}{     1.609}{     5.975}
\Sendtable

\Sinstance{flow6}\Salias{walking}\Soptimum{-1625.85}\\
\Sstarttable
\SgenE{dd-ls0    }{\Soptimal{  -1625.85}}
\Sgent{     0.689}{     1.216}{     0.065}{     0.070}{     1.095}{     1.098}{     1.840}
\SfusE{\Soptimal{  -1625.85}}{\Soptimal{  -1625.85}}{\Soptimal{  -1625.85}}{\Soptimal{  -1625.85}}{\Soptimal{  -1625.85}}{\Soptimal{  -1625.85}}
\Sfust{     1.216}{     0.065}{     0.070}{     1.095}{     1.098}{     1.840}
\SgenE{dd-ls3    }{\Soptimal{  -1625.85}}
\Sgent{     2.085}{     3.077}{     0.695}{     0.703}{     2.379}{     2.387}{     3.048}
\SfusE{\Soptimal{  -1625.85}}{\Soptimal{  -1625.85}}{\Soptimal{  -1625.85}}{\Soptimal{  -1625.85}}{\Soptimal{  -1625.85}}{\Soptimal{  -1625.85}}
\Sfust{     3.077}{     0.695}{     0.703}{     2.379}{     2.387}{     3.048}
\SgenE{dd-ls4    }{\Soptimal{  -1625.85}}
\Sgent{     4.902}{     3.348}{     1.310}{     1.315}{     5.215}{     5.188}{     5.119}
\SfusE{\Soptimal{  -1625.85}}{\Soptimal{  -1625.85}}{\Soptimal{  -1625.85}}{\Soptimal{  -1625.85}}{\Soptimal{  -1625.85}}{\Soptimal{  -1625.85}}
\Sfust{     3.348}{     1.310}{     1.315}{     5.215}{     5.188}{     5.119}
\SgenE{bca-lap   }{  -1625.55}
\Sgent{     2.122}{     1.202}{     1.077}{     1.082}{     1.086}{     1.092}{     1.130}
\SfusE{  -1625.55}{  -1625.55}{  -1625.55}{  -1625.55}{  -1625.55}{  -1625.55}
\Sfust{     1.202}{     1.077}{     1.082}{     1.086}{     1.092}{     1.130}
\SgenE{bca-greedy}{\Soptimal{  -1625.85}}
\Sgent{     0.607}{     0.099}{     0.028}{     0.029}{     0.262}{     0.250}{     0.325}
\SfusE{\Soptimal{  -1625.85}}{\Soptimal{  -1625.85}}{\Soptimal{  -1625.85}}{\Soptimal{  -1625.85}}{\Soptimal{  -1625.85}}{\Soptimal{  -1625.85}}
\Sfust{     0.099}{     0.028}{     0.029}{     0.262}{     0.250}{     0.325}
\SgenE{greedy    }{  -1623.27}
\Sgent{     0.280}{     0.214}{     0.014}{     0.016}{     1.257}{     1.161}{     0.706}
\SfusE{\Soptimal{  -1625.85}}{\Soptimal{  -1625.85}}{\Soptimal{  -1625.85}}{  -1623.27}{  -1623.27}{\Soptimal{  -1625.85}}
\Sfust{     0.214}{     0.014}{     0.016}{     1.257}{     1.161}{     1.056}
\Sendtable

\Ssubsection{opengm}

\Snrinst{4}

\Snriter{3000}

\Sinstance{opengm0}\Salias{matching0}\Soptimum{19.36}\\
\Sstarttable
\SgenE{dd-ls0    }{     60.14}
\Sgent{     0.958}{     3.030}{     0.069}{     0.062}{     1.394}{     1.393}{     0.070}
\SfusE{     32.93}{     33.95}{     39.20}{     60.14}{     60.14}{     39.20}
\Sfust{    24.895}{     0.695}{     0.455}{     1.394}{     1.393}{     0.437}
\SgenE{dd-ls3    }{\Soptimal{     19.36}}
\Sgent{     3.264}{    69.549}{     3.528}{     3.560}{     3.682}{     3.694}{     3.760}
\SfusE{\Soptimal{     19.36}}{\Soptimal{     19.36}}{\Soptimal{     19.36}}{\Soptimal{     19.36}}{\Soptimal{     19.36}}{\Soptimal{     19.36}}
\Sfust{    69.549}{     3.528}{     3.560}{     3.682}{     3.694}{     3.760}
\SgenE{dd-ls4    }{\Soptimal{     19.36}}
\Sgent{    25.738}{    80.629}{    25.433}{    22.885}{    26.121}{    26.119}{    22.988}
\SfusE{\Soptimal{     19.36}}{\Soptimal{     19.36}}{\Soptimal{     19.36}}{\Soptimal{     19.36}}{\Soptimal{     19.36}}{\Soptimal{     19.36}}
\Sfust{    80.629}{    25.433}{    22.885}{    26.121}{    26.119}{    22.988}
\SgenE{bca-lap   }{     72.62}
\Sgent{     0.397}{     0.897}{     0.386}{     0.388}{     0.388}{     0.387}{     0.392}
\SfusE{     72.57}{     72.57}{     72.57}{     72.62}{     72.62}{     72.57}
\Sfust{     4.306}{     1.697}{     1.707}{     0.388}{     0.387}{     1.731}
\SgenE{bca-greedy}{\Soptimal{     19.36}}
\Sgent{     0.006}{     0.429}{     0.009}{     0.009}{     0.010}{     0.010}{     0.020}
\SfusE{\Soptimal{     19.36}}{\Soptimal{     19.36}}{\Soptimal{     19.36}}{\Soptimal{     19.36}}{\Soptimal{     19.36}}{\Soptimal{     19.36}}
\Sfust{     0.429}{     0.009}{     0.009}{     0.010}{     0.010}{     0.020}
\SgenE{greedy    }{\Soptimal{     19.36}}
\Sgent{     0.004}{     2.277}{     0.017}{     0.016}{     0.017}{     0.016}{     0.121}
\SfusE{\Soptimal{     19.36}}{\Soptimal{     19.36}}{\Soptimal{     19.36}}{\Soptimal{     19.36}}{\Soptimal{     19.36}}{\Soptimal{     19.36}}
\Sfust{     2.277}{     0.017}{     0.016}{     0.017}{     0.016}{     0.121}
\Sendtable

\Sinstance{opengm1}\Salias{matching1}\Soptimum{23.58}\\
\Sstarttable
\SgenE{dd-ls0    }{\Soptimal{     23.58}}
\Sgent{     0.763}{    22.826}{     0.779}{     0.793}{     1.003}{     1.002}{     0.313}
\SfusE{\Soptimal{     23.58}}{\Soptimal{     23.58}}{\Soptimal{     23.58}}{\Soptimal{     23.58}}{\Soptimal{     23.58}}{\Soptimal{     23.58}}
\Sfust{    22.826}{     0.779}{     0.793}{     1.003}{     1.002}{     0.313}
\SgenE{dd-ls3    }{\Soptimal{     23.58}}
\Sgent{     1.858}{     5.147}{     0.330}{     0.171}{     2.162}{     2.163}{     1.509}
\SfusE{\Soptimal{     23.58}}{\Soptimal{     23.58}}{\Soptimal{     23.58}}{\Soptimal{     23.58}}{\Soptimal{     23.58}}{\Soptimal{     23.58}}
\Sfust{     5.147}{     0.330}{     0.171}{     2.162}{     2.163}{     1.509}
\SgenE{dd-ls4    }{\Soptimal{     23.58}}
\Sgent{     9.441}{    15.191}{     5.507}{     7.109}{     9.726}{     9.725}{     7.212}
\SfusE{\Soptimal{     23.58}}{\Soptimal{     23.58}}{\Soptimal{     23.58}}{\Soptimal{     23.58}}{\Soptimal{     23.58}}{\Soptimal{     23.58}}
\Sfust{    15.191}{     5.507}{     7.109}{     9.726}{     9.725}{     7.212}
\SgenE{bca-lap   }{\Soptimal{     23.58}}
\Sgent{    16.040}{     0.477}{     0.120}{     0.121}{     0.149}{     0.149}{     0.121}
\SfusE{\Soptimal{     23.58}}{\Soptimal{     23.58}}{\Soptimal{     23.58}}{\Soptimal{     23.58}}{\Soptimal{     23.58}}{\Soptimal{     23.58}}
\Sfust{     0.477}{     0.120}{     0.121}{     0.149}{     0.149}{     0.121}
\SgenE{bca-greedy}{\Soptimal{     23.58}}
\Sgent{     2.426}{     0.801}{     0.009}{     0.009}{     0.009}{     0.009}{     0.029}
\SfusE{\Soptimal{     23.58}}{\Soptimal{     23.58}}{\Soptimal{     23.58}}{\Soptimal{     23.58}}{\Soptimal{     23.58}}{\Soptimal{     23.58}}
\Sfust{     0.801}{     0.009}{     0.009}{     0.009}{     0.009}{     0.029}
\SgenE{greedy    }{\Soptimal{     23.58}}
\Sgent{     0.074}{     3.091}{     0.009}{     0.009}{     0.269}{     0.257}{     0.118}
\SfusE{\Soptimal{     23.58}}{\Soptimal{     23.58}}{\Soptimal{     23.58}}{\Soptimal{     23.58}}{\Soptimal{     23.58}}{\Soptimal{     23.58}}
\Sfust{     3.091}{     0.009}{     0.009}{     0.269}{     0.257}{     0.118}
\Sendtable

\Sinstance{opengm2}\Salias{matching2}\Soptimum{26.08}\\
\Sstarttable
\SgenE{dd-ls0    }{\Soptimal{     26.08}}
\Sgent{     1.001}{    31.515}{     1.170}{     1.193}{     1.280}{     1.277}{     0.440}
\SfusE{\Soptimal{     26.08}}{\Soptimal{     26.08}}{\Soptimal{     26.08}}{\Soptimal{     26.08}}{\Soptimal{     26.08}}{\Soptimal{     26.08}}
\Sfust{    31.515}{     1.170}{     1.193}{     1.280}{     1.277}{     0.440}
\SgenE{dd-ls3    }{\Soptimal{     26.08}}
\Sgent{     2.032}{    24.450}{     1.647}{     1.668}{     2.268}{     2.266}{     2.231}
\SfusE{\Soptimal{     26.08}}{\Soptimal{     26.08}}{\Soptimal{     26.08}}{\Soptimal{     26.08}}{\Soptimal{     26.08}}{\Soptimal{     26.08}}
\Sfust{    24.450}{     1.647}{     1.668}{     2.268}{     2.266}{     2.231}
\SgenE{dd-ls4    }{\Soptimal{     26.08}}
\Sgent{    15.703}{    34.218}{     8.934}{     8.961}{    15.998}{    16.008}{     9.049}
\SfusE{\Soptimal{     26.08}}{\Soptimal{     26.08}}{\Soptimal{     26.08}}{\Soptimal{     26.08}}{\Soptimal{     26.08}}{\Soptimal{     26.08}}
\Sfust{    34.218}{     8.934}{     8.961}{    15.998}{    16.008}{     9.049}
\SgenE{bca-lap   }{     26.32}
\Sgent{    16.621}{     0.604}{     0.102}{     0.103}{     0.103}{     0.103}{     0.104}
\SfusE{     26.32}{     26.32}{     26.32}{     26.32}{     26.32}{     26.32}
\Sfust{     0.604}{     0.102}{     0.103}{     0.103}{     0.103}{     0.104}
\SgenE{bca-greedy}{\Soptimal{     26.08}}
\Sgent{     2.640}{     0.526}{     0.013}{     0.013}{     0.013}{     0.013}{     0.027}
\SfusE{\Soptimal{     26.08}}{\Soptimal{     26.08}}{\Soptimal{     26.08}}{\Soptimal{     26.08}}{\Soptimal{     26.08}}{\Soptimal{     26.08}}
\Sfust{     0.526}{     0.013}{     0.013}{     0.013}{     0.013}{     0.027}
\SgenE{greedy    }{\Soptimal{     26.08}}
\Sgent{     0.087}{     2.599}{     0.012}{     0.014}{     0.294}{     0.294}{     0.082}
\SfusE{\Soptimal{     26.08}}{\Soptimal{     26.08}}{\Soptimal{     26.08}}{\Soptimal{     26.08}}{\Soptimal{     26.08}}{\Soptimal{     26.08}}
\Sfust{     2.599}{     0.012}{     0.014}{     0.294}{     0.294}{     0.082}
\Sendtable

\Sinstance{opengm3}\Salias{matching3}\Soptimum{15.86}\\
\Sstarttable
\SgenE{dd-ls0    }{\Soptimal{     15.86}}
\Sgent{     1.023}{    54.705}{     1.035}{     1.039}{     1.438}{     1.441}{     1.273}
\SfusE{\Soptimal{     15.86}}{\Soptimal{     15.86}}{\Soptimal{     15.86}}{\Soptimal{     15.86}}{\Soptimal{     15.86}}{\Soptimal{     15.86}}
\Sfust{    54.705}{     1.035}{     1.039}{     1.438}{     1.441}{     1.273}
\SgenE{dd-ls3    }{\Soptimal{     15.86}}
\Sgent{     3.105}{    59.497}{     2.926}{     2.955}{     3.497}{     3.491}{     3.115}
\SfusE{\Soptimal{     15.86}}{\Soptimal{     15.86}}{\Soptimal{     15.86}}{\Soptimal{     15.86}}{\Soptimal{     15.86}}{\Soptimal{     15.86}}
\Sfust{    59.497}{     2.926}{     2.955}{     3.497}{     3.491}{     3.115}
\SgenE{dd-ls4    }{\Soptimal{     15.86}}
\Sgent{    23.475}{    74.308}{    23.409}{     3.624}{    23.865}{    23.869}{    23.562}
\SfusE{\Soptimal{     15.86}}{\Soptimal{     15.86}}{\Soptimal{     15.86}}{\Soptimal{     15.86}}{\Soptimal{     15.86}}{\Soptimal{     15.86}}
\Sfust{    74.308}{    23.409}{     3.624}{    23.865}{    23.869}{    23.562}
\SgenE{bca-lap   }{     77.44}
\Sgent{     0.405}{     0.503}{     0.248}{     0.249}{     0.423}{     0.423}{     0.252}
\SfusE{     76.52}{     77.24}{     74.42}{     77.44}{     77.44}{     76.52}
\Sfust{     0.835}{     0.248}{     0.921}{     0.423}{     0.423}{     0.406}
\SgenE{bca-greedy}{\Soptimal{     15.86}}
\Sgent{     0.094}{     5.234}{     0.105}{     0.109}{     0.010}{     0.010}{     0.372}
\SfusE{\Soptimal{     15.86}}{\Soptimal{     15.86}}{\Soptimal{     15.86}}{\Soptimal{     15.86}}{\Soptimal{     15.86}}{\Soptimal{     15.86}}
\Sfust{     5.234}{     0.105}{     0.109}{     0.010}{     0.010}{     0.372}
\SgenE{greedy    }{\Soptimal{     15.86}}
\Sgent{     0.023}{     2.010}{     0.011}{     0.013}{     0.085}{     0.080}{     0.073}
\SfusE{\Soptimal{     15.86}}{\Soptimal{     15.86}}{\Soptimal{     15.86}}{\Soptimal{     15.86}}{\Soptimal{     15.86}}{\Soptimal{     15.86}}
\Sfust{     2.010}{     0.011}{     0.013}{     0.085}{     0.080}{     0.073}
\Sendtable

\Ssubsection{worms}\label{supp:subsec-worms}

\Snrinst{30}

\Snriter{5000}

\Sinstance{worms1}\Salias{C18G1\_2L1\_1}\Soptimum{-46310.55}\\
\Sstarttable
\SgenE{dd-ls0    }{ -39123.13}
\Sgent{   321.057}{   193.197}{    28.161}{    26.462}{   334.728}{   330.858}{   333.326}
\SfusE{ -46110.89}{ -46140.40}{ -46140.40}{ -39123.13}{ -39123.13}{ -45652.79}
\Sfust{  1428.422}{   327.855}{   332.943}{   334.728}{   330.858}{   454.881}
\SgenE{dd-ls3    }{ -41028.12}
\Sgent{   718.373}{   342.020}{    74.922}{    67.186}{   728.688}{   728.070}{   636.536}
\SfusE{ -46177.87}{ -46197.14}{ -46177.87}{ -41028.12}{ -41028.12}{ -45677.03}
\Sfust{  1581.419}{   763.133}{   612.877}{   728.688}{   728.070}{   916.310}
\SgenE{dd-ls4    }{ -42447.88}
\Sgent{  2993.446}{   591.860}{   388.972}{   250.156}{  3008.302}{  3005.457}{  1707.042}
\SfusE{ -46303.15}{ -46310.40}{ -46297.96}{ -42447.88}{ -42447.88}{ -46044.28}
\Sfust{  4161.920}{  3042.246}{  2659.009}{  3008.302}{  3005.457}{  2926.553}
\SgenE{bca-lap   }{ -46256.37}
\Sgent{   409.225}{    44.866}{    39.980}{    39.901}{   387.634}{   388.060}{   101.367}
\SfusE{ -46284.76}{ -46284.76}{ -46284.76}{ -46256.37}{ -46256.37}{ -46272.53}
\Sfust{   111.940}{   101.069}{   100.836}{   387.634}{   388.060}{   320.515}
\SgenE{bca-greedy}{ -46300.66}
\Sgent{    57.720}{     3.886}{     1.283}{     1.390}{    63.877}{    67.049}{     7.384}
\SfusE{\Soptimal{ -46310.55}}{\Soptimal{ -46310.55}}{\Soptimal{ -46310.55}}{ -46300.66}{ -46300.66}{ -46310.51}
\Sfust{     6.193}{     2.340}{     2.534}{    63.877}{    67.049}{    14.504}
\SgenE{greedy    }{ -43178.01}
\Sgent{     5.367}{     1.448}{     0.028}{     0.029}{     9.583}{     9.624}{     3.698}
\SfusE{\Soptimal{ -46310.55}}{\Soptimal{ -46310.55}}{\Soptimal{ -46310.55}}{ -43178.01}{ -43178.01}{ -45989.36}
\Sfust{   895.717}{    21.984}{    22.712}{     9.583}{     9.624}{   171.191}
\Sendtable

\Sinstance{worms2}\Salias{cnd1threeL1\_1213061}\Soptimum{-49998.17}\\
\Sstarttable
\SgenE{dd-ls0    }{ -48691.75}
\Sgent{   433.962}{   183.566}{    44.886}{    44.330}{   446.760}{   443.881}{   393.566}
\SfusE{\Soptimal{ -49998.17}}{\Soptimal{ -49998.17}}{\Soptimal{ -49998.17}}{ -48691.75}{ -48691.75}{ -49971.36}
\Sfust{  1163.376}{   461.577}{   459.958}{   446.760}{   443.881}{   548.800}
\SgenE{dd-ls3    }{ -48777.45}
\Sgent{   721.248}{   144.342}{   143.115}{    43.783}{   730.584}{   730.996}{   606.015}
\SfusE{\Soptimal{ -49998.17}}{ -49996.36}{ -49996.36}{ -48777.45}{ -48777.45}{ -49985.90}
\Sfust{  1336.586}{   643.136}{   369.809}{   730.584}{   730.996}{   913.895}
\SgenE{dd-ls4    }{ -48596.61}
\Sgent{  2628.949}{   562.969}{   159.621}{   131.331}{  2639.493}{  2643.540}{  1752.037}
\SfusE{\Soptimal{ -49998.17}}{ -49996.36}{ -49996.36}{ -48596.61}{ -48596.61}{ -49978.42}
\Sfust{  1991.114}{  1315.869}{  1316.266}{  2639.493}{  2643.540}{  3079.617}
\SgenE{bca-lap   }{ -49996.82}
\Sgent{    77.999}{    51.714}{    47.628}{    47.546}{    55.456}{    55.441}{    33.200}
\SfusE{ -49996.82}{ -49996.82}{ -49996.82}{ -49996.82}{ -49996.82}{ -49996.82}
\Sfust{    51.714}{    47.628}{    47.546}{    55.456}{    55.441}{    33.200}
\SgenE{bca-greedy}{\Soptimal{ -49998.17}}
\Sgent{   118.332}{   231.913}{   134.989}{   139.474}{   131.642}{   138.316}{   150.605}
\SfusE{\Soptimal{ -49998.17}}{\Soptimal{ -49998.17}}{\Soptimal{ -49998.17}}{\Soptimal{ -49998.17}}{\Soptimal{ -49998.17}}{\Soptimal{ -49998.17}}
\Sfust{   231.913}{   134.989}{   139.474}{   131.642}{   138.316}{   150.605}
\SgenE{greedy    }{ -48177.87}
\Sgent{     8.884}{     1.374}{     0.026}{     0.029}{    15.642}{    15.638}{     1.088}
\SfusE{\Soptimal{ -49998.17}}{\Soptimal{ -49998.17}}{\Soptimal{ -49998.17}}{ -48177.87}{ -48177.87}{ -49870.22}
\Sfust{    12.723}{     0.280}{     0.312}{    15.642}{    15.638}{   113.519}
\Sendtable

\Sinstance{worms3}\Salias{cnd1threeL1\_1228061}\Soptimum{-50553.88}\\
\Sstarttable
\SgenE{dd-ls0    }{ -49392.28}
\Sgent{   474.198}{   136.306}{    32.931}{    35.155}{   487.153}{   485.371}{   415.597}
\SfusE{\Soptimal{ -50553.88}}{ -50553.55}{ -50553.55}{ -49392.28}{ -49392.28}{ -50395.56}
\Sfust{   911.263}{   285.523}{   289.201}{   487.153}{   485.371}{   604.231}
\SgenE{dd-ls3    }{ -49490.58}
\Sgent{   779.423}{   225.968}{    64.694}{    64.836}{   791.110}{   789.975}{   666.117}
\SfusE{ -50553.00}{\Soptimal{ -50553.88}}{ -50553.00}{ -49490.58}{ -49490.58}{ -50527.60}
\Sfust{  1230.522}{   519.134}{   520.292}{   791.110}{   789.975}{   943.460}
\SgenE{dd-ls4    }{ -49900.52}
\Sgent{  2689.057}{   588.075}{   296.214}{   296.717}{  2703.118}{  2704.609}{  2058.103}
\SfusE{\Soptimal{ -50553.88}}{\Soptimal{ -50553.88}}{\Soptimal{ -50553.88}}{ -49900.52}{ -49900.52}{ -50550.83}
\Sfust{  2666.419}{  1868.033}{  1870.549}{  2703.118}{  2704.609}{  3107.303}
\SgenE{bca-lap   }{ -50529.05}
\Sgent{  1206.781}{    16.370}{    14.787}{    14.809}{  1190.562}{  1191.026}{    24.200}
\SfusE{ -50546.78}{ -50546.78}{ -50546.78}{ -50529.05}{ -50529.05}{ -50533.11}
\Sfust{   381.478}{   351.370}{   351.895}{  1190.562}{  1191.026}{    26.193}
\SgenE{bca-greedy}{ -50553.58}
\Sgent{    97.724}{     1.626}{     3.928}{     4.224}{   112.919}{   114.402}{     3.086}
\SfusE{\Soptimal{ -50553.88}}{\Soptimal{ -50553.88}}{\Soptimal{ -50553.88}}{ -50553.58}{ -50553.58}{\Soptimal{ -50553.88}}
\Sfust{     1.885}{     3.928}{     4.224}{   112.919}{   114.402}{     3.086}
\SgenE{greedy    }{ -49365.34}
\Sgent{     3.100}{     1.379}{     0.031}{     0.033}{     5.784}{     5.853}{     0.966}
\SfusE{\Soptimal{ -50553.88}}{\Soptimal{ -50553.88}}{\Soptimal{ -50553.88}}{ -49365.34}{ -49365.34}{ -50486.13}
\Sfust{    48.775}{     1.650}{     1.753}{     5.784}{     5.853}{   109.296}
\Sendtable

\Sinstance{worms4}\Salias{cnd1threeL1\_1229061}\Soptimum{-49113.44}\\
\Sstarttable
\SgenE{dd-ls0    }{ -44484.95}
\Sgent{   488.240}{    62.074}{    12.786}{    11.921}{   501.508}{   497.841}{   389.189}
\SfusE{ -49060.21}{ -49059.06}{ -49059.06}{ -44484.95}{ -44484.95}{ -48512.97}
\Sfust{  1185.184}{   485.719}{   487.959}{   501.508}{   497.841}{   618.574}
\SgenE{dd-ls3    }{ -45623.39}
\Sgent{   702.342}{    95.457}{    22.418}{    21.922}{   712.637}{   712.029}{   588.154}
\SfusE{ -49078.71}{ -49073.97}{ -49071.13}{ -45623.39}{ -45623.39}{ -48748.03}
\Sfust{  1358.181}{   746.631}{   714.440}{   712.637}{   712.029}{   857.982}
\SgenE{dd-ls4    }{ -45598.32}
\Sgent{  2386.137}{   107.118}{    41.621}{    45.346}{  2395.856}{  2395.795}{  1610.971}
\SfusE{ -49078.71}{ -49078.62}{ -49078.62}{ -45598.32}{ -45598.32}{ -48530.96}
\Sfust{  3218.362}{  2096.595}{  2097.337}{  2395.856}{  2395.795}{  2937.980}
\SgenE{bca-lap   }{ -49075.54}
\Sgent{   634.797}{    60.836}{    55.493}{    55.537}{   641.711}{   641.439}{    93.879}
\SfusE{ -49104.71}{ -49096.29}{ -49104.71}{ -49075.54}{ -49075.54}{ -49093.55}
\Sfust{   483.800}{   231.019}{   447.201}{   641.711}{   641.439}{   304.018}
\SgenE{bca-greedy}{ -49110.36}
\Sgent{    61.101}{     1.849}{     0.290}{     0.295}{    70.929}{    67.811}{    12.284}
\SfusE{\Soptimal{ -49113.44}}{\Soptimal{ -49113.44}}{\Soptimal{ -49113.44}}{ -49110.36}{ -49110.36}{ -49113.04}
\Sfust{     2.222}{     0.443}{     0.449}{    70.929}{    67.811}{    36.247}
\SgenE{greedy    }{ -47420.86}
\Sgent{     3.488}{     0.906}{     0.029}{     0.026}{     6.212}{     6.589}{     3.179}
\SfusE{ -49113.35}{ -49113.35}{ -49113.35}{ -47420.86}{ -47420.86}{ -48791.20}
\Sfust{   555.895}{    19.969}{    18.627}{     6.212}{     6.589}{   111.499}
\Sendtable

\Sinstance{worms5}\Salias{cnd1threeL1\_1229062}\Soptimum{-49419.53}\\
\Sstarttable
\SgenE{dd-ls0    }{ -43007.12}
\Sgent{   464.143}{    45.731}{     7.908}{     8.076}{   475.077}{   473.294}{   373.936}
\SfusE{ -49303.37}{ -49299.95}{ -49304.59}{ -43007.12}{ -43007.12}{ -48594.38}
\Sfust{  1206.866}{   432.643}{   430.799}{   475.077}{   473.294}{   595.230}
\SgenE{dd-ls3    }{ -45274.84}
\Sgent{   696.314}{    58.325}{    14.789}{    15.572}{   706.081}{   706.364}{   515.169}
\SfusE{ -49399.09}{ -49399.09}{ -49399.09}{ -45274.84}{ -45274.84}{ -49066.58}
\Sfust{  1304.514}{   583.618}{   585.512}{   706.081}{   706.364}{   805.792}
\SgenE{dd-ls4    }{ -45310.57}
\Sgent{  2592.202}{    93.218}{    33.887}{    33.882}{  2603.710}{  2603.026}{  1301.383}
\SfusE{ -49397.69}{ -49396.50}{ -49396.50}{ -45310.57}{ -45310.57}{ -48882.40}
\Sfust{  3074.849}{  2229.263}{  2228.953}{  2603.710}{  2603.026}{  2640.210}
\SgenE{bca-lap   }{ -49299.77}
\Sgent{   137.259}{    13.077}{    11.080}{    11.095}{   138.979}{   139.023}{    63.365}
\SfusE{ -49405.38}{ -49405.38}{ -49405.38}{ -49299.77}{ -49299.77}{ -49381.19}
\Sfust{   175.107}{   157.892}{   158.107}{   138.979}{   139.023}{    90.794}
\SgenE{bca-greedy}{ -49415.52}
\Sgent{    91.879}{     1.823}{     1.142}{     1.205}{   106.063}{   102.147}{    23.071}
\SfusE{\Soptimal{ -49419.53}}{\Soptimal{ -49419.53}}{\Soptimal{ -49419.53}}{ -49415.52}{ -49415.52}{ -49418.68}
\Sfust{     2.813}{     1.905}{     2.009}{   106.063}{   102.147}{   130.723}
\SgenE{greedy    }{ -47722.06}
\Sgent{    16.188}{     0.754}{     0.020}{     0.022}{    29.763}{    28.813}{     2.074}
\SfusE{ -49412.94}{ -49412.05}{ -49412.05}{ -47722.06}{ -47722.06}{ -49271.68}
\Sfust{   211.444}{     0.374}{     0.381}{    29.763}{    28.813}{   165.332}
\Sendtable

\Sinstance{worms6}\Salias{cnd1threeL1\_1229063}\Soptimum{-50480.36}\\
\Sstarttable
\SgenE{dd-ls0    }{ -48283.46}
\Sgent{   671.578}{   500.338}{   150.355}{   200.262}{   683.117}{   681.605}{   571.469}
\SfusE{ -50427.62}{ -50427.62}{ -50427.62}{ -48283.46}{ -48283.46}{ -50233.49}
\Sfust{  1365.224}{   572.513}{   574.085}{   683.117}{   681.605}{   800.780}
\SgenE{dd-ls3    }{ -49059.67}
\Sgent{   690.313}{   876.681}{   278.411}{   279.379}{   699.672}{   699.937}{   646.841}
\SfusE{ -50430.25}{ -50430.25}{ -50430.25}{ -49059.67}{ -49059.67}{ -50303.85}
\Sfust{  1525.641}{   663.222}{   665.131}{   699.672}{   699.937}{   844.897}
\SgenE{dd-ls4    }{ -50283.72}
\Sgent{  3436.307}{   987.044}{   523.262}{   524.222}{  3451.071}{  3453.746}{  2001.204}
\SfusE{\Soptimal{ -50480.36}}{\Soptimal{ -50480.36}}{ -50479.70}{ -50283.72}{ -50283.72}{ -50471.68}
\Sfust{  2854.578}{  2079.596}{  2526.645}{  3451.071}{  3453.746}{  3636.511}
\SgenE{bca-lap   }{ -50460.50}
\Sgent{    24.279}{    20.296}{    18.430}{    18.462}{    24.554}{    24.547}{    18.761}
\SfusE{ -50476.63}{ -50476.63}{ -50476.63}{ -50460.50}{ -50460.50}{ -50476.63}
\Sfust{   141.772}{   131.006}{   131.197}{    24.554}{    24.547}{   141.486}
\SgenE{bca-greedy}{\Soptimal{ -50480.36}}
\Sgent{    85.555}{   152.219}{    90.222}{    89.574}{    92.567}{    92.898}{    96.366}
\SfusE{\Soptimal{ -50480.36}}{\Soptimal{ -50480.36}}{\Soptimal{ -50480.36}}{\Soptimal{ -50480.36}}{\Soptimal{ -50480.36}}{\Soptimal{ -50480.36}}
\Sfust{   152.219}{    90.222}{    89.574}{    92.567}{    92.898}{    96.366}
\SgenE{greedy    }{ -49203.46}
\Sgent{     1.870}{     1.236}{     0.034}{     0.036}{     3.335}{     4.245}{     1.908}
\SfusE{\Soptimal{ -50480.36}}{\Soptimal{ -50480.36}}{\Soptimal{ -50480.36}}{ -49203.46}{ -49203.46}{ -50334.19}
\Sfust{    60.803}{     1.784}{     1.911}{     3.335}{     4.245}{   113.826}
\Sendtable

\Sinstance{worms7}\Salias{eft3RW10035L1\_0125071}\Soptimum{-47057.33}\\
\Sstarttable
\SgenE{dd-ls0    }{ -39794.13}
\Sgent{   603.511}{    52.171}{    11.520}{    11.605}{   614.393}{   612.429}{   530.374}
\SfusE{ -46693.42}{ -46720.21}{ -46720.21}{ -39794.13}{ -39794.13}{ -44624.72}
\Sfust{  1295.296}{   470.684}{   473.115}{   614.393}{   612.429}{   728.913}
\SgenE{dd-ls3    }{ -43972.38}
\Sgent{   837.726}{   132.285}{    27.478}{    26.628}{   850.431}{   849.992}{   674.645}
\SfusE{ -47035.51}{ -46923.69}{ -46919.71}{ -43972.38}{ -43972.38}{ -46397.24}
\Sfust{  1682.279}{   712.851}{   750.217}{   850.431}{   849.992}{   917.964}
\SgenE{dd-ls4    }{ -42646.19}
\Sgent{  2598.876}{   151.150}{    38.390}{    55.939}{  2610.883}{  2609.511}{  1734.126}
\SfusE{ -46945.80}{ -46945.80}{ -46945.80}{ -42646.19}{ -42646.19}{ -46301.98}
\Sfust{  2747.707}{  1826.032}{  1825.613}{  2610.883}{  2609.511}{  2625.940}
\SgenE{bca-lap   }{ -46964.53}
\Sgent{   515.751}{    65.346}{    57.485}{    57.596}{   520.940}{   521.843}{   190.737}
\SfusE{ -46988.44}{ -46988.44}{ -46988.44}{ -46964.53}{ -46964.53}{ -46966.99}
\Sfust{   119.083}{   106.142}{   106.327}{   520.940}{   521.843}{   190.737}
\SgenE{bca-greedy}{ -47030.24}
\Sgent{    24.571}{     5.859}{     0.602}{     0.645}{    27.239}{    28.392}{    19.101}
\SfusE{\Soptimal{ -47057.33}}{ -47057.10}{ -47057.10}{ -47030.24}{ -47030.24}{ -47057.10}
\Sfust{    21.306}{     6.776}{     7.319}{    27.239}{    28.392}{    70.863}
\SgenE{greedy    }{ -44561.77}
\Sgent{     2.633}{     1.578}{     0.028}{     0.029}{     4.802}{     6.017}{     3.368}
\SfusE{ -47050.21}{ -47050.21}{ -47050.21}{ -44561.77}{ -44561.77}{ -46716.50}
\Sfust{    46.300}{     0.960}{     0.963}{     4.802}{     6.017}{   178.876}
\Sendtable

\Sinstance{worms8}\Salias{eft3RW10035L1\_0125072}\Soptimum{-49244.01}\\
\Sstarttable
\SgenE{dd-ls0    }{ -48177.51}
\Sgent{   660.238}{   500.325}{   161.867}{   162.228}{   669.858}{   672.693}{   545.689}
\SfusE{ -49243.90}{ -49243.61}{ -49243.61}{ -48177.51}{ -48177.51}{ -49171.54}
\Sfust{  1397.104}{   509.709}{   510.694}{   669.858}{   672.693}{   810.971}
\SgenE{dd-ls3    }{ -48598.61}
\Sgent{   767.035}{   219.224}{    78.717}{    78.820}{   776.301}{   776.267}{   620.616}
\SfusE{ -49243.72}{ -49243.72}{ -49243.72}{ -48598.61}{ -48598.61}{ -49223.78}
\Sfust{  1066.395}{   519.570}{   520.259}{   776.301}{   776.267}{  1005.263}
\SgenE{dd-ls4    }{ -48640.11}
\Sgent{  2729.297}{   366.882}{   159.708}{   200.834}{  2740.769}{  2741.423}{  1919.238}
\SfusE{\Soptimal{ -49244.01}}{ -49243.72}{ -49243.72}{ -48640.11}{ -48640.11}{ -49240.92}
\Sfust{  2815.227}{  2092.486}{  2093.290}{  2740.769}{  2741.423}{  2956.309}
\SgenE{bca-lap   }{ -49221.47}
\Sgent{  1269.452}{   105.568}{  1208.500}{  1207.127}{  1206.569}{  1206.537}{  1223.603}
\SfusE{ -49229.38}{ -49221.47}{ -49221.47}{ -49221.47}{ -49221.47}{ -49221.47}
\Sfust{   105.568}{  1208.500}{  1207.127}{  1206.569}{  1206.537}{  1223.603}
\SgenE{bca-greedy}{ -49241.94}
\Sgent{    24.932}{     2.181}{     0.824}{     0.778}{    28.453}{    28.946}{    11.729}
\SfusE{\Soptimal{ -49244.01}}{\Soptimal{ -49244.01}}{\Soptimal{ -49244.01}}{ -49241.94}{ -49241.94}{ -49243.21}
\Sfust{    24.259}{     1.853}{     1.749}{    28.453}{    28.946}{    12.042}
\SgenE{greedy    }{ -48063.58}
\Sgent{     3.477}{     0.770}{     0.025}{     0.024}{     6.496}{     7.802}{     2.173}
\SfusE{\Soptimal{ -49244.01}}{\Soptimal{ -49244.01}}{\Soptimal{ -49244.01}}{ -48063.58}{ -48063.58}{ -49133.64}
\Sfust{   195.644}{     2.736}{     2.527}{     6.496}{     7.802}{   150.250}
\Sendtable

\Sinstance{worms9}\Salias{eft3RW10035L1\_0125073}\Soptimum{-45149.41}\\
\Sstarttable
\SgenE{dd-ls0    }{ -34536.21}
\Sgent{   657.078}{    44.690}{     9.047}{     7.792}{   666.505}{   667.321}{   635.141}
\SfusE{ -44627.82}{ -44628.12}{ -44628.12}{ -34536.21}{ -34536.21}{ -41035.73}
\Sfust{  1733.546}{   661.378}{   661.607}{   666.505}{   667.321}{   782.214}
\SgenE{dd-ls3    }{ -40047.61}
\Sgent{   817.318}{   203.074}{    42.874}{    43.822}{   828.860}{   827.511}{   693.818}
\SfusE{ -45120.72}{ -45120.72}{ -45120.72}{ -40047.61}{ -40047.61}{ -44212.99}
\Sfust{  1583.269}{   611.956}{   614.296}{   828.860}{   827.511}{  1005.641}
\SgenE{dd-ls4    }{ -40471.85}
\Sgent{  2853.635}{   287.657}{   194.102}{   137.304}{  2867.155}{  2868.561}{  1791.886}
\SfusE{ -45120.72}{ -45120.72}{ -45120.72}{ -40471.85}{ -40471.85}{ -44661.73}
\Sfust{  3187.456}{  1674.704}{  2216.292}{  2867.155}{  2868.561}{  3176.951}
\SgenE{bca-lap   }{ -45096.98}
\Sgent{   208.579}{    45.191}{    41.282}{    41.367}{   211.261}{   210.902}{    91.004}
\SfusE{ -45147.73}{ -45147.73}{ -45147.73}{ -45096.98}{ -45096.98}{ -45147.24}
\Sfust{   642.538}{   601.305}{   602.439}{   211.261}{   210.902}{   612.540}
\SgenE{bca-greedy}{ -45149.13}
\Sgent{    90.037}{     6.980}{     2.210}{     2.362}{    99.725}{   105.187}{   115.300}
\SfusE{\Soptimal{ -45149.41}}{\Soptimal{ -45149.41}}{\Soptimal{ -45149.41}}{ -45149.13}{ -45149.13}{\Soptimal{ -45149.41}}
\Sfust{     8.329}{     2.761}{     2.950}{    99.725}{   105.187}{   136.032}
\SgenE{greedy    }{ -41403.04}
\Sgent{     9.670}{     2.405}{     0.046}{     0.042}{    17.425}{    17.739}{     6.730}
\SfusE{\Soptimal{ -45149.41}}{\Soptimal{ -45149.41}}{\Soptimal{ -45149.41}}{ -41403.04}{ -41403.04}{ -44427.72}
\Sfust{  1202.275}{    10.479}{     9.992}{    17.425}{    17.739}{   241.062}
\Sendtable

\Sinstance{worms10}\Salias{egl5L1\_0606074}\\
\Sstarttable
\SgenE{dd-ls0    }{ -14345.06}
\Sgent{   567.829}{    20.224}{     2.938}{     2.675}{   581.041}{   577.744}{   296.904}
\SfusE{ -41973.41}{ -41862.94}{ -41865.87}{ -14345.06}{ -14345.06}{ -32929.86}
\Sfust{  1811.659}{   573.637}{   572.878}{   581.041}{   577.744}{   690.105}
\SgenE{dd-ls3    }{ -23433.16}
\Sgent{   844.026}{    37.442}{     5.366}{     5.380}{   854.295}{   853.555}{   530.818}
\SfusE{ -42070.04}{ -42062.85}{ -42065.74}{ -23433.16}{ -23433.16}{ -35760.78}
\Sfust{  1932.712}{   794.207}{   686.877}{   854.295}{   853.555}{  1003.357}
\SgenE{dd-ls4    }{ -21041.73}
\Sgent{  3215.666}{    38.575}{     8.079}{     8.100}{  3226.837}{  3226.587}{   903.208}
\SfusE{ -42115.59}{ -42069.53}{ -42117.07}{ -21041.73}{ -21041.73}{ -37669.83}
\Sfust{  3377.292}{  2218.603}{  2283.128}{  3226.837}{  3226.587}{  3459.221}
\SgenE{bca-lap   }{ -41090.97}
\Sgent{   281.695}{    12.951}{    11.215}{    11.228}{   284.810}{   284.935}{   270.344}
\SfusE{ -41868.42}{ -41868.42}{ -41868.42}{ -41090.97}{ -41090.97}{ -41421.86}
\Sfust{   323.794}{   291.647}{   291.941}{   284.810}{   284.935}{   734.324}
\SgenE{bca-greedy}{ -42227.79}
\Sgent{    31.501}{     5.699}{     3.474}{     3.537}{    35.196}{    37.163}{    18.598}
\SfusE{ -42383.83}{ -42383.83}{ -42383.83}{ -42227.79}{ -42227.79}{ -42372.76}
\Sfust{    52.896}{    24.314}{    24.707}{    35.196}{    37.163}{    94.162}
\SgenE{greedy    }{ -38024.79}
\Sgent{    16.459}{     1.739}{     0.021}{     0.027}{    30.949}{    31.835}{     1.154}
\SfusE{ -42306.99}{ -42280.37}{ -42279.90}{ -38024.79}{ -38024.79}{ -40770.96}
\Sfust{  1017.486}{    13.184}{     2.996}{    30.949}{    31.835}{   277.948}
\Sendtable

\Sinstance{worms11}\Salias{elt3L1\_0503071}\Soptimum{-48664.98}\\
\Sstarttable
\SgenE{dd-ls0    }{ -44518.81}
\Sgent{   371.537}{   186.565}{    30.376}{    29.024}{   386.940}{   385.492}{   352.548}
\SfusE{ -48570.89}{ -48570.00}{ -48570.89}{ -44518.81}{ -44518.81}{ -48123.06}
\Sfust{  1235.114}{   242.879}{   295.715}{   386.940}{   385.492}{   504.517}
\SgenE{dd-ls3    }{ -45354.56}
\Sgent{   800.895}{   226.259}{    72.218}{    97.410}{   811.212}{   810.712}{   614.360}
\SfusE{ -48570.89}{ -48572.85}{ -48572.85}{ -45354.56}{ -45354.56}{ -47941.38}
\Sfust{  1192.565}{   437.533}{   437.471}{   811.212}{   810.712}{   958.440}
\SgenE{dd-ls4    }{ -46115.16}
\Sgent{  3006.709}{   200.185}{    78.713}{    79.370}{  3021.220}{  3023.128}{  1842.742}
\SfusE{ -48570.73}{ -48570.73}{ -48570.73}{ -46115.16}{ -46115.16}{ -48420.15}
\Sfust{  2543.062}{  2382.926}{  2383.999}{  3021.220}{  3023.128}{  3202.194}
\SgenE{bca-lap   }{ -48651.14}
\Sgent{    94.703}{    21.995}{    19.904}{    19.927}{    93.939}{    93.851}{    95.020}
\SfusE{ -48660.25}{ -48660.25}{ -48660.25}{ -48651.14}{ -48651.14}{ -48652.79}
\Sfust{  1060.917}{   984.426}{   985.593}{    93.939}{    93.851}{   190.793}
\SgenE{bca-greedy}{ -48661.44}
\Sgent{   118.429}{     2.073}{     6.148}{     6.419}{   131.927}{   139.353}{    10.576}
\SfusE{\Soptimal{ -48664.98}}{\Soptimal{ -48664.98}}{\Soptimal{ -48664.98}}{ -48661.44}{ -48661.44}{\Soptimal{ -48664.98}}
\Sfust{     2.744}{     6.465}{     6.747}{   131.927}{   139.353}{    66.810}
\SgenE{greedy    }{ -46929.79}
\Sgent{    10.616}{     1.735}{     0.039}{     0.041}{    19.209}{    19.290}{     1.257}
\SfusE{\Soptimal{ -48664.98}}{\Soptimal{ -48664.98}}{\Soptimal{ -48664.98}}{ -46929.79}{ -46929.79}{ -48404.61}
\Sfust{   131.724}{     1.723}{     1.865}{    19.209}{    19.290}{   162.345}
\Sendtable

\Sinstance{worms12}\Salias{elt3L1\_0503072}\Soptimum{-50403.60}\\
\Sstarttable
\SgenE{dd-ls0    }{ -46749.89}
\Sgent{   324.732}{    98.627}{    18.023}{    18.328}{   337.450}{   336.679}{   285.491}
\SfusE{ -50344.86}{ -50332.70}{ -50344.86}{ -46749.89}{ -46749.89}{ -50031.73}
\Sfust{   992.719}{   254.068}{   257.863}{   337.450}{   336.679}{   448.001}
\SgenE{dd-ls3    }{ -48038.89}
\Sgent{   970.963}{   379.208}{    59.019}{    31.284}{   982.077}{   984.390}{   753.755}
\SfusE{\Soptimal{ -50403.60}}{\Soptimal{ -50403.60}}{\Soptimal{ -50403.60}}{ -48038.89}{ -48038.89}{ -50252.17}
\Sfust{  1789.850}{   921.928}{   921.454}{   982.077}{   984.390}{  1083.110}
\SgenE{dd-ls4    }{ -47601.36}
\Sgent{  2475.406}{   233.266}{    91.838}{    91.431}{  2485.868}{  2489.959}{  1401.733}
\SfusE{ -50403.56}{\Soptimal{ -50403.60}}{\Soptimal{ -50403.60}}{ -47601.36}{ -47601.36}{ -50299.94}
\Sfust{  3418.181}{  2045.375}{  2045.618}{  2485.868}{  2489.959}{  2705.931}
\SgenE{bca-lap   }{ -50395.02}
\Sgent{   893.992}{    44.785}{    41.266}{    41.291}{   884.422}{   882.254}{    83.866}
\SfusE{ -50403.36}{ -50403.36}{ -50403.36}{ -50395.02}{ -50395.02}{ -50401.25}
\Sfust{   244.924}{   221.908}{   221.735}{   884.422}{   882.254}{   146.041}
\SgenE{bca-greedy}{\Soptimal{ -50403.60}}
\Sgent{    38.415}{     5.290}{     1.857}{     1.994}{    44.158}{    44.478}{    12.122}
\SfusE{\Soptimal{ -50403.60}}{\Soptimal{ -50403.60}}{\Soptimal{ -50403.60}}{\Soptimal{ -50403.60}}{\Soptimal{ -50403.60}}{\Soptimal{ -50403.60}}
\Sfust{     5.290}{     1.857}{     1.994}{    44.158}{    44.478}{    12.122}
\SgenE{greedy    }{ -49064.36}
\Sgent{     6.267}{     1.743}{     0.033}{     0.036}{    12.425}{    12.581}{     1.123}
\SfusE{\Soptimal{ -50403.60}}{\Soptimal{ -50403.60}}{\Soptimal{ -50403.60}}{ -49064.36}{ -49064.36}{ -50250.40}
\Sfust{    62.359}{     1.309}{     1.395}{    12.425}{    12.581}{    90.020}
\Sendtable

\Sinstance{worms13}\Salias{elt3L1\_0504073}\Soptimum{-45831.06}\\
\Sstarttable
\SgenE{dd-ls0    }{ -40499.68}
\Sgent{   359.235}{   194.890}{    33.383}{    33.658}{   370.886}{   370.494}{   375.823}
\SfusE{ -45672.10}{ -45672.10}{ -45672.10}{ -40499.68}{ -40499.68}{ -44956.88}
\Sfust{  1317.725}{   302.841}{   306.001}{   370.886}{   370.494}{   538.387}
\SgenE{dd-ls3    }{ -43689.43}
\Sgent{   941.279}{   220.272}{    58.228}{    57.358}{   952.579}{   954.577}{   798.540}
\SfusE{ -45829.01}{ -45829.01}{ -45829.01}{ -43689.43}{ -43689.43}{ -45767.96}
\Sfust{  1603.918}{   740.253}{   764.143}{   952.579}{   954.577}{  1109.242}
\SgenE{dd-ls4    }{ -44253.29}
\Sgent{  3063.117}{  1595.778}{   886.715}{   827.077}{  3074.967}{  3079.116}{  2056.076}
\SfusE{\Soptimal{ -45831.06}}{ -45829.01}{ -45829.01}{ -44253.29}{ -44253.29}{ -45768.68}
\Sfust{  2973.587}{  1807.671}{  2057.633}{  3074.967}{  3079.116}{  3369.290}
\SgenE{bca-lap   }{ -45828.00}
\Sgent{   122.761}{   111.869}{   101.978}{   102.043}{   124.016}{   124.045}{   126.057}
\SfusE{ -45829.75}{ -45829.75}{ -45829.75}{ -45828.00}{ -45828.00}{ -45829.60}
\Sfust{   218.605}{   200.804}{   200.944}{   124.016}{   124.045}{   203.294}
\SgenE{bca-greedy}{ -45830.99}
\Sgent{    14.787}{     6.045}{     1.998}{     1.934}{    17.210}{    16.491}{    13.828}
\SfusE{\Soptimal{ -45831.06}}{\Soptimal{ -45831.06}}{\Soptimal{ -45831.06}}{ -45830.99}{ -45830.99}{\Soptimal{ -45831.06}}
\Sfust{     6.045}{     1.998}{     1.934}{    17.210}{    16.491}{    13.828}
\SgenE{greedy    }{ -43207.33}
\Sgent{    15.413}{     1.075}{     0.026}{     0.026}{    29.916}{    29.111}{     3.227}
\SfusE{\Soptimal{ -45831.06}}{\Soptimal{ -45831.06}}{\Soptimal{ -45831.06}}{ -43207.33}{ -43207.33}{ -45190.00}
\Sfust{    71.184}{     1.844}{     1.812}{    29.916}{    29.111}{   201.481}
\Sendtable

\Sinstance{worms14}\Salias{hlh1fourL1\_0417071}\\
\Sstarttable
\SgenE{dd-ls0    }{ -41614.43}
\Sgent{   336.211}{    71.000}{    15.230}{    12.973}{   345.040}{   348.163}{   324.409}
\SfusE{ -46876.07}{ -46874.13}{ -46910.61}{ -41614.43}{ -41614.43}{ -46394.62}
\Sfust{  1226.721}{   386.777}{   386.181}{   345.040}{   348.163}{   504.794}
\SgenE{dd-ls3    }{ -42272.18}
\Sgent{   850.338}{   110.057}{    25.214}{    28.555}{   859.858}{   860.029}{   664.945}
\SfusE{ -46944.34}{ -46933.83}{ -46954.37}{ -42272.18}{ -42272.18}{ -46494.31}
\Sfust{  1683.886}{   725.473}{   743.403}{   859.858}{   860.029}{  1007.572}
\SgenE{dd-ls4    }{ -42534.80}
\Sgent{  2551.090}{   119.307}{    43.140}{    42.561}{  2562.390}{  2565.698}{  1558.107}
\SfusE{ -46934.41}{ -46940.86}{ -46985.92}{ -42534.80}{ -42534.80}{ -46384.22}
\Sfust{  3192.512}{  2689.922}{  2260.036}{  2562.390}{  2565.698}{  2851.542}
\SgenE{bca-lap   }{ -46380.32}
\Sgent{    15.205}{     5.160}{     3.986}{     3.992}{    15.621}{    15.606}{     6.775}
\SfusE{ -46935.06}{ -46899.80}{ -46935.06}{ -46380.32}{ -46380.32}{ -46802.40}
\Sfust{  1016.979}{   851.135}{   852.242}{    15.621}{    15.606}{   320.721}
\SgenE{bca-greedy}{ -46917.99}
\Sgent{    19.035}{     1.181}{     0.240}{     0.240}{    22.522}{    22.434}{     6.330}
\SfusE{ -46998.31}{ -46997.81}{ -46997.81}{ -46917.99}{ -46917.99}{ -46970.89}
\Sfust{    16.733}{     5.338}{    22.636}{    22.522}{    22.434}{   154.879}
\SgenE{greedy    }{ -44273.01}
\Sgent{    15.707}{     2.219}{     0.024}{     0.027}{    32.248}{    32.484}{     3.139}
\SfusE{ -46998.31}{ -46991.91}{ -46988.32}{ -44273.01}{ -44273.01}{ -46248.91}
\Sfust{  1427.409}{    24.485}{    26.625}{    32.248}{    32.484}{   180.703}
\Sendtable

\Sinstance{worms15}\Salias{hlh1fourL1\_0417075}\Soptimum{-49550.13}\\
\Sstarttable
\SgenE{dd-ls0    }{ -48073.05}
\Sgent{   431.349}{   200.535}{    40.867}{    42.001}{   444.829}{   442.908}{   377.551}
\SfusE{ -49546.59}{ -49549.31}{ -49549.31}{ -48073.05}{ -48073.05}{ -49419.17}
\Sfust{  1212.633}{   430.109}{   429.186}{   444.829}{   442.908}{   550.213}
\SgenE{dd-ls3    }{ -48216.50}
\Sgent{   942.422}{  1036.523}{   368.348}{    64.595}{   953.849}{   953.072}{   778.474}
\SfusE{\Soptimal{ -49550.13}}{\Soptimal{ -49550.13}}{\Soptimal{ -49550.13}}{ -48216.50}{ -48216.50}{ -49476.63}
\Sfust{  1508.307}{   546.803}{   648.759}{   953.849}{   953.072}{  1114.500}
\SgenE{dd-ls4    }{ -48397.90}
\Sgent{  2954.615}{   243.255}{    97.561}{    97.575}{  2966.946}{  2966.053}{  1742.250}
\SfusE{ -49546.59}{ -49546.59}{ -49546.59}{ -48397.90}{ -48397.90}{ -49535.50}
\Sfust{  2189.294}{  1463.352}{  1463.244}{  2966.946}{  2966.053}{  3218.521}
\SgenE{bca-lap   }{ -49548.60}
\Sgent{    61.130}{    55.176}{    50.089}{    50.158}{    61.720}{    61.716}{    62.352}
\SfusE{ -49549.12}{ -49549.12}{ -49549.12}{ -49548.60}{ -49548.60}{ -49548.69}
\Sfust{    61.781}{    56.206}{    56.282}{    61.720}{    61.716}{    71.390}
\SgenE{bca-greedy}{\Soptimal{ -49550.13}}
\Sgent{    34.218}{     1.667}{     0.482}{     0.516}{    37.988}{    39.566}{    12.545}
\SfusE{\Soptimal{ -49550.13}}{\Soptimal{ -49550.13}}{\Soptimal{ -49550.13}}{\Soptimal{ -49550.13}}{\Soptimal{ -49550.13}}{\Soptimal{ -49550.13}}
\Sfust{     1.667}{     0.482}{     0.516}{    37.988}{    39.566}{    12.545}
\SgenE{greedy    }{ -48152.12}
\Sgent{     9.219}{     1.465}{     0.028}{     0.031}{    18.553}{    23.397}{     2.495}
\SfusE{\Soptimal{ -49550.13}}{\Soptimal{ -49550.13}}{\Soptimal{ -49550.13}}{ -48152.12}{ -48152.12}{ -49398.55}
\Sfust{    55.797}{     1.244}{     1.411}{    18.553}{    23.397}{    75.286}
\Sendtable

\Sinstance{worms16}\Salias{hlh1fourL1\_0417076}\Soptimum{-48404.21}\\
\Sstarttable
\SgenE{dd-ls0    }{ -45279.31}
\Sgent{   574.848}{   154.722}{    30.553}{    30.427}{   587.111}{   589.333}{   500.795}
\SfusE{ -48369.03}{ -48369.03}{ -48369.03}{ -45279.31}{ -45279.31}{ -48117.97}
\Sfust{  1406.497}{   497.998}{   496.327}{   587.111}{   589.333}{   702.471}
\SgenE{dd-ls3    }{ -46037.61}
\Sgent{   981.553}{   235.150}{    68.346}{    58.978}{   993.652}{   992.896}{   815.875}
\SfusE{ -48400.22}{ -48400.22}{ -48401.68}{ -46037.61}{ -46037.61}{ -48189.36}
\Sfust{  1540.525}{   726.465}{   735.746}{   993.652}{   992.896}{  1131.073}
\SgenE{dd-ls4    }{ -45686.01}
\Sgent{  2752.931}{   485.012}{   221.258}{   221.432}{  2765.051}{  2764.773}{  1643.619}
\SfusE{ -48399.63}{ -48402.56}{ -48399.63}{ -45686.01}{ -45686.01}{ -48203.34}
\Sfust{  2921.456}{  1810.010}{  2087.837}{  2765.051}{  2764.773}{  2804.854}
\SgenE{bca-lap   }{ -48374.00}
\Sgent{    55.136}{    14.647}{    13.021}{    13.035}{    52.951}{    52.950}{    50.550}
\SfusE{ -48394.35}{ -48394.35}{ -48394.35}{ -48374.00}{ -48374.00}{ -48386.57}
\Sfust{   103.133}{    96.438}{    96.535}{    52.951}{    52.950}{   131.679}
\SgenE{bca-greedy}{ -48402.03}
\Sgent{    66.882}{     1.947}{     0.450}{     0.480}{    77.764}{    77.515}{     9.561}
\SfusE{\Soptimal{ -48404.21}}{\Soptimal{ -48404.21}}{\Soptimal{ -48404.21}}{ -48402.03}{ -48402.03}{\Soptimal{ -48404.21}}
\Sfust{     4.424}{     0.474}{     1.450}{    77.764}{    77.515}{    78.406}
\SgenE{greedy    }{ -46662.31}
\Sgent{     5.618}{     0.806}{     0.026}{     0.025}{    10.540}{    10.709}{     1.700}
\SfusE{\Soptimal{ -48404.21}}{\Soptimal{ -48404.21}}{\Soptimal{ -48404.21}}{ -46662.31}{ -46662.31}{ -48066.10}
\Sfust{    77.048}{     2.146}{     2.087}{    10.540}{    10.709}{   150.384}
\Sendtable

\Sinstance{worms17}\Salias{hlh1fourL1\_0417077}\Soptimum{-48071.87}\\
\Sstarttable
\SgenE{dd-ls0    }{ -44616.42}
\Sgent{   563.241}{   125.910}{    32.221}{    32.130}{   576.165}{   575.941}{   472.923}
\SfusE{ -48033.92}{ -48033.92}{ -48033.92}{ -44616.42}{ -44616.42}{ -47812.45}
\Sfust{  1407.484}{   518.885}{   518.750}{   576.165}{   575.941}{   707.418}
\SgenE{dd-ls3    }{ -46726.05}
\Sgent{   900.512}{   236.826}{   192.832}{    78.293}{   911.375}{   910.446}{   827.630}
\SfusE{ -48069.66}{ -48069.66}{ -48069.66}{ -46726.05}{ -46726.05}{ -48032.87}
\Sfust{  1702.243}{   819.555}{   819.120}{   911.375}{   910.446}{  1130.075}
\SgenE{dd-ls4    }{ -46617.70}
\Sgent{  2628.325}{  1331.204}{   717.928}{   717.964}{  2642.988}{  2642.184}{  1674.106}
\SfusE{\Soptimal{ -48071.87}}{\Soptimal{ -48071.87}}{\Soptimal{ -48071.87}}{ -46617.70}{ -46617.70}{ -48056.86}
\Sfust{  3086.905}{  2253.888}{  2254.393}{  2642.988}{  2642.184}{  2988.966}
\SgenE{bca-lap   }{ -48061.13}
\Sgent{    60.817}{    36.966}{    34.029}{    34.091}{    59.008}{    59.008}{    59.783}
\SfusE{ -48063.70}{ -48063.70}{ -48063.70}{ -48061.13}{ -48061.13}{ -48061.13}
\Sfust{    43.520}{    40.199}{    40.271}{    59.008}{    59.008}{    59.783}
\SgenE{bca-greedy}{ -48070.43}
\Sgent{   123.153}{     2.319}{     0.721}{     0.691}{   141.938}{   142.147}{     4.839}
\SfusE{\Soptimal{ -48071.87}}{\Soptimal{ -48071.87}}{\Soptimal{ -48071.87}}{ -48070.43}{ -48070.43}{\Soptimal{ -48071.87}}
\Sfust{     3.374}{     1.154}{     1.102}{   141.938}{   142.147}{     4.839}
\SgenE{greedy    }{ -45567.88}
\Sgent{     7.136}{     1.852}{     0.030}{     0.033}{    13.381}{    13.875}{     1.769}
\SfusE{ -48070.43}{ -48070.43}{ -48070.43}{ -45567.88}{ -45567.88}{ -47732.29}
\Sfust{    92.963}{     5.884}{     6.483}{    13.381}{    13.875}{   141.611}
\Sendtable

\Sinstance{worms18}\Salias{hlh1fourL1\_0417078}\Soptimum{-48236.29}\\
\Sstarttable
\SgenE{dd-ls0    }{ -45890.44}
\Sgent{   578.063}{   156.813}{    87.834}{    87.372}{   588.644}{   590.431}{   435.204}
\SfusE{\Soptimal{ -48236.29}}{ -48220.25}{ -48220.25}{ -45890.44}{ -45890.44}{ -47999.49}
\Sfust{  1277.060}{   559.718}{   556.304}{   588.644}{   590.431}{   693.465}
\SgenE{dd-ls3    }{ -46428.64}
\Sgent{   920.498}{   143.049}{    35.872}{    35.949}{   930.436}{   931.632}{   731.400}
\SfusE{ -48235.35}{ -48235.35}{\Soptimal{ -48236.29}}{ -46428.64}{ -46428.64}{ -48167.60}
\Sfust{  1780.136}{   962.231}{   964.373}{   930.436}{   931.632}{  1144.844}
\SgenE{dd-ls4    }{ -47145.61}
\Sgent{  3177.206}{   225.837}{    95.117}{    95.206}{  3194.189}{  3191.990}{  1953.547}
\SfusE{ -48235.35}{ -48235.35}{ -48235.35}{ -47145.61}{ -47145.61}{ -48207.49}
\Sfust{  2885.869}{  2181.460}{  2182.504}{  3194.189}{  3191.990}{  3278.166}
\SgenE{bca-lap   }{ -48228.98}
\Sgent{   131.208}{   121.552}{   111.592}{   111.695}{   123.373}{   123.224}{   113.129}
\SfusE{ -48230.22}{ -48230.22}{ -48230.22}{ -48228.98}{ -48228.98}{ -48230.22}
\Sfust{   134.097}{   123.186}{   123.304}{   123.373}{   123.224}{   124.894}
\SgenE{bca-greedy}{\Soptimal{ -48236.29}}
\Sgent{   112.496}{   218.989}{   123.585}{   130.487}{   128.852}{   131.115}{   133.720}
\SfusE{\Soptimal{ -48236.29}}{\Soptimal{ -48236.29}}{\Soptimal{ -48236.29}}{\Soptimal{ -48236.29}}{\Soptimal{ -48236.29}}{\Soptimal{ -48236.29}}
\Sfust{   218.989}{   123.585}{   130.487}{   128.852}{   131.115}{   133.720}
\SgenE{greedy    }{ -45826.53}
\Sgent{    11.204}{     1.127}{     0.017}{     0.020}{    21.605}{    22.383}{     1.295}
\SfusE{\Soptimal{ -48236.29}}{\Soptimal{ -48236.29}}{\Soptimal{ -48236.29}}{ -45826.53}{ -45826.53}{ -47961.24}
\Sfust{    40.263}{     0.943}{     0.934}{    21.605}{    22.383}{   151.662}
\Sendtable

\Sinstance{worms19}\Salias{mir61L1\_1228061}\Soptimum{-48787.16}\\
\Sstarttable
\SgenE{dd-ls0    }{ -46966.78}
\Sgent{   582.599}{   250.380}{    39.892}{    39.701}{   593.239}{   593.210}{   479.445}
\SfusE{ -48743.73}{ -48743.73}{ -48743.73}{ -46966.78}{ -46966.78}{ -48613.11}
\Sfust{  1096.605}{   560.534}{   559.544}{   593.239}{   593.210}{   705.778}
\SgenE{dd-ls3    }{ -46939.28}
\Sgent{   914.664}{   380.521}{    52.993}{    71.013}{   925.017}{   925.159}{   755.238}
\SfusE{ -48743.47}{ -48743.71}{ -48743.71}{ -46939.28}{ -46939.28}{ -48540.26}
\Sfust{  1801.899}{   530.417}{   531.176}{   925.017}{   925.159}{  1104.888}
\SgenE{dd-ls4    }{ -47265.20}
\Sgent{  3149.328}{   437.093}{   225.855}{    99.619}{  3165.845}{  3160.819}{  1858.809}
\SfusE{ -48761.71}{ -48761.71}{ -48761.71}{ -47265.20}{ -47265.20}{ -48634.51}
\Sfust{  2748.766}{  2066.410}{  2069.274}{  3165.845}{  3160.819}{  3280.411}
\SgenE{bca-lap   }{ -48762.12}
\Sgent{   284.423}{    24.977}{    22.089}{    22.116}{   219.489}{   219.412}{    49.868}
\SfusE{ -48784.65}{ -48784.65}{ -48784.65}{ -48762.12}{ -48762.12}{ -48772.83}
\Sfust{   127.587}{   117.046}{   117.175}{   219.489}{   219.412}{   119.513}
\SgenE{bca-greedy}{ -48785.57}
\Sgent{    46.765}{     2.189}{    52.198}{     0.593}{    54.692}{    55.364}{     4.714}
\SfusE{\Soptimal{ -48787.16}}{\Soptimal{ -48787.16}}{\Soptimal{ -48787.16}}{ -48785.57}{ -48785.57}{\Soptimal{ -48787.16}}
\Sfust{     2.767}{    56.845}{     0.776}{    54.692}{    55.364}{   146.800}
\SgenE{greedy    }{ -47406.62}
\Sgent{     4.206}{     2.402}{     0.033}{     0.035}{     8.455}{     8.524}{     0.896}
\SfusE{ -48784.83}{ -48784.83}{ -48784.83}{ -47406.62}{ -47406.62}{ -48503.03}
\Sfust{   128.563}{     2.909}{     3.017}{     8.455}{     8.524}{   162.455}
\Sendtable

\Sinstance{worms20}\Salias{mir61L1\_1228062}\Soptimum{-49416.46}\\
\Sstarttable
\SgenE{dd-ls0    }{ -44546.66}
\Sgent{   526.537}{    71.418}{    26.215}{    26.128}{   537.187}{   539.332}{   396.361}
\SfusE{ -49183.83}{ -49183.83}{ -49174.26}{ -44546.66}{ -44546.66}{ -48680.04}
\Sfust{  1160.714}{   479.870}{   479.554}{   537.187}{   539.332}{   649.211}
\SgenE{dd-ls3    }{ -47716.14}
\Sgent{   687.109}{   106.130}{    34.183}{    33.429}{   698.583}{   699.447}{   596.470}
\SfusE{ -49352.55}{ -49370.08}{ -49370.08}{ -47716.14}{ -47716.14}{ -49283.18}
\Sfust{  1303.301}{   465.766}{   467.716}{   698.583}{   699.447}{   817.997}
\SgenE{dd-ls4    }{ -48016.68}
\Sgent{  3183.582}{   198.700}{    81.926}{    82.108}{  3201.952}{  3196.147}{  1891.757}
\SfusE{ -49369.30}{ -49369.30}{ -49369.30}{ -48016.68}{ -48016.68}{ -49339.55}
\Sfust{  3048.064}{  2292.323}{  2295.038}{  3201.952}{  3196.147}{  3247.611}
\SgenE{bca-lap   }{ -49292.22}
\Sgent{    34.580}{     7.629}{     6.102}{     6.112}{    33.759}{    33.754}{    12.082}
\SfusE{ -49373.93}{ -49373.93}{ -49373.93}{ -49292.22}{ -49292.22}{ -49362.12}
\Sfust{   256.348}{   234.066}{   234.320}{    33.759}{    33.754}{   182.537}
\SgenE{bca-greedy}{ -49415.70}
\Sgent{    81.490}{     5.746}{     2.482}{     2.538}{    93.187}{    89.808}{    99.754}
\SfusE{\Soptimal{ -49416.46}}{\Soptimal{ -49416.46}}{\Soptimal{ -49416.46}}{ -49415.70}{ -49415.70}{\Soptimal{ -49416.46}}
\Sfust{     5.746}{     2.482}{     2.538}{    93.187}{    89.808}{   102.937}
\SgenE{greedy    }{ -47789.00}
\Sgent{     4.725}{     2.655}{     0.041}{     0.038}{     8.934}{     9.440}{     0.483}
\SfusE{\Soptimal{ -49416.46}}{\Soptimal{ -49416.46}}{\Soptimal{ -49416.46}}{ -47789.00}{ -47789.00}{ -49131.39}
\Sfust{   335.444}{     1.531}{     1.432}{     8.934}{     9.440}{   146.639}
\Sendtable

\Sinstance{worms21}\Salias{mir61L1\_1229062}\Soptimum{-49836.77}\\
\Sstarttable
\SgenE{dd-ls0    }{ -45033.15}
\Sgent{   523.981}{    60.758}{    18.821}{    19.581}{   534.392}{   535.321}{   393.310}
\SfusE{ -49508.21}{ -49508.21}{ -49508.21}{ -45033.15}{ -45033.15}{ -48904.90}
\Sfust{   983.461}{   360.086}{   357.783}{   534.392}{   535.321}{   644.039}
\SgenE{dd-ls3    }{ -48623.83}
\Sgent{   613.958}{   864.275}{   228.008}{   228.570}{   624.684}{   624.007}{   561.326}
\SfusE{ -49831.10}{ -49831.10}{ -49831.10}{ -48623.83}{ -48623.83}{ -49733.71}
\Sfust{  1328.646}{   475.604}{   476.521}{   624.684}{   624.007}{   785.956}
\SgenE{dd-ls4    }{ -48909.20}
\Sgent{  3000.582}{   401.120}{   192.584}{   192.706}{  3016.672}{  3015.087}{  2065.119}
\SfusE{\Soptimal{ -49836.77}}{\Soptimal{ -49836.77}}{\Soptimal{ -49836.77}}{ -48909.20}{ -48909.20}{ -49808.04}
\Sfust{  2953.910}{  2128.545}{  2129.529}{  3016.672}{  3015.087}{  3448.203}
\SgenE{bca-lap   }{ -49815.22}
\Sgent{   268.775}{    19.170}{    17.337}{    17.364}{   232.906}{   233.025}{    25.659}
\SfusE{\Soptimal{ -49836.77}}{\Soptimal{ -49836.77}}{\Soptimal{ -49836.77}}{ -49815.22}{ -49815.22}{ -49831.58}
\Sfust{    56.625}{    52.730}{    52.796}{   232.906}{   233.025}{    69.884}
\SgenE{bca-greedy}{ -49834.96}
\Sgent{    26.327}{     2.624}{     1.015}{     1.081}{    29.095}{    29.089}{    36.188}
\SfusE{\Soptimal{ -49836.77}}{\Soptimal{ -49836.77}}{\Soptimal{ -49836.77}}{ -49834.96}{ -49834.96}{\Soptimal{ -49836.77}}
\Sfust{     2.624}{     1.015}{     1.081}{    29.095}{    29.089}{    53.806}
\SgenE{greedy    }{ -48573.41}
\Sgent{     5.292}{     0.897}{     0.019}{     0.019}{    10.907}{    11.100}{     1.117}
\SfusE{\Soptimal{ -49836.77}}{\Soptimal{ -49836.77}}{\Soptimal{ -49836.77}}{ -48573.41}{ -48573.41}{ -49489.21}
\Sfust{   378.267}{    10.708}{    11.835}{    10.907}{    11.100}{   137.114}
\Sendtable

\Sinstance{worms22}\Salias{pha4A7L1\_1213061}\Soptimum{-47994.44}\\
\Sstarttable
\SgenE{dd-ls0    }{ -45481.44}
\Sgent{   572.524}{   287.183}{    40.788}{    40.809}{   583.410}{   585.000}{   480.667}
\SfusE{ -47952.23}{ -47956.71}{ -47954.68}{ -45481.44}{ -45481.44}{ -47735.72}
\Sfust{  1347.101}{   329.781}{   332.628}{   583.410}{   585.000}{   710.277}
\SgenE{dd-ls3    }{ -45717.32}
\Sgent{   562.605}{   134.302}{    34.712}{    32.834}{   572.377}{   572.539}{   513.376}
\SfusE{\Soptimal{ -47994.44}}{\Soptimal{ -47994.44}}{\Soptimal{ -47994.44}}{ -45717.32}{ -45717.32}{ -47791.05}
\Sfust{  1265.607}{   586.955}{   589.478}{   572.377}{   572.539}{   825.142}
\SgenE{dd-ls4    }{ -45412.16}
\Sgent{  2556.655}{   304.263}{   137.914}{    92.640}{  2571.283}{  2567.877}{  1714.549}
\SfusE{\Soptimal{ -47994.44}}{ -47992.32}{\Soptimal{ -47994.44}}{ -45412.16}{ -45412.16}{ -47849.12}
\Sfust{  2416.052}{  1827.242}{  1830.224}{  2571.283}{  2567.877}{  2996.687}
\SgenE{bca-lap   }{ -47923.74}
\Sgent{    86.533}{    12.151}{    10.031}{    10.036}{    87.505}{    87.524}{    88.186}
\SfusE{ -47973.32}{ -47972.67}{ -47973.32}{ -47923.74}{ -47923.74}{ -47924.53}
\Sfust{    95.516}{    57.979}{    85.491}{    87.505}{    87.524}{    88.186}
\SgenE{bca-greedy}{ -47988.73}
\Sgent{     6.117}{     1.322}{     0.320}{     0.350}{     7.437}{     7.265}{    50.234}
\SfusE{\Soptimal{ -47994.44}}{\Soptimal{ -47994.44}}{\Soptimal{ -47994.44}}{ -47988.73}{ -47988.73}{ -47994.35}
\Sfust{     2.826}{     0.773}{     0.842}{     7.437}{     7.265}{    71.324}
\SgenE{greedy    }{ -46246.78}
\Sgent{    12.519}{     1.977}{     0.023}{     0.025}{    23.854}{    24.091}{     2.669}
\SfusE{\Soptimal{ -47994.44}}{\Soptimal{ -47994.44}}{\Soptimal{ -47994.44}}{ -46246.78}{ -46246.78}{ -47635.51}
\Sfust{   195.424}{    21.442}{    24.164}{    23.854}{    24.091}{   149.466}
\Sendtable

\Sinstance{worms23}\Salias{pha4A7L1\_1213062}\Soptimum{-49985.66}\\
\Sstarttable
\SgenE{dd-ls0    }{ -48000.33}
\Sgent{   557.036}{   192.881}{    44.621}{    44.736}{   569.452}{   569.425}{   432.747}
\SfusE{ -49959.28}{ -49962.83}{ -49959.28}{ -48000.33}{ -48000.33}{ -49808.50}
\Sfust{  1186.093}{   444.961}{   445.900}{   569.452}{   569.425}{   655.254}
\SgenE{dd-ls3    }{ -49264.13}
\Sgent{   631.604}{   225.558}{    53.930}{    56.076}{   641.737}{   641.673}{   569.272}
\SfusE{ -49982.11}{ -49982.11}{ -49982.11}{ -49264.13}{ -49264.13}{ -49979.29}
\Sfust{  1235.472}{   525.810}{   526.380}{   641.737}{   641.673}{   853.384}
\SgenE{dd-ls4    }{ -49850.80}
\Sgent{  3126.844}{   734.455}{   387.469}{   445.299}{  3139.087}{  3139.111}{  2054.883}
\SfusE{\Soptimal{ -49985.66}}{\Soptimal{ -49985.66}}{\Soptimal{ -49985.66}}{ -49850.80}{ -49850.80}{ -49980.74}
\Sfust{  3341.181}{  2605.135}{  2606.480}{  3139.087}{  3139.111}{  3060.050}
\SgenE{bca-lap   }{ -49985.15}
\Sgent{  1063.597}{    66.771}{    60.973}{    61.092}{   318.880}{   319.100}{   322.157}
\SfusE{\Soptimal{ -49985.66}}{\Soptimal{ -49985.66}}{\Soptimal{ -49985.66}}{ -49985.15}{ -49985.15}{ -49985.15}
\Sfust{    66.771}{    60.973}{    61.092}{   318.880}{   319.100}{   322.157}
\SgenE{bca-greedy}{\Soptimal{ -49985.66}}
\Sgent{    91.143}{    51.404}{    30.641}{    30.958}{    30.831}{    30.916}{    34.922}
\SfusE{\Soptimal{ -49985.66}}{\Soptimal{ -49985.66}}{\Soptimal{ -49985.66}}{\Soptimal{ -49985.66}}{\Soptimal{ -49985.66}}{\Soptimal{ -49985.66}}
\Sfust{    51.404}{    30.641}{    30.958}{    30.831}{    30.916}{    34.922}
\SgenE{greedy    }{ -48718.69}
\Sgent{     1.617}{     1.196}{     0.031}{     0.034}{     3.103}{     3.756}{     3.012}
\SfusE{\Soptimal{ -49985.66}}{\Soptimal{ -49985.66}}{\Soptimal{ -49985.66}}{ -48718.69}{ -48718.69}{ -49798.42}
\Sfust{    55.549}{     1.319}{     1.443}{     3.103}{     3.756}{   127.122}
\Sendtable

\Sinstance{worms24}\Salias{pha4A7L1\_1213064}\Soptimum{-49309.50}\\
\Sstarttable
\SgenE{dd-ls0    }{ -44641.80}
\Sgent{   407.212}{   106.444}{    16.110}{    16.399}{   417.182}{   417.183}{   388.570}
\SfusE{ -49275.91}{ -49275.91}{ -49275.91}{ -44641.80}{ -44641.80}{ -48657.63}
\Sfust{  1240.790}{   366.920}{   367.607}{   417.182}{   417.183}{   574.624}
\SgenE{dd-ls3    }{ -46520.06}
\Sgent{   645.929}{   122.281}{    19.703}{    19.703}{   656.191}{   655.923}{   561.063}
\SfusE{\Soptimal{ -49309.50}}{\Soptimal{ -49309.50}}{\Soptimal{ -49309.50}}{ -46520.06}{ -46520.06}{ -49117.68}
\Sfust{  1474.548}{   633.044}{   633.575}{   656.191}{   655.923}{   782.722}
\SgenE{dd-ls4    }{ -46700.02}
\Sgent{  2635.186}{   164.904}{    67.713}{    59.517}{  2646.284}{  2646.462}{  1691.717}
\SfusE{\Soptimal{ -49309.50}}{\Soptimal{ -49309.50}}{\Soptimal{ -49309.50}}{ -46700.02}{ -46700.02}{ -48998.45}
\Sfust{  3013.487}{  2167.104}{  2168.750}{  2646.284}{  2646.462}{  2653.379}
\SgenE{bca-lap   }{ -49297.29}
\Sgent{   293.119}{   200.725}{   183.262}{   183.484}{   291.746}{   291.843}{   293.636}
\SfusE{ -49305.20}{ -49305.20}{ -49305.20}{ -49297.29}{ -49297.29}{ -49303.48}
\Sfust{   306.129}{   277.498}{   277.827}{   291.746}{   291.843}{   293.636}
\SgenE{bca-greedy}{ -49308.59}
\Sgent{    23.737}{     2.863}{     0.774}{     0.787}{    28.283}{    28.567}{    24.140}
\SfusE{\Soptimal{ -49309.50}}{\Soptimal{ -49309.50}}{\Soptimal{ -49309.50}}{ -49308.59}{ -49308.59}{\Soptimal{ -49309.50}}
\Sfust{     3.853}{     1.158}{     1.175}{    28.283}{    28.567}{    24.486}
\SgenE{greedy    }{ -48078.74}
\Sgent{    11.543}{     1.777}{     0.035}{     0.042}{    22.651}{    22.508}{     0.842}
\SfusE{\Soptimal{ -49309.50}}{\Soptimal{ -49309.50}}{\Soptimal{ -49309.50}}{ -48078.74}{ -48078.74}{ -49139.77}
\Sfust{    34.027}{     2.588}{     0.919}{    22.651}{    22.508}{    87.107}
\Sendtable

\Sinstance{worms25}\Salias{pha4B2L1\_0125072}\Soptimum{-47233.63}\\
\Sstarttable
\SgenE{dd-ls0    }{ -40410.39}
\Sgent{   340.798}{    55.153}{     6.455}{     6.979}{   349.420}{   349.353}{   384.886}
\SfusE{ -47036.03}{ -47027.59}{ -47021.02}{ -40410.39}{ -40410.39}{ -46264.64}
\Sfust{  1096.209}{   301.578}{   305.310}{   349.420}{   349.353}{   541.425}
\SgenE{dd-ls3    }{ -44264.60}
\Sgent{   815.747}{   151.617}{    36.779}{    32.424}{   826.969}{   827.345}{   701.906}
\SfusE{ -47200.61}{ -47199.55}{ -47199.55}{ -44264.60}{ -44264.60}{ -47018.62}
\Sfust{  1772.980}{   822.566}{   824.889}{   826.969}{   827.345}{   985.705}
\SgenE{dd-ls4    }{ -43584.31}
\Sgent{  2786.506}{   182.263}{    67.381}{    78.400}{  2797.568}{  2797.628}{  1759.186}
\SfusE{ -47162.16}{ -47162.16}{ -47162.16}{ -43584.31}{ -43584.31}{ -46631.48}
\Sfust{  2722.664}{  1918.991}{  1919.018}{  2797.568}{  2797.628}{  2970.915}
\SgenE{bca-lap   }{ -47209.74}
\Sgent{   257.904}{   128.951}{   117.030}{   116.925}{   257.448}{   257.230}{   194.927}
\SfusE{ -47219.96}{ -47219.96}{ -47219.96}{ -47209.74}{ -47209.74}{ -47211.57}
\Sfust{   202.315}{   183.285}{   183.123}{   257.448}{   257.230}{   260.131}
\SgenE{bca-greedy}{ -47230.06}
\Sgent{    35.330}{     1.765}{     0.603}{     0.646}{    41.792}{    42.283}{     9.377}
\SfusE{\Soptimal{ -47233.63}}{\Soptimal{ -47233.63}}{\Soptimal{ -47233.63}}{ -47230.06}{ -47230.06}{ -47232.59}
\Sfust{     2.417}{     0.853}{     0.904}{    41.792}{    42.283}{    17.071}
\SgenE{greedy    }{ -44278.65}
\Sgent{    14.745}{     2.073}{     0.025}{     0.026}{    29.349}{    28.298}{     0.535}
\SfusE{\Soptimal{ -47233.63}}{\Soptimal{ -47233.63}}{\Soptimal{ -47233.63}}{ -44278.65}{ -44278.65}{ -46870.37}
\Sfust{    96.835}{     0.586}{     0.606}{    29.349}{    28.298}{   230.675}
\Sendtable

\Sinstance{worms26}\Salias{pha4I2L\_0408071}\Soptimum{-46119.91}\\
\Sstarttable
\SgenE{dd-ls0    }{ -40580.74}
\Sgent{   435.120}{   224.247}{    11.220}{    11.291}{   447.634}{   446.009}{   456.403}
\SfusE{ -45825.06}{ -45825.06}{ -45825.06}{ -40580.74}{ -40580.74}{ -45158.28}
\Sfust{  1350.394}{   377.839}{   377.541}{   447.634}{   446.009}{   597.417}
\SgenE{dd-ls3    }{ -40892.62}
\Sgent{   720.006}{   123.178}{    32.417}{    23.432}{   731.813}{   730.919}{   579.037}
\SfusE{ -45966.15}{ -45967.06}{ -45965.67}{ -40892.62}{ -40892.62}{ -45524.26}
\Sfust{  1625.674}{   729.055}{   507.497}{   731.813}{   730.919}{   899.628}
\SgenE{dd-ls4    }{ -42700.91}
\Sgent{  3070.955}{   154.228}{    47.900}{    47.909}{  3087.622}{  3083.382}{  1926.849}
\SfusE{ -46077.51}{ -46077.51}{ -46076.55}{ -42700.91}{ -42700.91}{ -45895.73}
\Sfust{  3592.625}{  2686.924}{  2714.268}{  3087.622}{  3083.382}{  3263.146}
\SgenE{bca-lap   }{ -46041.98}
\Sgent{  1100.639}{    63.144}{    59.019}{    59.085}{  1039.943}{  1039.555}{    68.987}
\SfusE{ -46106.35}{ -46102.53}{ -46106.35}{ -46041.98}{ -46041.98}{ -46079.45}
\Sfust{   245.960}{   231.284}{   231.550}{  1039.943}{  1039.555}{   330.331}
\SgenE{bca-greedy}{ -46112.46}
\Sgent{    12.875}{     2.217}{     0.647}{     0.631}{    15.359}{    15.649}{    14.183}
\SfusE{\Soptimal{ -46119.91}}{\Soptimal{ -46119.91}}{\Soptimal{ -46119.91}}{ -46112.46}{ -46112.46}{ -46118.77}
\Sfust{    13.050}{     5.150}{     5.018}{    15.359}{    15.649}{    84.005}
\SgenE{greedy    }{ -42787.68}
\Sgent{     1.357}{     1.770}{     0.024}{     0.030}{     2.617}{     2.597}{     0.966}
\SfusE{\Soptimal{ -46119.91}}{ -46119.69}{ -46119.69}{ -42787.68}{ -42787.68}{ -45535.70}
\Sfust{  1354.701}{     4.679}{     5.989}{     2.617}{     2.597}{   163.991}
\Sendtable

\Sinstance{worms27}\Salias{pha4I2L\_0408072}\Soptimum{-50062.40}\\
\Sstarttable
\SgenE{dd-ls0    }{ -48332.15}
\Sgent{   393.033}{   159.685}{    27.238}{    48.467}{   404.275}{   405.523}{   363.271}
\SfusE{ -50061.94}{\Soptimal{ -50062.40}}{ -50061.94}{ -48332.15}{ -48332.15}{ -49910.65}
\Sfust{   981.308}{   365.781}{   303.880}{   404.275}{   405.523}{   524.622}
\SgenE{dd-ls3    }{ -49422.69}
\Sgent{   819.145}{   149.040}{    43.851}{    43.994}{   830.340}{   830.322}{   695.310}
\SfusE{ -50061.94}{ -50061.94}{ -50061.94}{ -49422.69}{ -49422.69}{ -49953.92}
\Sfust{  1195.216}{   498.235}{   499.714}{   830.340}{   830.322}{   947.681}
\SgenE{dd-ls4    }{ -49547.02}
\Sgent{  2583.131}{   169.090}{    83.600}{    81.460}{  2598.679}{  2594.276}{  1943.079}
\SfusE{ -50061.94}{ -50061.94}{ -50061.94}{ -49547.02}{ -49547.02}{ -49952.83}
\Sfust{  2171.674}{  1472.129}{  1473.243}{  2598.679}{  2594.276}{  2713.363}
\SgenE{bca-lap   }{ -50061.94}
\Sgent{   173.708}{   191.013}{   175.557}{   175.393}{   175.358}{   175.511}{   177.018}
\SfusE{ -50061.94}{ -50061.94}{ -50061.94}{ -50061.94}{ -50061.94}{ -50061.94}
\Sfust{   191.013}{   175.557}{   175.393}{   175.358}{   175.511}{   177.018}
\SgenE{bca-greedy}{\Soptimal{ -50062.40}}
\Sgent{    12.892}{     0.996}{     0.302}{     0.330}{    13.626}{    13.986}{     1.236}
\SfusE{\Soptimal{ -50062.40}}{\Soptimal{ -50062.40}}{\Soptimal{ -50062.40}}{\Soptimal{ -50062.40}}{\Soptimal{ -50062.40}}{\Soptimal{ -50062.40}}
\Sfust{     0.996}{     0.302}{     0.330}{    13.626}{    13.986}{     1.236}
\SgenE{greedy    }{ -48506.93}
\Sgent{    16.697}{     3.503}{     0.061}{     0.069}{    31.454}{    31.228}{     1.556}
\SfusE{\Soptimal{ -50062.40}}{\Soptimal{ -50062.40}}{\Soptimal{ -50062.40}}{ -48506.93}{ -48506.93}{ -49999.89}
\Sfust{    25.991}{     0.538}{     0.613}{    31.454}{    31.228}{   102.813}
\Sendtable

\Sinstance{worms28}\Salias{pha4I2L\_0408073}\Soptimum{-49497.10}\\
\Sstarttable
\SgenE{dd-ls0    }{ -47484.31}
\Sgent{   551.011}{   188.621}{    41.942}{    41.284}{   562.249}{   563.737}{   491.462}
\SfusE{ -49458.97}{ -49463.31}{ -49458.92}{ -47484.31}{ -47484.31}{ -49254.19}
\Sfust{  1513.142}{   577.574}{   574.185}{   562.249}{   563.737}{   665.630}
\SgenE{dd-ls3    }{ -48039.83}
\Sgent{   806.546}{  1073.655}{   361.952}{   363.000}{   819.083}{   822.326}{   658.798}
\SfusE{ -49463.05}{ -49463.05}{ -49463.05}{ -48039.83}{ -48039.83}{ -49430.66}
\Sfust{  1606.667}{   709.488}{   711.238}{   819.083}{   822.326}{   931.339}
\SgenE{dd-ls4    }{ -48855.75}
\Sgent{  3229.234}{  1528.114}{   838.059}{   838.208}{  3247.030}{  3243.361}{  1911.992}
\SfusE{\Soptimal{ -49497.10}}{\Soptimal{ -49497.10}}{\Soptimal{ -49497.10}}{ -48855.75}{ -48855.75}{ -49463.77}
\Sfust{  3091.279}{  2147.542}{  2147.830}{  3247.030}{  3243.361}{  3539.742}
\SgenE{bca-lap   }{ -49479.43}
\Sgent{    55.114}{    52.465}{    46.902}{    46.965}{    54.601}{    54.595}{    53.286}
\SfusE{ -49484.61}{ -49484.61}{ -49484.61}{ -49479.43}{ -49479.43}{ -49482.31}
\Sfust{   390.506}{   357.231}{   357.610}{    54.601}{    54.595}{    55.200}
\SgenE{bca-greedy}{ -49496.97}
\Sgent{    34.356}{     2.449}{     0.845}{     0.896}{    38.953}{    38.893}{     7.078}
\SfusE{\Soptimal{ -49497.10}}{\Soptimal{ -49497.10}}{\Soptimal{ -49497.10}}{ -49496.97}{ -49496.97}{\Soptimal{ -49497.10}}
\Sfust{     2.449}{     0.845}{     0.896}{    38.953}{    38.893}{     7.078}
\SgenE{greedy    }{ -47702.79}
\Sgent{     1.377}{     1.245}{     0.024}{     0.026}{     2.664}{     2.666}{     0.506}
\SfusE{\Soptimal{ -49497.10}}{\Soptimal{ -49497.10}}{\Soptimal{ -49497.10}}{ -47702.79}{ -47702.79}{ -49430.49}
\Sfust{   272.938}{     6.803}{     7.696}{     2.664}{     2.666}{    64.438}
\Sendtable

\Sinstance{worms29}\Salias{unc54L1\_0123071}\Soptimum{-50069.17}\\
\Sstarttable
\SgenE{dd-ls0    }{ -48795.38}
\Sgent{   486.719}{   614.054}{   113.230}{   107.785}{   498.147}{   496.551}{   395.867}
\SfusE{ -50063.09}{ -50047.69}{ -50047.69}{ -48795.38}{ -48795.38}{ -49956.65}
\Sfust{  1124.328}{   427.754}{   427.190}{   498.147}{   496.551}{   609.759}
\SgenE{dd-ls3    }{ -49509.08}
\Sgent{   661.707}{   590.318}{   287.043}{   286.896}{   671.262}{   675.757}{   612.254}
\SfusE{ -50068.14}{ -50068.14}{ -50068.14}{ -49509.08}{ -49509.08}{ -50036.53}
\Sfust{  1416.590}{   660.565}{   660.088}{   671.262}{   675.757}{   906.000}
\SgenE{dd-ls4    }{ -48109.92}
\Sgent{  2505.508}{  1137.220}{   580.631}{   581.708}{  2519.815}{  2518.604}{  1461.610}
\SfusE{\Soptimal{ -50069.17}}{\Soptimal{ -50069.17}}{\Soptimal{ -50069.17}}{ -48109.92}{ -48109.92}{ -50031.33}
\Sfust{  3068.765}{  2356.842}{  2264.104}{  2519.815}{  2518.604}{  2695.304}
\SgenE{bca-lap   }{\Soptimal{ -50069.17}}
\Sgent{   127.831}{    58.354}{    54.448}{    54.518}{   127.779}{   127.730}{   128.493}
\SfusE{\Soptimal{ -50069.17}}{\Soptimal{ -50069.17}}{\Soptimal{ -50069.17}}{\Soptimal{ -50069.17}}{\Soptimal{ -50069.17}}{\Soptimal{ -50069.17}}
\Sfust{    58.354}{    54.448}{    54.518}{   127.779}{   127.730}{   128.493}
\SgenE{bca-greedy}{\Soptimal{ -50069.17}}
\Sgent{    13.624}{     4.511}{     2.063}{     2.116}{     5.844}{     5.924}{     3.110}
\SfusE{\Soptimal{ -50069.17}}{\Soptimal{ -50069.17}}{\Soptimal{ -50069.17}}{\Soptimal{ -50069.17}}{\Soptimal{ -50069.17}}{\Soptimal{ -50069.17}}
\Sfust{     4.511}{     2.063}{     2.116}{     5.844}{     5.924}{     3.110}
\SgenE{greedy    }{ -49034.56}
\Sgent{     2.093}{     0.932}{     0.017}{     0.020}{     4.222}{     4.283}{     0.977}
\SfusE{\Soptimal{ -50069.17}}{\Soptimal{ -50069.17}}{\Soptimal{ -50069.17}}{ -49034.56}{ -49034.56}{ -49983.59}
\Sfust{    54.033}{     1.363}{     1.521}{     4.222}{     4.283}{    99.180}
\Sendtable

\Sinstance{worms30}\Salias{unc54L1\_0123072}\Soptimum{-49775.89}\\
\Sstarttable
\SgenE{dd-ls0    }{ -47371.72}
\Sgent{   502.234}{   191.032}{    41.544}{    47.562}{   512.461}{   515.250}{   425.754}
\SfusE{ -49769.96}{ -49769.96}{ -49769.96}{ -47371.72}{ -47371.72}{ -49707.00}
\Sfust{  1293.836}{   322.683}{   292.977}{   512.461}{   515.250}{   656.779}
\SgenE{dd-ls3    }{ -49308.89}
\Sgent{   734.261}{   885.267}{   303.447}{   303.682}{   745.303}{   744.405}{   656.583}
\SfusE{ -49775.85}{ -49769.47}{ -49775.85}{ -49308.89}{ -49308.89}{ -49767.45}
\Sfust{  1211.961}{   493.059}{   493.421}{   745.303}{   744.405}{   973.388}
\SgenE{dd-ls4    }{ -49354.42}
\Sgent{  2683.330}{   978.124}{   854.857}{   539.742}{  2698.155}{  2694.190}{  1703.415}
\SfusE{\Soptimal{ -49775.89}}{\Soptimal{ -49775.89}}{ -49774.99}{ -49354.42}{ -49354.42}{ -49754.83}
\Sfust{  2031.596}{  1580.836}{  2368.479}{  2698.155}{  2694.190}{  2203.605}
\SgenE{bca-lap   }{ -49765.18}
\Sgent{   498.829}{    38.954}{    35.120}{    35.160}{   467.845}{   467.549}{   177.916}
\SfusE{ -49766.92}{ -49766.92}{ -49766.92}{ -49765.18}{ -49765.18}{ -49766.17}
\Sfust{   242.640}{   221.678}{   221.975}{   467.845}{   467.549}{   224.388}
\SgenE{bca-greedy}{\Soptimal{ -49775.89}}
\Sgent{    40.474}{    81.312}{    47.165}{    44.540}{    46.952}{    45.526}{    51.366}
\SfusE{\Soptimal{ -49775.89}}{\Soptimal{ -49775.89}}{\Soptimal{ -49775.89}}{\Soptimal{ -49775.89}}{\Soptimal{ -49775.89}}{\Soptimal{ -49775.89}}
\Sfust{    81.312}{    47.165}{    44.540}{    46.952}{    45.526}{    51.366}
\SgenE{greedy    }{ -48268.19}
\Sgent{    16.528}{     1.677}{     0.029}{     0.030}{    28.780}{    28.996}{     3.102}
\SfusE{\Soptimal{ -49775.89}}{\Soptimal{ -49775.89}}{\Soptimal{ -49775.89}}{ -48268.19}{ -48268.19}{ -49660.21}
\Sfust{   191.868}{     6.789}{     6.956}{    28.780}{    28.996}{    89.892}
\Sendtable

\Ssubsection{pairs}

\Snrinst{16}

\Snriter{5000}

\Sinstance{pairs1}\Salias{worm01-worm04}\\
\Sstarttable
\SgenE{dd-ls0    }{ -62323.61}
\Sgent{   225.385}{     0.789}{     0.209}{     0.238}{   344.081}{   346.997}{    55.198}
\SfusE{ -66198.81}{ -66123.76}{ -66159.73}{ -62323.61}{ -62323.61}{ -63359.04}
\Sfust{   633.439}{   236.402}{   310.527}{   344.081}{   346.997}{   302.616}
\SgenE{dd-ls3    }{ -64038.59}
\Sgent{   660.568}{    20.089}{     8.807}{     8.877}{   770.297}{   772.007}{   814.549}
\SfusE{ -66237.05}{ -66207.37}{ -66207.37}{ -64038.59}{ -64038.59}{ -65027.35}
\Sfust{  1054.250}{   486.425}{   492.454}{   770.297}{   772.007}{   934.969}
\SgenE{dd-ls4    }{ -64349.51}
\Sgent{  1330.636}{    36.998}{    12.734}{    19.426}{  1456.461}{  1436.319}{   908.343}
\SfusE{ -66255.68}{ -66107.07}{ -65994.56}{ -64349.51}{ -64349.51}{ -65097.66}
\Sfust{  1425.181}{   968.407}{  1414.788}{  1456.461}{  1436.319}{  1604.907}
\SgenE{bca-lap   }{ -63148.52}
\Sgent{     0.644}{     1.240}{     1.101}{     1.110}{     1.141}{     1.142}{     1.216}
\SfusE{ -64726.00}{ -64139.84}{ -64329.25}{ -63148.52}{ -63148.52}{ -63250.79}
\Sfust{  1934.312}{   917.798}{   922.416}{     1.141}{     1.142}{     1.663}
\SgenE{bca-greedy}{ -60737.06}
\Sgent{   756.951}{     3.958}{     1.979}{     1.344}{   879.970}{   900.166}{   175.436}
\SfusE{ -66179.08}{ -66141.78}{ -66197.67}{ -60737.06}{ -60737.06}{ -62488.32}
\Sfust{  2216.929}{   572.275}{   875.892}{   879.970}{   900.166}{  1837.377}
\SgenE{greedy    }{ -57559.71}
\Sgent{   103.682}{     1.665}{     0.284}{     0.248}{   473.068}{   458.942}{    19.730}
\SfusE{ -65856.68}{ -65920.84}{ -65865.44}{ -57559.71}{ -57559.71}{ -60883.07}
\Sfust{  1323.739}{   210.445}{   147.323}{   473.068}{   458.942}{   770.574}
\Sendtable

\Sinstance{pairs2}\Salias{worm02-worm22}\\
\Sstarttable
\SgenE{dd-ls0    }{ -68763.98}
\Sgent{   178.628}{    50.637}{    14.415}{    15.165}{   275.714}{   274.005}{   276.356}
\SfusE{ -69989.17}{ -69948.49}{ -69966.07}{ -68763.98}{ -68763.98}{ -69075.53}
\Sfust{   599.427}{   311.029}{   362.007}{   275.714}{   274.005}{   474.033}
\SgenE{dd-ls3    }{ -69385.61}
\Sgent{   637.464}{   167.233}{    89.626}{    90.348}{   756.857}{   757.852}{   790.979}
\SfusE{ -70024.48}{ -69927.47}{ -69857.92}{ -69385.61}{ -69385.61}{ -69385.61}
\Sfust{   912.930}{   851.295}{   461.332}{   756.857}{   757.852}{   790.979}
\SgenE{dd-ls4    }{ -69477.29}
\Sgent{   986.763}{   236.789}{   181.445}{   213.632}{  1110.407}{  1105.436}{  1148.672}
\SfusE{ -69968.22}{ -70033.55}{ -69977.55}{ -69477.29}{ -69477.29}{ -69477.29}
\Sfust{  1754.051}{  1459.559}{  1501.382}{  1110.407}{  1105.436}{  1148.672}
\SgenE{bca-lap   }{ -68634.64}
\Sgent{    28.024}{    10.598}{     9.920}{    10.140}{    31.316}{    31.390}{    31.402}
\SfusE{ -69330.11}{ -69283.69}{ -69283.69}{ -68634.64}{ -68634.64}{ -68717.80}
\Sfust{  1008.914}{   844.748}{   872.235}{    31.316}{    31.390}{    33.477}
\SgenE{bca-greedy}{ -67943.14}
\Sgent{  1205.613}{     4.997}{     2.174}{     2.702}{  1395.785}{  1397.557}{   839.094}
\SfusE{ -69980.76}{ -69940.80}{ -69984.35}{ -67943.14}{ -67943.14}{ -68349.39}
\Sfust{  1264.270}{   425.481}{  1485.110}{  1395.785}{  1397.557}{  1518.680}
\SgenE{greedy    }{ -62662.39}
\Sgent{    55.527}{     1.200}{     0.174}{     0.196}{   200.481}{   229.538}{    67.227}
\SfusE{ -69777.90}{ -69868.19}{ -69800.63}{ -62662.39}{ -62662.39}{ -64573.11}
\Sfust{   706.087}{   139.012}{   232.847}{   200.481}{   229.538}{   440.063}
\Sendtable

\Sinstance{pairs3}\Salias{worm03-worm11}\\
\Sstarttable
\SgenE{dd-ls0    }{ -65443.70}
\Sgent{   270.021}{    10.934}{     7.609}{     9.973}{   418.594}{   425.106}{   330.433}
\SfusE{ -67904.95}{ -67781.37}{ -67781.37}{ -65443.70}{ -65443.70}{ -66329.54}
\Sfust{   668.487}{   319.626}{   331.806}{   418.594}{   425.106}{   525.852}
\SgenE{dd-ls3    }{ -66860.76}
\Sgent{   830.851}{    83.154}{    46.424}{    71.692}{  1008.972}{  1014.992}{   925.419}
\SfusE{ -67917.65}{ -67843.37}{ -67839.52}{ -66860.76}{ -66860.76}{ -66989.23}
\Sfust{  1149.491}{   705.862}{   640.080}{  1008.972}{  1014.992}{  1167.527}
\SgenE{dd-ls4    }{ -67286.85}
\Sgent{  1759.710}{   203.340}{    97.741}{   187.420}{  1929.131}{  1949.353}{  1871.347}
\SfusE{ -67925.74}{ -67889.35}{ -67796.13}{ -67286.85}{ -67286.85}{ -67358.39}
\Sfust{  1463.081}{   837.859}{   850.326}{  1929.131}{  1949.353}{  1927.942}
\SgenE{bca-lap   }{ -64152.33}
\Sgent{    31.709}{     1.964}{     2.981}{     3.082}{    35.354}{    34.932}{     4.505}
\SfusE{ -66169.69}{ -66001.40}{ -66005.73}{ -64152.33}{ -64152.33}{ -64538.70}
\Sfust{   435.495}{   317.031}{   330.001}{    35.354}{    34.932}{   140.513}
\SgenE{bca-greedy}{ -63518.69}
\Sgent{   669.350}{     3.440}{     2.333}{     1.425}{   780.056}{   781.197}{   664.823}
\SfusE{ -67719.33}{ -67842.99}{ -67807.46}{ -63518.69}{ -63518.69}{ -64360.09}
\Sfust{  2291.774}{   791.752}{   431.626}{   780.056}{   781.197}{  1854.605}
\SgenE{greedy    }{ -59697.93}
\Sgent{     7.313}{     1.251}{     0.151}{     0.229}{    39.058}{    37.515}{    73.969}
\SfusE{ -67710.97}{ -67727.42}{ -67641.96}{ -59697.93}{ -59697.93}{ -62864.89}
\Sfust{  1199.464}{   157.238}{   215.090}{    39.058}{    37.515}{   806.957}
\Sendtable

\Sinstance{pairs4}\Salias{worm07-worm20}\\
\Sstarttable
\SgenE{dd-ls0    }{ -59010.48}
\Sgent{   205.891}{     1.040}{     0.326}{     0.345}{   313.819}{   322.118}{    85.697}
\SfusE{ -62956.25}{ -62870.73}{ -62848.42}{ -59010.48}{ -59010.48}{ -61005.01}
\Sfust{   690.392}{   233.220}{   255.797}{   313.819}{   322.118}{   314.826}
\SgenE{dd-ls3    }{ -61417.09}
\Sgent{   733.935}{    28.027}{    15.415}{    14.458}{   884.296}{   881.834}{   916.289}
\SfusE{ -62963.73}{ -62843.26}{ -62885.11}{ -61417.09}{ -61417.09}{ -61417.09}
\Sfust{  1067.861}{   660.836}{   811.921}{   884.296}{   881.834}{   916.289}
\SgenE{dd-ls4    }{ -61235.69}
\Sgent{   885.647}{    30.033}{    15.352}{    14.443}{   986.723}{   983.553}{   704.103}
\SfusE{ -62912.23}{ -63008.66}{ -62883.80}{ -61235.69}{ -61235.69}{ -61699.48}
\Sfust{   889.365}{   792.350}{  1099.446}{   986.723}{   983.553}{  1432.758}
\SgenE{bca-lap   }{ -60684.24}
\Sgent{    40.045}{    13.060}{    42.169}{    17.551}{    44.122}{    44.059}{    44.857}
\SfusE{ -61356.83}{ -61200.16}{ -61272.79}{ -60684.24}{ -60684.24}{ -60823.51}
\Sfust{   888.772}{   723.064}{   750.636}{    44.122}{    44.059}{    50.674}
\SgenE{bca-greedy}{ -59553.80}
\Sgent{   156.434}{     5.772}{     2.606}{     1.784}{   184.301}{   180.881}{   197.565}
\SfusE{ -62970.80}{ -62943.73}{ -62939.94}{ -59553.80}{ -59553.80}{ -60225.88}
\Sfust{  2220.470}{  1519.703}{   982.066}{   184.301}{   180.881}{  1716.330}
\SgenE{greedy    }{ -55209.09}
\Sgent{    72.492}{     1.720}{     0.147}{     0.141}{   379.908}{   367.298}{    33.206}
\SfusE{ -62824.56}{ -62825.58}{ -62845.67}{ -55209.09}{ -55209.09}{ -57835.72}
\Sfust{  1199.586}{   188.246}{   217.494}{   379.908}{   367.298}{   638.929}
\Sendtable

\Sinstance{pairs5}\Salias{worm10-worm27}\\
\Sstarttable
\SgenE{dd-ls0    }{ -58435.60}
\Sgent{   268.482}{     7.813}{     1.983}{     2.005}{   434.164}{   437.703}{   351.897}
\SfusE{ -61121.07}{ -61091.26}{ -61113.72}{ -58435.60}{ -58435.60}{ -59564.10}
\Sfust{   579.552}{   181.005}{   308.953}{   434.164}{   437.703}{   521.349}
\SgenE{dd-ls3    }{ -59255.12}
\Sgent{   780.535}{    26.767}{    13.296}{    12.735}{   964.316}{   966.621}{   695.707}
\SfusE{ -61170.68}{ -61140.14}{ -61172.94}{ -59255.12}{ -59255.12}{ -59789.30}
\Sfust{  1217.305}{   838.567}{   795.395}{   964.316}{   966.621}{   939.185}
\SgenE{dd-ls4    }{ -59695.72}
\Sgent{  1727.199}{    36.492}{    22.627}{    22.692}{  1938.863}{  1935.184}{  1042.383}
\SfusE{ -61201.75}{ -61187.97}{ -61168.12}{ -59695.72}{ -59695.72}{ -60202.64}
\Sfust{  1597.650}{   835.693}{   836.607}{  1938.863}{  1935.184}{  1962.239}
\SgenE{bca-lap   }{ -58487.47}
\Sgent{    40.791}{    14.513}{    12.235}{    12.421}{    46.181}{    46.873}{    46.392}
\SfusE{ -59825.23}{ -59810.02}{ -59756.95}{ -58487.47}{ -58487.47}{ -58516.37}
\Sfust{  2006.337}{   817.417}{   831.562}{    46.181}{    46.873}{    49.411}
\SgenE{bca-greedy}{ -56690.16}
\Sgent{  1582.648}{     4.180}{     2.951}{     2.431}{  1925.784}{  1916.374}{   287.581}
\SfusE{ -61180.03}{ -61181.13}{ -61163.78}{ -56690.16}{ -56690.16}{ -57564.56}
\Sfust{  2408.642}{  1431.679}{  1686.886}{  1925.784}{  1916.374}{  2194.520}
\SgenE{greedy    }{ -51709.07}
\Sgent{    32.473}{     1.761}{     0.376}{     0.377}{   231.201}{   228.583}{    27.348}
\SfusE{ -61009.39}{ -60773.76}{ -60759.88}{ -51709.07}{ -51709.07}{ -54008.31}
\Sfust{  1370.492}{   200.203}{   227.219}{   231.201}{   228.583}{   913.723}
\Sendtable

\Sinstance{pairs6}\Salias{worm12-worm16}\\
\Sstarttable
\SgenE{dd-ls0    }{ -67396.46}
\Sgent{   438.319}{    71.732}{    24.981}{    24.644}{   544.162}{   552.163}{   604.722}
\SfusE{ -68320.05}{ -68295.53}{ -68304.57}{ -67396.46}{ -67396.46}{ -67593.53}
\Sfust{   767.133}{   211.874}{   424.658}{   544.162}{   552.163}{   723.749}
\SgenE{dd-ls3    }{ -68039.40}
\Sgent{   780.602}{   308.208}{   175.041}{   177.356}{   917.714}{   915.514}{   949.190}
\SfusE{ -68322.27}{ -68329.48}{ -68329.48}{ -68039.40}{ -68039.40}{ -68089.85}
\Sfust{   509.833}{   334.011}{   337.953}{   917.714}{   915.514}{  1103.800}
\SgenE{dd-ls4    }{ -68119.09}
\Sgent{   769.139}{   327.356}{   247.241}{   222.152}{   846.738}{   841.731}{   886.232}
\SfusE{ -68351.14}{ -68349.27}{ -68349.27}{ -68119.09}{ -68119.09}{ -68214.79}
\Sfust{   849.907}{   667.642}{   658.805}{   846.738}{   841.731}{  2086.937}
\SgenE{bca-lap   }{ -66758.26}
\Sgent{    35.649}{    10.223}{    11.092}{     9.455}{    38.470}{    38.465}{    40.461}
\SfusE{ -67721.88}{ -67711.49}{ -67721.32}{ -66758.26}{ -66758.26}{ -66964.88}
\Sfust{  1322.699}{  1417.802}{   939.010}{    38.470}{    38.465}{   127.916}
\SgenE{bca-greedy}{ -65640.46}
\Sgent{   631.523}{     3.755}{     1.503}{     1.826}{   718.160}{   707.236}{   388.793}
\SfusE{ -68321.83}{ -68256.61}{ -68351.14}{ -65640.46}{ -65640.46}{ -66326.67}
\Sfust{  1403.606}{   885.821}{  1064.277}{   718.160}{   707.236}{  1580.749}
\SgenE{greedy    }{ -60585.96}
\Sgent{    31.997}{     1.559}{     0.149}{     0.158}{   110.383}{   124.811}{    21.974}
\SfusE{ -68153.09}{ -68188.90}{ -68263.90}{ -60585.96}{ -60585.96}{ -63930.45}
\Sfust{   913.857}{   164.190}{   157.618}{   110.383}{   124.811}{   660.373}
\Sendtable

\Sinstance{pairs7}\Salias{worm14-worm06}\\
\Sstarttable
\SgenE{dd-ls0    }{ -59352.55}
\Sgent{   338.629}{     0.743}{     0.394}{     0.392}{   418.114}{   425.538}{    75.492}
\SfusE{ -63237.39}{ -63187.69}{ -63207.92}{ -59352.55}{ -59352.55}{ -61944.98}
\Sfust{   837.611}{   409.681}{   409.405}{   418.114}{   425.538}{   628.412}
\SgenE{dd-ls3    }{ -61058.18}
\Sgent{   601.017}{     4.271}{     2.544}{     1.883}{   700.177}{   699.493}{   330.922}
\SfusE{ -63277.90}{ -63252.17}{ -63206.73}{ -61058.18}{ -61058.18}{ -62176.50}
\Sfust{  1072.027}{   398.011}{   352.796}{   700.177}{   699.493}{   852.730}
\SgenE{dd-ls4    }{ -61842.39}
\Sgent{  1242.737}{    22.313}{    16.857}{    12.887}{  1356.511}{  1356.788}{   700.796}
\SfusE{ -63262.89}{ -63324.10}{ -63183.70}{ -61842.39}{ -61842.39}{ -62628.08}
\Sfust{  1055.336}{   962.342}{  1365.450}{  1356.511}{  1356.788}{  1456.051}
\SgenE{bca-lap   }{ -61703.64}
\Sgent{    96.569}{    17.315}{    15.551}{    15.260}{   109.181}{   108.547}{   116.200}
\SfusE{ -62548.24}{ -62478.34}{ -62529.85}{ -61703.64}{ -61703.64}{ -61849.99}
\Sfust{  1536.105}{  1549.989}{  1588.130}{   109.181}{   108.547}{   981.058}
\SgenE{bca-greedy}{ -60165.25}
\Sgent{   993.164}{     2.348}{     1.762}{     0.917}{  1138.079}{  1120.450}{   146.853}
\SfusE{ -63324.54}{ -63243.29}{ -63303.73}{ -60165.25}{ -60165.25}{ -61478.40}
\Sfust{  1081.270}{   552.985}{   848.346}{  1138.079}{  1120.450}{  1673.180}
\SgenE{greedy    }{ -56970.78}
\Sgent{    76.805}{     2.411}{     0.132}{     0.145}{   350.549}{   359.920}{    14.813}
\SfusE{ -63266.84}{ -63194.24}{ -63248.96}{ -56970.78}{ -56970.78}{ -59277.43}
\Sfust{   770.996}{   111.424}{   111.079}{   350.549}{   359.920}{   612.391}
\Sendtable

\Sinstance{pairs8}\Salias{worm15-worm05}\\
\Sstarttable
\SgenE{dd-ls0    }{ -69219.38}
\Sgent{   392.194}{    29.756}{    18.219}{    15.276}{   493.199}{   476.609}{   418.554}
\SfusE{ -70464.01}{ -70438.71}{ -70401.17}{ -69219.38}{ -69219.38}{ -69566.25}
\Sfust{   608.414}{   391.505}{   358.736}{   493.199}{   476.609}{   619.198}
\SgenE{dd-ls3    }{ -69728.12}
\Sgent{   742.695}{    34.686}{    32.466}{    22.038}{   881.202}{   882.046}{   610.304}
\SfusE{ -70438.71}{ -70388.43}{ -70464.01}{ -69728.12}{ -69728.12}{ -69942.72}
\Sfust{   541.526}{   351.471}{   339.745}{   881.202}{   882.046}{  1013.475}
\SgenE{dd-ls4    }{ -69944.90}
\Sgent{  1376.478}{    73.708}{    34.570}{    36.915}{  1523.706}{  1528.337}{  1090.141}
\SfusE{ -70486.60}{ -70464.01}{ -70486.60}{ -69944.90}{ -69944.90}{ -70120.24}
\Sfust{   539.547}{  1005.352}{   517.176}{  1523.706}{  1528.337}{  1366.377}
\SgenE{bca-lap   }{ -69316.14}
\Sgent{    44.604}{    10.015}{     9.241}{     9.438}{    50.034}{    49.577}{    10.651}
\SfusE{ -70116.29}{ -70019.11}{ -70098.15}{ -69316.14}{ -69316.14}{ -69558.89}
\Sfust{  1647.297}{  1357.019}{   870.557}{    50.034}{    49.577}{    11.089}
\SgenE{bca-greedy}{ -68577.86}
\Sgent{   888.776}{     5.557}{     1.930}{     2.015}{   992.529}{  1022.140}{   865.959}
\SfusE{ -70460.11}{ -70475.03}{ -70451.63}{ -68577.86}{ -68577.86}{ -68974.88}
\Sfust{  1639.294}{  1234.486}{  1496.547}{   992.529}{  1022.140}{  1511.708}
\SgenE{greedy    }{ -63453.69}
\Sgent{     3.228}{     2.077}{     0.131}{     0.166}{    13.320}{    12.994}{    24.012}
\SfusE{ -70393.00}{ -70406.36}{ -70393.68}{ -63453.69}{ -63453.69}{ -65663.07}
\Sfust{   417.241}{   121.938}{   214.136}{    13.320}{    12.994}{   631.432}
\Sendtable

\Sinstance{pairs9}\Salias{worm16-worm03}\\
\Sstarttable
\SgenE{dd-ls0    }{ -66312.79}
\Sgent{   333.338}{     0.505}{     0.140}{     0.156}{   429.161}{   425.367}{    96.084}
\SfusE{ -68944.90}{ -68866.74}{ -68872.99}{ -66312.79}{ -66312.79}{ -68022.41}
\Sfust{   834.976}{   370.566}{   242.576}{   429.161}{   425.367}{   649.992}
\SgenE{dd-ls3    }{ -67719.90}
\Sgent{   655.959}{    14.353}{     6.814}{     7.728}{   767.343}{   761.918}{   432.375}
\SfusE{ -68957.89}{ -68910.91}{ -68926.82}{ -67719.90}{ -67719.90}{ -68605.09}
\Sfust{  1034.926}{   555.687}{   407.711}{   767.343}{   761.918}{   913.084}
\SgenE{dd-ls4    }{ -68004.11}
\Sgent{  1029.736}{    29.132}{    17.614}{    16.091}{  1132.563}{  1139.842}{   598.167}
\SfusE{ -68971.13}{ -68914.45}{ -68951.41}{ -68004.11}{ -68004.11}{ -68541.56}
\Sfust{  1119.128}{   531.828}{  1370.011}{  1132.563}{  1139.842}{  1306.457}
\SgenE{bca-lap   }{ -68179.87}
\Sgent{   482.039}{    55.476}{    50.066}{    54.112}{   506.009}{   515.939}{   304.200}
\SfusE{ -68610.11}{ -68411.64}{ -68573.56}{ -68179.87}{ -68179.87}{ -68295.31}
\Sfust{   699.792}{   927.525}{   561.486}{   506.009}{   515.939}{   432.099}
\SgenE{bca-greedy}{ -66485.55}
\Sgent{  1400.920}{     3.647}{     2.158}{     1.952}{  1585.196}{  1594.090}{   441.102}
\SfusE{ -68955.16}{ -68954.00}{ -68954.00}{ -66485.55}{ -66485.55}{ -67642.65}
\Sfust{   405.537}{   764.113}{   770.915}{  1585.196}{  1594.090}{  1738.323}
\SgenE{greedy    }{ -62640.09}
\Sgent{    48.367}{     1.926}{     0.125}{     0.161}{   200.439}{   205.336}{    22.670}
\SfusE{ -68774.87}{ -68827.25}{ -68737.12}{ -62640.09}{ -62640.09}{ -64746.81}
\Sfust{   571.845}{   143.519}{   157.604}{   200.439}{   205.336}{   598.252}
\Sendtable

\Sinstance{pairs10}\Salias{worm19-worm26}\\
\Sstarttable
\SgenE{dd-ls0    }{ -57872.80}
\Sgent{   431.910}{     8.865}{     3.534}{     3.587}{   570.076}{   566.124}{   527.260}
\SfusE{ -60580.40}{ -60570.45}{ -60533.28}{ -57872.80}{ -57872.80}{ -58943.66}
\Sfust{   937.652}{   427.210}{   438.228}{   570.076}{   566.124}{   640.770}
\SgenE{dd-ls3    }{ -59205.04}
\Sgent{   718.726}{    41.690}{    17.336}{    17.943}{   869.505}{   875.416}{   858.961}
\SfusE{ -60669.27}{ -60656.95}{ -60682.05}{ -59205.04}{ -59205.04}{ -59319.45}
\Sfust{  1104.943}{   660.968}{   699.440}{   869.505}{   875.416}{   908.932}
\SgenE{dd-ls4    }{ -59781.61}
\Sgent{  1541.061}{   169.747}{    73.973}{   103.797}{  1708.237}{  1713.150}{  1742.680}
\SfusE{ -60777.12}{ -60778.25}{ -60785.31}{ -59781.61}{ -59781.61}{ -59781.61}
\Sfust{  1611.885}{  1643.122}{  1346.195}{  1708.237}{  1713.150}{  1742.680}
\SgenE{bca-lap   }{ -57767.81}
\Sgent{    20.351}{     1.857}{     1.671}{     1.704}{    23.642}{    23.404}{    25.391}
\SfusE{ -59159.71}{ -59057.77}{ -59232.16}{ -57767.81}{ -57767.81}{ -57997.52}
\Sfust{   818.056}{   928.044}{  1641.269}{    23.642}{    23.404}{    40.751}
\SgenE{bca-greedy}{ -56382.86}
\Sgent{   646.693}{     5.631}{     3.232}{     2.820}{   778.382}{   778.525}{   306.116}
\SfusE{ -60751.30}{ -60612.29}{ -60749.08}{ -56382.86}{ -56382.86}{ -57138.20}
\Sfust{  1933.949}{  1375.631}{  1543.716}{   778.382}{   778.525}{  2019.253}
\SgenE{greedy    }{ -52177.25}
\Sgent{    37.315}{     1.828}{     0.224}{     0.215}{   365.618}{   379.232}{   121.754}
\SfusE{ -60619.59}{ -60499.54}{ -60540.62}{ -52177.25}{ -52177.25}{ -54614.53}
\Sfust{  1309.688}{   155.011}{   193.781}{   365.618}{   379.232}{   842.068}
\Sendtable

\Sinstance{pairs11}\Salias{worm20-worm12}\\
\Sstarttable
\SgenE{dd-ls0    }{ -60802.88}
\Sgent{   330.941}{     3.671}{     1.830}{     1.508}{   418.376}{   420.285}{   171.939}
\SfusE{ -64134.22}{ -64340.09}{ -64191.32}{ -60802.88}{ -60802.88}{ -62778.96}
\Sfust{   970.470}{   387.710}{   457.250}{   418.376}{   420.285}{   565.803}
\SgenE{dd-ls3    }{ -62002.67}
\Sgent{   559.477}{    12.771}{     5.522}{     4.156}{   671.021}{   668.065}{   354.293}
\SfusE{ -64291.58}{ -64323.05}{ -64264.71}{ -62002.67}{ -62002.67}{ -63290.01}
\Sfust{   871.892}{   657.581}{   672.558}{   671.021}{   668.065}{   793.246}
\SgenE{dd-ls4    }{ -62301.95}
\Sgent{  1093.928}{    21.210}{     9.599}{     9.885}{  1199.823}{  1217.818}{   549.420}
\SfusE{ -64355.30}{ -64390.31}{ -64412.51}{ -62301.95}{ -62301.95}{ -63345.58}
\Sfust{  1755.903}{   746.586}{  1062.277}{  1199.823}{  1217.818}{  1361.393}
\SgenE{bca-lap   }{ -62733.78}
\Sgent{   835.742}{    51.686}{    61.227}{    63.002}{   916.504}{   915.236}{  1019.754}
\SfusE{ -63490.63}{ -63407.72}{ -63516.47}{ -62733.78}{ -62733.78}{ -62743.01}
\Sfust{  1719.886}{  1704.608}{  1742.465}{   916.504}{   915.236}{  1939.395}
\SgenE{bca-greedy}{ -61486.39}
\Sgent{  1433.497}{     7.142}{     2.247}{     2.308}{  1648.800}{  1659.977}{   691.630}
\SfusE{ -64336.77}{ -64435.32}{ -64352.63}{ -61486.39}{ -61486.39}{ -62433.38}
\Sfust{   488.656}{  1196.386}{  1296.391}{  1648.800}{  1659.977}{  1838.179}
\SgenE{greedy    }{ -57614.80}
\Sgent{    57.227}{     1.106}{     0.150}{     0.131}{   291.254}{   297.717}{    87.555}
\SfusE{ -64256.21}{ -64166.27}{ -64078.77}{ -57614.80}{ -57614.80}{ -59753.30}
\Sfust{  1221.986}{   129.106}{   119.412}{   291.254}{   297.717}{   627.903}
\Sendtable

\Sinstance{pairs12}\Salias{worm21-worm09}\\
\Sstarttable
\SgenE{dd-ls0    }{ -65274.10}
\Sgent{   248.566}{    22.907}{     5.245}{     4.905}{   320.159}{   320.208}{   212.249}
\SfusE{ -67293.17}{ -67232.68}{ -67259.04}{ -65274.10}{ -65274.10}{ -66106.76}
\Sfust{   838.330}{   409.547}{   335.734}{   320.159}{   320.208}{   605.705}
\SgenE{dd-ls3    }{ -66015.95}
\Sgent{   508.817}{    53.020}{    28.214}{    14.049}{   611.729}{   607.693}{   391.031}
\SfusE{ -67305.35}{ -67221.32}{ -67270.68}{ -66015.95}{ -66015.95}{ -66753.40}
\Sfust{   864.686}{   752.699}{   455.992}{   611.729}{   607.693}{   880.806}
\SgenE{dd-ls4    }{ -66334.92}
\Sgent{  1210.430}{    58.381}{    77.254}{    75.373}{  1337.444}{  1347.318}{   555.752}
\SfusE{ -67288.77}{ -67271.91}{ -67286.65}{ -66334.92}{ -66334.92}{ -66646.17}
\Sfust{  1684.150}{  1316.769}{   844.502}{  1337.444}{  1347.318}{   587.659}
\SgenE{bca-lap   }{ -65477.34}
\Sgent{  1687.904}{    36.424}{    34.507}{    35.110}{  1875.205}{  1868.586}{   135.297}
\SfusE{ -66644.80}{ -66410.55}{ -66415.11}{ -65477.34}{ -65477.34}{ -65922.99}
\Sfust{  2037.196}{  1765.756}{   795.764}{  1875.205}{  1868.586}{  1213.000}
\SgenE{bca-greedy}{ -64487.02}
\Sgent{   378.702}{     5.748}{     3.081}{     3.562}{   431.830}{   424.579}{   350.748}
\SfusE{ -67290.28}{ -67305.96}{ -67258.27}{ -64487.02}{ -64487.02}{ -65412.90}
\Sfust{   772.453}{   953.557}{   511.162}{   431.830}{   424.579}{  1798.255}
\SgenE{greedy    }{ -59760.49}
\Sgent{    23.185}{     2.123}{     0.147}{     0.138}{   121.054}{   106.381}{    50.835}
\SfusE{ -67173.74}{ -67180.21}{ -67209.45}{ -59760.49}{ -59760.49}{ -62512.52}
\Sfust{   944.718}{   139.685}{   207.766}{   121.054}{   106.381}{   591.734}
\Sendtable

\Sinstance{pairs13}\Salias{worm22-worm25}\\
\Sstarttable
\SgenE{dd-ls0    }{ -64495.98}
\Sgent{   540.401}{    60.378}{    34.852}{    33.200}{   675.971}{   673.895}{   350.397}
\SfusE{ -66220.33}{ -66253.84}{ -66234.37}{ -64495.98}{ -64495.98}{ -65128.05}
\Sfust{   631.110}{   381.943}{   416.161}{   675.971}{   673.895}{   491.076}
\SgenE{dd-ls3    }{ -65546.07}
\Sgent{   529.794}{   319.195}{   123.500}{    98.584}{   659.461}{   655.666}{   708.440}
\SfusE{ -66252.78}{ -66255.77}{ -66259.46}{ -65546.07}{ -65546.07}{ -65703.41}
\Sfust{  1006.101}{   565.063}{   337.241}{   659.461}{   655.666}{   907.220}
\SgenE{dd-ls4    }{ -65558.80}
\Sgent{   949.005}{   189.952}{   105.550}{   181.181}{  1040.542}{  1060.899}{  1096.406}
\SfusE{ -66259.55}{ -66210.18}{ -66232.76}{ -65558.80}{ -65558.80}{ -65676.68}
\Sfust{   616.748}{   537.547}{  1520.614}{  1040.542}{  1060.899}{  1310.651}
\SgenE{bca-lap   }{ -63891.20}
\Sgent{    37.465}{    13.309}{    12.923}{    13.038}{    41.273}{    41.605}{    42.567}
\SfusE{ -64782.00}{ -64967.18}{ -64828.55}{ -63891.20}{ -63891.20}{ -63894.34}
\Sfust{  1427.607}{  1205.359}{  1219.947}{    41.273}{    41.605}{    47.366}
\SgenE{bca-greedy}{ -63038.58}
\Sgent{   453.532}{     3.558}{     1.681}{     1.738}{   550.922}{   562.715}{   238.336}
\SfusE{ -66215.82}{ -66252.94}{ -66211.31}{ -63038.58}{ -63038.58}{ -64002.03}
\Sfust{  1219.435}{   363.373}{   651.267}{   550.922}{   562.715}{  1333.721}
\SgenE{greedy    }{ -58097.29}
\Sgent{    30.317}{     1.603}{     0.113}{     0.190}{   182.284}{   184.989}{    22.548}
\SfusE{ -66052.12}{ -66106.75}{ -66130.10}{ -58097.29}{ -58097.29}{ -61089.59}
\Sfust{  1134.013}{   119.947}{   169.513}{   182.284}{   184.989}{   607.387}
\Sendtable

\Sinstance{pairs14}\Salias{worm25-worm18}\\
\Sstarttable
\SgenE{dd-ls0    }{ -65124.56}
\Sgent{   522.739}{     1.931}{     1.545}{     0.883}{   639.249}{   637.212}{   228.965}
\SfusE{ -67123.00}{ -67168.97}{ -67123.00}{ -65124.56}{ -65124.56}{ -66246.07}
\Sfust{   624.042}{   375.501}{   335.753}{   639.249}{   637.212}{   740.169}
\SgenE{dd-ls3    }{ -65548.37}
\Sgent{   417.892}{    14.986}{     8.110}{     8.315}{   513.129}{   509.933}{   310.830}
\SfusE{ -67195.54}{ -67139.54}{ -66975.36}{ -65548.37}{ -65548.37}{ -66482.83}
\Sfust{   835.798}{   441.866}{   304.855}{   513.129}{   509.933}{   609.074}
\SgenE{dd-ls4    }{ -65715.62}
\Sgent{   882.842}{    22.434}{    12.854}{    12.829}{   958.442}{   976.172}{   418.606}
\SfusE{ -67212.82}{ -67171.71}{ -67210.17}{ -65715.62}{ -65715.62}{ -66622.77}
\Sfust{  1138.770}{   767.227}{   834.028}{   958.442}{   976.172}{   915.050}
\SgenE{bca-lap   }{ -65871.31}
\Sgent{   214.748}{    63.884}{    94.066}{    61.214}{   218.582}{   220.823}{   129.606}
\SfusE{ -66345.87}{ -66253.42}{ -66297.69}{ -65871.31}{ -65871.31}{ -66027.37}
\Sfust{   852.173}{  1339.132}{  1567.298}{   218.582}{   220.823}{   477.196}
\SgenE{bca-greedy}{ -64192.01}
\Sgent{   150.080}{     3.977}{     1.968}{     1.210}{   175.414}{   172.774}{   123.999}
\SfusE{ -67227.47}{ -67098.63}{ -67212.30}{ -64192.01}{ -64192.01}{ -65649.79}
\Sfust{  1484.417}{   796.738}{   605.433}{   175.414}{   172.774}{  1001.729}
\SgenE{greedy    }{ -60923.98}
\Sgent{    19.891}{     0.868}{     0.093}{     0.124}{   101.165}{   101.483}{    10.062}
\SfusE{ -67048.48}{ -67176.09}{ -67114.78}{ -60923.98}{ -60923.98}{ -62825.28}
\Sfust{   979.478}{   152.441}{    31.225}{   101.165}{   101.483}{   532.903}
\Sendtable

\Sinstance{pairs15}\Salias{worm29-worm10}\\
\Sstarttable
\SgenE{dd-ls0    }{ -58501.06}
\Sgent{   474.105}{     2.234}{     0.618}{     0.666}{   594.999}{   574.853}{   120.148}
\SfusE{ -62604.08}{ -62517.94}{ -62491.69}{ -58501.06}{ -58501.06}{ -61293.51}
\Sfust{   850.289}{   274.979}{   285.274}{   594.999}{   574.853}{   630.379}
\SgenE{dd-ls3    }{ -60651.86}
\Sgent{   519.567}{    17.447}{     2.904}{     2.594}{   644.994}{   643.189}{   311.264}
\SfusE{ -62664.12}{ -62611.25}{ -62660.99}{ -60651.86}{ -60651.86}{ -61901.19}
\Sfust{  1080.282}{   650.914}{   519.575}{   644.994}{   643.189}{   839.806}
\SgenE{dd-ls4    }{ -61251.32}
\Sgent{   985.329}{    70.954}{    23.989}{    19.208}{  1074.985}{  1085.390}{   745.036}
\SfusE{ -62737.08}{ -62731.48}{ -62719.54}{ -61251.32}{ -61251.32}{ -62136.28}
\Sfust{  1561.512}{  1489.977}{  1098.439}{  1074.985}{  1085.390}{  1460.789}
\SgenE{bca-lap   }{ -59411.17}
\Sgent{    63.852}{     3.260}{     3.395}{     3.088}{    70.366}{    69.820}{    38.897}
\SfusE{ -61118.31}{ -61170.12}{ -61134.69}{ -59411.17}{ -59411.17}{ -59942.45}
\Sfust{  1430.041}{  1429.066}{  1105.566}{    70.366}{    69.820}{   267.852}
\SgenE{bca-greedy}{ -58260.32}
\Sgent{   799.335}{     4.654}{     2.385}{     1.232}{  1017.211}{  1031.738}{   247.867}
\SfusE{ -62602.72}{ -62559.20}{ -62632.57}{ -58260.32}{ -58260.32}{ -59634.95}
\Sfust{  1719.295}{   381.981}{   594.603}{  1017.211}{  1031.738}{  1113.447}
\SgenE{greedy    }{ -54572.31}
\Sgent{    16.089}{     2.162}{     0.152}{     0.192}{   127.075}{   138.153}{    47.721}
\SfusE{ -62421.94}{ -62317.54}{ -62351.75}{ -54572.31}{ -54572.31}{ -56910.62}
\Sfust{  1239.378}{   122.924}{   165.537}{   127.075}{   138.153}{   810.663}
\Sendtable

\Sinstance{pairs16}\Salias{worm30-worm24}\\
\Sstarttable
\SgenE{dd-ls0    }{ -66930.42}
\Sgent{   373.689}{    51.973}{    14.517}{    13.055}{   465.735}{   466.303}{   295.429}
\SfusE{ -68514.68}{ -68519.91}{ -68501.62}{ -66930.42}{ -66930.42}{ -67811.09}
\Sfust{   962.066}{   211.720}{   222.974}{   465.735}{   466.303}{   779.842}
\SgenE{dd-ls3    }{ -67880.07}
\Sgent{   365.432}{   111.097}{    54.730}{    50.762}{   467.932}{   463.630}{   294.497}
\SfusE{ -68641.31}{ -68609.32}{ -68606.48}{ -67880.07}{ -67880.07}{ -68168.43}
\Sfust{   870.010}{   552.864}{   577.833}{   467.932}{   463.630}{   599.154}
\SgenE{dd-ls4    }{ -68066.52}
\Sgent{  1366.473}{   201.251}{   116.252}{    83.422}{  1506.379}{  1520.751}{   591.036}
\SfusE{ -68535.31}{ -68600.26}{ -68592.56}{ -68066.52}{ -68066.52}{ -68280.73}
\Sfust{   937.398}{   603.115}{   347.027}{  1506.379}{  1520.751}{  1510.095}
\SgenE{bca-lap   }{ -67310.94}
\Sgent{    79.041}{    59.877}{    55.809}{    78.309}{    86.435}{    87.235}{    89.291}
\SfusE{ -67938.04}{ -67877.61}{ -67906.20}{ -67310.94}{ -67310.94}{ -67341.79}
\Sfust{   831.601}{  1041.442}{  1029.391}{    86.435}{    87.235}{    97.349}
\SgenE{bca-greedy}{ -65981.34}
\Sgent{   394.109}{     3.814}{     2.016}{     2.097}{   482.736}{   490.634}{   169.687}
\SfusE{ -68602.77}{ -68514.06}{ -68518.90}{ -65981.34}{ -65981.34}{ -67023.57}
\Sfust{  1230.825}{   248.996}{   405.268}{   482.736}{   490.634}{   956.900}
\SgenE{greedy    }{ -60968.82}
\Sgent{    17.928}{     0.973}{     0.108}{     0.171}{    97.081}{   108.983}{    30.767}
\SfusE{ -68452.62}{ -68380.37}{ -68453.19}{ -60968.82}{ -60968.82}{ -63176.06}
\Sfust{   734.007}{    98.381}{   177.845}{    97.081}{   108.983}{   581.942}
\Sendtable

\newpage
\section{Qualitative results}
\label{supp:qualitative-results}

Below we show some qualitative results for the \Sdata{worms} and \Sdata{pairs} datasets. As the instances of the datasets
\Sdata{hotel}, \Sdata{house}, \Sdata{car}, \Sdata{motor}
are largely solved to optimality by all state-of-the-art methods, \cf Tables~\ref{tab:comparison-detailled} and~\ref{tab:comparison-detailled-full}, we do not show qualitative results for these as they would be essentially identical for all methods. Unfortunately, for the datasets \Sdata{opengm}, \Sdata{flow} to the best of our knowledge no visualization corresponding to the models is publicly available.

\subsection{\Sdata{worms}}

Below we show qualitative results for the \Sdata{worms} dataset for the methods and time budgets also stated in Table~\ref{tab:comparison-detailled-full}. Each column corresponds to one of the methods \Salg{dd-ls0}~\cite{GraphMatchingDDTorresaniEtAl}, \Salg{dd-ls3}~\cite{GraphMatchingDDTorresaniEtAl}, \Salg{AMP}~\cite{swoboda2017study}, \Salg{AMP-tight}~\cite{swoboda2017study}, and our proposed method \Salg{bca-greedy+qpbo-i}. As \Salg{HBP}~\cite{HungarianBP} did not yield any results within the given time budget we omit it here.

Each line corresponds to an instance of the dataset numbered in the same manner as in Section~\ref{supp:subsec-worms}.
Nodes labeled correctly according to the atlas are marked in green, nodes labeled with the wrong label are marked in orange, and labeled nodes with no ground truth known are shown in grey. Simply put, the more green, the better the result.

\textit{Results for a time budget of 1 second.}

\begin{tikzpicture}
\begin{scope}[shift={(1.65,0)}]
  \node at (0,0) {\scriptsize\Salg{dd-ls0}};
  \node at (3.3,0) {\scriptsize\Salg{dd-ls3}};
  \node at (6.6,0) {\scriptsize\Salg{AMP}};
  \node at (9.9,0) {\scriptsize\Salg{AMP-tight}};
  \node at (13.2,0) {\scriptsize\textit{our} (\Salg{bca-greedy+qpbo-i})};
\end{scope}
\foreach \i/\j in {9/1,21/2}
  {
    \node [anchor=west, draw, inner sep=0pt] at (0,-0.55*\j)
      {\includegraphics[trim=1cm 4.4cm 1cm 4.4cm, clip, height=0.45cm]{images/qualitative-atlas/dd-ls0-1-\i.pdf}};
    \node [anchor=west, draw, inner sep=0pt] at (3.3,-0.55*\j)
      {\includegraphics[trim=1cm 4.4cm 1cm 4.4cm, clip, height=0.45cm]{images/qualitative-atlas/dd-ls2-1-\i.pdf}};
    \node [anchor=west, draw, inner sep=0pt] at (13.2,-0.55*\j)
      {\includegraphics[trim=1cm 4.4cm 1cm 4.4cm, clip, height=0.45cm]{images/qualitative-atlas/bca-greedy+qpbo-i-1-\i.pdf}};

    \node [anchor=east] at (0,-0.55*\j) {\scriptsize\i};
  }
\end{tikzpicture}

\textit{Results for a time budget of 10 seconds.}

\begin{tikzpicture}
\begin{scope}[shift={(1.65,0)}]
  \node at (0,0) {\scriptsize\Salg{dd-ls0}};
  \node at (3.3,0) {\scriptsize\Salg{dd-ls3}};
  \node at (6.6,0) {\scriptsize\Salg{AMP}};
  \node at (9.9,0) {\scriptsize\Salg{AMP-tight}};
  \node at (13.2,0) {\scriptsize\textit{our} (\Salg{bca-greedy+qpbo-i})};
\end{scope}
\foreach \i/\j in {9/1,21/2}
  {
    \node [anchor=west, draw, inner sep=0pt] at (0,-0.55*\j)
      {\includegraphics[trim=1cm 4.4cm 1cm 4.4cm, clip, height=0.45cm]{images/qualitative-atlas/dd-ls0-10-\i.pdf}};
    \node [anchor=west, draw, inner sep=0pt] at (3.3,-0.55*\j)
      {\includegraphics[trim=1cm 4.4cm 1cm 4.4cm, clip, height=0.45cm]{images/qualitative-atlas/dd-ls2-10-\i.pdf}};
    \node [anchor=west, draw, inner sep=0pt] at (6.6,-0.55*\j)
      {\includegraphics[trim=1cm 4.4cm 1cm 4.4cm, clip, height=0.45cm]{images/qualitative-atlas/AMP-10-\i.pdf}};
    \node [anchor=west, draw, inner sep=0pt] at (13.2,-0.55*\j)
      {\includegraphics[trim=1cm 4.4cm 1cm 4.4cm, clip, height=0.45cm]{images/qualitative-atlas/bca-greedy+qpbo-i-10-\i.pdf}};

    \node [anchor=east] at (0,-0.55*\j) {\scriptsize\i};
  }
\end{tikzpicture}

\textit{Results for a time budget of 30 seconds.}

\begin{tikzpicture}
\begin{scope}[shift={(1.65,0)}]
  \node at (0,0) {\scriptsize\Salg{dd-ls0}};
  \node at (3.3,0) {\scriptsize\Salg{dd-ls3}};
  \node at (6.6,0) {\scriptsize\Salg{AMP}};
  \node at (9.9,0) {\scriptsize\Salg{AMP-tight}};
  \node at (13.2,0) {\scriptsize\textit{our} (\Salg{bca-greedy+qpbo-i})};
\end{scope}
\foreach \i/\j in {9/1,21/2}
  {
    \node [anchor=west, draw, inner sep=0pt] at (0,-0.55*\j)
      {\includegraphics[trim=1cm 4.4cm 1cm 4.4cm, clip, height=0.45cm]{images/qualitative-atlas/dd-ls0-30-\i.pdf}};
    \node [anchor=west, draw, inner sep=0pt] at (3.3,-0.55*\j)
      {\includegraphics[trim=1cm 4.4cm 1cm 4.4cm, clip, height=0.45cm]{images/qualitative-atlas/dd-ls2-30-\i.pdf}};
    \node [anchor=west, draw, inner sep=0pt] at (6.6,-0.55*\j)
      {\includegraphics[trim=1cm 4.4cm 1cm 4.4cm, clip, height=0.45cm]{images/qualitative-atlas/AMP-30-\i.pdf}};
    \node [anchor=west, draw, inner sep=0pt] at (9.9,-0.55*\j)
      {\includegraphics[trim=1cm 4.4cm 1cm 4.4cm, clip, height=0.45cm]{images/qualitative-atlas/AMP-tightening-30-\i.pdf}};
    \node [anchor=west, draw, inner sep=0pt] at (13.2,-0.55*\j)
      {\includegraphics[trim=1cm 4.4cm 1cm 4.4cm, clip, height=0.45cm]{images/qualitative-atlas/bca-greedy+qpbo-i-30-\i.pdf}};

    \node [anchor=east] at (0,-0.55*\j) {\scriptsize\i};
  }
\end{tikzpicture}

\textit{Results for a time budget of 180 seconds.}

\begin{tikzpicture}
\begin{scope}[shift={(1.65,0)}]
  \node at (0,0) {\scriptsize\Salg{dd-ls0}};
  \node at (3.3,0) {\scriptsize\Salg{dd-ls3}};
  \node at (6.6,0) {\scriptsize\Salg{AMP}};
  \node at (9.9,0) {\scriptsize\Salg{AMP-tight}};
  \node at (13.2,0) {\scriptsize\textit{our} (\Salg{bca-greedy+qpbo-i})};
\end{scope}
\foreach \i/\j in {9/1,21/2}
  {
    \node [anchor=west, draw, inner sep=0pt] at (0,-0.55*\j)
      {\includegraphics[trim=1cm 4.4cm 1cm 4.4cm, clip, height=0.45cm]{images/qualitative-atlas/dd-ls0-180-\i.pdf}};
    \node [anchor=west, draw, inner sep=0pt] at (3.3,-0.55*\j)
      {\includegraphics[trim=1cm 4.4cm 1cm 4.4cm, clip, height=0.45cm]{images/qualitative-atlas/dd-ls2-180-\i.pdf}};
    \node [anchor=west, draw, inner sep=0pt] at (6.6,-0.55*\j)
      {\includegraphics[trim=1cm 4.4cm 1cm 4.4cm, clip, height=0.45cm]{images/qualitative-atlas/AMP-180-\i.pdf}};
    \node [anchor=west, draw, inner sep=0pt] at (9.9,-0.55*\j)
      {\includegraphics[trim=1cm 4.4cm 1cm 4.4cm, clip, height=0.45cm]{images/qualitative-atlas/AMP-tightening-180-\i.pdf}};
    \node [anchor=west, draw, inner sep=0pt] at (13.2,-0.55*\j)
      {\includegraphics[trim=1cm 4.4cm 1cm 4.4cm, clip, height=0.45cm]{images/qualitative-atlas/bca-greedy+qpbo-i-180-\i.pdf}};

    \node [anchor=east] at (0,-0.55*\j) {\scriptsize\i};
  }
\end{tikzpicture}

\textit{Results for a time budget of 300 seconds.}

\begin{tikzpicture}
\begin{scope}[shift={(1.65,0)}]
  \node at (0,0) {\scriptsize\Salg{dd-ls0}};
  \node at (3.3,0) {\scriptsize\Salg{dd-ls3}};
  \node at (6.6,0) {\scriptsize\Salg{AMP}};
  \node at (9.9,0) {\scriptsize\Salg{AMP-tight}};
  \node at (13.2,0) {\scriptsize\textit{our} (\Salg{bca-greedy+qpbo-i})};
\end{scope}
\foreach \i/\j in {9/1,21/2}
  {
    \node [anchor=west, draw, inner sep=0pt] at (0,-0.55*\j)
      {\includegraphics[trim=1cm 4.4cm 1cm 4.4cm, clip, height=0.45cm]{images/qualitative-atlas/dd-ls0-300-\i.pdf}};
    \node [anchor=west, draw, inner sep=0pt] at (3.3,-0.55*\j)
      {\includegraphics[trim=1cm 4.4cm 1cm 4.4cm, clip, height=0.45cm]{images/qualitative-atlas/dd-ls2-300-\i.pdf}};
    \node [anchor=west, draw, inner sep=0pt] at (6.6,-0.55*\j)
      {\includegraphics[trim=1cm 4.4cm 1cm 4.4cm, clip, height=0.45cm]{images/qualitative-atlas/AMP-300-\i.pdf}};
    \node [anchor=west, draw, inner sep=0pt] at (9.9,-0.55*\j)
      {\includegraphics[trim=1cm 4.4cm 1cm 4.4cm, clip, height=0.45cm]{images/qualitative-atlas/AMP-tightening-300-\i.pdf}};
    \node [anchor=west, draw, inner sep=0pt] at (13.2,-0.55*\j)
      {\includegraphics[trim=1cm 4.4cm 1cm 4.4cm, clip, height=0.45cm]{images/qualitative-atlas/bca-greedy+qpbo-i-300-\i.pdf}};

    \node [anchor=east] at (0,-0.55*\j) {\scriptsize\i};
  }
\end{tikzpicture}

\clearpage
\subsection{\Sdata{pairs}}

Below we show qualitative data for the \Sdata{pairs} dataset for the methods \Salg{dd-ls0}~\cite{GraphMatchingDDTorresaniEtAl}, \Salg{AMP}~\cite{swoboda2017study}, and our proposed method \Salg{bca-greedy+qpbo-i}. Green lines indicate correct matchings when using the same labelings as for the atlas in the \Sdata{worms} dataset. Red lines indicate errors, and for yellow lines it is not known whether they are correct or not. Shown is the result for \Sdata{pairs} instance 12, which is also depicted in the teaser figure in the main paper.

\begin{center}
\rotatebox{90}{
\begin{tabular}{r*3c}\toprule
time budget & \Salg{dd-ls0} & \Salg{AMP} & \textit{our} (\Salg{bca-greedy+qpbo-i}) \\
\bottomrule
\end{tabular}}

\end{center}


\end{document}